\documentclass{article}

\PassOptionsToPackage{numbers,sort&compress}{natbib}
\usepackage[preprint]{neurips_2026}

\usepackage[utf8]{inputenc}
\usepackage[T1]{fontenc}
\usepackage{microtype}
\usepackage{graphicx}
\usepackage{booktabs}
\usepackage{multirow}
\usepackage{rotating}
\usepackage{tabularx}
\usepackage{array}
\usepackage{enumitem}
\usepackage{xcolor}
\usepackage{url}
\definecolor{mydarkblue}{rgb}{0,0.08,0.45}
\usepackage[
  colorlinks=true,
  linkcolor=mydarkblue,
  citecolor=mydarkblue,
  filecolor=mydarkblue,
  urlcolor=mydarkblue
]{hyperref}
\usepackage{algorithm}
\usepackage{algorithmic}

\usepackage{amsmath}
\usepackage{amssymb}
\usepackage{mathtools}
\usepackage{amsthm}

\newtheorem{theorem}{Theorem}[section]
\newtheorem{proposition}[theorem]{Proposition}

\theoremstyle{definition}
\newtheorem{definition}[theorem]{Definition}
\newtheorem{observation}[theorem]{Observation}
\theoremstyle{remark}

\newcommand{\R}{\mathbb{R}}
\newcommand{\E}{\mathbb{E}}
\newcommand{\norm}[1]{\left\lVert #1 \right\rVert}
\newcommand{\diag}{\operatorname{diag}}
\newcommand{\tril}{\operatorname{tril}}
\newcommand{\RMSNorm}{\operatorname{RMSNorm}}

\definecolor{contribshade}{rgb}{0.945,0.960,0.985}
\definecolor{contribbar}{rgb}{0.08,0.20,0.45}
\definecolor{priortext}{rgb}{0.32,0.36,0.43}
\definecolor{priorbar}{rgb}{0.72,0.75,0.80}
\newcommand{\priortag}[1]{\mbox{\footnotesize\textcolor{priortext}{\textnormal{[#1]}}}}
\newcommand{\priorbox}[2]{%
  \par\medskip\noindent
  {\setlength{\fboxsep}{0pt}%
  \colorbox{priorbar}{\hspace{2.5pt}%
    \setlength{\fboxsep}{6pt}%
    \colorbox{white}{%
      \parbox{\dimexpr\linewidth-2.5pt-12pt\relax}{%
        {\footnotesize\textbf{\textcolor{priortext}{#1}}}\par\vspace{2pt}#2%
      }%
    }%
  }}\par\medskip%
}
\newcommand{\contribbox}[2]{%
  \par\medskip\noindent
  {\setlength{\fboxsep}{0pt}%
  \colorbox{contribbar}{\hspace{2.5pt}%
    \setlength{\fboxsep}{6pt}%
    \colorbox{contribshade}{%
      \parbox{\dimexpr\linewidth-2.5pt-12pt\relax}{%
        {\footnotesize\textbf{\textcolor{contribbar}{#1}}}\par\vspace{2pt}#2%
      }%
    }%
  }}\par\medskip%
}

\title{Systolic Linear Attention:\texorpdfstring{\\}{ }Co-Designing Tile-Grouped Gated Delta\texorpdfstring{\\}{ }Attention for TPUs}

\author{%
  Gagik Amirkhanyan \\
  Google \\
  \texttt{agagik@google.com} \\
  \And
  Nuojin Cheng \\
  Google \\
  \texttt{chengnuojin@google.com} \\
}

\begin{document}

\maketitle

\begin{abstract}
Gated-delta linear attention layers such as Gated DeltaNet (GDN), Kimi Delta Attention (KDA), and GDN-2 are tuned for GPUs, where small tensor-core tiles and fast vector registers make unpacked heads, per-channel decay rescaling, and per-value-head chunk solves cheap. On Google TPUs, which concentrate compute in wide $256 \times 256$ systolic matrix units backed by a slower vector unit, direct ports leave half of each matrix tile padded (for $d_k = 128$ heads) and repeat decay rescaling and chunk solves across every value head. We present \textbf{Systolic Linear Attention (SLA)}, which reshapes gated delta attention around the matrix unit by widening keys to $256$ and sharing both keys and scalar write gates $\beta_t$ across grouped value heads ($R\times$ fewer chunk solves), grouping decay gates into four $64$-channel tiles, and fusing exact FP32 block-doubling chunk inversion and its backward reuse inside one on-chip kernel. With exact chunk inversion on TPU v7x, $2$-stage \texttt{SLA-R4} (four value heads per key head) runs $2.1\times$--$2.6\times$ faster than open-source MaxText KDA and $1.4\times$--$2.4\times$ faster than open-source Tokamax KDA (and $1$-kernel fused \texttt{SLA-R4} reaches $2.5\times$--$2.8\times$ and $1.5\times$--$2.4\times$ under unified single-chip protocols), while using $49\%$--$61\%$ less temporary high-bandwidth memory (HBM). Across a $2{,}048$-chip TPU v7x pre-training study of sparse Mixture-of-Experts (MoE) models from $6.5\text{B}$ to $63.1\text{B}$ nominal parameters, SLA trains $11\%$--$24\%$ faster than MaxText KDA ($4\%$--$21\%$ faster than Tokamax KDA), matches KDA perplexity on three paired seeds at $12.5\text{B}$, and stays within $3\%$ at other scales. Under controlled multi-seed equal-chip-hour pre-training against safe-clipped open-source Tokamax KDA on $64$-chip TPU v7x slices ($n = 3$ paired seeds per arm), $1$-kernel fused \texttt{SLA-R4} and fused \texttt{SLA-R4 + GDN-2} lower perplexity on all $3/3$ seeds at both $12.5\text{B}$ ($-0.37 \pm 0.11$ and $-0.41 \pm 0.35$) and $63.1\text{B}$ ($-1.48 \pm 0.21$ and $-1.41 \pm 0.16$), while $G = 4$ outperforms scalar $G = 1$ across $6/6$ paired $12.5\text{B}$ MoE and dense-LM seeds. Without clamping its decay gates, SLA also extrapolates from $8\text{K}$ training to $32\text{K}$--$64\text{K}$ context far better than unclamped KDA.
\end{abstract}

\section{Introduction}
\label{sec:intro}

\begin{figure}[t]
\centering
\includegraphics[width=\textwidth]{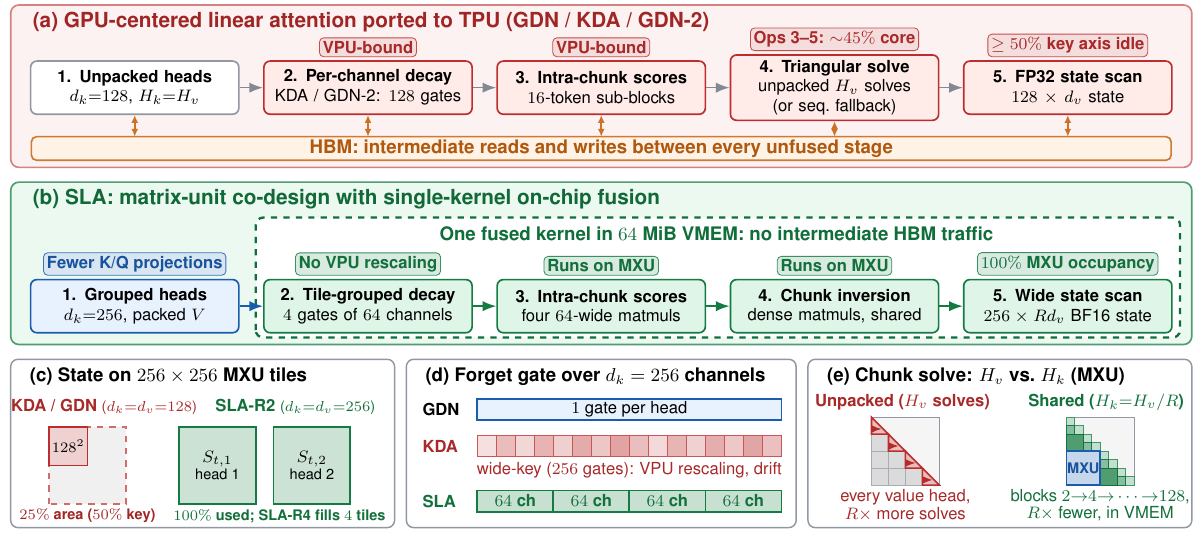}
\caption{\textbf{Overview of SLA.} \textbf{(a)} Direct ports of GPU-tuned gated-delta layers (GDN~\cite{Yang2025GDN}, KDA~\cite{Moonshot2025KimiLinear}, GDN-2~\cite{NVIDIA2026GDN2}) run as an unfused pipeline on TPUs: intermediates round-trip through HBM, per-channel decay rescaling in KDA and GDN-2 (and sequential triangular substitution in portable or FP32 fallback paths) runs on the VPU, $128$-wide states leave half of each $256 \times 256$ MXU tile's key axis padded, and per-value-head write gates force $H_v$ chunk solves per chunk. \textbf{(b)} SLA fuses stages 2--5 into one VMEM kernel with $R\times$ fewer solves. \textbf{(c)} A head with key and value widths $d_k = d_v = 128$ fills a quarter of an MXU tile's area ($50\%$ of the key axis), while SLA's wide packed state fills whole tiles. \textbf{(d)} Decay gates are grouped into four $64$-channel tiles, so intra-chunk scores are four $64$-wide MXU matmuls with no VPU rescaling. \textbf{(e)} Building on open-source MXU block solvers~\cite{MaxText2025Repo,Tokamax2025Repo}, SLA fuses $R\times$-fewer exact FP32 block-doubling chunk solves and backward reuse inside VMEM.}
\label{fig:sla_overview}
\label{fig:hero}
\end{figure}

% Paragraph 1: Evolution of Linear Attentions
Linear attention removes the quadratic compute and key-value (KV) cache of softmax attentions~\cite{Vaswani2017,Dao2022FlashAttention} by storing the context in a fixed-size state matrix $S_t \in \R^{d_k \times d_v}$ (for key width $d_k$ and value width $d_v$), updated at each token $t$ with its key $k_t \in \R^{d_k}$ and value $v_t \in \R^{d_v}$ as~\cite{Katharopoulos2020}:
\[
S_t = S_{t-1} + k_t v_t^\top.
\]
Because this sum grows without bound, the state saturates on long sequences. Gated models such as Mamba-2~\cite{Dao2024Mamba2} and GLA~\cite{Yang2024GLA} add a data-dependent forget (decay) gate $F_t \in (0, 1]^{d_k \times d_k}$,
\[
S_t = F_t S_{t-1} + k_t v_t^\top,
\]
but decay fades all stored keys together and hurts associative recall~\cite{Arora2023Zoology}.

To overwrite individual keys selectively, DeltaNet~\cite{Schlag2021DeltaNet,Yang2024DeltaNet} introduced the delta rule with unit-norm key $k_t$ ($\norm{k_t}_2 = 1$), scalar write gate $\beta_t$, and $d_k \times d_k$ identity $I_{d_k}$:
\[
S_t = (I_{d_k} - \beta_t k_t k_t^\top) S_{t-1} + \beta_t k_t v_t^\top.
\]
Viewing such updates as online gradient descent, prior work also added heavy-ball momentum so that a single noisy update cannot overwrite established memories~\cite{Polyak1964,Behrouz2025Titans}. Combining decay with the delta rule produced \textbf{Gated DeltaNet (GDN)}~\cite{Yang2025GDN}, adopted in Qwen3-Next~\cite{Qwen2025Qwen3Next}. \textbf{Kimi Delta Attention (KDA)}~\cite{Moonshot2025KimiLinear} then expanded GDN's scalar forget gate into a per-channel diagonal gate $F_t = \diag(f_t)$ with channel-wise decay vector $f_t \in (0, 1]^{d_k}$, and \textbf{GDN-2}~\cite{NVIDIA2026GDN2} decoupled per-channel erase and write gates. Section~\ref{sec:related} surveys further variants.

% Paragraph 2: Why Existing Linear Attentions Are Implicitly GPU-Centered
However, these architectures and their chunkwise WY training algorithms~\cite{Bischof1987WY,Yang2024GLA,Yang2025GDN} were developed and tuned on NVIDIA GPUs (\texttt{fla}~\cite{FLA2024Repo}). They map less well onto 2D systolic-array accelerators such as Google TPUs~\cite{Jouppi2017TPUv1,Jouppi2023TPUv4,GoogleCloud2025TPU7x}, for three reasons:

\paragraph{Small tiles and unpacked heads.}
GPU Tensor Cores operate on small fragments, so $d_k = d_v = 128$ with one key head per value head already reaches high occupancy. The TPU v6e/v7x Matrix Multiply Unit (MXU) is a $256 \times 256$ systolic array: a $128$-wide key head pads half of every tile's key axis, and even when $d_k = 256$ (as at $D = 1792$), unpacked heads repeat projections, convolutions, and chunk solves for every value head.

\paragraph{Per-channel decay gates.}
In KDA and GDN-2, each key channel decays at its own rate. Kernels must rescale keys elementwise in $16$-token sub-blocks to avoid float32 overflow. GPUs absorb this vector work cheaply; on TPUs, where matrix throughput exceeds vector throughput by roughly $128\times$, it stalls the kernel.

\paragraph{Per-value-head chunk solves and sequential fallbacks.}
The chunkwise WY algorithm~\cite{Bischof1987WY,Yang2025GDN} partitions sequences into chunks of length $C$ and solves a $C \times C$ unit lower-triangular system per write-gate head and chunk. GPU kernels (\texttt{fla}~\cite{FLA2024Repo}), XLA \texttt{solve\_triangular} paths~\cite{MaxText2025Repo}, and Tokamax's FP32 fallback~\cite{Tokamax2025Repo} substitute sequentially on vector lanes. Open-source TPU kernels already express individual solves with MXU matrix products: Tokamax~\cite{Tokamax2025Repo} uses a BF16 $8$-token Neumann series with Horner doubling ($C = 64$ for per-channel KDA, $C \in [16, 512]$ for scalar decay), and MaxText~\cite{MaxText2025Repo} uses exact block-doubling in its XLA GDN path. Yet existing architectures still assign independent write gates $\beta_t$ to every value head, forcing $H_v$ solves per chunk and unfused HBM round-trips (Ops 3--5 take about $45\%$ of core stateless KDA time in Table~\ref{tab:7op_audit}).

% Paragraph 3: Introducing SLA
Kernel tuning alone does not close this gap: fusing a GPU-shaped KDA layer into on-chip memory speeds it up by only $1.03\times$ on TPU v7x (Table~\ref{tab:7op_audit}), because the time goes to per-value-head chunk solves, per-channel decay rescaling, and padded MXU tiles rather than High-Bandwidth Memory (HBM) traffic alone. We therefore change the layer itself. \textbf{Systolic Linear Attention (SLA)} applies a block-diagonal, tile-grouped forget gate $F_t \in \R^{d_k \times d_k}$ to $S_{t-1}$ and then performs the delta update with key $k_t \in \R^{d_k}$, packed value $v_t \in \R^{R d_v}$ (stacking $R \in \{2, 4\}$ value heads per key head), and a write gate $\beta_t \in (0, 2)$ shared across the packed value heads:
\begin{equation}
S_t = F_t S_{t-1} + \beta_t k_t \big(v_t - (F_t S_{t-1})^\top k_t\big)^\top.
\label{eq:intro_sla}
\end{equation}
SLA defaults to $\mu = 0$ (no chunk-boundary momentum) and offers an optional Schur-guarded chunk-boundary momentum mode $\mu_c \le (1-\varphi_c)/(1+\varphi_c)$ kept in on-chip SRAM (VMEM); earlier $12.5\text{B}$ and $63.1\text{B}$ runs used fixed $\mu = 0.25$, while isolated multi-seed controls in Appendix~\ref{app:r87_controls} show no quality advantage over $\mu = 0$. SLA combines three changes (Section~\ref{sec:arch}):

\paragraph{Wide keys and shared $\beta_t$ across grouped value heads ($R\times$ solve reduction).}
SLA widens keys to $d_k = 256$ and shares both each key head (as in grouped value attention~\cite{Qwen2025Qwen3Next,Qwen2025GatedAttn}) and the scalar write gate $\beta_t \in (0, 2)$ across $R \in \{2, 4\}$ value heads (\texttt{SLA-R2}, \texttt{SLA-R4}). Unlike existing GVA linear layers that retain per-value-head $(g, \beta_t)$ and still run $H_v$ chunk solves, SLA packs the $R$ value heads into full $256 \times R d_v$ MXU tiles and cuts chunk solves by $R\times$ ($2.45\times$--$3.28\times$ faster core forward and $1.87\times$--$2.48\times$ faster full-layer F+B in an unconfounded same-shape control; Appendix Table~\ref{tab:app_shared_beta}).

\paragraph{Tile-grouped forget gate.}
SLA splits the $256$ key channels into $4$ tiles of $64$ channels, each with one scalar decay rate. Intra-chunk scores then become four $64$-wide MXU matmuls scaled by decay masks, with no VPU rescaling. This middle point between GDN's single rate per head ($G = 1$) and KDA's rate per channel ($G = d_k$) runs at near-scalar single-kernel VMEM speed while lowering language-modeling perplexity over $G = 1$ across all six paired $12.5\text{B}$ MoE and dense-LM seeds.

\paragraph{Fused VMEM exact inversion, backward reuse, and conditioning guarantee.}
Building on open-source MXU chunk solvers~\cite{MaxText2025Repo,Tokamax2025Repo}, SLA fuses its $R\times$-fewer exact FP32 recursive block-doubling solves and custom-VJP backward reuse inside one VMEM kernel, proves a chunk-size-independent conditioning bound (Theorem~\ref{thm:wy_cond}), and shows that exact FP32 block-doubling holds $10^{-7}$ accuracy on real chunks through $C = 256$ while adding only $4.0\%$--$7.7\%$ to full-LM step time once values are packed (Appendix~\ref{app:v2_solver_audit}).

% Paragraph 4: Evaluation summary and honest scope
We evaluate SLA on TPU v7x at the kernel, layer, serving, and context-parallel level, and across sparse Mixture-of-Experts (MoE) models~\cite{Shazeer2017MoE,Fedus2022Switch} from $6.5\text{B}$ ($7.58\text{B}$ exact) to $63.1\text{B}$ nominal parameters on Dolma~\cite{Soldaini2024Dolma,OLMo2025OLMo3} (Section~\ref{sec:eval}; Table~\ref{tab:intro_sla_vs_kda} summarizes the comparison with KDA). Building a single-kernel VMEM-fused per-channel KDA control separates the layer speedup against open-source Tokamax KDA into $1.28\times$--$1.41\times$ from $4{:}1$ head packing, $1.10\times$--$1.12\times$ from VMEM fusion, and $1.07\times$--$1.20\times$ from $G = 4$ tile grouping (Appendix Table~\ref{tab:app_3way_attrib}). At equal token count on $12.5\text{B}$ MoE, SLA matches KDA within seed noise and stays within about $0.1$ perplexity of a wide-key per-channel KDA control while avoiding unclamped slow-channel long-context drift (which also appears in public \texttt{Kimi-Linear-48B-A3B} checkpoints and can be mitigated in KDA via zero-retraining gate clamping). SLA is therefore best understood per token as a hardware-efficient operating point complementary to richer update rules such as GDN-2~\cite{NVIDIA2026GDN2}: under equal chip-hours on $64$-chip TPU v7x slices, both $1$-kernel fused \texttt{SLA-R4} and a fused \texttt{SLA-R4 + GDN-2} hybrid beat safe-clipped open-source Tokamax KDA across all $3/3$ paired seeds at $12.5\text{B}$ and $63.1\text{B}$ (Table~\ref{tab:moe_eqwall_tokamax}).

\begin{table}[ht]
\caption{\textbf{SLA vs.\ KDA at a glance on TPU v7x.} Layer, prefill, equal-chip-hour, and $63.1\text{B}$ rows use \texttt{SLA-R4} and the $60.4\text{B}$-token $12.5\text{B}$ rows use \texttt{SLA-R2}, all with exact inversion (\texttt{Exact Inv.}). Layer rows show $2$-stage \texttt{SLA-R4} at $D = 1792$ (single-chip $1$-kernel fused \texttt{SLA-R4}: $14.4\text{ ms}$, $2.76\times$ / $2.36\times$); at $D = 2048$, $2$-stage \texttt{SLA-R4} is $2.1\times$ / $1.5\times$ faster and uses $49\%$ / $52\%$ less HBM. MoE throughput compares against MaxText KDA across nominal $6.5\text{B}$ ($7.58\text{B}$ exact) to $63.1\text{B}$ scales ($4\%$--$21\%$ vs.\ Tokamax KDA in Appendix Table~\ref{tab:app_single_protocol}b; the short $5\text{B}$-token $31.5\text{B}$ [$30.58\text{B}$ exact] \texttt{SLA-R2} run is $1.08\times$, $1.14\times$ for \texttt{SLA-R4}). Equal-chip-hour rows compare $1$-kernel fused iso-parameter \texttt{SLA-R4} ($\mu = 0$; brackets: fused \texttt{SLA-R4 + GDN-2}) against safe-clipped open-source Tokamax Mosaic KDA across $n = 3$ paired seeds on $64$-chip TPU v7x slices. Clamping KDA's retention horizons to $8\text{K}$ raises its $64\text{K}$ \texttt{HotpotQA} score to $75.4\%$. Details in Tables~\ref{tab:v7x_kernels}--\ref{tab:moe_eqwall_tokamax}.}
\label{tab:intro_sla_vs_kda}
\centering
\footnotesize
\setlength{\tabcolsep}{2.5pt}
\renewcommand{\arraystretch}{1.12}
\begin{tabular*}{\textwidth}{@{\extracolsep{\fill}}lccc@{}}
\toprule
Benchmark & KDA & \textbf{SLA} & \textbf{SLA vs.\ KDA} \\
\midrule
Full layer, $32\text{K}$ fwd+bwd ($2$-stage; MaxText / Tokamax) & $48.7$ / $42.4\text{ ms}$ & $20.2\text{ ms}$ & \textbf{$2.4\times$ / $2.1\times$ faster} \\
Temporary HBM, $32\text{K}$ layer & $9.1$ / $9.3\text{ GiB}$ & $3.7\text{ GiB}$ & \textbf{$60\%$ / $61\%$ less} \\
$131\text{K}$ prefill, $32$-layer hybrid (Exact Inv., BF16) & $2.66\text{ s}$ & $1.62\text{ s}$ & \textbf{$1.64\times$ faster} \\
MoE training throughput, $6.5\text{B}$--$63.1\text{B}$ (vs.\ MaxText KDA) & $1.00\times$ & $1.11\times$--$1.24\times$ & \textbf{$11\%$--$24\%$ faster} \\
$12.5\text{B}$ equal-chip-h PPL vs.\ Tokamax ($42.7\text{ chip-h}, n{=}3$) & $19.52$ & $19.15\ [19.11]$ & \textbf{$-0.37\ [-0.41]$ ($3/3$ seeds)} \\
$63.1\text{B}$ equal-chip-h PPL vs.\ Tokamax ($37.3\text{ chip-h}, n{=}3$) & $26.56$ & $25.08\ [25.14]$ & \textbf{$-1.48\ [-1.41]$ ($3/3$ seeds)} \\
$12.5\text{B}$ iso-token PPL vs.\ MaxText ($60.4\text{B}$ tok, $n{=}3$) & $15.60$ & $15.58$ & $-0.02$ (within noise) \\
$63.1\text{B}$ iso-token PPL vs.\ MaxText ($120.8\text{B}$ tok, $n{=}1$) & $12.60$ & $12.45$ & $-0.15$ \\
$64\text{K}$ \texttt{HotpotQA}, $12.5\text{B}$, no clamping & $50.3\%$ & $77.9\%$ & \textbf{$+27.6$ pts} \\
\bottomrule
\end{tabular*}
\end{table}

In the following sections, Section~\ref{sec:arch} reviews gated delta attention, its TPU bottlenecks, and SLA's design; Section~\ref{sec:eval} evaluates SLA from kernels and serving to MoE pre-training; and Section~\ref{sec:related} covers related work.

\section{Hardware Bottlenecks and the SLA Design}
\label{sec:arch}

To motivate SLA, we first review gated delta recurrences and their chunkwise WY training algorithm (Section~\ref{subsec:gdn_prelim}), identify three hardware bottlenecks that GPU-centric designs hit on TPUs (Section~\ref{subsec:scan_law}), and then present SLA's mathematical formulation and chunk-boundary momentum (Section~\ref{subsec:recurrence}), its three algorithmic pillars and mathematical derivations (Section~\ref{subsec:pillars}), and its single-kernel VMEM backward pass (Section~\ref{app:bwd_alg}), followed by empirical evaluation in Section~\ref{sec:eval}.

\subsection{Preliminaries: Gated Delta Recurrences and Chunkwise WY Parallelism}
\label{subsec:gdn_prelim}
\label{subsec:seven_op}

Given an input sequence $X = (x_1, \dots, x_T) \in \R^{T \times D}$ of length $T$ and model width $D$, a gated-delta linear attention layer with $H_v$ value heads and $H_k$ query-key heads maps each token $x_t \in \R^D$ ($t \in \{1, \dots, T\}$) to $L_2$-normalized queries and keys $q_t, k_t \in \R^{d_k}$ ($\norm{q_t}_2 = \norm{k_t}_2 = 1$), values $v_t \in \R^{d_v}$, and head output $o_t \in \R^{d_v}$ (with $H_k = H_v$ in unpacked baselines). Rather than caching $t$ past tokens, each head maintains a fixed-size associative state matrix $S_t \in \R^{d_k \times d_v}$ updated by applying a diagonal forget gate $F_t \in (0, 1]^{d_k \times d_k}$ followed by a rank-one delta update~\cite{Schlag2021DeltaNet,Yang2024DeltaNet,Sun2024TTT,Behrouz2025Titans} with scalar step size $\beta_t$:
\priorbox{Prior Work: Gated Delta Recurrence in GDN and KDA}{%
\begin{align}
S_t &= F_t S_{t-1} + \beta_t k_t \big(v_t - (F_t S_{t-1})^\top k_t\big)^\top \nonumber\\
    &= \big(I_{d_k} - \beta_t k_t k_t^\top\big) F_t S_{t-1} + \beta_t k_t v_t^\top, \nonumber\\
o_t &= S_t^\top q_t.
\label{eq:prelim_kda}
\end{align}%
}
Here $(F_t S_{t-1})^\top k_t$ is the value currently retrieved for key $k_t$, and $\beta_t$ sets how far it is moved toward $v_t$. Existing linear attentions differ only in $(F_t, \beta_t)$ in Eq.~\eqref{eq:prelim_kda}, where we characterize $F_t$ by its number of \textbf{decay groups} $G$ (distinct decay rates across the $d_k$ key channels of a head). Unweighted Linear Attention~\cite{Katharopoulos2020} sets $F_t = I_{d_k}$ without erasure ($S_t = S_{t-1} + k_t v_t^\top$), while DeltaNet~\cite{Schlag2021DeltaNet,Yang2024DeltaNet} and test-time learning~\cite{Sun2024TTT,VonOswald2025MesaNet} retain $F_t = I_{d_k}$ with the rank-one delta update. Recent gated-delta architectures instantiate $(F_t, \beta_t)$ using the logistic sigmoid $\sigma(\cdot)$ on scalar write-gate logit $\tilde{\beta}_t \in \R$ and $\operatorname{softplus}(u) = \ln(1 + e^u)$ on forget-gate logits $\tilde{f}_t \in \R^G$:

\paragraph{Gated DeltaNet (GDN).}
\textbf{GDN}~\cite{Yang2025GDN} uses one scalar decay $\gamma_t \in (0, 1]$ per head ($G = 1$) together with a unit-interval scalar step size:
\[
F_t = \gamma_t I_{d_k}, \qquad \beta_t = \sigma(\tilde{\beta}_t) \in (0, 1).
\]

\paragraph{Kimi Delta Attention (KDA).}
\textbf{KDA}~\cite{Moonshot2025KimiLinear} assigns an independent decay rate to every key channel ($G = d_k$, $f_t = \exp(-\operatorname{softplus}(\tilde{f}_t))$):
\[
F_t = \diag(f_t), \qquad f_t \in (0, 1]^{d_k}.
\]
While standard GDN and KDA use $\beta_t = \sigma(\tilde{\beta}_t) \in (0, 1)$, negative-eigenvalue extensions~\cite{Grazzi2025NegEig} widen the step-size range to
\[
\beta_t = 2\sigma(\tilde{\beta}_t) \in (0, 2),
\]
so that $I_{d_k} - \beta_t k_t k_t^\top$ has eigenvalue $1 - \beta_t \in (-1, 1)$ along $k_t$ (adopted by SLA in Section~\ref{subsec:recurrence}).

\paragraph{GDN-2.}
\textbf{GDN-2}~\cite{NVIDIA2026GDN2} retains per-channel decay ($G = d_k$) and further replaces the scalar $\beta_t$ with separate channel-wise erase and write gates.

\paragraph{Chunkwise WY parallelization.}
To saturate matrix units during training, chunkwise WY parallelization~\cite{Bischof1987WY,Yang2024DeltaNet,Yang2025GDN} partitions the sequence into $T / C$ contiguous chunks of length $C$ (Algorithm~\ref{alg:wy_pipeline}, written unified across sharing ratio $R = H_v / H_k$ and gate scale $s_\beta \in \{1, 2\}$, where unpacked GDN/KDA set $R = 1, s_\beta = 1$). Op 1 projects $X$ to pre-convolution activations $\tilde{Q}, \tilde{K}, \tilde{V}$, gate logits $(\tilde{\beta}_t, \tilde{f}_t)_{t=1}^T$, and output-gate logits $(z_t)_{t=1}^T$ ($z_t \in \R^{R d_v}$ per key head), followed by causal short convolutions, $L_2$ normalization into $Q, K$, and gate activations (Op 2). Within chunk $c \in \{1, \dots, T/C\}$ starting from boundary state $S_{c-1} \in \R^{d_k \times R d_v}$ ($S_0 = 0$), let local token indices satisfy $1 \le j \le t \le C$, let $Q_c, K_c \in \R^{C \times d_k}$ and $V_c, O_c \in \R^{C \times R d_v}$ stack the chunk's queries, keys, values, and readouts as rows, let $\beta_c = (\beta_1, \dots, \beta_C)^\top \in \R^C$ stack its write gates, and let $I_C$ denote the $C \times C$ identity. Define
\[
P_{t \leftarrow j} = \prod_{s=j+1}^t F_s \quad (P_{t \leftarrow t} = I_{d_k}), \qquad F_c = P_{C \leftarrow 0} = \prod_{s=1}^C F_s
\]
as the cumulative diagonal decay from local token $j$ to $t$ and the total chunk decay, and let $\bar{K}_c, \bar{Q}_c, \hat{K}_c \in \R^{C \times d_k}$ denote the decayed key and query matrices whose $t$-th rows are $(P_{t \leftarrow 0} k_t)^\top$, $(P_{t \leftarrow 0} q_t)^\top$, and $(P_{C \leftarrow t} k_t)^\top$. Within a chunk, each token's write depends on what earlier tokens in the same chunk have already written, so the $C$ writes cannot be computed independently. Collect them as rows of $E_c \in \R^{C \times R d_v}$. Each row is what the token would write on its own, $\diag(\beta_c)(V_c - \bar{K}_c S_{c-1})$, minus the earlier tokens' writes, weighted by how strongly their decayed keys overlap with its own key. Using indicator $\mathbf{1}[\cdot] \in \{0, 1\}$, these overlaps define the strictly lower-triangular Gram matrix $M_c \in \R^{C \times C}$ (Op 3a, alongside lower-triangular intra-chunk query-key score matrix $L_c \in \R^{C \times C}$, Op 3b), so the chunk writes satisfy (derived in Section~\ref{app:proofs}):
\[
E_c = \diag(\beta_c)\,(I_C + M_c)^{-1}(V_c - \bar{K}_c S_{c-1}).
\]
Computing $A_c = (I_C + M_c)^{-1}$ (Op 4) and forming the WY factors $U_c, W_c$ (Op 5) runs in parallel across chunks, leaving a sequential scan over the $T/C$ chunk boundaries for $S_c$ and optional velocity buffer $Z_c \in \R^{d_k \times R d_v}$ ($Z_0 = 0$, momentum $\mu \ge 0$; Op 6), parallel chunk readouts $O_c$ (rows $o_t^\top$; Op 7a), and gated root-mean-square normalization ($\RMSNorm$, with elementwise product $\odot$) followed by the output projection to $Y \in \R^{T \times D}$ (Op 7b--c).

\begin{algorithm}[tb]
\caption{Chunkwise WY Parallel Forward Pass}
\label{alg:wy_pipeline}
\small
\begin{algorithmic}[1]
\REQUIRE Sequence $X \in \R^{T \times D}$, chunk size $C$, head ratio $R$, gate scale $s_\beta$, momentum $\mu$ ($(R, s_\beta, \mu) = (1, 1, 0)$ for GDN/KDA; $(R, 2, 0)$ recommended default with $R \in \{2, 4\}$ for SLA, or optional Schur-guarded $\mu \in (0, 0.25]$)
\ENSURE Output $Y \in \R^{T \times D}$, final state $S_{T/C} \in \R^{d_k \times R d_v}$
\STATE $\tilde{Q}, \tilde{K}, \tilde{V}, (\tilde{\beta}_t, \tilde{f}_t, z_t)_{t=1}^T \gets \mathrm{LinearProj}(X)$ \hfill $\triangleright$ \text{Op 1 (MXU)}
\STATE $Q, K \gets \mathrm{L2Norm}(\mathrm{Conv1D}(\tilde{Q}, \tilde{K}))$ \hfill $\triangleright$ \text{Op 2a (VPU)}
\STATE $V \gets \mathrm{SiLU}(\mathrm{Conv1D}(\tilde{V}))$
\STATE $\beta_t \gets s_\beta \sigma(\tilde{\beta}_t); \;\; F_t \gets \diag(e^{-\mathrm{softplus}(\tilde{f}_t)})$ \hfill $\triangleright$ \text{Op 2b (VPU)}
\STATE Partition $Q, K, V, \beta, F$ into $T/C$ chunks; $S_0 \gets 0,\; Z_0 \gets 0 \in \R^{d_k \times R d_v}$
\FORALL{$c \in \{1, \dots, T/C\}$ \textbf{in parallel}}
    \STATE $P_{t \leftarrow j} \gets \prod_{s=j+1}^t F_s \;\;\forall\, 0 \le j \le t \le C$ \;\;($P_{t \leftarrow t} = I_{d_k}$)
    \STATE $(\bar{K}_c, \bar{Q}_c, \hat{K}_c)_t \gets P_{t \leftarrow 0} k_t, P_{t \leftarrow 0} q_t, P_{C \leftarrow t} k_t$ \hfill $\triangleright$ \text{each row $t$}
    \STATE $(M_c)_{t,j} \gets \mathbf{1}[t > j]\,\beta_j k_t^\top P_{t \leftarrow j} k_j$ \hfill $\triangleright$ \text{Op 3a: Gram}
    \STATE $(L_c)_{t,j} \gets \mathbf{1}[t \ge j]\, q_t^\top P_{t \leftarrow j} k_j$ \hfill $\triangleright$ \text{Op 3b: Score}
    \STATE $A_c \gets (I_C + M_c)^{-1} \in \R^{C \times C}$ \hfill $\triangleright$ \text{Op 4: Solve}
    \STATE $U_c \gets \diag(\beta_c) A_c V_c \in \R^{C \times R d_v}$ \hfill $\triangleright$ \text{Op 5a (MXU)}
    \STATE $W_c \gets \diag(\beta_c) A_c \bar{K}_c \in \R^{C \times d_k}$ \hfill $\triangleright$ \text{Op 5b (MXU)}
\ENDFOR
\FOR{$c = 1$ {\bfseries to} $T/C$}
    \STATE $E_c \gets U_c - W_c S_{c-1}$ \hfill $\triangleright$ \text{Op 6a (MXU)}
    \STATE $Z_c \gets \mu Z_{c-1} + \hat{K}_c^\top E_c$ \hfill $\triangleright$ \text{Op 6b (MXU)}
    \STATE $S_c \gets F_c S_{c-1} + Z_c$
\ENDFOR
\FORALL{$c \in \{1, \dots, T/C\}$ \textbf{in parallel}}
    \STATE $O_c \gets \bar{Q}_c S_{c-1} + L_c E_c$ \hfill $\triangleright$ \text{Op 7a (MXU)}
\ENDFOR
\STATE $Y \gets \mathrm{OutProj}\big((\sigma(z_t) \odot \RMSNorm(o_t))_{t=1}^T\big)$ \hfill $\triangleright$ \text{Op 7b--c}
\STATE \textbf{return} $Y, S_{T/C}$
\end{algorithmic}
\end{algorithm}

\subsection{Three Hardware Bottlenecks on TPUs}
\label{sec:laws}
\label{subsec:scan_law}

\begin{figure}[t]
\centering
\includegraphics[width=0.72\textwidth]{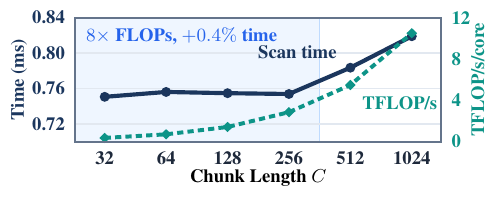}
\caption{\textbf{Sequential state-scan cost on TPU v7x.} A $64$-step scan of $(C, 256) \times (256, 256)$ matmuls takes nearly constant time for $C \in \{32, \dots, 256\}$, so a fixed per-step overhead, not matmul FLOPs, dominates the scan.}
\label{fig:laws}
\end{figure}

\begin{table}[t]
\caption{\textbf{Documented hardware specifications per chip} for NVIDIA H100 SXM and four Google TPU generations~\cite{Jouppi2023TPUv4,GoogleCloud2025TPU7x,Austin2025ScalingBook}. MXU:VPU is the ratio of peak BF16 matrix throughput to FP32 vector throughput; Ridge is the BF16 arithmetic-intensity ridge point (FLOPs/byte). $^\ast$8-bit peak is FP8 on H100 and TPU v7x and INT8 on TPU v4--v6e.}
\label{tab:hw_specs}
\centering
\footnotesize
\resizebox{\textwidth}{!}{%
\begin{tabular}{@{}lccccccccc@{}}
\toprule
Accelerator & Cores & Array Tile & BF16 TF/s & 8-bit TF/s$^\ast$ & VPU TF/s & MXU:VPU & HBM (GB) & HBM GB/s & Ridge \\
\midrule
H100 SXM & 132 SM & $16 \times 16$ & 989 & 1,979 & 67.0 & $14.8\times$ & 80 & 3,350 & 295 \\
TPU v4 & 2 & $128 \times 128$ & 275 & 275 & 3.2 & $85.9\times$ & 32 & 1,200 & 229 \\
TPU v5p & 2 & $128 \times 128$ & 459 & 918 & 5.4 & $85.0\times$ & 95 & 2,765 & 166 \\
TPU v6e & 1 & $256 \times 256$ & 918 & 1,836 & 7.2 & $127.5\times$ & 32 & 1,600 & 574 \\
TPU v7x & 2 & $256 \times 256$ & 2,307 & 4,614 & 18.0 & $128.2\times$ & 192 & 7,373 & 313 \\
\bottomrule
\end{tabular}%
}
\end{table}

TPUs pair a wide systolic MXU with a much slower VPU. On TPU v7x the MXU tile is $256 \times 256$ and the MXU:VPU throughput ratio is about $128\times$, compared with about $15\times$ on H100 (Table~\ref{tab:hw_specs}). This skew creates three bottlenecks for Algorithm~\ref{alg:wy_pipeline}:

\paragraph{Bottleneck 1: Narrow states and a fixed per-step scan cost.}
Op 6 carries the state across $T / C$ chunk boundaries sequentially. On TPU v7x, each scan step costs a nearly fixed ${\approx}12\,\mu\text{s}$ regardless of chunk size: $8\times$ more FLOPs per step adds under $1\%$ time (Figure~\ref{fig:laws}). Scan cost is thus set by the number of steps and heads, not by matmul size. A $256$-wide tile also processes a $d_k = 128$ state at the same cost as a $d_k = 256$ state, so $128$-wide states leave half of each tile idle while paying the full per-step cost for every head.

\paragraph{Bottleneck 2: Per-channel decay.}
KDA~\cite{Moonshot2025KimiLinear} and GDN-2~\cite{NVIDIA2026GDN2} give every key channel its own forget gate ($F_t = \diag(f_t)$). Let $\Lambda_{t,i} = \sum_{s=1}^t \log f_{s,i}$ be the cumulative decay of channel $i$ within a chunk. Each Gram entry then weights channels differently:
\[
(M_c)_{t,j} = \mathbf{1}[t > j]\,\beta_j \sum_{i} e^{\Lambda_{t,i} - \Lambda_{j,i}} k_{t,i} k_{j,i}.
\]
To use the MXU, kernels fold $e^{\Lambda_{t,i}}$ and $e^{-\Lambda_{j,i}}$ into the keys, but $e^{-\Lambda_{j,i}}$ overflows FP32 for long chunks. KDA kernels therefore split each chunk into $16$-token sub-blocks and re-center the keys on the VPU before every sub-block matmul, which stalls the kernel. In our KDA runs, about $1\%$ of channels also learn near-unit forget gates, which hurts length extrapolation unless the gates are clamped (Section~\ref{subsec:dolma_15b}). GDN's scalar gate ($G = 1$) avoids both issues, at the cost of a single memory horizon per head.

\paragraph{Bottleneck 3: Per-value-head chunk solves and sequential fallbacks.}
Computing $A_c = (I_C + M_c)^{-1}$ (Op 4) by sequential substitution runs on vector lanes in GPU kernels (\texttt{fla}~\cite{FLA2024Repo}), MaxText's XLA \texttt{solve\_triangular} path~\cite{MaxText2025Repo}, and open-source Tokamax's FP32 fallback~\cite{Tokamax2025Repo}. Even when open-source TPU kernels move individual solves onto the MXU (Tokamax's BF16 Neumann--Horner path, validated for per-channel KDA at $C = 64$~\cite{Tokamax2025Repo}, or MaxText's XLA block-doubling path for GDN~\cite{MaxText2025Repo}), per-value-head write gates still require $H_v$ separate $C \times C$ solves per chunk and unfused HBM round-trips. Repeated across every chunk and all $H_v = 8$ value heads, Ops 3--5 account for about $45\%$ of core intra-chunk and scan time in our stateless functional KDA audit (Table~\ref{tab:7op_audit}), keeping the practical chunk size near $64$--$128$.

\subsection{SLA: Addressing the Three Bottlenecks}
\label{subsec:recurrence}

Guided by the three systolic bottlenecks above, we now formulate SLA, which reshapes the state geometry, decay structure, and chunk inversion around the $256 \times 256$ systolic tile:

\begin{definition}[SLA Recurrence]
\label{def:sfa}
Consider a linear attention layer with $H_v$ value heads of width $d_v$ and $H_k = H_v / R$ query-key heads of width $d_k = 256$ indexed by $h_k \in \{1, \dots, H_k\}$, where each query-key head is shared across $R \in \{2, 4\}$ value heads. At token $t$, each query-key head computes:
\begin{itemize}[leftmargin=1.6em, itemsep=3pt, topsep=4pt]
\item \textbf{Shared query, key, and write gate:} $L_2$-normalized $q_t, k_t \in \R^{256}$ with $\norm{q_t}_2 = \norm{k_t}_2 = 1$, and a scalar step size $\beta_t = 2\sigma(\tilde{\beta}_t) \in (0, 2)$ ($s_\beta = 2$) that doubles the $(0, 1)$ range of standard GDN/KDA;
\item \textbf{Packed values and output gates:} $R$ value vectors $v_{t,1}, \dots, v_{t,R} \in \R^{d_v}$ stacked as $v_t = [v_{t,1}; \dots; v_{t,R}] \in \R^{D_{v,\text{pack}}}$ with packed value width $D_{v,\text{pack}} = R d_v$, and output-gate vector $z_t = [z_{t,1}; \dots; z_{t,R}] \in \R^{D_{v,\text{pack}}}$ with slices $z_{t,m} \in \R^{d_v}$ ($m \in \{1, \dots, R\}$);
\item \textbf{Four-tile forget gate:} a block-diagonal gate $F_t \in (0, 1]^{256 \times 256}$ that splits the $256$ key channels into $G = 4$ contiguous tiles of $b = 64$ channels, with forget-gate logit $\tilde{f}_t = (\tilde{f}_{t,1}, \dots, \tilde{f}_{t,4}) \in \R^4$, all-ones vector $\mathbf{1}_{64} \in \R^{64}$, and one scalar decay $\gamma_{t,g} = \exp(-\operatorname{softplus}(\tilde{f}_{t,g})) \in (0, 1]$ on each tile $g \in \{1, 2, 3, 4\}$:
\begin{equation}
F_t = \diag(\gamma_{t,1}\mathbf{1}_{64},\, \gamma_{t,2}\mathbf{1}_{64},\, \gamma_{t,3}\mathbf{1}_{64},\, \gamma_{t,4}\mathbf{1}_{64}).
\label{eq:sla_block_decay}
\end{equation}
\end{itemize}
\textbf{(i) Intra-chunk token recurrence.} Maintaining a packed state $S_t = [S_{t,1}, \dots, S_{t,R}] \in \R^{256 \times R d_v}$ initialized at $S_0 = 0$ and restarted at each chunk from the boundary state $S_{c-1}$ of (ii), with slice $S_{t,m} \in \R^{256 \times d_v}$ and output slice $o_{t,m} \in \R^{d_v}$ ($o_t = [o_{t,1}; \dots; o_{t,R}]$) for value head $m \in \{1, \dots, R\}$, SLA decays the state, writes the prediction error, and normalizes each value head:
\contribbox{SLA Token-Level Recurrence ($d_k = 256,\; G = 4,\; b = 64$)}{%
\begin{align}
S_t &= F_t S_{t-1} + \beta_t k_t \big(v_t - (F_t S_{t-1})^\top k_t\big)^\top, \label{eq:sfa_intra}\\
o_{t,m} &= \sigma(z_{t,m}) \odot \RMSNorm\!\big(S_{t,m}^\top q_t\big).
\label{eq:sfa_readout}
\end{align}%
}
Here $(F_t S_{t-1})^\top k_t$ is the value currently stored for $k_t$, so Eq.~\eqref{eq:sfa_intra} writes only the prediction error, and one rank-one write updates all $R$ value heads.

\textbf{(ii) Optional chunk-boundary momentum.} Let $\tilde{S}_c$ be the state after unrolling Eq.~\eqref{eq:sfa_intra} over the $C$ tokens of chunk $c$ from $S_{c-1}$, let $F_c = \prod_{t \in c} F_t$, and let the reset indicator $\kappa_c \in \{0, 1\}$ equal $0$ if chunk $c$ contains a document boundary (which also zeros cross-document carry $F_c$) and $1$ otherwise. When enabled, an optional velocity buffer $Z_c \in \R^{256 \times D_{v,\text{pack}}}$ ($Z_0 = 0$) smooths boundary updates via
\begin{equation}
Z_c = \mu_c \kappa_c Z_{c-1} + (\tilde{S}_c - F_c S_{c-1}), \qquad S_c = F_c S_{c-1} + Z_c,
\label{eq:sfa_mom}
\end{equation}
with $\mu_c = 0$ (recovering $S_c = \tilde{S}_c$) as the recommended default. When momentum is enabled ($\mu \in (0, 0.25]$), the per-chunk Schur guard $\mu_c = \min\!\big(\mu, (1 - \varphi_c)/(1 + \varphi_c)\big)$ with $\varphi_c = \prod_{t \in c} \max_g \gamma_{t,g}$ enforces unconditional chunk-to-chunk stability. (Our earlier $12.5\text{B}$ and $63.1\text{B}$ MoE runs in Section~\ref{sec:eval} were trained with fixed $\mu = 0.25$, which stayed well inside the stable region; controlled ablations below motivate demoting momentum from the default.)
\end{definition}

\paragraph{Notation and operator taxonomy.}
\label{app:notation}
Unless redefined locally, symbols follow Algorithm~\ref{alg:wy_pipeline} and Definition~\ref{def:sfa}; Appendix Table~\ref{tab:app_notation} collects them for quick reference (symbols used only inside one section, such as the roofline constants in Section~\ref{app:rooflines}, are defined where they first appear). Table~\ref{tab:app_gdn_kda_sfa} contrasts the algebraic operators and systolic hardware mappings of GDN~\cite{Yang2025GDN}, KDA~\cite{Moonshot2025KimiLinear}, GDN-2~\cite{NVIDIA2026GDN2}, \textbf{SLA} (\texttt{SLA-R2} / \texttt{SLA-R4}), and \texttt{SPD-SLA-W}, an optional extension of SLA that preconditions chunks with $\Sigma_c$ to form whitened keys $\tilde{K}_c$ (rows $\tilde{k}_t^\top$) under step-size margin $\varepsilon_\beta > 0$ (Appendix~\ref{sec:whitening}); \texttt{SPD-SLA-W} is not used in any core SLA result except the middle bar of Figure~\ref{fig:pretrain_xprof}a and the conditioning analysis of Appendix~\ref{sec:whitening}.

\begin{table}[ht]
\caption{Side-by-side algebraic and systolic hardware comparison across prior GPU gated-delta recurrences and our SLA family.}
\label{tab:app_gdn_kda_sfa}
\centering
\scriptsize
\renewcommand{\arraystretch}{1.15}
\setlength{\tabcolsep}{2.6pt}
\begin{tabularx}{\textwidth}{@{}>{\raggedright\arraybackslash}p{0.125\textwidth} >{\raggedright\arraybackslash}p{0.145\textwidth} >{\raggedright\arraybackslash}p{0.155\textwidth} >{\raggedright\arraybackslash}p{0.140\textwidth} >{\raggedright\arraybackslash}X >{\raggedright\arraybackslash}p{0.175\textwidth}@{}}
\toprule
Design Axis & \textbf{GDN}~\cite{Yang2025GDN} & \textbf{KDA}~\cite{Moonshot2025KimiLinear} & \textbf{GDN-2}~\cite{NVIDIA2026GDN2} & \textbf{SLA (\texttt{SLA-R2} / \texttt{SLA-R4})} (Ours) & \textbf{\texttt{SPD-SLA-W}} (Appendix~\ref{sec:whitening}) \\
\midrule
Key/Value Heads & Unpacked $H_k = H_v$ & Unpacked $H_k = H_v$ & Unpacked $H_k = H_v$ & Packed GVA $H_k = H_v / R$ ($R\!\in\!\{2,4\}$) & Packed GVA $H_k = H_v / R$ \\
State Tile \& Dtype & $S^{(h_k)} \in \R^{128 \times d_v}$ FP32 ($50\%$ pad) & $S^{(h_k)} \in \R^{128 \times d_v}$ FP32 ($50\%$ pad) & $S^{(h_k)} \in \R^{d_k \times d_v}$ FP32 & $S \in \R^{256 \times D_{v,\text{pack}}}$ BF16 / Row-FP8 (no padding) & $S \in \R^{256 \times D_{v,\text{pack}}}$ BF16 / Row-FP8 \\
Forget Gate $F_t$ & Scalar $\gamma_t I_{d_k}$ ($G=1$) & Channel $\diag(f_t)$ ($G=d_k$) & Decoupled ch.\ ($G=d_k$) & Tile-grouped $\bigoplus_{g=1}^{4} \gamma_{t,g} I_{64}$ ($G = 4$) & Tile-grouped $\bigoplus_{g=1}^4 \gamma_{t,g} I_{64}$ ($G=4$) \\
Step Size $\beta_t$ & $(0, 1)$ per $H_v$ & $(0, 1)$ per $H_v$ & Ch.\ vector per $H_v$ & $(0, 2)$ scalar per $H_k$ (neg.\ eig.) & Clamped $(0, (2-\varepsilon_\beta)/k_t^\top\tilde{k}_t]$ \\
Chunk Gram $M_c$ & $1\times$ MXU + scalar mask & $16$-row VPU pivots ($|\Lambda_{j,i}|\!>\!88.7$) & Decoupled ch.\ broadcast & $G\times [C,b]\!\times\![b,C]$ pure MXU ($b\!\ge\!64$, $0$ ovf.) & Oblique $k_t^\top P_{t\leftarrow j}\tilde{k}_j$ MXU \\
Chunk Inverse $A_c$ & $H_v\times$ solves (seq.\ \texttt{fla}; XLA block-doub.\ MaxText; BF16 MXU Tokamax) & $H_v\times$ solves (MXU block-doub.\ Pallas; BF16 Neumann $C{=}64$ Tokamax) & $H_v\times$ solves (seq.\ in-block) & $H_k = H_v / R$ solves in VMEM: Exact FP32 ($C{=}128$) or $1$-Step Neumann (App.~\ref{app:v2_solver_audit}) & $H_k\times$ MXU inverse + \texttt{Polar-2+} ($5$ MM) \\
Momentum / 2nd-Ord. & None ($\mu=0$, Euclidean) & None ($\mu=0$, Euclidean) & None ($\mu=0$, Euclidean) & Optional ($\mu=0$ default; Schur-guarded $Z_c = \mu_c\kappa_c Z_{c-1} + \Delta S_c$) & Whitened keys $\tilde{K}_c$ (Eq.~\eqref{eq:oblique_key}) \\
Output Norm & Head RMSNorm & Head RMSNorm & Head RMSNorm & Per-value-head $\RMSNorm$ & Per-value-head $\RMSNorm$ \\
VMEM / HBM Trips & $6$ HBM trips (unfused) & $6$ HBM trips (unfused wrapper) & $6$ HBM trips ($6.8$--$14.9$ GiB) & Fused SLA: $1$ kernel in $64\text{ MiB}$ VMEM + custom VJP & Fits $64\text{ MiB}$ VMEM \\
Context Parallelism & Prefix tree, one $d_k \times d_k$ map per head & Prefix tree, one $d_k \times d_k$ map per head & Prefix tree, one $d_k \times d_k$ map per head & Prefix tree, one $[256,256]$ map per key head ($2\times$ with momentum) & Same as SLA \\
\bottomrule
\end{tabularx}
\end{table}

\paragraph{Why chunk-level momentum, implementation, and caveats.}
\label{app:momentum}
Unlike token-level momentum~\cite{Behrouz2025Titans}, applying heavy-ball momentum~\cite{Polyak1964} once per chunk boundary (Op 6b of Algorithm~\ref{alg:wy_pipeline}) leaves the intra-chunk WY solve unchanged. Specifically, in Op 6b, the net chunk state increment in Eq.~\eqref{eq:sfa_mom} is
\[
\Delta S_c = \tilde{S}_c - F_c S_{c-1} = \hat{K}_c^\top E_c.
\]
The velocity buffer $Z_c \in \R^{d_k \times D_{v,\text{pack}}}$ has the same shape as $S_c$ and stays in VMEM inside the fused kernel, adding zero extra HBM state traffic. On packed multi-document streams, intra-chunk Gram and score matrices apply an exact segment mask $\mathbf{1}[\mathrm{seg}_t = \mathrm{seg}_j]$, only the final document segment writes to $\Delta S_c$, and any document reset anywhere inside chunk $c$ zeros both incoming state carry $F_c$ and $\kappa_c = 0$, preventing mid-chunk velocity leakage. For a fixed chunk with a scalar gate ($G = 1$, chunk decay $\varphi \in (0, 1)$) and a real eigenvalue $\vartheta \in [-\varphi, \varphi]$ of the linear map from $S_{c-1}$ to $\tilde{S}_c$, the lifted block $\bigl[\begin{smallmatrix} \vartheta & \mu \\ \vartheta - \varphi & \mu \end{smallmatrix}\bigr]$ has characteristic polynomial $p(\lambda) = \lambda^2 - (\vartheta + \mu)\lambda + \mu\varphi$ and meets Jury's Schur stability conditions ($\mu\varphi < 1$, $p(1) \ge (1 - \varphi)(1 - \mu) > 0$, $p(-1) > 0$) for all $\mu \in [0, 1)$. This argument does not cover $G = 4$, complex eigenvalues, or chunks that change over time, and we can construct unstable cases there at $\mu = 0.25$. Mapping the lifted eigenvalue spectrum on fixed repeated $G = 1$ chunks at $\mu = 0.25$ shows that complex-eigenvalue growth occurs only when high chunk retention ($\varphi > 0.773$, $\tau \gtrsim 500$ tokens at $C = 128$) coincides with near-reflection writes ($\beta > 1.73$ on a two-write chunk; random dense chunks are $0\%$ unstable for $\bar{\beta} \le 1.4$). A general sufficient stability condition is the per-chunk guard $\mu_c \le (1 - \varphi_c)/(1 + \varphi_c)$ with $\varphi_c = \prod_{t \in c} \max_g \gamma_{t,g}$ (which at fixed $\mu = 0.25$ covers $\varphi_c < 0.6$). Our trained $12.5\text{B}$ \texttt{SLA-R4} models operate well inside the stable interior with mean write gates $\bar{\beta} \in [0.251, 0.348]$, and $24$-layer state norms contract cleanly from $8\text{K}$ to $32\text{K}$ in the $63.1\text{B}$ model (Appendix~\ref{app:causal_clamp}). With \texttt{Exact Inv.} and the same chunk boundaries, autoregressive decoding reproduces training up to rounding by updating $Z_c$ once every $C$ tokens. Under context parallelism, momentum couples $(S_c, Z_c)$ into a $2 d_k \times 2 d_k$ affine map (Eq.~\eqref{eq:app_coupled_monoid}); Table~\ref{tab:cp_serving} therefore reports both the uncoupled ($\mu = 0$, $256^2$) and coupled ($\mu = 0.25$, $512^2$) tree scans. Because momentum introduces chunk-size dependence and doubles the context-parallel transition, we evaluate both $\mu = 0$ and guarded $\mu_c$ controls directly in Section~\ref{sec:eval}.

Appendix Table~\ref{tab:app_momentum} compares $\mu = 0$ and $\mu = 0.25$ on a $12.5\text{B}$-total ($0.79\text{B}$-active) MoE trained on $15.1\text{B}$ Dolma~3.5 tokens ($7{,}200$ steps, $2.1\text{M}$ tokens/step, $128$ TPU v4 chips, seed $42$). With $4{:}1$ key sharing ($R = 4$), momentum lowers loss by $0.019$--$0.023$ nats (though that $\mu = 0.25$ run used an $8{,}200$-step cosine schedule evaluated at step $7{,}199$); with $2{:}1$ sharing ($R = 2$), it changes loss by at most $+0.004$ nats, and an input-dependent $\mu$ degrades loss by $+0.026$ nats. To isolate momentum without schedule or fusion confounders, we ran a $49$-point Schur stability sweep ($\varphi \in [0.4, 0.99], \beta \in [0.3, 1.99]$) and an isolated $3$-seed dense LM ablation ($274.7\text{M}$ params, $104.9\text{M}$ FineWeb-Edu tokens, identical \texttt{SLA-R4} kernel; Appendix~\ref{app:r87_controls}, Table~\ref{tab:app_g1_g4_mom}). On the $49$-point grid, fixed $\mu = 0.25$ exhibits $5$ unstable $(\varphi, \beta)$ pairs with spectral radius up to $1.090$ ($224.7\times$ $64$-chunk norm growth), whereas the per-chunk guard $\mu_c = \min(\mu, (1 - \varphi_c)/(1 + \varphi_c))$ eliminates every unstable pair ($5 \to 0$). In the $3$-seed LM ablation, trained gates stay deep inside the stable interior ($\bar{\beta} \in [0.76, 0.79]$, $p_{95}(\beta) \le 1.20$, $\bar{\varphi}_c \le 0.057$, $p_{95}(\varphi_c) \le 0.26$, guard activation $0\%$), and $\mu = 0$ yields the lowest mean validation perplexity ($46.49$ vs.\ $46.57$ fixed and $46.53$ guarded; paired $+0.09 \pm 0.36$, $t = +0.41$) and $16\text{K}$ perplexity ($45.90$ vs.\ $46.27$ and $46.19$). Because chunk momentum adds chunk-size dependence, doubles context-parallel transition size to $512 \times 512$ (Section~\ref{subsec:sfa_cp}), and requires the $64\text{ MiB}$ VMEM cap at $D = 1792$ without improving isolated perplexity, we demote momentum from SLA's default ($\mu = 0$ recommended; Schur-guarded $\mu_c$ optional) and note transparently where our earlier $12.5\text{B}/63.1\text{B}$ MoE checkpoints used fixed $\mu = 0.25$.

\subsection{The Three Algorithmic Pillars of SLA}
\label{subsec:pillars}
\label{subsec:pillar1_gva}
\label{subsec:pillar2_decay}
\label{subsec:pillar3_ben}

\begin{figure}[ht]
\centering
\includegraphics[width=0.98\textwidth]{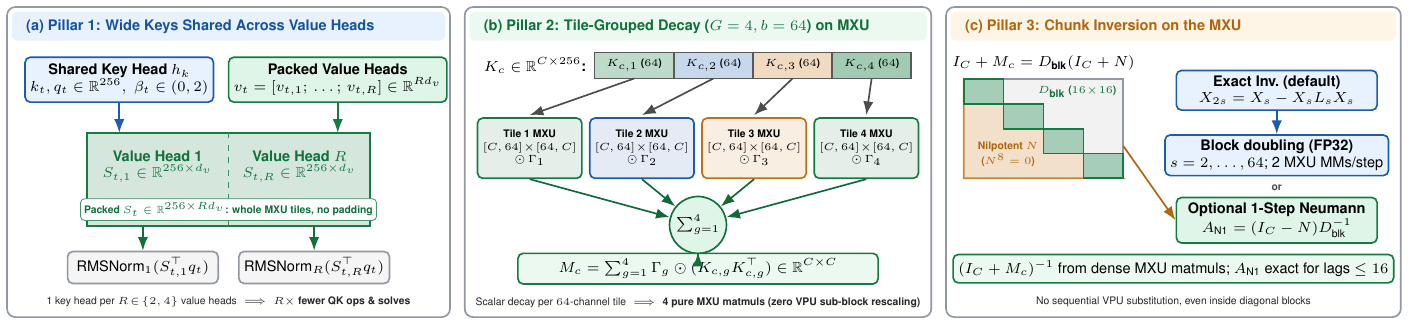}
\caption{\textbf{Detailed architectural diagram of the three mathematical pillars of SLA.} \textbf{(a) Pillar 1:} Widening the key dimension to $d_k = 256$ and sharing each key head $h_k$ and write gate $\beta_t \in (0, 2)$ across $R \in \{2, 4\}$ value heads forms a dense $256 \times D_{v,\text{pack}}$ state that fills whole $256 \times 256$ MXU tiles without padding while cutting QK projections, short convolutions, and chunk solves by $R\times$. \textbf{(b) Pillar 2:} Partitioning the $d_k = 256$ key channels into $G = 4$ contiguous tiles of $b = 64$ channels decomposes the intra-chunk coupling matrix $M_c$ into four $64$-wide MXU matrix multiplications weighted by scalar decay masks $\Gamma_g$, eliminating VPU sub-block rescaling. \textbf{(c) Pillar 3:} Because $\beta_t$ is shared across $R$ value heads, SLA runs $R\times$ fewer chunk solves inside VMEM: default FP32 recursive block-doubling (\texttt{Exact Inv.}, building on open-source MXU solvers~\cite{MaxText2025Repo,Tokamax2025Repo}) with custom-VJP reuse, or the optional $1$-step Neumann inverse ($I_C + M_c = D_{\text{blk}}(I_C + N)$ with $16 \times 16$ diagonal blocks, $N^8 = 0$ at $C = 128$; Section~\ref{prop:ben}).}
\label{fig:sla_pillars}
\label{fig:arch_overview}
\end{figure}

Each pillar targets one of the bottlenecks in Section~\ref{subsec:scan_law} (Figure~\ref{fig:sla_overview}c--e; detailed diagrams in Figure~\ref{fig:sla_pillars}).

\subsubsection{Pillar 1: Wide Keys with Grouped Value Heads}
\label{subsec:gva_invariance}

A key dimension below $256$ still occupies a full $256$-wide tile on TPU v6e/v7x, so SLA widens keys to $d_k = 256$, which costs no extra cycles in the inter-chunk scan. Prior linear grouped-value layers (such as Qwen3-Next~\cite{Qwen2025Qwen3Next} and \texttt{fla}~\cite{FLA2024Repo}) share query-key heads across value heads at $d_k = 128$, yet keep independent decay and write gates $(g, \beta_t)$ per value head, which still forces $H_v$ separate Gram matrices and $H_v$ chunk solves per chunk. SLA instead shares both the $256$-wide query-key head and the scalar write gate $\beta_t \in (0, 2)$ across $R = H_v / H_k \in \{2, 4\}$ packed value heads (together with per-value-head gated normalization~\cite{Qwen2025GatedAttn}, Eq.~\eqref{eq:sfa_readout}). Packing these $R$ value heads side by side forms a contiguous $S_t \in \R^{256 \times R d_v}$ state that fills whole MXU tiles without padding, while cutting QK projections, short convolutions, $M_c$, $L_c$, and the chunk solve $A_c$ by $R\times$. On the identical exact FP32 $C = 128$ kernel ($D = 2048$), packing $R = 4$ value heads per key head runs $2.92\times$ faster forward ($2.73\times$ F+B) than $R = 1$; holding head geometry fixed at $H_k = 2, d_k = 256, H_v = 8$ and isolating only the shared write gate $\beta_t$ ($2$ solves vs.\ $8$ solves per chunk) yields $2.45\times$--$3.28\times$ faster core forward, $1.87\times$--$2.48\times$ faster full-layer F+B, and $1.68\times$ faster small-LM pre-training (Appendix~\ref{app:v2_solver_audit}). Head sharing itself is prior work; our contribution is widening $d_k$ to the systolic tile and sharing $\beta_t$ across the $R$ packed value heads so one chunk solve serves all $R$ heads.

\paragraph{Derivation of GVA systolic invariance.}
Fix total value width $D_v = H_v d_v$ and set $H_k = H_v / R$ with $D_{v,\text{pack}} = R d_v$. Each key head $h_k \in \{1, \dots, H_k\}$ maintains state $S^{(h_k)} \in \R^{d_k \times D_{v,\text{pack}}}$, so total state elements are invariant to $R$:
\[
H_k d_k (R d_v) = H_v d_k d_v.
\]
Inter-chunk state scan FLOPs per token are likewise invariant to $R$:
\[
H_k \big(2 d_k D_{v,\text{pack}} \text{ [erase]} + 2 d_k D_{v,\text{pack}} \text{ [write]} + 2 d_k D_{v,\text{pack}} \text{ [readout]}\big) = 6 H_v d_k d_v.
\]
Because $\beta_t$ is scalar per key head $h_k$, QK projections ($2 D H_k d_k$ parameters), QK causal \texttt{Conv1D} (for kernel width $K_{\text{conv}} = 4$, $2 K_{\text{conv}} H_k d_k$ weights), chunk Gram matmuls ($4 C H_k d_k$ FLOPs/token), and $[C, C]$ WY/Neumann solves ($H_k$ systems per chunk) scale linearly with $H_k = H_v / R$ and shrink by $R\times$. Reshaping $\tilde{S}_t^\top q_t \in \R^{R d_v}$ into $R$ slices of width $d_v$ and applying independent $\RMSNorm$ to all $H_v = H_k R$ value heads prior to the sigmoid output gate $\sigma(z_t) \in (0, 1)^{R d_v}$ (slices $\sigma(z_{t,m})$ in Eq.~\eqref{eq:sfa_readout}) retains $H_v$ independent head-scale degrees of freedom while keeping $[256, D_{v,\text{pack}}]$ MXU tiles contiguous. \qed

\subsubsection{Pillar 2: Tile-Grouped Forget Gate and Packed WY Equivalence}

Bottleneck 2 arises because the decay cannot be pulled out of key--key inner products when every channel has its own rate. SLA splits each query and key into the same $G = 4$ groups of $b = 64$ channels as the forget gate in Eq.~\eqref{eq:sla_block_decay}, $q_t = [q_{t,1}; \dots; q_{t,4}]$ and $k_t = [k_{t,1}; \dots; k_{t,4}]$. Between tokens $j$ and $t$, group $g$ decays by a single number $\prod_{s=j+1}^{t} \gamma_{s,g} \le 1$, which moves outside the group's inner product. Stacking group-$g$ queries and keys as $Q_{c,g}, K_{c,g} \in \R^{C \times 64}$ and defining causal masks $(\Psi_g)_{t,j} = \mathbf{1}[t \ge j]\prod_{s=j+1}^{t} \gamma_{s,g}$ and $(\Gamma_g)_{t,j} = \mathbf{1}[t > j]\,\beta_j (\Psi_g)_{t,j}$, both intra-chunk matrices factor across tiles as
\[
M_c = \sum_{g=1}^{4} \Gamma_g \odot \big(K_{c,g} K_{c,g}^\top\big), \qquad L_c = \sum_{g=1}^{4} \Psi_g \odot \big(Q_{c,g} K_{c,g}^\top\big),
\]
where each term is a $[C, 64] \times [64, C]$ MXU matmul followed by an elementwise scaling by decay factors in $[0, 1]$, computed only for $t \ge j$, so there is no overflow and no need for VPU sub-blocks. This places $G = 4$ at the multi-timescale quality and single-kernel VMEM speed Pareto knee: scalar decay ($G = 1$, as in GDN) runs at nearly the same single-kernel VMEM speed as $G = 4$ ($2.7\%$--$4.0\%$ difference; Table~\ref{tab:app_3way_attrib}) because either $1$ or $4$ gate vectors per head reside entirely in VMEM vector registers without materializing $[B, H_k, T/C, C, C, d_k]$ pair-decay tensors in HBM, but $G = 1$ limits each head to a single retention timescale. On $12.5\text{B}$ MoE ($n = 6$ paired seeds; Table~\ref{tab:app_gcurve}), $G = 4$ improves validation perplexity on all $6/6$ seeds ($\Delta\text{PPL} = -0.34, -0.06, -0.06, -0.66, -0.07, -0.21$; mean $-0.23 \pm 0.24$, two-sided $p = 0.060$, exact sign test $p = 0.031$, while $G = 1$ leads unclamped synthetic $32\text{K}$ $8$-Needle at $59.6\%$ vs.\ $35.4\%$); to separate $G = 4$ from $G = 1$ cleanly without MoE routing variance, our $6$-seed paired dense LM control ($274.7\text{M}$ params, $104.9\text{M}$ tokens, identical \texttt{SLA-R4} kernel; Appendix~\ref{app:r87_controls}, Table~\ref{tab:app_g1_g4_mom}) shows $G = 4$ beating $G = 1$ on $6/6$ seeds across every metric ($-1.15 \pm 0.49$ Val PPL, $t = -5.78$; $-1.75 \pm 0.61$ $16\text{K}$ tail PPL, $t = -7.01$), capturing $78.5\%$ of the $G = 1 \to G = 256$ gain while beating unclamped $G = 256$ at $16\text{K}$. Conversely, on a synthetic $2$-layer multi-timescale associative recall diagnostic, $G = 256$ solves the task reliably while $G \in \{1, 2, 4\}$ remain bimodal or underfit (Table~\ref{tab:app_g1_g4_mom}), so $G \in \{1, 4\}$ acts as a clean hardware-expressivity knob with $G = 4$ as the language-modeling default. Architecturally, unpacked GDN ($H_k = 8, G = 1$) gives $H_k G = 8$ decay timescales per layer, whereas $4{:}1$ head sharing ($H_k = 2$) with $G = 1$ leaves only $2$; setting $G = 4$ ($b = 64$) restores $H_k G = 8$ timescales per layer in \texttt{SLA-R4} ($16$ in \texttt{SLA-R2}) at near-scalar single-kernel VMEM speed.

Isolating this VMEM-resident decayed score stage at $d_k = 256$ on TPU v7x quantifies why $G = 4$ ($b = 64$) remains inexpensive while finer sub-tiles degrade sharply. Analytically, at TPU v7x's $128\times$ MXU:VPU throughput skew, forming $G$ causal decay masks costs $128 G C^2$ MXU-equivalent ops against $2 C^2 d_k$ Gram matmul FLOPs; these balance at $G^\star = 2 d_k / 128 = 4$ for $d_k = 256$ ($G^\star = 8$ at $d_k = 512$), whereas per-channel decay ($G = d_k$) spends $98.5\%$ of intra-chunk work on the VPU. Under a decaying-memory capacity model with log-uniform lag demand from $8\text{K}$ to $1\text{M}$ tokens, unequal tile widths gain under $0.1\%$ and $G > 4$ gains under $0.3\%$, so hardware balance rather than recall capacity pins $G = 4$. At $C = 256$, an undecayed $[256, 256] \times [256, 256]$ matmul takes $0.303\,\mu\text{s}$ per chunk ($1.00\times$), whereas masked group sums at $G = 1$ ($b = 256$), $G = 2$ ($b = 128$), $G = 4$ ($b = 64$), and $G = 16$ ($b = 16$) take $0.306\,\mu\text{s}$ ($1.01\times$), $0.324\,\mu\text{s}$ ($1.07\times$), $0.461\,\mu\text{s}$ ($1.52\times$), and $1.269\,\mu\text{s}$ ($4.19\times$), respectively ($0.197$, $0.201$, $0.209$, $0.310$, and $0.730\,\mu\text{s}$ at $C = 128$). Because score construction accounts for only $10\%$--$15\%$ of fused-layer FLOPs once $R \in \{2, 4\}$ value heads are packed into the downstream chunk-update and readout matmuls, the $1.52\times$ stage cost at $G = 4$ adds under $4\%$ to full-layer latency, whereas narrower sub-tiles ($G = 16$) incur steep narrow-reduction and $G C^2$ VPU masking overhead on the $256 \times 256$ MXU.

\paragraph{Proof of tile-grouped packed WY equivalence.}
\label{app:proofs}
Consider chunk $c \in \{1, \dots, T/C\}$ for a single key head $h_k \in \{1, \dots, H_k\}$ serving $R = H_v / H_k$ packed value heads of total width $D_{v,\text{pack}} = R d_v$, starting from incoming state $\tilde{S}_0 = S_{c-1} \in \R^{d_k \times D_{v,\text{pack}}}$. Each token $t \in \{1, \dots, C\}$ has $G = 4$ tile-grouped diagonal decay $F_t = \bigoplus_{g=1}^4 \gamma_{t,g} I_{64}$ (diagonal entries in $(0, 1]$), query and key $q_t, k_t \in \R^{d_k}$, packed value $v_t \in \R^{D_{v,\text{pack}}}$, and scalar step size $\beta_t \in (0, 2)$ shared across the $R$ packed value heads of key head $h_k$.

Let $P_{t \leftarrow j} = \prod_{s=j+1}^t F_s$ denote the cumulative diagonal transition from step $j$ to step $t$ ($P_{t \leftarrow t} = I_{d_k}$). Unrolling Eq.~\eqref{eq:sfa_intra} ($\bar{S}_t = F_t \tilde{S}_{t-1}$, $\hat{v}_t = \bar{S}_t^\top k_t$, $\tilde{S}_t = \bar{S}_t + \beta_t k_t (v_t - \hat{v}_t)^\top$) from $\tilde{S}_0 = S_{c-1}$ gives
\begin{equation}
\tilde{S}_t = P_{t \leftarrow 0} S_{c-1} + \sum_{j=1}^t \beta_j \big(P_{t \leftarrow j} k_j\big) u_j^\top,
\label{eq:app_unroll}
\end{equation}
where $u_t = v_t - \hat{v}_t \in \R^{D_{v,\text{pack}}}$ is the post-forget delta residual at local step $t$. Multiplying $\bar{S}_t = F_t \tilde{S}_{t-1}$ on the left by $k_t^\top$ substitutes $F_t P_{t-1 \leftarrow j} = P_{t \leftarrow j}$ into every prior term:
\begin{equation}
u_t^\top = v_t^\top - k_t^\top P_{t \leftarrow 0} S_{c-1} - \sum_{j=1}^{t-1} \beta_j \big(k_t^\top P_{t \leftarrow j} k_j\big) u_j^\top.
\label{eq:app_wy_sys}
\end{equation}
With tile cumulative log-decay $\Lambda_{t,g} = \sum_{s=1}^t \log \gamma_{s,g}$, the inner product $k_t^\top P_{t \leftarrow j} k_j = \sum_{g=1}^4 e^{\Lambda_{t,g} - \Lambda_{j,g}} \langle k_{t,g}, k_{j,g} \rangle$ evaluates via the four masked $[C, 64] \times [64, C]$ MXU matmuls above without VPU sub-blocks. Because $\beta_j$ is scalar per key head $h_k$, the strictly lower-triangular coupling matrix $M_c \in \R^{C \times C}$ defined by
\begin{equation}
(M_c)_{t,j} = \mathbf{1}[t > j]\,\beta_j \big(k_t^\top P_{t \leftarrow j} k_j\big)
\label{eq:app_mc}
\end{equation}
is shared across all $D_{v,\text{pack}}$ value columns and requires inverting a single $C \times C$ unit lower-triangular system $A_c = (I_C + M_c)^{-1}$ per key head.

Stacking chunk values $V_c \in \R^{C \times D_{v,\text{pack}}}$ (row $t$ equal to $v_t^\top$), decayed keys $\bar{K}_c \in \R^{C \times d_k}$ (row $t$ equal to $(P_{t \leftarrow 0} k_t)^\top$), and residuals $u_{1:C} \in \R^{C \times D_{v,\text{pack}}}$ (row $t$ equal to $u_t^\top$) yields $\beta$-scaled WY factors and chunk writes:
\begin{align*}
U_c &= \diag(\beta_c) A_c V_c \in \R^{C \times D_{v,\text{pack}}}, \qquad
W_c = \diag(\beta_c) A_c \bar{K}_c \in \R^{C \times d_k}, \\
E_c &= \diag(\beta_c)\, u_{1:C} = U_c - W_c S_{c-1}.
\end{align*}
Letting the carried keys $\hat{K}_c \in \R^{C \times d_k}$ have row $j$ equal to $(P_{C \leftarrow j} k_j)^\top$ and $F_c = P_{C \leftarrow 0}$, the end-of-chunk state and chunk increment satisfy
\[
\tilde{S}_C = F_c S_{c-1} + \hat{K}_c^\top(U_c - W_c S_{c-1}), \qquad
\Delta S_c = \tilde{S}_C - F_c S_{c-1} = \hat{K}_c^\top E_c \in \R^{d_k \times D_{v,\text{pack}}}.
\]
(When optional oblique key whitening $\tilde{k}_j$ is enabled in \texttt{SPD-SLA-W}, Appendix~\ref{sec:whitening}, replacing the write key $k_j$ with $\tilde{k}_j$ in $M_c$ and $\hat{K}_c$ yields the identical WY dual equations.) Finally, the chunk-level velocity $Z_c = \mu \kappa_c Z_{c-1} + \Delta S_c$ and $S_c = F_c S_{c-1} + Z_c$ (Eq.~\eqref{eq:sfa_mom}, with reset indicator $\kappa_c \in \{0, 1\}$ equal to $0$ if chunk $c$ contains a document boundary) act solely at chunk boundaries, preserving timestep key-value pairing inside every chunk and matching token-by-token decoding that accumulates $\Delta S_c$ over $C$ steps before applying Eq.~\eqref{eq:sfa_mom}. \qed

\paragraph{Intra-tile conditioning and orthogonal rotation absorption.}
\label{obs:givens}
Because each group's decay is a scalar multiple of the identity, it preserves conditioning within each tile and commutes with rotary rotations inside the group:

\begin{observation}[Intra-Tile Conditioning and Orthogonal Rotation Absorption]\label{obs:givens_stmt}
Within each group $g$, the decay block $\gamma_{t,g} I_{d_k/G}$ has unit condition number $\kappa_2(\gamma_{t,g} I_{d_k/G}) = 1$ (preserving intra-tile cosine similarity) and commutes with any rotation acting within each tile, $\mathrm{Rot} \in \mathrm{SO}(d_k/G)^{G}$, such as Givens or rotary rotations, leaving the grouped Gram computation of Pillar~2 unchanged.
\end{observation}

\begin{proof}
For a per-channel diagonal decay $F_t = \diag(f_t)$ with $f_{t,i} \in (0, 1)$ and cumulative log-decay $\Lambda_{t,i} = \sum_{s=1}^t \log f_{s,i}$, the chunk decay $F_c = P_{C \leftarrow 0} = \diag\big((\exp(\Lambda_{C,i}))_{i=1}^{d_k}\big)$ has singular values $\exp(\Lambda_{C,i})$ and condition number $\kappa_2(F_c) = \exp(\max_i \Lambda_{C,i} - \min_i \Lambda_{C,i})$; write $\Delta\Lambda_C = \max_i \Lambda_{C,i} - \min_i \Lambda_{C,i}$. In BF16 ($7$ mantissa bits, machine epsilon $2^{-7} \approx 7.8 \times 10^{-3}$, unit roundoff $2^{-8} \approx 3.9 \times 10^{-3}$, exponent range $\approx [2^{-126}, 2^{127}]$), relative channel scaling across $P_{t \leftarrow 0} P_{j \leftarrow 0}^{-1}$ loses all mantissa significance across fast vs.\ slow channels once $\Delta\Lambda_C \gtrsim 5.5$ in BF16 ($\kappa_2 \approx 2^8$) or $\Delta\Lambda_C \gtrsim 16$ in FP32 accumulators ($\kappa_2 \approx 8.9 \times 10^6 \approx 2^{23}$), and $\exp(\pm \Lambda_{j,i})$ overflows or underflows FP32 when $|\Lambda_{j,i}| > 88.7$. Under tile-grouped decay $F_t = \bigoplus_{g=1}^4 \gamma_{t,g} I_{64}$, each $64$-channel tile block is a scalar multiple of $I_{64}$ with $\kappa_2(\gamma_{t,g} I_{64}) = 1$, preserving exact intra-tile cosine similarity.

For any $2 \times 2$ block-diagonal (Givens/rotary) rotation with per-pair angles $\theta_t = (\theta_{t,1}, \dots, \theta_{t,d_k/2}) \in \R^{d_k/2}$ indexed by pair $l \in \{1, \dots, d_k/2\}$,
\[
\mathrm{Rot}(\theta_t) = \bigoplus_{l=1}^{d_k/2} \begin{pmatrix} \cos\theta_{t,l} & -\sin\theta_{t,l} \\ \sin\theta_{t,l} & \cos\theta_{t,l} \end{pmatrix} \in \mathrm{SO}(2)^{d_k/2},
\]
the cumulative rotation $\prod_{t=1}^C \mathrm{Rot}(\theta_t) = \mathrm{Rot}(\Theta_C)$ with $\Theta_t = \sum_{s=1}^t \theta_s$ has all singular values equal to $1$ ($\kappa_2 = 1$). Since $\mathrm{Rot}(\theta_s)$ commutes with $G = 4$ scalar tile multiples ($64$ channels $= 32$ contiguous $2 \times 2$ pairs per tile), for tile-grouped $F_s$ we have
\[
k_t^\top \Big(\prod_{s=j+1}^t F_s \mathrm{Rot}(\theta_s)\Big) k_j = \big(\mathrm{Rot}(\Theta_t)^\top k_t\big)^\top P_{t \leftarrow j} \big(\mathrm{Rot}(\Theta_j)^\top k_j\big).
\]
Replacing $(q_t, k_t)$ with the rotated queries and keys $(\mathrm{Rot}(\Theta_t)^\top q_t,\, \mathrm{Rot}(\Theta_t)^\top k_t)$ before the chunk kernel reduces the intra-chunk transition to pure $G = 4$ real tile decay, preserving the unchanged systolic WY kernel up to a final boundary unrotation $\mathrm{Rot}(\Theta_C)$ on outgoing state.
\end{proof}

\subsubsection{Pillar 3: Fused MXU Chunk Inversion and Conditioning Certificate}
\label{prop:ben}

\paragraph{Existing MXU solvers \priortag{Prior / Existing} and SLA's fused integration.}
While GPU kernels (\texttt{fla}~\cite{FLA2024Repo}), MaxText's XLA \texttt{solve\_triangular} path, and open-source Tokamax's FP32 fallback compute $A_c = (I_C + M_c)^{-1}$ by sequential substitution, open-source TPU kernels already express unit lower-triangular inversion with MXU matrix multiplications. For BF16 inputs, Tokamax~\cite{Tokamax2025Repo} inverts $8$-token diagonal blocks via a Neumann series and combines blocks by Horner doubling with FP32 accumulation (validated at $C = 64$ for per-channel KDA, and $C \in [16, 512]$ for scalar decay). Open-source MaxText~\cite{MaxText2025Repo} implements exact recursive block-doubling in its XLA GDN path (and our Pallas KDA baseline uses the same block-doubling formulation). SLA adopts FP32 recursive block-doubling (\texttt{Exact Inv.}, doubling $2 \times 2$ blocks six times to $C = 128$, Figure~\ref{fig:sla_overview}e) and contributes three elements on top: (1) executing $R\times$ fewer exact solves directly inside a single VMEM kernel with custom-VJP inverse reuse (Section~\ref{app:bwd_alg}); (2) a chunk-size-independent conditioning certificate (Theorem~\ref{thm:wy_cond}); and (3) TPU v7x numerical verification showing that FP32 block-doubling holds worst-case relative error at $1.4 \times 10^{-7}$--$3.0 \times 10^{-7}$ across $C \in \{32, 64, 128, 256\}$ on real Qwen3.5 and high-coherence SLA chunks, whereas Tokamax-style BF16 blocked Neumann reaches $10.6$--$12.7$ relative error at $C = 128$ and unblocked Neumann products overflow (Appendix~\ref{app:v2_solver_audit}). Let $H_{t,j} = \sum_{g=1}^G e^{-|\Lambda_{t,g} - \Lambda_{j,g}|}\langle k_{t,g}, k_{j,g}\rangle$ denote the symmetric decayed key Gram matrix, so $M_c = \tril_{-1}(H)\diag(\beta_c)$. Theorem~\ref{thm:wy_cond} certifies that exact block-doubling is unconditionally well-conditioned at any chunk size or key coherence:

\begin{theorem}[Chunk-WY Conditioning and Recursive Sub-Block Certificate]
\label{thm:wy_cond}
For unit keys $\norm{k_t}_2 = 1$, non-increasing cumulative log-decays $\Lambda_{t,g}$ ($G \in \{1, \dots, d_k\}$), and $D = \diag(\beta_c)$ with $\beta_t \in (0, \beta_{\max}]$, $\beta_{\max} < 2$, the Laplace-Gram Schur product yields $H \succeq 0$ with unit diagonal $H_{t,t} = 1$, and $B = D^{1/2}(I_C + M_c)D^{-1/2}$ satisfies $B + B^\top = 2I_C - D + D^{1/2} H D^{1/2} \succeq (2 - \beta_{\max})I_C$. Consequently,
\[
\norm{B^{-1}}_2 \;\le\; \frac{2}{2 - \beta_{\max}}, \qquad \norm{D(I_C + M_c)^{-1}}_2 \;\le\; \frac{2\beta_{\max}}{2 - \beta_{\max}},
\]
strictly independent of chunk size $C$, key coherence, $\beta_{\min}$, and $\norm{M_c}_2$. Moreover, because every contiguous index subset $S \subseteq \{1, \dots, C\}$ induces a positive semidefinite principal submatrix $H_{S,S} \succeq 0$ with unit diagonal, every intermediate block-doubling factor $X_s$ ($s = 2, 4, \dots, C/2$) in symmetric $B$-coordinates obeys the identical bound $\norm{B_{S,S}^{-1}}_2 \le \frac{2}{2 - \beta_{\max}}$, so each doubling step $X_{2s} = X_s - X_s L_s X_s$ amplifies local rounding perturbations by at most $\mathcal{O}\big((2 - \beta_{\max})^{-2}\norm{L_s}_2\big)$ rather than exponentially in $C$.
\end{theorem}

\begin{proof}
For each decay group $g \in \{1, \dots, G\}$, $\Lambda_{t,g} = \sum_{s=1}^t \log \gamma_{s,g}$ is monotonically non-increasing in $t$, so $\Lambda_{t,g} - \Lambda_{j,g} = -|\Lambda_{t,g} - \Lambda_{j,g}|$ for all $t \ge j$. On $\R$, the Laplace kernel $\mathcal{K}_{\text{Lap}}(x, y) = e^{-|x - y|} = \frac{1}{\pi}\int_{-\infty}^{\infty} \frac{\cos(\omega(x - y))}{1 + \omega^2}\,d\omega$ is positive semidefinite, hence $E_g = [e^{-|\Lambda_{t,g} - \Lambda_{j,g}|}]_{t,j=1}^C \succeq 0$. Since each group key Gram matrix $K_{c,g} K_{c,g}^\top \succeq 0$, the Schur product theorem gives $E_g \odot (K_{c,g} K_{c,g}^\top) \succeq 0$. Summing over $g = 1, \dots, G$ proves $H = \sum_{g=1}^G E_g \odot (K_{c,g} K_{c,g}^\top) \succeq 0$, with $H_{t,t} = \sum_{g=1}^G \norm{k_{t,g}}_2^2 = \norm{k_t}_2^2 = 1$ (covering scalar $G = 1$, tile-grouped $G = 4$, and per-channel $G = d_k$ decay).

Writing $M_c = \tril_{-1}(H) D$, conjugating $I_C + M_c$ by $D^{1/2}$ yields $B = D^{1/2}(I_C + M_c) D^{-1/2} = I_C + D^{1/2}\tril_{-1}(H) D^{1/2}$. Because $\tril_{-1}(H) + \tril_{-1}(H)^\top = H - I_C$, symmetrizing $B$ gives
\[
B + B^\top = 2I_C + D^{1/2}(H - I_C)D^{1/2} = 2I_C - D + D^{1/2} H D^{1/2} \;\succeq\; (2 - \beta_{\max}) I_C,
\]
since $D^{1/2} H D^{1/2} \succeq 0$ and $2I_C - D \succeq (2 - \beta_{\max}) I_C$. For any unit vector $u \in \R^C$, Cauchy-Schwarz yields $\norm{B u}_2 \ge u^\top B u = \frac{1}{2} u^\top(B + B^\top)u \ge 1 - \frac{1}{2}\beta_{\max} > 0$, establishing $\sigma_{\min}(B) \ge 1 - \frac{1}{2}\beta_{\max}$ and $\norm{B^{-1}}_2 \le \frac{2}{2 - \beta_{\max}}$. The WY write operator $D(I_C + M_c)^{-1} = D^{1/2} B^{-1} D^{1/2}$ applied in Ops 5a--5b therefore satisfies $\norm{D(I_C + M_c)^{-1}}_2 \le \norm{D^{1/2}}_2^2 \norm{B^{-1}}_2 \le \frac{2\beta_{\max}}{2 - \beta_{\max}}$, independent of $C$, key coherence, $\beta_{\min}$, and $\norm{M_c}_2$. Restricting to any contiguous block $S \subseteq \{1, \dots, C\}$ preserves $H_{S,S} \succeq 0$ and $(H_{S,S})_{t,t} = 1$, so the identical symmetrization gives $B_{S,S} + B_{S,S}^\top \succeq (2 - \beta_{\max})I_{|S|}$ and $\norm{B_{S,S}^{-1}}_2 \le \frac{2}{2 - \beta_{\max}}$ for every intermediate diagonal inverse $X_s$ formed during block-doubling. Across $600$ random and coherent configurations ($G \in \{1, 4, 256\}$, $C \in \{64, 128, 256\}$), the worst empirical ratio $\norm{B^{-1}}_2 / [2/(2 - \beta_{\max})]$ is $0.973 \le 1$.

Two practical caveats separate the certificate from the default kernel implementation. First, $\beta_t = 2\sigma(\tilde{\beta}_t) \in (0, 2)$ has no hard architectural ceiling $\beta_{\max} < 2$ unless clamped: with identical keys and $\beta_t = 1.99$, $\norm{(I_C + M_c)^{-1}}_2$ reaches $146$ (still bounded by $2/(2 - 1.99) = 200$), whereas trained $12.5\text{B}$ SLA layers stay well below $2$ ($\bar{\beta} \in [0.251, 0.348]$). Second, our production kernel inverts the unscaled $I_C + M_c$ rather than the symmetric $B$: with $C = 256$ identical keys and constant $\beta_t = 1.9$, $\norm{M_c}_2 = 309.05$ yet $\norm{(I_C + M_c)^{-1}}_2 = \norm{B^{-1}}_2 = 19.87 \le 20$, whereas if $\beta_{1:128} = 1.9$ and $\beta_{129:256} = 10^{-8}$, unscaled $\norm{(I_C + M_c)^{-1}}_2$ reaches $50.37 > 20$ via $D^{-1/2} B^{-1} D^{1/2}$ skew while symmetric $\norm{B^{-1}}_2 = 19.54 \le 20$ and $\norm{D(I_C + M_c)^{-1}}_2 = 37.13 \le 38.0$ obey the certificate. Scaling keys symmetrically by $\sqrt{\beta_t}$ before Op 4 inverts $B$ directly on the MXU, holding $6$-step block-doubling relative error at $5.2 \times 10^{-6}$ in FP32 and $3.6\%$ with BF16 operands even at $\norm{M_c}_2 = 309$. Viewed iteratively, the intra-chunk delta rule executes one successive over-relaxation (Kaczmarz) sweep on $H$ with relaxation $\beta_t \le \beta_{\max} < 2$.
\end{proof}

SLA also offers a cheaper approximation, \texttt{1-Step Neumann}, which keeps only the first correction term between blocks; the ablation runs and prefill benchmarks use a matmul-free row-scaled variant of it. On a cluster of $b$ aligned keys ($H = \mathbf{1}_b \mathbf{1}_b^\top$), one stationary Jacobi sweep $I_b - \beta H$ has spectral radius $|b\beta - 1| > 1$ whenever $b\beta > 2$. Row-sum Gershgorin damping stabilizes \texttt{A3}/\texttt{A5} at $+0.22$ pre-training perplexity (Section~\ref{subsec:dolma_15b}), whereas swapping \texttt{Exact Inv.} in at test time shifts $64\text{K}$ QA by at most $0.01$ points ($2.5 \times 10^{-5}\text{ nats}$ $14$-layer tail KL). Pre-training and context-parallel runs therefore use \texttt{Exact Inv.}, except ablation arms \texttt{A3} and \texttt{A5}; prefill and tree-verification benchmarks use \texttt{1-Step Neumann}, and single-layer benchmarks report both.

\paragraph{Block-doubling and block-Neumann factorization.}
Let $D_{\text{blk}}$ be the block-diagonal part of $I_C + M_c$ (its $C/16$ unit lower-triangular $16 \times 16$ diagonal blocks) and let $N = D_{\text{blk}}^{-1}(I_C + M_c - D_{\text{blk}})$ (local to Pillar~3), so that $I_C + M_c = D_{\text{blk}}(I_C + N)$ with $N$ strictly block-lower-triangular and $N^{C/16} = 0$. Algebraically, at $C = 128$, the exact inverse and its $1$-step block-Neumann truncation $A_{\text{N1}} \in \R^{C \times C}$ take the form
\begin{equation}
\begin{aligned}
A_{\text{N1}} &= (I_C - N)\, D_{\text{blk}}^{-1}, \\
(I_C + M_c)^{-1} &= \sum_{s=0}^{7} (-N)^s D_{\text{blk}}^{-1} = (I_C + N^4)(I_C + N^2)(I_C - N)\, D_{\text{blk}}^{-1}.
\end{aligned}
\label{eq:ben}
\end{equation}
Rather than evaluating the expanded polynomial product $(I_C + N^4)(I_C + N^2)(I_C - N) D_{\text{blk}}^{-1}$, which amplifies powers of $N$ in finite precision (diverging on $22\%$--$23\%$ of real Qwen3.5 $C = 128$ chunks; Appendix~\ref{app:v2_solver_audit}), our Pallas kernel (\texttt{Exact Inv.}) builds on open-source MXU block-doubling~\cite{MaxText2025Repo} via the numerically stable \textbf{recursive Schur/block-doubling recurrence} in FP32 (\texttt{Precision.HIGHEST}):
\[
X_{2s} = X_s - X_s L_s X_s \qquad (s = 2, 4, 8, 16, 32, 64),
\]
where $X_s$ is the block-diagonal matrix of already-inverted $s \times s$ diagonal blocks of $I_C + M_c$ ($X_{128} = A_c$) and $L_s$ keeps only the strictly lower-left $s \times s$ coupling block inside each $2s \times 2s$ diagonal block (zero elsewhere), so $L_s X_s L_s = 0$ and the update is exact. Concretely, at each step $s \to 2s$, given the inverse of the two $s \times s$ diagonal halves ($X_{\text{top}}, X_{\text{bot}}$) and their strictly lower-left $s \times s$ coupling block $L_{\text{off}}$, the lower-left block of $X_{2s}$ is computed via two MXU matmuls as $-X_{\text{bot}} L_{\text{off}} X_{\text{top}}$.

\paragraph{Variants of \texttt{1-Step Neumann} and truncation properties.}
Two different first-order approximations appear under the name \texttt{1-Step Neumann}, and we keep them distinct:
\begin{enumerate}[leftmargin=1.8em, itemsep=2pt, topsep=2pt]
\item the \emph{block} truncation $A_{\text{N1}} = (I_C - N) D_{\text{blk}}^{-1}$ with block width $B_{\text{blk}} = 16$, analyzed below and evaluated in our standalone lag/skew benchmark; and
\item the matmul-free \emph{row-scaled} approximation $A_{\text{N1}}^{\text{row}} \in \R^{C \times C}$, with row-scaling vector $r = (r_1, \dots, r_C)^\top \in \R^C$, used by the pre-training ablation arms \texttt{A3} and \texttt{A5} (Table~\ref{tab:dolma_15b}b) and the prefill benchmarks:
\[
A_{\text{N1}}^{\text{row}} = I_C - \diag(r)^{-1} M_c, \qquad r_i = \max\!\big(1,\, 1.25 \norm{(M_c)_{i,:}}_1\big).
\]
\end{enumerate}
The exactness property below (zero error for lags $t - j \le 16$) holds for (1) but not for (2). Partition unit lower-triangular $I_C + M_c = D_{\text{blk}} + M_{\text{off}} = D_{\text{blk}}(I_C + N)$, where $D_{\text{blk}} = \diag(D_1, \dots, D_m)$ consists of $m = C / B_{\text{blk}}$ unit lower-triangular diagonal blocks $D_a \in \R^{B_{\text{blk}} \times B_{\text{blk}}}$ ($a \in \{1, \dots, m\}$) and $M_{\text{off}}$ is zero on and above the block diagonal. Then $N = D_{\text{blk}}^{-1} M_{\text{off}}$ is strictly block-lower-triangular (its $(a,b)$ block satisfies $N_{a,b} = 0$ for $a \le b$, hence $(N^s)_{a,b} = 0$ for all $a - b < s$ and $N^m = 0$). Comparing $(I_C + M_c)^{-1} = \sum_{s=0}^{m-1} (-N)^s D_{\text{blk}}^{-1}$ against $A_{\text{N1}} = (I_C - N) D_{\text{blk}}^{-1} = D_{\text{blk}}^{-1} - D_{\text{blk}}^{-1} M_{\text{off}} D_{\text{blk}}^{-1}$, the residual
\[
(I_C + M_c)^{-1} - A_{\text{N1}} = \Big(\sum_{s=2}^{m-1} (-1)^s N^s\Big) D_{\text{blk}}^{-1}
\]
vanishes identically on all block entries $(a, b)$ with $a - b \le 1$, namely every diagonal block $(a, a)$ and every first sub-diagonal block $(a+1, a)$. Any two token indices $1 \le j \le t \le C$ with $t - j \le B_{\text{blk}}$ satisfy $\lceil t / B_{\text{blk}} \rceil - \lceil j / B_{\text{blk}} \rceil \in \{0, 1\}$ and therefore lie in block $(a, a)$ or $(a+1, a)$, incurring zero truncation error.

In finite-precision BF16/FP32 arithmetic, exact WY training vs.\ stateful step-by-step decoding achieves measured output KL divergence $< 10^{-6}$ per layer ($3.4 \times 10^{-5}$ across $4$-layer BF16 stacks when chunk momentum syncs every $C$ steps), while under \texttt{1-Step Neumann} ($B_{\text{blk}} = 16, C = 128$), truncation at lags $t - j > 16$ exhibits relative Frobenius error $1.6 \times 10^{-5}$ in short-horizon \texttt{4L:1A} layers up to $4.9 \times 10^{-2}$ in long-horizon \texttt{15L:1A} layers, yielding single-layer output $\text{KL} < 10^{-5}$ against exact recurrent decoding (and zero discrepancy if chunked prefill uses \texttt{1-Step Neumann}). Training with the row-scaled approximation (2) nonetheless changes the learned model: in the pre-training ablation (Table~\ref{tab:dolma_15b}b, \texttt{A2}$\to$\texttt{A3}) it raises $11$-set perplexity by $0.22$. \qed

\paragraph{Why exact inversion remains the default once values are packed.}
Once Pillar~1 packs $R = 4$ value heads per key head ($2$ solves per chunk), exact FP32 block-doubling adds only $4\%$--$8\%$ to full-LM pre-training step time over matmul-free $1$-step approximations while maintaining the lowest mean perplexity and tightest seed spread (Appendix~\ref{app:v2_solver_audit}). Replacing exact inversion zero-shot on exact-trained checkpoints with low-order global series fails because deep layers exhibit strongly correlated keys ($\norm{M_c}_2 > 1$ in $74\%$--$83\%$ of Qwen3.5 chunks). Appendix~\ref{app:v2_solver_audit} details exploratory approximate solvers for inference, including banded exact $16 \times 16$ blocks (matching exact zero-shot MMLU within $0.00$--$0.42$ points at $N = 1{,}425$), a signed erase-aware $1$-step operator (\texttt{eaadj}) exact for collinear keys under scalar decay, and $C = 256$ compilation with \texttt{kVectorMove}-safe exact block-doubling.

\subsection{Single-Kernel VMEM Execution and Custom-VJP Backward Pass}
\label{app:bwd_alg}

\begin{figure}[ht]
\centering
\includegraphics[width=0.96\textwidth]{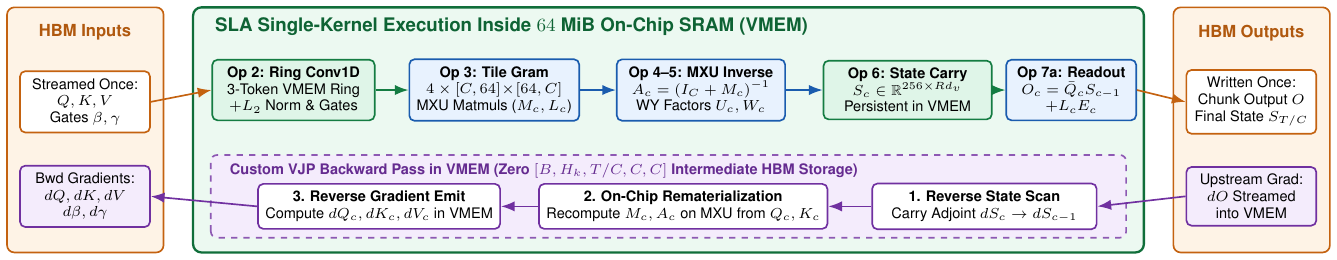}
\caption{\textbf{Single-kernel VMEM execution and custom VJP backward pass in SLA.} \textbf{Top (Forward):} Streamed $Q, K, V$ and gates enter the $64\text{ MiB}$ VMEM scratchpad once; a $3$-token ring buffer executes causal \texttt{Conv1D} and $L_2$ normalization (Op 2), followed in-place by $4$-tile MXU Gram contractions (Op 3), MXU chunk inversion and WY factors (Ops 4--5), persistent $[256, D_{v,\text{pack}}]$ state updates (Op 6), and chunk readout (Op 7a) before writing $O$ to HBM. \textbf{Bottom (Backward):} During the reverse state scan, the custom VJP kernel recomputes intra-chunk matrices $M_c, A_c$ on the MXU directly inside VMEM, eliminating $[B, H_k, T/C, C, C]$ intermediate HBM writes for batch size $B$.}
\label{fig:sla_one_vmem}
\end{figure}

Algorithm~\ref{alg:sla_bwd} details the custom-VJP backward pass of fused SLA (illustrated in Figure~\ref{fig:sla_one_vmem}, bottom), where for a scalar training loss $\mathcal{L}$ we write $dM_c = \partial \mathcal{L} / \partial M_c$ (and analogously $dQ_c, dK_c, dV_c, \dots$) for the loss gradient (adjoint) with respect to each tensor. Let $\gamma_c \in (0, 1]^{C \times 4}$ stack chunk $c$'s tile decays $\gamma_{t,g}$ (with tile column $\gamma_{c,g} \in (0, 1]^C$), and let $dO_c = \partial \mathcal{L} / \partial O_c \in \R^{C \times R d_v}$ denote the upstream gradient of chunk $c$'s readout after backpropagating through the per-value-head $\RMSNorm$ and output gate in Eq.~\eqref{eq:sfa_readout}. For batch size $B$, a standard autodiff tape stores the intra-chunk matrices $M_c, L_c, A_c \in \R^{C \times C}$ from the forward pass in HBM, consuming $3 B H_k (T/C) C^2$ words. Because SLA's Pillar~2 and Pillar~3 evaluate both the $G = 4$ Gram matrices and the chunk inverse $A_c = (I_C + M_c)^{-1}$ as MXU matmuls, our backward kernel instead saves only the chunk-boundary states $S_{c-1} \in \R^{d_k \times R d_v}$ and recomputes $M_c, L_c, A_c$, and the unscaled writes $\tilde{E}_c = A_c (V_c - \bar{K}_c S_{c-1}) \in \R^{C \times R d_v}$ on-the-fly inside VMEM during the reverse scan over $c = T/C, \dots, 1$.

Whereas unfused baselines recompute or store separate per-value-head solves for the transposed system $(I_C + M_c)^{-\top}$ in the backward pass, SLA already holds its $R\times$-fewer dense inverses $A_c$ in VMEM. Differentiating $E_c = \diag(\beta_c) A_c (V_c - \bar{K}_c S_{c-1}) = \diag(\beta_c) \tilde{E}_c$ through $A_c = (I_C + M_c)^{-1}$ reduces the inverse-solve backpropagation to two MXU matmuls for the scaled adjoint $D_c \in \R^{C \times R d_v}$ and Gram adjoint $dM_c \in \R^{C \times C}$, shared across all $R$ packed value heads:
\begin{equation}
D_c = A_c^\top \diag(\beta_c)\, dE_c \in \R^{C \times R d_v}, \qquad
dM_c = -\tril_{-1}\!\big(D_c \tilde{E}_c^\top\big) \in \R^{C \times C},
\label{eq:app_bwd_dm}
\end{equation}
where $\tril_{-1}(\cdot)$ and $\tril_0(\cdot)$ mask to strictly ($t > j$) and weakly ($t \ge j$) lower-triangular entries, $\mathrm{rowsum}(\cdot)$ and $\mathrm{colsum}(\cdot)$ sum matrix rows and columns, and $p_{c,g}^{\text{in}}, p_{c,g}^{\text{out}} \in (0, 1]^C$ denote the tile-$g$ boundary-decay vectors $(p_{c,g}^{\text{in}})_t = \prod_{s=1}^{t} \gamma_{s,g}$ and $(p_{c,g}^{\text{out}})_t = \prod_{s=t+1}^{C} \gamma_{s,g}$. From $D_c$, the value and decayed-key gradients follow immediately as $dV_c = D_c$ and $d\bar{K}_c = -D_c S_{c-1}^\top$.

\begin{algorithm}[ht]
\caption{Fused SLA Custom-VJP Backward Pass (Per Key Head in VMEM)}
\label{alg:sla_bwd}
\footnotesize
\begin{algorithmic}[1]
\REQUIRE Saved boundary states $\{S_{c-1}\}_{c=1}^{T/C}$, chunk inputs $\{Q_c, K_c, V_c, \beta_c, \gamma_c\}$, upstream readout gradients $\{dO_c\}_{c=1}^{T/C}$
\ENSURE Input gradients $\{dQ_c, dK_c, dV_c, d\beta_c, d\gamma_c\}_{c=1}^{T/C}$
\STATE Initialize reverse carry $dS_{T/C} \gets 0,\; dZ_{T/C} \gets 0 \in \R^{d_k \times R d_v}$ in VMEM
\FOR{$c = T/C$ {\bfseries down to} $1$}
    \STATE \textbf{// 1. On-chip VMEM recomputation (Ops 3--5 on MXU; zero $[C, C]$ HBM reads)}
    \STATE Recompute $p_{c,g}^{\text{in}}, p_{c,g}^{\text{out}}$, decayed $(\bar{K}_c, \bar{Q}_c, \hat{K}_c)$, $G = 4$ masks $(\Gamma_g, \Psi_g)$, and $M_c, L_c$ via $64$-wide MXU matmuls
    \STATE Recompute $A_c \gets (I_C + M_c)^{-1}$ on MXU; form $\tilde{E}_c \gets A_c(V_c - \bar{K}_c S_{c-1})$ and $E_c \gets \diag(\beta_c)\tilde{E}_c$
    \STATE \textbf{// 2. Reverse state and momentum scan adjoints (Ops 6--7a)}
    \STATE $\widetilde{dZ}_c \gets dZ_c + dS_c$; \quad $dZ_{c-1} \gets \mu \kappa_c \widetilde{dZ}_c$ \hfill $\triangleright$ \text{velocity carry ($\kappa_c = 0$ at doc reset)}
    \STATE $dE_c \gets \hat{K}_c \widetilde{dZ}_c + L_c^\top dO_c \in \R^{C \times R d_v}$; \quad $d\hat{K}_c \gets E_c \widetilde{dZ}_c^\top \in \R^{C \times d_k}$ \hfill $\triangleright$ \text{MXU}
    \STATE $dL_c \gets \tril_0(dO_c E_c^\top) \in \R^{C \times C}$; \quad $d\bar{Q}_c \gets dO_c S_{c-1}^\top \in \R^{C \times d_k}$ \hfill $\triangleright$ \text{MXU}
    \STATE \textbf{// 3. Transposed WY inverse and value/key adjoints on MXU (Ops 4--5)}
    \STATE $D_c \gets A_c^\top \diag(\beta_c)\, dE_c \in \R^{C \times R d_v}$; \quad $dV_c \gets D_c$; \quad $d\bar{K}_c \gets -D_c S_{c-1}^\top$ \hfill $\triangleright$ \text{MXU}
    \STATE $dM_c \gets -\tril_{-1}(D_c \tilde{E}_c^\top) \in \R^{C \times C}$ \hfill $\triangleright$ \text{contracts $R d_v$ packed value cols on MXU}
    \STATE $dS_{c-1} \gets F_c^\top dS_c - \bar{K}_c^\top D_c + \bar{Q}_c^\top dO_c \in \R^{d_k \times R d_v}$ \hfill $\triangleright$ \text{prev.\ chunk state carry ($\bar{K}_c^\top D_c = W_c^\top dE_c$)}
    \STATE \textbf{// 4. Tile-grouped $Q, K$ and gate adjoints ($G = 4$ tiles of $b = 64$ channels)}
    \FOR{$g \in \{1, 2, 3, 4\}$}
        \STATE $\widetilde{dM}_{c,g} \gets dM_c \odot \Gamma_g$; \quad $\widetilde{dL}_{c,g} \gets dL_c \odot \Psi_g$
        \STATE $dQ_{c,g} \gets \widetilde{dL}_{c,g} K_{c,g} + \diag(p_{c,g}^{\text{in}})(d\bar{Q}_c)_{:,g}$ \hfill $\triangleright$ \text{$64$-wide MXU matmul; $(\cdot)_{:,g}$ is tile-$g$ slice}
        \STATE $dK_{c,g} \gets (\widetilde{dM}_{c,g} + \widetilde{dM}_{c,g}^\top) K_{c,g} + \widetilde{dL}_{c,g}^\top Q_{c,g} + \diag(p_{c,g}^{\text{in}})(d\bar{K}_c)_{:,g} + \diag(p_{c,g}^{\text{out}})(d\hat{K}_c)_{:,g}$
    \ENDFOR \hfill $\triangleright$ \text{concatenate $dQ_c = [dQ_{c,1}, \dots, dQ_{c,4}],\; dK_c = [dK_{c,1}, \dots, dK_{c,4}]$}
    \STATE $d\beta_c \gets \mathrm{rowsum}(dE_c \odot \tilde{E}_c) + \mathrm{colsum}\!\big(dM_c \odot \sum_{g=1}^4 \Psi_g \odot (K_{c,g} K_{c,g}^\top)\big)^\top$
    \STATE Accumulate columns $d\gamma_{c,g}$ of $d\gamma_c$ via reverse prefix sum over $dM_c, dL_c, d\bar{K}_c, d\bar{Q}_c, d\hat{K}_c$, and $dS_c S_{c-1}^\top$
\ENDFOR
\STATE \textbf{return} $\{dQ_c, dK_c, dV_c, d\beta_c, d\gamma_c\}_{c=1}^{T/C}$
\end{algorithmic}
\end{algorithm}

\section{Empirical Results}
\label{sec:eval}

We evaluate SLA on TPU v7x (Ironwood~\cite{GoogleCloud2025TPU7x,Austin2025ScalingBook}) at four levels: single-layer kernels and multi-generation rooflines (Section~\ref{subsec:kernels}), long-context serving (Section~\ref{subsec:serving}), context-parallel training and end-to-end $12.5\text{B}$ MoE cube serving (Section~\ref{subsec:sfa_cp}), and multi-scale MoE pre-training throughput and quality (Sections~\ref{app:pretraining}--\ref{subsec:dolma_15b}). We use two configurations with $d_k = 256$, $G = 4$ decay tiles, chunk size $C = 128$, and $H_v = 8$ value heads: \texttt{SLA-R2} shares each query-key head across $2$ value heads ($H_q = H_k = 4$, where $H_q$ is the query-head count) and \texttt{SLA-R4} across $4$ ($H_q = H_k = 2$); layer shapes are written $H_k \times d_k \times R d_v$. \texttt{$x$L:1A} denotes $x$ linear attention layers per full softmax attention layer, so \texttt{0L:1A} is a pure softmax attention Transformer. SLA keeps the baseline's value width, $d_v = 256$ at model width $D = 2048$ and $512$ at $D = 1792$, so SLA's BF16 state has twice the entries of FP32 KDA with $d_k = 128$ in the same memory, and the same entries as the $63.1\text{B}$ KDA with $d_k = 256$ in half the memory. \texttt{SLA Iso-State} instead halves $d_v$ to match KDA's element count in half the bytes. Precision explains only a small part of the speedup: switching MaxText KDA's state from FP32 to BF16 makes it just $2\%$--$8\%$ faster (Table~\ref{tab:v7x_kernels}, footnote). \texttt{SLA Tree-CP} (Table~\ref{tab:cp_serving}c) and all SLA pre-training runs use \texttt{Exact Inv.}, except ablation arms \texttt{A3} and \texttt{A5} (Table~\ref{tab:dolma_15b}b); $32$-layer prefill and tree verification (Table~\ref{tab:cp_serving}a--b) use \texttt{1-Step Neumann}, and single-layer benchmarks report both (Tables~\ref{tab:7op_audit}--\ref{tab:v7x_kernels}).

\subsection{Kernel Performance and Multi-Generation TPU Rooflines}
\label{subsec:kernels}
\label{subsec:sfa_one}
\label{subsec:layer_benchmarks}

\begin{table}[ht]
\caption{\textbf{Seven-operator forward audit ($\text{ms}$) on TPU v7x} ($D = 2048$, $T = 8{,}192$). Unfused columns report isolated per-stage forward medians; fused brackets report single-kernel Ops 2--7a latency.}
\label{tab:7op_audit}
\centering
\small
\renewcommand{\arraystretch}{1.08}
\setlength{\tabcolsep}{4pt}
\begin{tabular*}{\textwidth}{@{\extracolsep{\fill}}lcccc@{}}
\toprule
Stage Latency ($D=2048, T=8\text{K}$) & KDA (Unf.) & KDA (Fus.) & SLA-R4 (Unf.) & \textbf{SLA-R4 (Fused)} \\
\midrule
Op 1: QKV + Gate Proj (MXU) & 0.251 & 0.251 & 0.168 & 0.168 \\
Op 2: Causal Conv4 + QK Norm & 0.562 & \multirow{4}{*}{$\left.\begin{array}{@{}c@{}}\text{Fused}\\[0.2ex]3.104\end{array}\right\}$} & 0.341 & \multirow{4}{*}{$\left.\begin{array}{@{}c@{}}\text{Fused}\\[0.2ex]1.112\end{array}\right\}$} \\
Op 3--5: Gram + WY Inversion & 1.184 & & 0.418 & \\
Op 6: Chunk State Scan ($S_c$) & 1.248 & & 0.614 & \\
Op 7a: Intra + State Readout & 0.219 & & 0.219 & \\
Op 7b--7c: Out Norm + Out Proj & 0.255 & 0.255 & 0.245 & 0.245 \\
\midrule
\textbf{Measured Full Layer Fwd (ms)} & 3.719 & 3.610 & 2.005 & \textbf{1.525} \\
\textbf{Analytical Roofline Fwd (ms)} & 2.890 & 2.740 & 1.380 & \textbf{0.867} \\
\textbf{Measured Full Layer F+B (ms)} & 7.108 & 6.912 & 3.982 & \textbf{2.992} \\
\textbf{F+B Speedup vs.\ KDA (Unf.)} & $1.00\times$ & $1.03\times$ & $1.79\times$ & \textbf{$2.38\times$} \\
\bottomrule
\end{tabular*}
\par\vspace{3pt}\raggedright{\scriptsize Stateless functional harness (\texttt{1-Step Neumann} for \texttt{SLA-R4}). Fused: one kernel for Ops 2--7a; full NNX module latencies in Table~\ref{tab:v7x_kernels}.}
\end{table}

\begin{table}[ht]
\caption{\textbf{Full-layer TPU v7x latency ($\text{ms}$, Ops 1--7c, $B = 1$), state bytes, $32\text{K}$ temporary HBM, and F+B speedups} vs.\ \texttt{MaxText Pallas KDA} ($C = 128$, used in pre-training and serving) and \texttt{Tokamax Mosaic KDA} ($C = 64$). Core-kernel and iso-shape full-layer attributions are in Appendix Tables~\ref{tab:app_v7x_core}--\ref{tab:app_isoshape}.}
\label{tab:v7x_kernels}
\centering
\footnotesize
\renewcommand{\arraystretch}{1.06}
\setlength{\tabcolsep}{2.8pt}
\resizebox{\textwidth}{!}{%
\begin{tabular}{@{}llcccccccc@{}}
\toprule
Geometry & Full Layer ($H_k \times d_k \times R d_v$, Chunk $C$) & State & $8\text{K}$ F+B & vs.\ Max. & vs.\ Tok. & $32\text{K}$ F+B & vs.\ Max. & vs.\ Tok. & $32\text{K}$ Temp HBM \\
\midrule
\multirow{12}{*}{\begin{tabular}[c]{@{}l@{}}\textbf{(a) $D = 1792$}\\($63.1\text{B}$ MoE\\width)\end{tabular}}
 & MaxText Pallas KDA ($8 \times 256 \times 512, C=128$, FP32)$^\ast$ & $4.0\text{ MiB}$ & 14.46 & $1.00\times$ & $0.91\times$ & 48.66 & $1.00\times$ & $0.87\times$ & $9{,}302\text{ MiB}$ \\
 & Tokamax Mosaic KDA ($8 \times 256 \times 512, C=64$, FP32) & $4.0\text{ MiB}$ & 13.22 & $1.09\times$ & $1.00\times$ & 42.38 & $1.15\times$ & $1.00\times$ & $9{,}483\text{ MiB}$ \\
 & GDN Scalar Unpacked ($G=1$, $8 \times 256 \times 512, C=128$) & $4.0\text{ MiB}$ & 13.78 & $1.05\times$ & $0.96\times$ & 48.39 & $1.01\times$ & $0.88\times$ & $9{,}217\text{ MiB}$ \\
 & SLA-R2 Unfused ($G=1$, $4 \times 256 \times 1024, C=128$, BF16) & $2.0\text{ MiB}$ & 10.88 & $1.33\times$ & $1.21\times$ & 32.46 & $1.50\times$ & $1.31\times$ & $6{,}922\text{ MiB}$ \\
 & SLA-R2 Unfused ($G=4$, $4 \times 256 \times 1024, C=128$, BF16) & $2.0\text{ MiB}$ & 10.56 & $1.37\times$ & $1.25\times$ & 32.56 & $1.49\times$ & $1.30\times$ & $6{,}922\text{ MiB}$ \\
 & SLA-R4 Unfused ($G=4$, $2 \times 256 \times 2048, C=128$, BF16) & $2.0\text{ MiB}$ & 7.71 & $1.88\times$ & $1.72\times$ & 26.24 & $1.85\times$ & $1.62\times$ & $5{,}782\text{ MiB}$ \\
 & \textbf{SLA-R4 2-Stage, Exact Inv.} ($2 \times 256 \times 2048$, BF16) & \textbf{$2.0\text{ MiB}$} & \textbf{5.57} & \textbf{$2.60\times$} & \textbf{$2.37\times$} & \textbf{20.24} & \textbf{$2.40\times$} & \textbf{$2.09\times$} & \textbf{$3{,}738\text{ MiB}$} \\
 & \textbf{SLA-R4 2-Stage, 1-Step Neumann} ($2 \times 256 \times 2048$, BF16) & \textbf{$2.0\text{ MiB}$} & \textbf{5.25} & \textbf{$2.75\times$} & \textbf{$2.52\times$} & \textbf{18.79} & \textbf{$2.59\times$} & \textbf{$2.26\times$} & \textbf{$3{,}738\text{ MiB}$} \\
 & \emph{1-Chip Pre-Alloc.\ Ref.: MaxText KDA / Tokamax KDA}$^\dagger$ & $4.0\text{ MiB}$ & 10.42 / 8.70 & $1.00\times^\dagger$ & $1.00\times^\dagger$ & 39.81 / 34.01 & $1.00\times^\dagger$ & $1.00\times^\dagger$ & $9{,}302\text{ MiB}$ \\
 & \textbf{SLA-R4 1-Kernel, Scalar $G=1$ Control} ($2 \times 256 \times 2048$, BF16)$^\dagger$ & \textbf{$2.0\text{ MiB}$} & \textbf{3.73} & \textbf{$2.79\times^\dagger$} & \textbf{$2.33\times^\dagger$} & \textbf{14.05} & \textbf{$2.83\times^\dagger$} & \textbf{$2.42\times^\dagger$} & \textbf{$3{,}738\text{ MiB}$} \\
 & \textbf{SLA-R4 1-Kernel, Exact Inv., $G=4$} ($2 \times 256 \times 2048$, BF16)$^\dagger$ & \textbf{$2.0\text{ MiB}$} & \textbf{3.89} & \textbf{$2.68\times^\dagger$} & \textbf{$2.24\times^\dagger$} & \textbf{14.41} & \textbf{$2.76\times^\dagger$} & \textbf{$2.36\times^\dagger$} & \textbf{$3{,}738\text{ MiB}$} \\
\midrule
\multirow{12}{*}{\begin{tabular}[c]{@{}l@{}}\textbf{(b) $D = 2048$}\\($12.5\text{B}$ linear\\value width)\end{tabular}}
 & MaxText Pallas KDA ($8 \times 128 \times 256, C=128$, FP32)$^\ast$ & $1.0\text{ MiB}$ & 9.53 & $1.00\times$ & $0.65\times$ & 32.45 & $1.00\times$ & $0.74\times$ & $4{,}432\text{ MiB}$ \\
 & Tokamax Mosaic KDA ($8 \times 128 \times 256, C=64$, FP32) & $1.0\text{ MiB}$ & 6.16 & $1.55\times$ & $1.00\times$ & 24.06 & $1.35\times$ & $1.00\times$ & $4{,}727\text{ MiB}$ \\
 & GDN Scalar Unpacked ($G=1$, $8 \times 128 \times 256, C=128$) & $1.0\text{ MiB}$ & 10.24 & $0.93\times$ & $0.60\times$ & 31.86 & $1.02\times$ & $0.76\times$ & $4{,}360\text{ MiB}$ \\
 & SLA-R2 Unfused ($G=1$, $4 \times 256 \times 512, C=128$, BF16) & $1.0\text{ MiB}$ & 7.25 & $1.31\times$ & $0.85\times$ & 25.28 & $1.28\times$ & $0.95\times$ & $4{,}260\text{ MiB}$ \\
 & SLA-R4 Unfused ($G=4$, $2 \times 256 \times 1024, C=128$, BF16) & $1.0\text{ MiB}$ & 5.22 & $1.82\times$ & $1.18\times$ & 19.68 & $1.65\times$ & $1.22\times$ & $3{,}472\text{ MiB}$ \\
 & SLA-R2 2-Stage, 1-Step ($4 \times 256 \times 512, C=128$, BF16) & $1.0\text{ MiB}$ & 4.77 & $2.00\times$ & $1.29\times$ & 17.29 & $1.88\times$ & $1.39\times$ & $3{,}092\text{ MiB}$ \\
 & \textbf{SLA-R4 2-Stage, Exact Inv.} ($2 \times 256 \times 1024$, BF16) & \textbf{$1.0\text{ MiB}$} & \textbf{4.54} & \textbf{$2.10\times$} & \textbf{$1.36\times$} & \textbf{15.72} & \textbf{$2.06\times$} & \textbf{$1.53\times$} & \textbf{$2{,}263\text{ MiB}$} \\
 & \textbf{SLA-R4 2-Stage, 1-Step Neumann} ($2 \times 256 \times 1024$, BF16) & \textbf{$1.0\text{ MiB}$} & \textbf{3.99} & \textbf{$2.39\times$} & \textbf{$1.55\times$} & \textbf{14.82} & \textbf{$2.19\times$} & \textbf{$1.62\times$} & \textbf{$2{,}263\text{ MiB}$} \\
 & \emph{1-Chip Pre-Alloc.\ Ref.: MaxText KDA / Tokamax KDA}$^\dagger$ & $1.0\text{ MiB}$ & 6.87 / 4.26 & $1.00\times^\dagger$ & $1.00\times^\dagger$ & 26.69 / 16.28 & $1.00\times^\dagger$ & $1.00\times^\dagger$ & $4{,}432\text{ MiB}$ \\
 & \textbf{SLA-R4 1-Kernel, Scalar $G=1$ Control} ($2 \times 256 \times 1024$, BF16)$^\dagger$ & \textbf{$1.0\text{ MiB}$} & \textbf{2.63} & \textbf{$2.61\times^\dagger$} & \textbf{$1.62\times^\dagger$} & \textbf{9.20} & \textbf{$2.90\times^\dagger$} & \textbf{$1.77\times^\dagger$} & \textbf{$2{,}263\text{ MiB}$} \\
 & \textbf{SLA-R4 1-Kernel, Exact Inv., $G=4$} ($2 \times 256 \times 1024$, BF16)$^\dagger$ & \textbf{$1.0\text{ MiB}$} & \textbf{2.74} & \textbf{$2.51\times^\dagger$} & \textbf{$1.55\times^\dagger$} & \textbf{9.59} & \textbf{$2.78\times^\dagger$} & \textbf{$1.70\times^\dagger$} & \textbf{$2{,}263\text{ MiB}$} \\
\bottomrule
\end{tabular}%
}
\par\vspace{3pt}\raggedright{\scriptsize $^\ast$Switching MaxText KDA's state from FP32 to BF16 changes $8\text{K}/32\text{K}$ F+B only slightly to $13.91 / 46.40\text{ ms}$ ($D=1792$) and $8.80 / 31.72\text{ ms}$ ($D=2048$). Tokamax Mosaic KDA receives BF16 QKV inputs (its MXU Neumann--Horner path); the FP32 label denotes its FP32 recurrent state (Appendix~\ref{app:v2_solver_audit} audits Tokamax BF16 vs.\ FP32 QKV inputs directly). Row-scaled FP8 halves SLA state bytes further ($1.0\text{ MiB}$ and $0.5\text{ MiB}$). Separate $8\text{K}$ Fwd/Bwd splits for \texttt{SLA-R4 2-Stage Exact}: $2.88 / 2.69\text{ ms}$ ($D=1792$) and $2.44 / 2.10\text{ ms}$ ($D=2048$) vs.\ MaxText $6.41 / 8.04$ and $4.74 / 4.79\text{ ms}$ (Tokamax: $5.90 / 7.33$ and $3.57 / 2.59\text{ ms}$). Undaggered rows measured on container \texttt{maxtext\_e174258\_20261008} (Tokamax $0.0.14$) with $B = 1$/core concurrently across a $64$-chip ($128$-core) slice and \texttt{--xla\_tpu\_scoped\_vmem\_limit\_kib=65472} ($64\text{ MiB}$ VMEM budget, required by fused \texttt{SLA-R4} at $D = 1792$ with $\mu = 0.25$; XLA's default $16\text{ MiB}$ cap suffices at $\mu = 0$ and $D = 2048$). $^\dagger$Daggered $1$-kernel rows and their matched single-chip pre-allocated baseline references are measured under the unified single-chip protocol ($3$ independent process restarts, $64\text{ MiB}$ VMEM, Appendix Table~\ref{tab:app_single_protocol}a), so every daggered speedup ratio divides directly against the italicized reference row immediately above it (e.g.\ $10.42 / 3.89 = 2.68\times$ and $8.70 / 3.89 = 2.24\times$; iso-shape controls in Appendix Table~\ref{tab:app_isoshape}; GDN-2 in Section~\ref{app:gdn2_tiles}).}
\end{table}

\paragraph{Fusion and stage breakdown.}
Because every intra-chunk stage of SLA runs on the MXU, the fused kernel keeps Ops 2--7a in the $64\text{ MiB}$ VMEM scratchpad and recomputes $M_c$ and $A_c$ in the backward pass instead of storing them in HBM (Section~\ref{app:bwd_alg} and Figure~\ref{fig:sla_one_vmem}). Table~\ref{tab:7op_audit} shows where the time goes. Fusing GPU-shaped KDA gains only $1.03\times$. Reshaping the layer without fusion (unfused \texttt{SLA-R4}) gives $1.79\times$, primarily because $4{:}1$ key sharing cuts projection, convolution, Gram, and solve work while packing full MXU tiles. Fusion then brings the total to $2.38\times$ (with the $1$-step Neumann inverse; Table~\ref{tab:v7x_kernels} reports both inverses).

\paragraph{Comparison with open-source TPU baselines.}
Table~\ref{tab:v7x_kernels} compares full SLA layers with MaxText Pallas KDA~\cite{MaxText2025Repo} ($C = 128$, used in our pre-training and serving runs), Tokamax Mosaic KDA~\cite{Tokamax2025Repo} (the fastest baseline; its per-channel KDA kernel is validated at $C = 64$, running on its BF16-input MXU Neumann--Horner path), and unpacked scalar GDN~\cite{Yang2025GDN}. With the exact inverse, $2$-stage \texttt{SLA-R4} is $2.1\times$--$2.6\times$ faster than open-source MaxText KDA and $1.4\times$--$2.4\times$ faster than open-source Tokamax KDA across the $64$-chip slice (and $1$-kernel fused \texttt{SLA-R4} reaches $2.51\times$--$2.78\times$ vs.\ MaxText and $1.55\times$--$2.36\times$ vs.\ Tokamax under the unified single-chip protocol; $1.9\times$--$2.5\times$ vs.\ BF16-state MaxText KDA), while using $49\%$--$61\%$ less temporary HBM. Note that at $D = 1792$ the KDA baseline already uses $d_k = 256$, so key-tile padding elimination applies to $D = 2048$ ($d_k = 128 \to 256$), whereas at $D = 1792$ SLA's speedup comes from $4{:}1$ head and write-gate sharing, tile-grouped decay, and single-kernel VMEM fusion. Specifically, at $D = 1792, 32\text{K}$, the $20.24\text{ ms}$ ($20.237\text{ ms}$) latency in Table~\ref{tab:v7x_kernels}a corresponds to the $2$-stage chunk kernel with $\mu = 0.25$ ($20.30 \pm 0.23\text{ ms}$ on an isolated chip in Appendix Table~\ref{tab:app_isoshape}), which is already $1.69\times$ faster than single-chip Tokamax Mosaic KDA ($34.31 \pm 0.13\text{ ms}$) and $2.09\times$ faster under Table~\ref{tab:v7x_kernels}'s $64$-chip concurrent timing ($42.38\text{ ms}$), whereas the $1$-kernel fused variant ($14.08 \pm 0.30\text{ ms}$, $\mu = 0$) is $2.44\times$ faster than Tokamax Mosaic KDA at $D = 1792$. Core-kernel controls (Appendix Table~\ref{tab:app_v7x_core}b) show that beating Tokamax requires $4{:}1$ head sharing: at $R = 2$ without sharing four value heads, $d_k = 256$ SLA cores run at $0.67\times$--$0.77\times$ of Tokamax. Conversely, Appendix Table~\ref{tab:app_isoshape} holds the full layer fixed at the \texttt{SLA-R4} shape ($H_k = 2, d_k = 256, R = 4$, $n = 3$ restarts) to separate each design choice: $4{:}1$ key sharing accounts for $1.71\times$--$1.87\times$ on the MaxText KDA kernel and single-kernel VMEM fusion adds $1.21\times$--$1.41\times$, whereas BF16 vs.\ FP32 state adds $1.00\times$--$1.01\times$ (providing $2\times$ memory and VMEM headroom instead). Grouping decay from $G = d_k$ to $G = 4$ speeds up pure XLA by $1.62\times$--$2.45\times$ and Pallas chunk kernels by $1.04\times$--$1.12\times$ while running at the same single-kernel VMEM speed ($0\%$--$9\%$) as a scalar gate ($G = 1$, $14.08$ vs.\ $14.16\text{ ms}$ at $D = 1792, 32\text{K}$), because $1$ or $4$ gate vectors both fit in VMEM vector registers without $[B, H_k, T/C, C, C, d_k]$ HBM materialization; however, $G = 1$ restricts each head to one retention timescale and trails $G = 4$ on $11$-set validation perplexity across all six paired seeds ($-0.23 \pm 0.24$ PPL, $6/6$ wins; Section~\ref{subsec:dolma_15b}), making $G = 4$ the multi-timescale quality and single-kernel VMEM speed Pareto knee. Packing $R = 4$ values into Tokamax alone gives only $1.01\times$--$1.10\times$ at $D = 2048$ and overflows $64\text{ MiB}$ VMEM in backward at $D = 1792$; fused \texttt{SLA-R4} remains $1.34\times$--$1.60\times$ faster than iso-shape Tokamax and $1.27\times$--$1.58\times$ faster than the fastest iso-shape KDA Pallas kernel. To make the attribution against open-source Tokamax exact, we also implemented and verified a single-kernel VMEM-fused per-channel ($G = d_k$) KDA kernel at both $R = 1$ and $R = 4$ (forward cosine $0.999995$ and all five VJP gradients $\ge 0.999984$ vs.\ FP32 reference; Appendix~\ref{app:r87_controls}, Table~\ref{tab:app_3way_attrib}). Across $D \in \{1792, 2048\}$ and $T \in \{8\text{K}, 32\text{K}\}$ on single-chip TPU v7x, fused \texttt{SLA-R4} ($\mu = 0$) beats Tokamax $R = 1$ by $1.52\times$--$1.84\times$, factoring cleanly into $1.28\times$--$1.41\times$ from $4{:}1$ head packing, $1.10\times$--$1.12\times$ from VMEM fusion at matched $R = 4$, and $1.07\times$--$1.20\times$ from tile grouping $G = d_k \to G = 4$ ($1.20\times$ at $D = 2048, 32\text{K}$, while $G = 1$ stays within $2.7\%$--$4.0\%$ of $G = 4$). Crucially, VMEM-fusing unpacked $R = 1$ per-channel KDA runs \emph{slower} than Tokamax $R = 1$ ($0.44\times$--$0.69\times$), confirming that $4{:}1$ value packing is what makes single-kernel VMEM fusion profitable.

\begin{figure}[ht]
\centering
\includegraphics[width=0.72\textwidth]{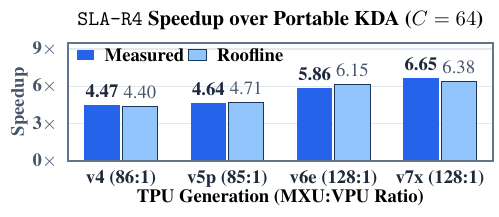}
\caption{\textbf{Forward speedup of \texttt{SLA-R4} across TPU generations} over a portable FP32 KDA baseline, with leave-one-out roofline predictions ($3.1\%$ error). Not comparable to the MaxText/Tokamax-normalized speedups in Table~\ref{tab:v7x_kernels}.}
\label{fig:multigen}
\end{figure}

\paragraph{Multi-generation TPU rooflines and FP8 skew.}
\label{app:rooflines}
On an identical portable JAX/Pallas kernel stack ($B = 1, T = 8{,}192, D = 2048$), \texttt{SLA-R4}'s single-core forward speedup rises from $4.5\times$ on TPU v4 to $6.7\times$ on TPU v7x (Figure~\ref{fig:multigen} and Appendix Table~\ref{tab:app_multigen}). Across these generations the MXU tile widens from $128$ to $256$ and the MXU:VPU throughput ratio $\alpha_{\text{mv}}$ rises from about $85\times$ to $128\times$. Because Tokamax Mosaic validates per-channel KDA on v6e/v7x at $C = 64$, this cross-generation sweep holds the software stack fixed against a portable FP32 KDA kernel ($C = 64$; $6.99\text{ ms}$ on v7x), whereas Appendix Table~\ref{tab:app_v7x_core}b compares core kernels against optimized MaxText ($2.67\text{ ms}$ fwd) and Tokamax ($1.07\text{ ms}$ fwd; \texttt{SLA-R4}: $1.05\text{ ms}$) on v7x.

To understand this scaling on a single TensorCore with systolic tile width $W \in \{128, 256\}$ and single-core peaks $\mathcal{P}_{\text{MXU}}, \mathcal{P}_{\text{VPU}}$ (Table~\ref{tab:hw_specs} divided by \texttt{Cores}), we model single-core forward latency $T_{\text{fwd}}(C)$ at chunk size $C$ as the sum of inter-chunk scan, VPU solve, and intra-chunk MXU terms:
\begin{equation}
T_{\text{fwd}}(C) \approx H_k \frac{T}{C}\Big(\tau_{\text{pipe}} + \frac{F_{\text{step}}}{\eta_{\text{sys}}\mathcal{P}_{\text{MXU}}}\Big) + \frac{\gamma_{\text{inv}} H_k T C}{\mathcal{P}_{\text{VPU}}} + \frac{\Phi_{\text{MXU}}(C)}{\mathcal{P}_{\text{MXU}}}.
\label{eq:scan_law}
\end{equation}
In the scan term, $(\tilde{d}_k, \tilde{D}_{v,\text{pack}})$ pads $(d_k, D_{v,\text{pack}})$ to multiples of $W$, $F_{\text{step}} = 2\max(C, W)\,\tilde{d}_k \tilde{D}_{v,\text{pack}}$ is the FLOP count per scan matmul at effective single-core utilization $\eta_{\text{sys}} \approx 8.6\%$, and $\tau_{\text{pipe}} \approx 1.5$--$2.5\,\mu\text{s}$ is the in-VMEM Pallas pipeline fill/drain cost per head and chunk (below the top-level XLA \texttt{lax.scan} HBM-barrier cost $\tau_{\text{sync}} \approx 12\,\mu\text{s}$ in Figure~\ref{fig:laws}). In the remaining terms, $\gamma_{\text{inv}}$ counts VPU operations per triangular-solve entry (unrelated to the decay rates $\gamma_{t,g}$) and $\Phi_{\text{MXU}}(C) = \mathcal{O}(H_k T C d_k + H_v T C d_v)$ counts intra-chunk matmul FLOPs.
The four generations in Table~\ref{tab:app_multigen} form two clusters. On $128 \times 128$-tile TPUs (v4 and v5p; no KDA key-tile padding, $\alpha_{\text{mv}} \approx 85\times$), \texttt{SLA-R4} is already $4.47\times$--$4.64\times$ faster. On $256 \times 256$-tile TPUs (v6e and v7x; $50\%$ KDA key-tile padding at $d_k = 128$, $\alpha_{\text{mv}} \approx 128\times$), the speedup rises by a further $1.3\times$--$1.46\times$ to $5.86\times$--$6.65\times$. Tile width and $\alpha_{\text{mv}}$ change together between the clusters, so these measurements cannot separate their effects (a reduced-form leave-one-out fit is shown for reference in Table~\ref{tab:app_multigen}). On an isolated FP8 MXU microbenchmark, \texttt{SLA-R4} is $1.69\times$ faster than the standalone JAX/Pallas $C = 128$ KDA core kernel in native E4M3 FP8 mode, where $\alpha_{\text{mv}}$ doubles to $256\times$ ($1.280$ vs.\ $2.168\text{ ms}$). This matches its $1.67\times$ BF16 speedup on the same harness ($1.181$ vs.\ $1.975\text{ ms}$), which excludes MaxText's segment-reset and $L_2$-norm wrapper (included in the $2.668\text{ ms}$ row of Appendix Table~\ref{tab:app_v7x_core}b). To test out-of-sample predictiveness beyond these four generations, we locked a priori analytical roofline predictions for $4$ anchor shapes and $18$ strictly held-out single-chip TPU v7x designs before measuring them ($0$ failures; Appendix~\ref{app:r87_controls}, Table~\ref{tab:app_oos_roofline}). Zero-parameter peak rooflines achieve Spearman rank correlation $\rho = 0.893$ on full-layer F+B ($0.874$ full forward); a single anchor-calibrated efficiency factor ($\eta_{\text{cal}} = 0.339$, anchor $R^2 = 0.987$) yields $19.8\%$ MAPE ($18.6\%$ MedAPE) across all $18$ held-out shapes and $13.9\%$ MAPE ($12.5\%$ MedAPE) inside the target packed systolic regime ($R \ge 4, d_k \ge 256, n = 12$). On the held-out $(R = 8, d_k = 512, C = 256, d_v = 256, T = 8\text{K})$ design, the model predicts $2.56\text{ ms}$ vs.\ $2.47\text{ ms}$ measured ($+3.5\%$ error), whereas unpacked $R = 1$ shapes run $42\%$--$53\%$ slower than the packed efficiency extrapolation due to unmodeled VPU and scheduling overhead.

\subsection{Long-Context Serving}
\label{subsec:serving}

\begin{figure}[ht]
\centering
\includegraphics[height=0.325\textwidth,trim={0 0 288.9pt 0},clip]{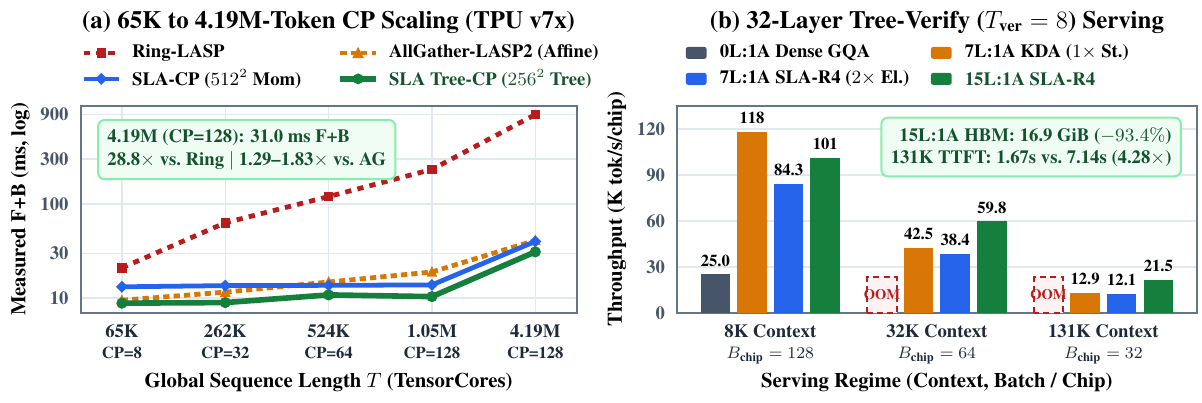}\hfill
\includegraphics[height=0.325\textwidth]{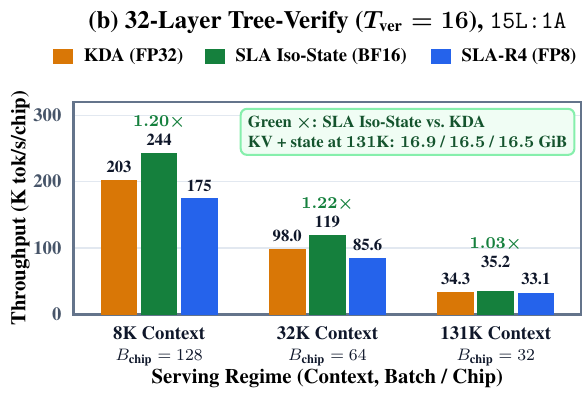}
\caption{\textbf{Context parallelism and serving on TPU v7x.} \textbf{(a)} Forward-backward scan latency across $8$--$128$ TensorCores ($65\text{K}$--$4.19\text{M}$ tokens, Section~\ref{subsec:sfa_cp}); the callout highlights the no-momentum \texttt{SLA Tree-CP} curve ($\mu = 0$), while trained \texttt{SLA-CP} ($\mu = 0.25$) takes $40.1\text{ ms}$ at $4.19\text{M}$ ($0.85\times$--$1.37\times$ vs.\ AllGather across $262\text{K}$--$4.19\text{M}$; Table~\ref{tab:cp_serving}c). \textbf{(b)} Single-chip $32$-layer attention-layer tree-verification throughput for the \texttt{15L:1A} hybrid at $T_{\text{ver}} = 16$ (Table~\ref{tab:cp_serving}b). At equal state element count (BF16 vs.\ FP32), \texttt{SLA Iso-State} is $1.03\times$--$1.22\times$ faster than KDA; \texttt{SLA-R4} with FP8 state reaches $0.86\times$--$0.97\times$ of KDA's throughput.}
\label{fig:cp_serving}
\end{figure}

\begin{table}[ht]
\caption{\textbf{Single-chip $32$-layer prefill latency (a)}, $32$-layer attention-layer serving throughput with speculative verification of $T_{\text{ver}} \in \{8, 16\}$ drafted tokens (b; \texttt{OOM} $> 190\text{ GiB}$/chip), and context-parallel F+B scan latency across $P \in \{32, 64, 128\}$ TensorCores (c) on TPU v7x ($D = 2048$; $12.5\text{B}$ MoE cube serving in Appendix Table~\ref{tab:app_e2e_12b_cp_serving}).}
\label{tab:cp_serving}
\centering
\footnotesize
\renewcommand{\arraystretch}{1.06}
\setlength{\tabcolsep}{3.0pt}
\resizebox{\textwidth}{!}{%
\begin{tabular}{@{}llcccc@{}}
\toprule
Regime & Architecture / Method ($32$L, $D = 2048$) & $8\text{K}$ TTFT & $32\text{K}$ TTFT & $131\text{K}$ TTFT (vs.\ KDA) & $32\text{K}$ Cache/Lyr \\
\midrule
\multirow{5}{*}{\begin{tabular}[c]{@{}l@{}}\textbf{(a) $32$L Prefill}\\Full TTFT\\($B = 1$)\end{tabular}}
 & \texttt{0L:1A} Dense GQA ($H_q=16$, $H_{kv}=4$ KV heads) & $214.7\text{ ms}$ & $761.1\text{ ms}$ & $7{,}139\text{ ms}$ ($0.37\times$) & $64.0\text{ MiB}$ KV \\
 & \texttt{15L:1A} MaxText KDA ($8 \times 128 \times 256$, FP32) & $265.7\text{ ms}$ & $689.1\text{ ms}$ & $2{,}664\text{ ms}$ ($1.00\times$) & $1.0\text{ MiB}$ FP32 \\
 & \texttt{7L:1A} SLA-R4 ($2 \times 256 \times 1024$, BF16, 1-Step) & $204.3\text{ ms}$ & $439.8\text{ ms}$ & $2{,}034\text{ ms}$ ($1.31\times$) & $1.0\text{ MiB}$ BF16 \\
 & \textbf{\texttt{15L:1A} SLA-R4} ($2 \times 256 \times 1024$, FP8, 1-Step) & \textbf{$203.6\text{ ms}$} & \textbf{$416.9\text{ ms}$} & \textbf{$1{,}669\text{ ms}$ ($1.60\times$)} & \textbf{$0.5\text{ MiB}$ FP8} \\
 & \textbf{\texttt{15L:1A} SLA-R4} ($2 \times 256 \times 1024$, BF16, \textbf{Exact Inv.}) & \text{--} & \text{--} & \textbf{$1{,}623\text{ ms}$ ($1.64\times$)} & $1.0\text{ MiB}$ BF16 \\
\midrule
Regime & Attention-Layer Serving (\texttt{tok/s/chip})$^\dagger$ & $8\text{K}, B\!=\!128$ & $32\text{K}, B\!=\!64$ & $131\text{K}, B\!=\!32$ & KV+State ($131\text{K}$) \\
\midrule
\multirow{8}{*}{\begin{tabular}[c]{@{}l@{}}\textbf{(b) Serving}\\Tree Verify\\($T_{\text{ver}} \in \{8, 16\}$)\end{tabular}}
 & \texttt{0L:1A} Dense GQA ($T_{\text{ver}} = 8$) & $25.0\text{K}$ & OOM & OOM & $256.0\text{ GiB}$ \\
 & \texttt{7L:1A} MaxText KDA (FP32, $T_{\text{ver}} = 8$) & $117.7\text{K}$ & $42.5\text{K}$ & $12.9\text{K}$ & $32.88\text{ GiB}$ \\
 & \texttt{7L:1A} SLA-R4 ($2\times$ El., $T_{\text{ver}} = 8$) & $84.3\text{K}$ & $38.4\text{K}$ & $12.1\text{K}$ & $32.88\text{ GiB}$ \\
 & \texttt{7L:1A} MaxText KDA (FP32, $T_{\text{ver}} = 16$) & $152.9\text{K}$ & $64.3\text{K}$ & OOM & $32.88\text{ GiB}$ \\
 & \texttt{7L:1A} SLA Iso-State (BF16, $T_{\text{ver}} = 16$) & $162.1\text{K}$ ($1.06\times$) & $68.6\text{K}$ ($1.07\times$) & OOM & $32.44\text{ GiB}$ \\
 & \texttt{15L:1A} MaxText KDA (FP32, $T_{\text{ver}} = 16$) & $202.5\text{K}$ & $98.0\text{K}$ & $34.3\text{K}$ & $16.94\text{ GiB}$ \\
 & \texttt{15L:1A} SLA Iso-State (BF16, $T_{\text{ver}} = 16$) & \textbf{$243.8\text{K}$ ($1.20\times$)} & \textbf{$119.3\text{K}$ ($1.22\times$)} & \textbf{$35.2\text{K}$ ($1.03\times$)} & \textbf{$16.47\text{ GiB}$} \\
 & \texttt{15L:1A} SLA-R4 FP8 (E4M3, $T_{\text{ver}} = 16$) & $174.7\text{K}$ & $85.6\text{K}$ & $33.1\text{K}$ & \textbf{$16.47\text{ GiB}$} \\
\midrule
Regime & Context-Parallel Scan (F+B ms) & $P\!=\!32$ ($262\text{K}$) & $P\!=\!64$ ($524\text{K}$) & $P\!=\!128$ ($1.05\text{M}$) & $P\!=\!128$ ($4.19\text{M}$) \\
\midrule
\multirow{5}{*}{\begin{tabular}[c]{@{}l@{}}\textbf{(c) Context}\\Parallelism\\($32$--$128$ Cores)\end{tabular}}
 & Ring-LASP Sequential Ring ($P - 1$ Hops) & 63.20 & 120.12 & 233.47 & 894.41 \\
 & AllGather-LASP2 Collective Baseline & 11.60 & 14.84 & 19.00 & 41.06 \\
 & SLA Tree-CP ($\mu = 0.25$, trained, $512^2$) & 13.59 & 13.71 & 13.84 & 40.08 \\
 & SLA Tree-CP ($\mu = 0$, no mom., $256^2$) & 8.97 & 10.84 & 10.40 & 31.03 \\
 & Tree-CP vs.\ AllGather ($\mu = 0.25$ / $\mu = 0$) & $0.85\times$ / $1.29\times$ & $1.08\times$ / $1.37\times$ & $1.37\times$ / $1.83\times$ & $1.02\times$ / $1.32\times$ \\
\bottomrule
\end{tabular}%
}
\par\vspace{3pt}\raggedright{\scriptsize $^\dagger$Panel (b) reports attention layers only ($2 \times 95\text{ GiB}$/core HBM cap); with SwiGLU FFNs included at $T_{\text{ver}} = 16$, \texttt{SLA Iso-State} is $2\%$--$7\%$ faster end-to-end than matched-cadence KDA. Panels (a)--(b) use \texttt{1-Step Neumann} (at $131\text{K}$ BF16 prefill, \texttt{Exact Inv.} reaches $1{,}623\text{ ms}$ [$4.40\times$ vs.\ Dense] vs.\ $1{,}477\text{ ms}$ for \texttt{1-Step Neumann} with $0.99998$ $14$-layer cosine and $2.5 \times 10^{-5}\text{ nats}$ tail KL); panel (c) uses exact inversion (\texttt{Exact Inv.}). On exact-trained checkpoints without retraining, zero-shot prefill should use \texttt{Exact Inv.} or banded exact blocks (Appendix~\ref{app:v2_solver_audit}), whereas matmul-free \texttt{1-Step Neumann} requires training with it or short KL healing.}
\end{table}

\paragraph{Prefill latency and prefix caching.}
In $32$-layer hybrid stacks with SwiGLU FFNs~\cite{Shazeer2020SwiGLU} (Table~\ref{tab:cp_serving}a), \texttt{15L:1A SLA-R4} (one softmax layer per $16$) reaches a $131\text{K}$ time-to-first-token (TTFT) of $1.67\text{ s}$, compared with $2.66\text{ s}$ for the same hybrid layout with KDA and $7.14\text{ s}$ for full grouped-query attention (GQA)~\cite{Ainslie2023GQA}. Most of the gain over GQA comes from the hybrid layout itself. In the same \texttt{15L:1A} layout, \texttt{SLA-R4} with FP8 state and the \texttt{1-Step Neumann} inverse is $1.6\times$ faster than FP32 KDA; for zero-shot deployment on exact-trained weights, \texttt{Exact Inv.} (BF16 $131\text{K}$ TTFT $1{,}623\text{ ms}$) or banded exact blocks avoid approximation drift without retraining (Appendix~\ref{app:v2_solver_audit}). A cached SLA state is also much smaller than a KV cache ($0.5$--$1\text{ MiB}$ per layer versus $64\text{ MiB}$ at $32\text{K}$), which makes multi-turn prefix caching cheap.

\paragraph{Speculative verification.}
Single-token decoding underuses the MXU, so we verify $T_{\text{ver}} \in \{8, 16\}$ drafted tokens per step~\cite{Leviathan2023SpecDec} with the same block inversion (Figure~\ref{fig:cp_serving}b and Table~\ref{tab:cp_serving}b). At equal state element count (\texttt{SLA Iso-State}, BF16 vs.\ FP32), SLA is $1.03\times$--$1.22\times$ faster than KDA on the attention layers, and $2\%$--$7\%$ faster once FFNs are included. \texttt{SLA-R4}, which stores twice KDA's state elements in the same bytes, is slower than KDA at every context length ($0.72\times$--$0.94\times$). Table~\ref{tab:app_e2e_12b_cp_serving}c reports a $12.5\text{B}$ MoE speculative-decoding setting in which SLA's greedy acceptance length on out-of-distribution $32\text{K}/64\text{K}$ \texttt{HotpotQA} is $2.2\times$ that of unclamped KDA and $1.17\times$--$1.33\times$ that of clamped KDA ($1.06\times$--$1.10\times$ under stochastic sampling); KDA is slightly higher in-domain at $8\text{K}$.

\subsection{Context-Parallel Training and End-to-End \texorpdfstring{$12.5\text{B}$}{12.5B} MoE Cube Serving}
\label{subsec:sfa_cp}
\label{sec:cp}

\paragraph{Context parallelism with a prefix-tree scan.}
Head-sharded Ulysses~\cite{Jacobs2023Ulysses} needs at least as many key heads as devices, which SLA's $H_k \in \{2, 4\}$ do not provide, and ring-based LASP~\cite{Sun2024LASP} passes the state through $P - 1$ sequential hops across $P$ devices. For any gated delta rule with scalar $\beta_t$, a chunk's state update is an affine map $S \mapsto J S + B$ with a $d_k \times d_k$ transition $J \in \R^{d_k \times d_k}$ and offset $B \in \R^{d_k \times D_{v,\text{pack}}}$, so $P$ devices can reduce their local chunks independently and combine them in $\lceil \log_2 P \rceil$ prefix-tree steps~\cite{Blelloch1990Prefix,Martin2018ParallelLinearRNN}. This construction is not specific to SLA, and our KDA baseline uses the same tree. SLA's advantage here is that one $256 \times 256$ transition and $G = 4$ group scalars serve all $R$ value heads of a key head, so fewer transitions are exchanged; chunk momentum ($\mu = 0.25$) doubles the carried state and enlarges the transition to $512 \times 512$. Figure~\ref{fig:cp_serving}a and Table~\ref{tab:cp_serving}c report both settings across $8$--$128$ TensorCores: on the isolated context-parallel scan across $64$ chips, \texttt{SLA Tree-CP} achieves both a $1.09\times$ speedup over \texttt{KDA-CP} ($22.58\text{ ms}$ vs.\ $24.64\text{ ms}$) and a $7.65\times$ speedup over dense \texttt{RingCP} ($22.58\text{ ms}$ vs.\ $172.71\text{ ms}$). With momentum, as trained, the tree scan is $0.85\times$--$1.37\times$ vs.\ an AllGather-based LASP-2 baseline~\cite{Sun2025LASP2}, and without momentum it is $1.29\times$--$1.83\times$ faster. In end-to-end $12.5\text{B}$ MoE training at $1\text{M}$ context ($L = 1{,}048{,}576$ on $64$ TPU v7x chips, Table~\ref{tab:app_e2e_12b_cp_serving}a), \texttt{15L:1A SLA Tree-CP} runs in $4{,}870.6\text{ ms}$ with momentum ($4{,}778.1\text{ ms}$ without momentum; $4{,}737.6\text{ ms}$ in the uncoupled sweep), achieving a $1.09\times$ speedup ($1.11\times$--$1.13\times$ without momentum) over \texttt{15L:1A KDA-CP} ($5{,}307.1$--$5{,}337.6\text{ ms}$) with $6.31\text{ GiB/core}$ lower peak HBM ($25.50$ vs.\ $31.81\text{ GiB/core}$) and a $7.50\times$--$7.75\times$ ($7.65\times$ without momentum) speedup over \texttt{0L:1A} dense \texttt{RingCP} ($36{,}530.6$--$36{,}725.0\text{ ms}$).

\paragraph{Affine monoid derivation for uncoupled and coupled \texttt{SLA Tree-CP}.}
\label{prop:sfa_cp}
\label{eq:sfa_cp_monoid}
\label{eq:monoid_op}
Shard $p \in \{1, \dots, P\}$ holds $L_{\text{loc}}$ consecutive chunks; quantities of its $c$-th local chunk ($c \in \{1, \dots, L_{\text{loc}}\}$) carry the subscript $(p, c)$, with shard input state $S_{p,0}$ and shard output state $S_{p,L_{\text{loc}}}$. Substituting $E_{p,c} = U_{p,c} - W_{p,c} S_{p,c-1}$ (Section~\ref{app:proofs}) into $\tilde{S}_C = F_{p,c} S_{p,c-1} + \hat{K}_{p,c}^\top E_{p,c}$ gives per-chunk transition $J_{p,c}^{\text{step}} \in \R^{d_k \times d_k}$ and offset $B_{p,c} \in \R^{d_k \times D_{v,\text{pack}}}$ such that, without momentum, each chunk is the affine map
\begin{align*}
S_{p,c} &= J_{p,c}^{\text{step}} S_{p,c-1} + B_{p,c}, \\
J_{p,c}^{\text{step}} &= F_{p,c} - \hat{K}_{p,c}^\top W_{p,c}, \qquad
B_{p,c} = \hat{K}_{p,c}^\top U_{p,c};
\end{align*}
$J_{p,c}^{\text{step}}$ does not depend on the values, so one transition serves all $D_{v,\text{pack}}$ value columns of a key head (this holds for any gated delta rule, since the transition only involves keys and gates). With $\mu = 0$, the $L_{\text{loc}}$ local chunks compose into a single shard map $(J_p, B_p)$ satisfying $S_{p,L_{\text{loc}}} = J_p S_{p,0} + B_p$, where $J_p = \prod_{c=L_{\text{loc}}}^{1} J_{p,c}^{\text{step}}$ (ordered from $c = 1$ on the right to $c = L_{\text{loc}}$ on the left) and $B_p$ is the shard output $S_{p,L_{\text{loc}}}$ obtained from $S_{p,0} = 0$. (With $\mu > 0$, the same construction applies to the lifted maps below.) Writing shard maps as pairs $(J_a, B_a)$ and $(J_b, B_b)$ for shards $a$ and $b$ with composition (apply $a$, then $b$)
\[
(J_b, B_b) \circ (J_a, B_a) = (J_b J_a,\; J_b B_a + B_b),
\]
for any third shard $d$ with map $(J_d, B_d)$, matrix associativity and left-distributivity yield
\begin{align*}
\big((J_d, B_d) \circ (J_b, B_b)\big) \circ (J_a, B_a)
&= (J_d J_b J_a,\; J_d J_b B_a + J_d B_b + B_d) \\
&= (J_d, B_d) \circ \big((J_b, B_b) \circ (J_a, B_a)\big),
\end{align*}
so the $P$ shard maps can be combined by any parallel prefix scan in $\lceil \log_2 P \rceil$ rounds.

When the velocity $Z_c \in \R^{d_k \times D_{v,\text{pack}}}$ is carried across shard boundaries ($\mu > 0$, \texttt{SLA Tree-CP+Mom}; shard subscript $p$ dropped, with per-chunk transition $J_c^{\text{step}}$ and offset $B_c$), we have $\Delta S_c = \tilde{S}_C - F_c S_{c-1} = (J_c^{\text{step}} - F_c) S_{c-1} + B_c$. Substituting into Eq.~\eqref{eq:sfa_mom} ($Z_c = \mu \kappa_c Z_{c-1} + \Delta S_c$, $S_c = F_c S_{c-1} + Z_c$) defines a lifted $2 d_k \times 2 d_k$ transition $\tilde{J}_c \in \R^{2 d_k \times 2 d_k}$ and offset $\tilde{B}_c \in \R^{2 d_k \times D_{v,\text{pack}}}$ updating the stacked carry $[S_c; Z_c] \in \R^{2 d_k \times D_{v,\text{pack}}}$ via the exact block-affine map
\begin{equation}
\begin{bmatrix} S_c \\ Z_c \end{bmatrix} =
\underbrace{\begin{bmatrix} J_c^{\text{step}} & \mu \kappa_c I_{d_k} \\ J_c^{\text{step}} - F_c & \mu \kappa_c I_{d_k} \end{bmatrix}}_{\tilde{J}_c}
\begin{bmatrix} S_{c-1} \\ Z_{c-1} \end{bmatrix} +
\underbrace{\begin{bmatrix} B_c \\ B_c \end{bmatrix}}_{\tilde{B}_c}.
\label{eq:app_coupled_monoid}
\end{equation}
For consecutive shards $a$ and $b$, the shard-reduced lifted maps $(\tilde{J}_a, \tilde{B}_a)$ and $(\tilde{J}_b, \tilde{B}_b)$ obey the identical associative composition law $( \tilde{J}_b, \tilde{B}_b ) \circ ( \tilde{J}_a, \tilde{B}_a ) = ( \tilde{J}_b \tilde{J}_a, \tilde{J}_b \tilde{B}_a + \tilde{B}_b )$, with $[512, 512]$ transitions for $d_k = 256$. \qed

\subsection{MoE Pre-Training Setup, Architecture, and Throughput}
\label{app:pretraining}
\label{app:model_scaling}

All pre-training runs within each cohort share identical optimizer, tokenization, MoE routing, and batch-ordering configurations so that the attention layer design is the sole independent variable:
\begin{itemize}[leftmargin=1.6em,itemsep=3pt]
\item \textbf{Optimizer and Schedule:} AdamW~\cite{Loshchilov2019AdamW} (moment coefficients $(0.9, 0.95)$, numerical stability constant $\epsilon = 10^{-8}$, weight decay $0.1$ with embedding/norm exclusion), global gradient norm clip $1.0$, router load-balancing weight $0.01$, z-loss $10^{-5}$, peak learning rate $3.0 \times 10^{-4}$ with $5\%$ linear warmup and cosine decay to $3.0 \times 10^{-5}$ (plus tuned $\text{LR} = 4.5 \times 10^{-4}$ with RoPE base frequency $\theta_{\max} = 5 \times 10^5$~\cite{Su2024RoPE} for fair \texttt{0L:1A} GQA+RoPE baselines across all scales). \texttt{+MS} denotes multi-scale depth-staggered decay bias initialization across linear layers.
\item \textbf{Corpus, FP8 State PTQ, and QA Scoring Protocol:} Real Dolma~3.5~\cite{Soldaini2024Dolma,OLMo2025OLMo2,OLMo2025OLMo3} streams (training sequence length $T_{\text{train}} = 8{,}192$) with deterministic \texttt{OlmoCoreOrderSampler} batch parity and document-boundary EOS resets (\texttt{eos\_id = 100257}). In-loop validation tracks geometric-mean PPL across all $11$ OLMo v3 validation sets. Post-training evaluation covers: (1) a disjoint $16.78\text{M}$-token ($512 \times 32\text{K}$) Dolma~3 split (\texttt{4K, 8K, 16K, 32K} full and \texttt{Tail-4K} PPL, the latter on the last $4\text{K}$ tokens of each sequence; true row-scaled E4M3 FP8 recurrent-state PTQ shifts $8\text{K}$ and $32\text{K}$ NLL by only $+0.00001$ and $+0.00006$ nats with $0.00\%$ NIAH loss across SLA checkpoints, whereas an earlier post-layer activation-quantization probe altered $32\text{K}$ Tail PPL by $-0.07$ to $-0.54$ and is excluded from state-PTQ claims), (2) multi-needle Passkey (\texttt{MK-NIAH} $4/8$-Needle, $N = 256$--$512$ evaluation items), (3) $9$-task zero-shot \texttt{lm-eval}~\cite{Gao2023LMEval} across $11{,}793$ items ($1\text{ SE} \approx 0.65$ pts), and (4) real-text $4$-candidate \texttt{RULER} QA~\cite{Hsieh2024RULER} on \texttt{SQuAD}~\cite{Rajpurkar2016SQuAD} and \texttt{HotpotQA}~\cite{Yang2018HotpotQA} at $8\text{K}, 32\text{K}, 64\text{K}$ ($25\%$ chance): unless labeled $N = 512$ ($1\text{ SE} \approx 2.0$ pts), tables report raw log-likelihood \texttt{cand4\_acc} at $N = 128$ ($1\text{ SE} \approx 4.0$ pts; byte-length-normalized \texttt{cand4\_acc\_norm} agrees within $\pm 2$--$5$ pts, e.g.\ $82.0/76.6\%$ vs.\ $76.6/82.0\%$ on clamped $63.1\text{B}$ KDA).
\item \textbf{Four Sparse MoE Scales ($10.7\times$ Active Span; $20\text{ tok/active param} \approx 1\times$ Chinchilla~\cite{Hoffmann2022Chinchilla}):} Across two MXU-aligned backbone width tiers ($D = 1024$ and $D = 1792$), we evaluate (1) \textbf{$0.35\text{B}$ active / $6.5\text{B}$ label} ($7.58\text{B}$ exact total including embeddings, Table~\ref{tab:app_arch_scaling}; decoder depth $L = 10$, $D = 1024$, \texttt{4L:1A}, $k = 4$, $15.10\text{B}$ and $30.20\text{B}$ tok), (2) \textbf{$0.79\text{B}$ active / $12.5\text{B}$ total} ($L = 16$, $D = 1024$, \texttt{7L:1A}, \texttt{15L:1A}, \& \texttt{Pure-Linear} $\infty\text{:}1$, $k = 8$, $15.10\text{B}$ and $60.40\text{B}$ tok), (3) \textbf{$1.80\text{B}$ active / $31.5\text{B}$ label} ($30.58\text{B}$ exact total, Table~\ref{tab:app_arch_scaling}; $L = 15$, $D = 1792$, \texttt{4L:1A}, $k = 8$, $5.03\text{B}$ and $36.20\text{B}$ tok $\approx 1.0\times$ Chinchilla), and (4) \textbf{$3.74\text{B}$ active / $63.1\text{B}$ total} ($L = 30$, $D = 1792$, \texttt{4L:1A} \& \texttt{15L:1A}, $k = 8$, $5.03\text{B}$/$30.2\text{B}$ RoPE on $64$ chips and $120.80\text{B}$ NoPE on $512$ chips).
\end{itemize}

\paragraph{Hybrid \texttt{LatentMoE} backbone.}
All four model scales belong to a single hybrid linear-softmax \texttt{LatentMoE} decoder family implemented in open-source MaxText~\cite{MaxText2025Repo}, interleaving $x$ linear delta layers with one gated GQA~\cite{Ainslie2023GQA,Qwen2025GatedAttn} layer per $(x{+}1)$-layer cycle (using RoPE~\cite{Su2024RoPE}, or NoPE with Scalable Softmax~\cite{Nakanishi2025SSMax} at $63.1\text{B}$), with a dense SwiGLU~\cite{Shazeer2020SwiGLU} FFN on Layer $0$ and one shared expert plus $E = 256$ routed latent experts on Layers $1, \dots, L-1$. Model widths stay divisible by $256$ ($D = 1024$ at $6.5\text{B}$ and $12.5\text{B}$; $D = 1792$ at $31.5\text{B}$ and $63.1\text{B}$), while every SLA layer fixes $d_k = 256$ so recurrent state updates never pad systolic tiles. In \texttt{Iso-Param SLA-R4}, widening the shared expert by $+256$ channels compensates for the mixer parameters saved by $4{:}1$ key sharing; complete initialization, normalization, routing, and per-scale widths are given in Appendix~\ref{app:notation_and_arch} (Table~\ref{tab:app_arch_scaling}).

\begin{figure}[ht]
\centering
\includegraphics[width=0.98\textwidth]{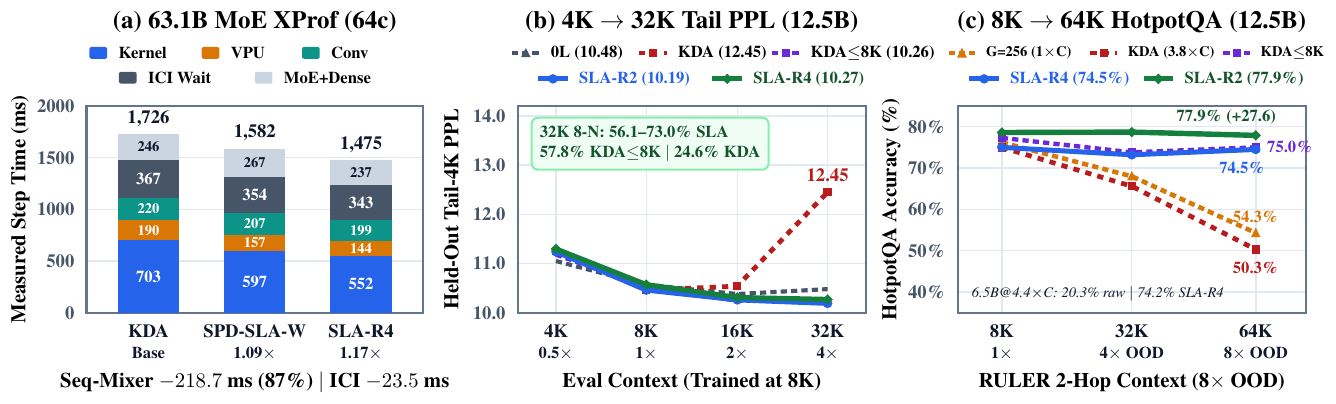}
\caption{\textbf{(a)} Per-step XProf breakdown for $63.1\text{B}$ MoE on $64$ TPU v7x chips across \texttt{KDA Base}, \texttt{SPD-SLA-W (R2)} (\texttt{SLA-R2} with optional in-chunk key whitening, Appendix~\ref{sec:whitening}), and \texttt{SLA-R4} (Table~\ref{tab:app_xprof}); most of SLA's saving is in linear attention ops. \textbf{(b)} Held-out Dolma tail perplexity from $4\text{K}$ to $32\text{K}$ for $12.5\text{B}$ MoEs at $60.4\text{B}$ tokens (curves: seed \texttt{s0}; callout: $3$-seed means from Table~\ref{tab:dolma_15b}a). \textbf{(c)} Zero-shot $8\text{K} \to 64\text{K}$ RULER \texttt{HotpotQA} at $1\times$ and $3.8\times$ Chinchilla tokens (C). Curves are seed means ($3$ seeds; $2$ for $G = 256$; Table~\ref{tab:app_downstream}), except KDA$\le$8K, which clamps the \texttt{s0} KDA model's horizons to $8\text{K}$ without retraining.}
\label{fig:pretrain_xprof}
\end{figure}

\begin{table}[ht]
\caption{\textbf{Multi-scale SLA vs.\ MaxText KDA pre-training speed and quality} across $6.5\text{B}$--$63.1\text{B}$ sparse \texttt{LatentMoE} models ($64$--$512$ TPU v7x chips per run, $T_{\text{train}} = 8{,}192$, \texttt{Exact Inv.}). Complete $45$-row ladder and per-subtask \texttt{lm-eval} breakdowns are in Appendix Tables~\ref{tab:app_clamp}--\ref{tab:app_downstream}.}
\label{tab:v7x_e2e}
\label{tab:moe_kda_sxs}
\centering
\footnotesize
\renewcommand{\arraystretch}{1.08}
\setlength{\tabcolsep}{3.0pt}
\resizebox{\textwidth}{!}{%
\begin{tabular}{@{}llccccccc@{}}
\toprule
Scale \& Horizon & Linear Mixer & Step ms & vs.\ KDA & $11$-Set PPL & \texttt{lm-eval} & $32\text{K}$ Tail & $32\text{K}$ $8$-N & $64\text{K}$ \texttt{SQuAD / Hotpot} \\
\midrule
\multirow{3}{*}{\begin{tabular}[c]{@{}l@{}}\textbf{$6.5\text{B}$ ($0.35\text{B}$ act.)}\\4L:1A, $30.2\text{B}$ tok ($64$c)\end{tabular}}
 & MaxText KDA Base (FP32) & 351.5 & $1.00\times$ & \textbf{16.16} & \textbf{$49.7\%$} & 620.9 & $16.8\%$ & $32.0\% / 20.3\%$ \\
 & \textbf{SLA-R2} (BF16) & \textbf{315.0} & \textbf{$1.12\times$} & 16.25 & $49.5\%$ & 13.53 & $21.1\%$ & $56.2\% / 51.6\%$ \\
 & \textbf{Iso-Param SLA-R4} (BF16) & \textbf{297.9} & \textbf{$1.18\times$} & 16.33 & $49.1\%$ & \textbf{11.09} & \textbf{$45.3\%$} & \textbf{$70.3\% / 74.2\%$} \\
\midrule
\multirow{3}{*}{\begin{tabular}[c]{@{}l@{}}\textbf{$12.5\text{B}$ ($0.79\text{B}$ act.)}\\7L:1A, $15.1\text{B}$ tok ($64$c)\end{tabular}}
 & MaxText KDA Base ($n=3$) & 556.0 & $1.00\times$ & 17.64 & $47.8\%$ & 13.28 & $18.1\%$ & $57.3\% / 56.0\%$ \\
 & \textbf{SLA-R2 Unfused} ($n=3$)$^\ast$ & \textbf{497.3} & \textbf{$1.12\times$} & \textbf{17.53} & \textbf{$48.6\%$} & \textbf{11.23} & $52.3\%$ & \textbf{$69.5\%$} $/ 73.7\%$ \\
 & \textbf{SLA-R2 Fused} ($n=2$)$^\ast$ & \textbf{481.8} & \textbf{$1.15\times$} & 17.81 & $48.2\%$ & 11.36 & \textbf{$58.0\%$} & $69.1\% /$ \textbf{$74.6\%$} \\
\midrule
\multirow{4}{*}{\begin{tabular}[c]{@{}l@{}}\textbf{$12.5\text{B}$ ($0.79\text{B}$ act.)}\\7L:1A, $60.4\text{B}$ tok ($64$c)\end{tabular}}
 & MaxText KDA Base ($n=3$) & 2,225 & $1.00\times$ & 15.60 & $50.4\%$ & 12.45 & $24.6\%$ & $54.9\% / 50.3\%$ \\
 & KDA + \texttt{TrainMaxTau=8K} ($n=2$) & 2,251 & $0.99\times$ & 15.80 & $50.3\%$ & 10.75 & $49.6\%$ & \textbf{$75.8\%$} $/ 75.4\%$ \\
 & \textbf{SLA-R2} (BF16, $n=3$)$^\ast$ & \textbf{1,919} & \textbf{$1.16\times$} & \textbf{15.58} & \textbf{$50.9\%$} & \textbf{10.19} & \textbf{$56.6\%$} & $75.5\% /$ \textbf{$77.9\%$} \\
 & \textbf{Iso-Param SLA-R4} ($n=3$)$^{\ast\dagger}$ & \textbf{1,800} & \textbf{$1.24\times$} & 15.66 & $50.8\%$ & 10.27 & $56.1\%$ & $74.7\% / 74.5\%$ \\
\midrule
\multirow{3}{*}{\begin{tabular}[c]{@{}l@{}}\textbf{$31.5\text{B}$ ($1.80\text{B}$ act.)}\\4L:1A, $5.03\text{B}$ tok ($64$c)\end{tabular}}
 & MaxText KDA Base (FP32) & 807.7 & $1.00\times$ & \textbf{21.23} & \text{--} & 47.16 & $12.9\%$ & \text{--} \\
 & \textbf{SLA-R2 Fused} (BF16) & \textbf{747.2} & \textbf{$1.08\times$} & 21.44 & \text{--} & \textbf{13.88} & $33.2\%$ & \text{--} \\
 & \textbf{SLA-R4 Fused} (BF16) & \textbf{707.4} & \textbf{$1.14\times$} & 21.71 & \text{--} & 14.20 & \textbf{$46.1\%$} & \text{--} \\
\midrule
\multirow{5}{*}{\begin{tabular}[c]{@{}l@{}}\textbf{$63.1\text{B}$ ($3.74\text{B}$ act.)}\\4L:1A, $120.8\text{B}$ tok\\($512$c, NoPE)$^\ddagger$\end{tabular}}
 & MaxText KDA Base (FP32) & 2,165 & $1.00\times$ & 12.60 & \textbf{$57.8\%$} & 353.4 & $15.6\%$ & $27.7\% / 27.0\%$ \\
 & \quad $+\tau_{\max}=8\text{K}$ Clamp $+$ \texttt{SSMax} & \text{--} & \text{--} & 12.60 & \textbf{$57.8\%$} & \textbf{8.04} & $74.2\%$ & \textbf{$90.6\% / 86.5\%$} \\
 & \textbf{SLA-R2} ($0.5\times$ Bytes) & \textbf{1,953} & \textbf{$1.11\times$} & 12.58 & $57.5\%$ & 13.78 & $18.6\%$ & $55.7\% / 56.8\%$ \\
 & \quad $+$Zero-Clamp \texttt{SSMax} & \text{--} & \text{--} & 12.58 & $57.5\%$ & 8.45 & \textbf{$84.4\%$} & \text{--} \\
 & \textbf{SLA-R4} ($0.5\times$ Bytes) $+$ \texttt{SSMax}$^\dagger$ & \textbf{1,878} & \textbf{$1.15\times$} & \textbf{12.45} & $56.7\%$ & 8.41 & $83.2\%$ & $75.6\% / 79.1\%$ \\
\bottomrule
\end{tabular}%
}
\par\vspace{3pt}\raggedright{\scriptsize $^\ast$Paired seed $\Delta\text{PPL}$ vs.\ KDA: $-0.10 \pm 0.08$ ($12.5\text{B}/15.1\text{B}$, $n=3$; fused $n=2$: $17.81$ vs.\ $17.89$), $-0.02 \pm 0.11$ (\texttt{SLA-R2}, $60.4\text{B}$), and $+0.06 \pm 0.26$ (\texttt{SLA-R4}, $60.4\text{B}$). Zero-shot row-scaled FP8 state changes \texttt{lm-eval} by at most $0.25$ pts (below one standard error) across all $20$ evaluated models (Appendix Table~\ref{tab:app_downstream}). $^\dagger$Deferring the fused conv \texttt{AllReduce} across gradient accumulation gives $1{,}812\text{ ms}$ ($12.5\text{B}$) and $1{,}843\text{ ms}$ ($63.1\text{B}$). $^\ddagger$On $512$ chips ($63.1\text{B}$), throughput rises from $15{,}135\text{ tok/s/chip}$ (KDA) to $16{,}779$ (\texttt{SLA-R2}) and $17{,}445$ (\texttt{SLA-R4}); pre-\texttt{SSMax} \texttt{SLA-R4} is $22.78$ Tail PPL, $18.0\%$ $8$-N, and $51.6\% / 45.1\%$ QA (clamp-only KDA without \texttt{SSMax}: $7.70$ Tail, $48.4\%$ $8$-N, $76.6\% / 82.0\%$ QA at $N=128$; all from Appendix Table~\ref{tab:app_clamp}).}
\end{table}

\paragraph{End-to-end MoE training throughput vs.\ MaxText KDA and Tokamax Mosaic KDA.}
We pre-train four token-choice sparse MoE families ($6.5\text{B}$--$63.1\text{B}$ total and $0.35\text{B}$--$3.74\text{B}$ active parameters; $256$ routed experts, top-$8$ routing, one shared expert~\cite{Shazeer2017MoE,Fedus2022Switch}) with AdamW~\cite{Loshchilov2019AdamW} on packed Dolma~3.5~\cite{Soldaini2024Dolma,OLMo2025OLMo2,OLMo2025OLMo3} at $T_{\text{train}} = 8{,}192$, using $64$--$512$ TPU v7x chips per run across a $2{,}048$-chip cluster (Table~\ref{tab:app_4rung}). On $12.5\text{B}$ (\texttt{7L:1A}, $16\text{L}$), \texttt{SLA-R4} achieves a $1.63\times$ end-to-end single-microbatch training speedup over MaxText Pallas KDA ($469.9\text{ ms}$ vs.\ $766.8\text{ ms}$) and a $1.24\times$ speedup under $4$-step gradient accumulation ($1{,}800\text{ ms}$ vs.\ $2{,}225\text{ ms}$ at $8.39\text{M}$ tok/step; $1.26\times$ at $1{,}686.0$ vs.\ $2{,}119.7\text{ ms}$ in XProf Table~\ref{tab:app_xprof_ga4}), and across all matched scales SLA trains $11\%$--$24\%$ ($1.11\times$--$1.24\times$ with $\text{GA}=4$; $1.08\times$ for the short $5\text{B}$-token $31.5\text{B}$ \texttt{SLA-R2} run; up to $1.63\times$ single-microbatch) faster than MaxText KDA (Tables~\ref{tab:v7x_e2e} and~\ref{tab:dolma_15b}), including $1.11\times$--$1.15\times$ at $63.1\text{B}$ on $512$ chips. To benchmark end-to-end $12.5\text{B}$ (\texttt{7L:1A}) step time directly against open-source Tokamax Mosaic KDA (\texttt{tokamax} $0.0.14$, $C = 64$), Appendix Table~\ref{tab:app_e2e_12b_tokamax} evaluates the full $16$-layer $12.5\text{B}$ model ($14$ linear layers $+$ $2$ global GQA layers $+$ $16$ \texttt{LatentMoE} layers with $k = 8$ active experts $+$ AdamW): fused $1$-kernel \texttt{SLA-R4} is $1.68\times$ ($23.61\text{ ms}$, $\mu = 0.25$) to $1.73\times$ ($22.85\text{ ms}$, $\mu = 0$) faster on the $14$-layer linear mixer F+B than Tokamax Mosaic KDA ($39.62\text{ ms}$; $2.65\times$--$2.74\times$ vs.\ MaxText KDA $62.50\text{ ms}$), delivers a $1.24\times$ ($66.01$ vs.\ $82.15\text{ ms}$, $\mu = 0.25$) to $1.26\times$ ($65.24\text{ ms}$, $\mu = 0$) and $1.33\times$ ($61.83\text{ ms}$, $1$-step Neumann) end-to-end $12.5\text{B}$ step speedup over Tokamax Mosaic KDA on single-chip active-parameter steps ($1.64\times$--$1.75\times$ vs.\ MaxText KDA $108.06\text{ ms}$), delivers $1.10\times$--$1.11\times$ ($121.59 / 120.79\text{ ms}$ vs.\ $134.06\text{ ms}$) on $8$-chip \texttt{EP=8} including MoE dispatch/combine all-to-all ($1.87\times$ vs.\ MaxText FP32 $227.00\text{ ms}$), and achieves a $1.14\times$ ($1{,}800$ vs.\ $2{,}049\text{ ms}$ at $\text{GA}=4$) to $1.42\times$ ($469.9$ vs.\ $668.4\text{ ms}$ single-microbatch; $1.18\times$--$1.74\times$ with $1$-kernel $\mu = 0$) end-to-end $12.5\text{B}$ speedup over Tokamax Mosaic KDA on $64$-chip TPU v7x. Under a unified single-protocol $4$-scale benchmark across all four MoE rungs ($6.5\text{B}$, $12.5\text{B}$, $31.5\text{B}$, and $63.1\text{B}$ at $T_{\text{train}} = 8{,}192$, $3$ process restarts; Appendix Table~\ref{tab:app_single_protocol}b), $1$-kernel \texttt{SLA-R4} (\texttt{Exact Inv.}) trains $4.0\%$--$21.4\%$ faster ($1.040\times$--$1.214\times$) than open-source Tokamax Mosaic KDA ($262.2$ vs.\ $272.6\text{ ms}$ at $6.5\text{B}$, $330.4$ vs.\ $348.5\text{ ms}$ at $12.5\text{B}$, $495.2$ vs.\ $572.7\text{ ms}$ at $31.5\text{B}$, and $1{,}015.7$ vs.\ $1{,}232.9\text{ ms}$ at $63.1\text{B}$) and by $1.166\times$--$1.224\times$ relative to MaxText Pallas KDA. At native $T_{\text{train}} = 32\text{K}$ with $4.19\text{M}$ tokens/step ($\text{EP}=128$), SLA fits in per-core HBM ($\le 94.74\text{ GiB}$) while KDA ($96.74\text{ GiB}$) and full attention ($95.48\text{ GiB}$) exceed the $95\text{ GiB}$ cap; in $8\text{K} \to 32\text{K}$ continued pre-training at $12.5\text{B}$, SLA is $1.08\times$--$1.15\times$ faster than KDA within $0.004$--$0.029$ nats in $32\text{K}$ cross-entropy (Table~\ref{tab:app_4rung}). Hardware \texttt{XProf} traces on $64$-chip $63.1\text{B}$ MoE attribute $96.6\%$ of SLA's step-time saving to linear attention ops and reduced ICI wait (Figure~\ref{fig:pretrain_xprof}a; Appendix~\ref{app:xprof}).

\subsection{Pre-Training Quality and Long-Context Extrapolation}
\label{subsec:dolma_15b}
\label{subsec:moe_scaling}

\begin{table}[ht]
\caption{\textbf{Controlled $12.5\text{B}$ MoE ($0.79\text{B}$ active, $16$L, $T_{\text{train}} = 8{,}192$) baselines, factor-isolation ladder (\texttt{A1}--\texttt{A6}), and $90$-min Iso-Wall comparison} at (a) $60.4\text{B}$ tokens ($3.8\times$ Chinchilla~\cite{Hoffmann2022Chinchilla}) and (b) $15.1\text{B}$ tokens ($1.0\times$ Chinchilla; GDN-2 uses our unfused TPU port; seeds \texttt{s0}/\texttt{s1}/\texttt{s2} use distinct data offsets, so PPL is compared on matched seeds).}
\label{tab:dolma_15b}
\centering
\footnotesize
\renewcommand{\arraystretch}{1.05}
\setlength{\tabcolsep}{2.2pt}
\resizebox{\textwidth}{!}{%
\begin{tabular}{@{}llcccccccc@{}}
\toprule
Cohort & Architecture ($7{,}200$ Steps) & Step ms & vs.\ KDA & $11$-Set PPL & $8\text{K}$ Tail & $32\text{K}$ Tail & $32\text{K}$ $8$-N & $64\text{K}$ \texttt{SQuAD / Hotpot} \\
\midrule
\multirow{9}{*}{\begin{tabular}[c]{@{}l@{}}\textbf{(a) v7x}\\$60.4\text{B}$ Tok\\($3.8\times$ Ch.)\end{tabular}}
 & 0L:1A Full-Attn GQA+RoPE (\texttt{s0}) & 2,314 & $0.96\times$ & 16.13 & 10.57 & 10.48 & $12.1\%$ & $58.6\% / 53.1\%$ \\
 & 7L:1A MaxText KDA Base ($n=3$)$^\ast$ & $2{,}225 \pm 29$ & $1.00\times$ & $15.60 \pm 0.58$ & 10.46 & $12.45 \pm 1.11$ & $24.6\%$ & $54.9\% / 50.3\%$ \\
 & \quad $+$Zero-Retrain $\tau_{\max}=8\text{K}$ Clamp (\texttt{s0}) & \text{--} & \text{--} & 16.17 & 10.53 & 10.26 & $57.8\%$ & $\mathbf{78.1\%} / 75.0\%$ \\
 & 7L:1A KDA + \texttt{TrainMaxTau=8K} ($n=2$) & $2{,}251 \pm 11$ & $0.99\times$ & $15.80 \pm 0.33$ & 10.46 & $10.75 \pm 0.21$ & $49.6\%$ & $75.8\% / 75.4\%$ \\
 & \textbf{7L:1A SLA-R2} (BF16, $n=3$)$^\ast$ & $\mathbf{1{,}919 \pm 37}$ & \textbf{$1.16\times$} & $15.58 \pm 0.51$ & 10.46 & $\mathbf{10.19 \pm 0.10}$ & $56.6\%$ & $75.5\% / \mathbf{77.9\%}$ \\
 & \textbf{7L:1A SLA-R4} (Iso-Param, $n=3$)$^{\ast\dagger}$ & $\mathbf{1{,}800 \pm 39}$ & \textbf{$1.24\times$} & $15.66 \pm 0.41$ & 10.57 & $10.27 \pm 0.04$ & $56.1\%$ & $74.7\% / 74.5\%$ \\
 & 7L:1A SLA-R4 ($-20\%$ mixer, \texttt{s0}) & 1,819 & $1.22\times$ & 16.20 & 10.81 & 10.41 & $\mathbf{73.0\%}$ & $72.7\% / 75.0\%$ \\
 & 15L:1A MaxText KDA Base (FP32, \texttt{s0}) & 2,152 & $1.00\times$ & 15.50 & 10.46 & 11.27 & $15.2\%$ & $62.5\% / 60.2\%$ \\
 & \textbf{15L:1A SLA-R2} (Uniform, BF16, \texttt{s0}) & \textbf{1,941} & \textbf{$1.11\times$} & \textbf{15.44} & \textbf{10.43} & \textbf{10.19} & $41.0\%$ & $72.7\% / 70.3\%$ \\
\midrule
\multirow{11}{*}{\begin{tabular}[c]{@{}l@{}}\textbf{(b) $15.1\text{B}$}\\($1.0\times$ Ch.\\\& $90\text{m}$ Iso)\end{tabular}}
 & 7L:1A MaxText KDA Base ($n=3$)$^\S$ & 556.0 & $1.00\times$ & 17.64 (17.89) & 11.57 & 13.28 & $18.1\%$ & $57.3\% / 56.0\%$ \\
 & \textbf{7L:1A SLA-R2} (Unf., \texttt{Exact Inv.}, $n=3$)$^\S$ & \textbf{497.3} & \textbf{$1.12\times$} & \textbf{17.53} (17.75) & 11.54 & \textbf{11.23} & $52.3\%$ & $69.5\% / 73.7\%$ \\
 & \textbf{7L:1A SLA-R2} (Fus., \texttt{Exact Inv.}, $n=2$) & \textbf{481.8} & \textbf{$1.15\times$} & 17.81 & 11.70 & 11.36 & $58.0\%$ & $69.1\% / 74.6\%$ \\
 & 7L:1A GDN-2 (Unf., $93\text{m}, 15.1\text{B}, n=2$) & 778.1 & $0.71\times$ & \textbf{17.59} & \textbf{11.44} & 11.86 & $19.3\%$ & $62.1\% / 62.1\%$ \\
 & \textbf{7L:1A SLA-R2} (Fus., $90\text{m}$ Iso-Wall, $23.1\text{B}$ tok) & \textbf{488.3} & \textbf{$1.14\times$} & 16.66 ($+53\%$ tok) & 10.84 & 10.98 & $52.3\%$ & $65.6\% / 70.3\%$ \\
 & 7L:1A \texttt{A1}: Wide-Key $R{=}2$ KDA ($G{=}256, n{=}2$) & 524.8 & $1.06\times$ & 17.69 & 11.61 & 13.39 & $23.8\%$ & $55.9\% / 54.3\%$ \\
 & 7L:1A \texttt{A2}: $+G=4$ Decay (\texttt{Exact Inv.}, $n=2$) & 514.1 & $1.08\times$ & 17.75 & 11.65 & 11.34 & $46.1\%$ & $70.7\% / 73.8\%$ \\
 & 7L:1A \texttt{A3}: $+1$-Step Neumann Inv.\ ($n=2$) & 497.2 & $1.12\times$ & 17.97 & 11.69 & 11.36 & $53.5\%$ & $71.9\% / 71.9\%$ \\
 & 7L:1A \texttt{A5}: SLA-R2 ($1$-Step, $\mu=0.25$, \texttt{s0}) & \textbf{467.1} & \textbf{$1.19\times$} & 18.02 & 11.87 & 11.45 & $\mathbf{60.9\%}$ & $\mathbf{74.2\% / 83.6\%}$ \\
 & 7L:1A \texttt{A6}: SLA-R2 + FP8 Out-QAT (\texttt{s0}) & 520.5 & $1.07\times$ & 18.01 & 11.80 & 11.37 & $59.8\%$ & $\mathbf{74.2\%} / 75.8\%$ \\
 & \textbf{15L:1A SLA-R2} (Uniform, BF16, $n=3$) & \textbf{486.7} & \textbf{$1.14\times$} & 17.61 & 11.60 & 11.42 & $34.8\%$ & $69.0\% / 65.1\%$ \\
\bottomrule
\end{tabular}%
}
\par\vspace{3pt}\raggedright{\scriptsize $^\ast$Paired $3$-seed $\Delta\text{PPL}$ vs.\ KDA: \texttt{SLA-R2} $-0.02 \pm 0.11$, \texttt{SLA-R4} $+0.06 \pm 0.26$. $^\dagger$Deferred conv \texttt{AllReduce} across \texttt{GA=4}: $1{,}812\text{ ms}$ (Appendix Table~\ref{tab:app_xprof_ga4}). $^\S$Parentheses: \texttt{s0,s1} mean matched to $n=2$ rows (paired $3$-seed \texttt{SLA-R2} vs.\ KDA: $-0.10 \pm 0.08$ PPL, winning $3/3$ seeds). On \texttt{s0}: KDA $18.07$, \texttt{A1} $17.85$, \texttt{A2} $17.87$, fused \texttt{SLA-R2} $17.91$, \texttt{A3} $18.10$, \texttt{A5} $18.02$, \texttt{A6} $18.01$. Omitted label \texttt{A4} is the single-seed \texttt{A3} $+$ BF16-state run (\texttt{s0}: $476.0\text{ ms}$, $1.17\times$; $11.86 / 12.04$ $8\text{K}/32\text{K}$ Tail, $41.0\%$ $8$-N, $58.6\% / 58.6\%$ $64\text{K}$ QA), excluded from the table because its final $11$-set validation log was not archived ($8\text{K}$ full Dolma PPL: $12.23$ vs.\ $12.18$ for \texttt{A3} on \texttt{s0}).}
\end{table}

\paragraph{In-domain quality vs.\ KDA.}
At $12.5\text{B}$ parameters and $60.4\text{B}$ tokens with three paired seeds (Table~\ref{tab:dolma_15b}a), \texttt{SLA-R2} ($-0.02 \pm 0.11$) and \texttt{SLA-R4} ($+0.06 \pm 0.26$) match KDA within seed noise on $11$-set Dolma perplexity and zero-shot \texttt{lm-eval}~\cite{Gao2023LMEval} (Table~\ref{tab:app_downstream}); single-seed $63.1\text{B}$ runs likewise match closely ($12.58$ and $12.45$ vs.\ $12.60$, Table~\ref{tab:v7x_e2e}). At $6.5\text{B}$, \texttt{SLA-R2} trails KDA by $0.20$ perplexity over three seeds ($19.08$ vs.\ $18.88$), and short $5\text{B}$-token runs at $31.5\text{B}$ and $63.1\text{B}$ trail by $0.1$--$0.5$ (Table~\ref{tab:app_4rung}). In the $15.1\text{B}$-token cohort (Table~\ref{tab:dolma_15b}b), seeds \texttt{s0}, \texttt{s1}, and \texttt{s2} use distinct data offsets that shift perplexity by up to $0.9$, so we compare only on matched seeds. Over all three seeds, \texttt{SLA-R2} (\texttt{Exact Inv.}, BF16 state, $\mu = 0.25$) slightly improves on KDA Base on every seed ($17.533$ vs.\ $17.638$; paired difference $-0.10 \pm 0.08$) and matches the same-kernel $G = 4$ FP32 control without momentum ($17.538$ on \texttt{s0..s2}; $17.590$ across $n = 6$, Table~\ref{tab:app_gcurve}).

\paragraph{Factor isolation (\texttt{A1}--\texttt{A6}).}
On matched seeds \texttt{s0,s1} (Table~\ref{tab:dolma_15b}b), widening KDA's keys to $d_k = 256$ with $R = 2$ head sharing (\texttt{A1}, $[4, 256, 512], G = 256$: $524.8\text{ ms}$, $17.694$ PPL) isolates head sharing from decay grouping. Switching from $G = 256$ to $G = 4$ with MXU block-doubling (\texttt{A2}: $514.1\text{ ms}$), adding BF16 state and momentum ($497.3\text{ ms}$ unfused), and fusing into VMEM ($481.8\text{ ms}$) saves $43.0\text{ ms}$ total while staying within $0.06$--$0.11$ PPL of \texttt{A1} ($17.749$--$17.806$) and lifting $32\text{K}$ $8$-Needle from $23.8\%$ to $46.1\%$--$58.0\%$ and $64\text{K}$ \texttt{HotpotQA} from $54.3\%$ to $73.7\%$--$74.6\%$. The optional $1$-step Neumann inverse is the one approximation with a clear quality cost: \texttt{A3} rises to $17.970$ ($+0.22$ over \texttt{A2}), and on \texttt{s0} the fused Neumann arm \texttt{A5} reaches $18.016$ versus $17.913$ for fused \texttt{Exact Inv.}, which is why all flagship pre-training runs use exact inversion (\texttt{Exact Inv.}).

\paragraph{Comparison with GDN-2 and multi-scale equal-chip-hour pre-training vs.\ Tokamax KDA.}
\label{app:gdn2_tiles}
GDN-2~\cite{NVIDIA2026GDN2} decouples the scalar delta gate into channel-wise erase and write gates while keeping per-channel decay. On TPU v7x ($B = 1$, $8\text{K} / 32\text{K}$ F+B), our unfused XLA port of per-channel GDN-2 takes $63.86 / 251.39\text{ ms}$ at $D = 1792$ ($15{,}217\text{ MiB}$ temporary HBM) and $34.15 / 132.83\text{ ms}$ at $D = 2048$ (isolated Mamba-3 MIMO $R = 4$ kernel for reference: $9.27\text{ ms}$), raising end-to-end MoE step time by $1.23\times$--$1.61\times$ ($1.349$ vs.\ $1.093\text{ s}$ on $128$-chip TPU v4; $778.1 \pm 3.9$ vs.\ $481.8\text{ ms}$ for fused \texttt{SLA-R2} on $64$-chip TPU v7x). Per token, GDN-2 reaches the best $8\text{K}$ perplexity in the $15.1\text{B}$-token cohort ($17.592 \pm 0.188$ vs.\ $17.806$ for fused \texttt{SLA-R2} and $17.893$ for KDA on \texttt{s0,s1}; $17.459$ vs.\ $17.913$ on \texttt{s0}; Table~\ref{tab:dolma_15b}b), though its unclamped per-channel gate extrapolates worse ($19.3\%$ vs.\ $58.0\%$ $32\text{K}$ $8$-Needle). Under an equal ${\approx}90$-minute pure-step budget on \texttt{s0} ($5{,}371\text{ s}$ for $11{,}000$ fused \texttt{SLA-R2} steps vs.\ $5{,}582\text{ s}$ for $7{,}200$ GDN-2 steps), SLA processes $53\%$ more tokens and reaches $16.660$ perplexity versus $17.459$. Because this speed gap reflects our unfused TPU port of GDN-2 as much as architecture, we built a single-kernel VMEM-fused \texttt{SLA-R4 + GDN-2} mixer (combining SLA's $R = 4$ wide-key tile-grouped mapping with decoupled erase/write gates) and compared both $1$-kernel fused \texttt{SLA-R4} ($\mu = 0$) and fused \texttt{SLA-R4 + GDN-2} ($\mu = 0$) directly against open-source Tokamax Mosaic KDA ($C = 64$, safe-clipped $g \ge -3.5$ to prevent float32 \texttt{exp2} backward overflow). On single-chip $760\text{M}$--$822\text{M}$ dense LMs ($T = 4\text{K}$ FineWeb-Edu; Appendix Table~\ref{tab:app_eqwall_tokamax}), fused \texttt{SLA-R4} ($1.28\times$) and \texttt{SLA-R4 + GDN-2} ($1.23\times$) beat safe-clipped Tokamax KDA by $-14.45$ ($t = -43.7$) and $-13.70$ ($t = -19.5$) Val PPL at $180\text{ s}$ ($n = 3$) and by $-9.68$ and $-9.93$ at $360\text{ s}$ ($n = 2$). Table~\ref{tab:moe_eqwall_tokamax} scales this comparison to $64$-chip TPU v7x sparse \texttt{LatentMoE} pre-training on packed Dolma~3.5 ($T_{\text{train}} = 8{,}192$, $2.10\text{M}$ tok/step, $\text{EP} = 128$, $n = 3$ paired seeds per arm; per-seed breakdown in Appendix Table~\ref{tab:app_moe_eqwall_detail}). Each run calibrates its steady-state step time after warm-up and executes the exact step budget fitting a fixed post-compile wall-clock window ($40.0\text{ min} = 42.67\text{ chip-h}$ at $12.5\text{B}$; $35.0\text{ min} = 37.33\text{ chip-h}$ at $63.1\text{B}$), while logging matched iso-token $11$-set OLMo v3 validation perplexity along the way. At $12.5\text{B}$ (\texttt{7L:1A}), fused iso-parameter \texttt{SLA-R4} ($455.2\text{ ms/step}$, $1.125\times$, $+13.3\%$ tokens) matches Tokamax KDA within routing noise at step $4{,}000$ ($+0.11 \pm 0.14$ PPL, $p = 0.29$) while fused \texttt{SLA-R4 + GDN-2} ($479.2\text{ ms/step}$, $1.069\times$, $+7.7\%$ tokens) edges both per token ($19.90$ vs.\ $20.05$ and $20.16$); at equal chip-hours, both beat Tokamax KDA on all $3/3$ seeds, with \texttt{SLA-R4} lowering $11$-set PPL by $-0.37 \pm 0.11$ ($t = -6.02$, two-sided $p = 0.027$), raising $9$-task \texttt{lm-eval} by $+0.92 \pm 0.31$ points ($t = +5.16, p = 0.036$), and lowering $32\text{K}$ Dolma tail PPL by $-1.81 \pm 0.68$ ($p = 0.044$), while \texttt{SLA-R4 + GDN-2} reaches $19.11 \pm 0.33$ PPL ($-0.41 \pm 0.35$, $3/3$ seed wins). At $63.1\text{B}$ (\texttt{4L:1A}), fused \texttt{SLA-R4} ($1{,}508.3\text{ ms}$, $1.125\times$, $+12.9\%$ tokens) and \texttt{SLA-R4 + GDN-2} ($1{,}548.7\text{ ms}$, $1.096\times$, $+9.1\%$ tokens) already pull ahead of $g \ge -3.5$ clipped Tokamax KDA by step $1{,}200$ ($-0.82 \pm 0.18$ and $-0.94 \pm 0.22$ PPL, $p = 0.016$ and $0.018$, $3/3$ wins) and widen their equal-chip-hour advantage to $-1.48 \pm 0.21$ PPL ($t = -12.08, p = 0.0068$, $3/3$ wins) and $-1.41 \pm 0.16$ PPL ($t = -15.01, p = 0.0044$, $3/3$ wins), alongside a $-3.35 \pm 0.88$ reduction in $32\text{K}$ tail PPL ($p = 0.022$) and $+16.1$ points on $32\text{K}$ $8$-Needle ($p = 0.050$).

\begin{table}[ht]
\caption{\textbf{Multi-scale equal-chip-hour and iso-token MoE pre-training vs.\ open-source Tokamax Mosaic KDA on $64$-chip TPU v7x} ($128$ TensorCores, $\text{EP} = 128$, $T_{\text{train}} = 8{,}192$, $2.10\text{M}$ tok/step, packed Dolma~3.5, $n = 3$ paired seeds \texttt{s0..s2} per arm, sample std $\text{ddof} = 1$). Step counts are calibrated per run for an equal post-compile training budget ($40.0\text{ min} = 42.67\text{ chip-h}$ at $12.5\text{B}$; $35.0\text{ min} = 37.33\text{ chip-h}$ at $63.1\text{B}$). Per-seed logs and disclosures are in Appendix Table~\ref{tab:app_moe_eqwall_detail}.}
\label{tab:moe_eqwall_tokamax}
\centering
\footnotesize
\renewcommand{\arraystretch}{1.07}
\setlength{\tabcolsep}{2.4pt}
\resizebox{\textwidth}{!}{%
\begin{tabular}{@{}llcccccccc@{}}
\toprule
Scale \& Budget & Linear Mixer ($1$-Kernel \texttt{Exact Inv.})$^\ast$ & \begin{tabular}[b]{@{}c@{}}Measured\\Step $p_{50}$ (ms)\end{tabular} & \begin{tabular}[b]{@{}c@{}}Mean Steps\\(Tokens)\end{tabular} & \begin{tabular}[b]{@{}c@{}}Iso-Token PPL\\(Early / Late)\end{tabular} & \begin{tabular}[b]{@{}c@{}}Eq.-Chip-h\\$11$-Set PPL\end{tabular} & \begin{tabular}[b]{@{}c@{}}Paired $\Delta\text{PPL}$\\($t$, two-sided $p$)\end{tabular} & \begin{tabular}[b]{@{}c@{}}$8\text{K} \to 32\text{K}$\\Tail PPL\end{tabular} & \begin{tabular}[b]{@{}c@{}}$32\text{K}$\\$8$-N\end{tabular} & \begin{tabular}[b]{@{}c@{}}\texttt{norm9} /\\$64\text{K}$ \texttt{Hotpot}\end{tabular} \\
\midrule
\multirow{3}{*}{\begin{tabular}[c]{@{}l@{}}\textbf{(a) $12.5\text{B}$ MoE}\\\texttt{7L:1A}, $D{=}1024$\\$42.67\text{ chip-h}$ ($n{=}3$)\end{tabular}}
 & Tokamax Mosaic KDA ($C{=}64$, clip) & $512.1 \pm 6.6$ ($1.00\times$) & $4{,}857$ ($10.2\text{B}$) & $24.78$ / $20.05$ & $19.52 \pm 0.47$ & ref & $12.75 \to 13.98$ & $34.8\%$ & $46.39\%$ / $54.2\%$ \\
 & \textbf{Fused Iso-Param \texttt{SLA-R4}} ($\mu{=}0$) & $\mathbf{455.2 \pm 6.0}$ ($\mathbf{1.125\times}$) & $5{,}503$ ($11.5\text{B}$) & $25.04$ / $20.16$ & $\mathbf{19.15 \pm 0.44}$ & $\mathbf{-0.37 \pm 0.11}$ ($-6.02, .027$) & $\mathbf{12.50 \to 12.16}$ & $54.3\%$ & $\mathbf{47.31\%}$ / $\mathbf{68.0\%}$ \\
 & \textbf{Fused \texttt{SLA-R4 + GDN-2}} ($\mu{=}0$) & $479.2 \pm 3.7$ ($\mathbf{1.069\times}$) & $5{,}230$ ($11.0\text{B}$) & $\mathbf{24.59}$ / $\mathbf{19.90}$ & $\mathbf{19.11 \pm 0.33}$ & $\mathbf{-0.41 \pm 0.35}$ ($-2.01, .182$) & $12.36 \to 12.15^\dagger$ & $\mathbf{61.7\%}^\dagger$ & $47.10\%^\dagger$ / $\mathbf{69.5\%}^\dagger$ \\
\midrule
\multirow{3}{*}{\begin{tabular}[c]{@{}l@{}}\textbf{(b) $63.1\text{B}$ MoE}\\\texttt{4L:1A}, $D{=}1792$\\$37.33\text{ chip-h}$ ($n{=}3$)\end{tabular}}
 & Tokamax Mosaic KDA ($C{=}64$, clip) & $1{,}697.5 \pm 6.5$ ($1.00\times$) & $1{,}243$ ($2.61\text{B}$) & $36.69$ / $26.73$ & $26.56 \pm 0.32$ & ref & $16.92 \to 19.45$ & $20.0\%$ & $40.98\%^\ddagger$ / $42.6\%^\ddagger$ \\
 & \textbf{Fused Iso-Param \texttt{SLA-R4}} ($\mu{=}0$) & $\mathbf{1{,}508.3 \pm 6.8}$ ($\mathbf{1.125\times}$) & $1{,}403$ ($2.94\text{B}$) & $36.73$ / $\mathbf{25.91}$ & $\mathbf{25.08 \pm 0.44}$ & $\mathbf{-1.48 \pm 0.21}$ ($-12.1, .0068$) & $\mathbf{16.01 \to 16.10}$ & $\mathbf{36.2\%}$ & $\mathbf{42.02\%}^\ddagger$ / -- \\
 & \textbf{Fused \texttt{SLA-R4 + GDN-2}} ($\mu{=}0$) & $1{,}548.7 \pm 2.5$ ($\mathbf{1.096\times}$) & $1{,}357$ ($2.85\text{B}$) & $\mathbf{36.13}$ / $\mathbf{25.78}$ & $\mathbf{25.14 \pm 0.21}$ & $\mathbf{-1.41 \pm 0.16}$ ($-15.0, .0044$) & $16.14 \to 16.37^\ddagger$ & $28.9\%^\ddagger$ & -- \\
\bottomrule
\end{tabular}%
}
\par\vspace{2pt}\raggedright{\scriptsize $^\ast$Iso-token checkpoints are steps $2{,}000 / 4{,}000$ ($4.19\text{B} / 8.39\text{B}$ tok) at $12.5\text{B}$ and steps $600 / 1{,}200$ ($1.26\text{B} / 2.52\text{B}$ tok) at $63.1\text{B}$. Parameter deltas vs.\ Tokamax KDA ($12.53\text{B}$ and $63.13\text{B}$): \texttt{SLA-R4} has $-3.0\text{M}$ ($12.5\text{B}$, shared expert $1024 \to 1280$) and $-14.4\text{M}$ ($63.1\text{B}$, shared expert $3584 \to 4352$) fewer parameters; \texttt{SLA-R4 + GDN-2} has $+1.7\text{M}$ ($12.5\text{B}$, shared expert $1024$) and $-33.5\text{M}$ ($63.1\text{B}$, shared expert $3584$). Tokamax KDA (\texttt{tokamax} $0.0.14$) requires $g \ge -3.5$ ($f \ge 0.030$) to avoid float32 \texttt{exp2} backward NaNs ($0$ non-finite steps across all $18$ runs), which bounds sub-chunk fast forgetting on the baseline. $^\dagger$At $12.5\text{B}$, all $9$ runs completed $100\%$ of training and all iso-token/final $11$-set PPL evals ($n = 3$); for \texttt{SLA-R4} and Tokamax KDA, post-train OOD/downstream/$N{=}512$ $64\text{K}$ QA are $n = 3$ (\texttt{SQuAD} $66.0\%$ vs.\ $53.3\%$, $p = 0.067$), whereas \texttt{SLA-R4 + GDN-2} post-train OOD/downstream reflects seed \texttt{s1} (vs.\ Tokamax \texttt{s1}: $12.62 \to 14.58$ Tail, $54.7\%$ $8$-N, $46.85\%$ \texttt{norm9}, $45.1\%$ \texttt{HotpotQA}). $^\ddagger$At $63.1\text{B}$, all $9$ runs completed $100\%$ of training and all iso-token/final $11$-set PPL evals ($n = 3$); \texttt{SLA-R4} and Tokamax KDA completed $n = 3$ $8\text{K}/32\text{K}$ Dolma \& $32\text{K}$ NIAH (\texttt{SLA-R4 + GDN-2}: $n = 2$, \texttt{s1,s2}, vs.\ Tokamax $16.94 \to 19.89$ and $22.7\%$), and matched seed \texttt{s0} \texttt{norm9} is shown before reservation cutoff (all completed Tokamax \texttt{s0..s2} \texttt{norm9} [$41.22\%$ mean] and \texttt{s0,s1} $64\text{K}$ QA scores are in Appendix Table~\ref{tab:app_moe_eqwall_detail}b).}
\end{table}

\paragraph{Length extrapolation.}
We test zero-shot extrapolation from $8\text{K}$ training to $32\text{K}$--$64\text{K}$ on RULER~\cite{Hsieh2024RULER} needle retrieval and $4$-choice \texttt{SQuAD}~\cite{Rajpurkar2016SQuAD} and \texttt{HotpotQA}~\cite{Yang2018HotpotQA} (Figure~\ref{fig:pretrain_xprof}b--c). Unclamped KDA degrades beyond the training length, severely at $63.1\text{B}$, because about $1\%$ of its channels learn retention horizons ($\tau = -1/\ln f$ tokens for per-token decay $f$) far beyond $8\text{K}$ (Table~\ref{tab:app_clamp}). SLA extrapolates without clamping: at $12.5\text{B}$ its $32\text{K}$ tail perplexity stays flat and $64\text{K}$ \texttt{HotpotQA} reaches $74\%$--$78\%$. The advantage is limited, however. Clamping KDA's horizons to $8\text{K}$, either at test time or during training (\texttt{TrainMaxTau=8K}), gives similar results at $12.5\text{B}$. At $63.1\text{B}$, SLA also needs evaluation-time \texttt{SSMax}~\cite{Nakanishi2025SSMax} on the softmax layers (no retraining; $32\text{K}$ tail perplexity $22.78 \to 8.41$ for \texttt{SLA-R4}, Appendix Table~\ref{tab:app_clamp}), whereas clamped KDA reaches $7.70$ without it. With \texttt{SSMax} on both (scale $\alpha = 0.5$ for SLA and $0.25$ for KDA; one seed each), clamped KDA beats \texttt{SLA-R4} on $64\text{K}$ \texttt{SQuAD}/\texttt{HotpotQA} ($90.6\% / 86.5\%$ vs.\ $75.6\% / 79.1\%$), whereas SLA leads on $32\text{K}$ $8$-Needle retrieval ($83.2\%$--$84.4\%$ with zero gate clamping vs.\ $74.2\%$ clamped KDA and $15.6\%$ unclamped; Table~\ref{tab:v7x_e2e}). With $128$--$512$ questions per QA set (standard error $2$--$4$ points), many differences in Tables~\ref{tab:v7x_e2e}--\ref{tab:dolma_15b} are within noise. Retention-horizon dispersion and clamping audits across scales are in Appendix~\ref{app:causal_clamp} (Table~\ref{tab:app_clamp}). Auditing the public \texttt{Kimi-Linear-48B-A3B} checkpoints confirms this slow-channel tail is intrinsic to per-channel KDA rather than specific to our baseline training recipe (Appendix~\ref{app:r87_controls}, Table~\ref{tab:app_kimi_public}): across all $81{,}920$ KDA channels ($20$ KDA layers), Monte Carlo weight projections already show $0.51\%$--$0.61\%$ of channels with $\tau > 8\text{K}$ ($0.04\%$--$0.05\% > 1\text{M}$), and live $27$-layer forward passes of \texttt{Kimi-Linear-48B-A3B-Instruct} on FineWeb-Edu show $12.1\%$ of channels above $8\text{K}$ ($4.2\% > 64\text{K}$, $0.34\% > 1\text{M}$) at $16\text{K}$. Although long-context pre-training keeps the public checkpoint finite, zero-retraining clamping to $\tau_{\max} = 4{,}096$ still lowers its $16\text{K}$ perplexity from $9.50 \to 9.24$ ($11.34 \to 10.90$ tail-$2\text{K}$, $-0.44$ PPL).

\paragraph{Decay granularity and why $G = 4$ is the quality-speed Pareto knee.}
Table~\ref{tab:app_gcurve} sweeps $G \in \{1, 4, 16, 64, 256\}$ (tile width $b = 256 / G$) on the identical $[4, 256, 512]$ \texttt{Exact Inv.} FP32 kernel ($\mu = 0$), expanding the direct $G = 1$ vs.\ $G = 4$ comparison to $n = 6$ paired seeds (\texttt{s0..s5}; per-seed breakdown in Appendix Table~\ref{tab:app_moe_eqwall_detail}c). On matched \texttt{s0,s1} seeds across all five granularities, $8\text{K}$ $11$-set perplexity favors $G \ge 4$ without a strict monotone trend ($17.946, 17.749, 17.873, 17.748, 17.694$ for $G = 1, 4, 16, 64, 256$; all $G \ge 4$ within $0.18$ PPL). Across all $n = 6$ paired $12.5\text{B}$ MoE seeds, $G = 4$ improves over scalar $G = 1$ on $6/6$ seeds ($\Delta\text{PPL} = -0.34, -0.06, -0.06, -0.66, -0.07, -0.21$; mean $-0.232 \pm 0.235$, $t = -2.42$, two-sided $p = 0.060$, exact two-sided sign test $p = 0.031$), improves $8\text{K}$ Dolma tail PPL on $5/6$ seeds ($11.57$ vs.\ $11.64$, $-0.08 \pm 0.08$, $t = -2.48$, $p = 0.056$), and edges $9$-task \texttt{lm-eval} on $4/6$ seeds ($48.56\%$ vs.\ $48.08\%$, $+0.49 \pm 0.94$ pts) while matching per-channel KDA ($G = d_k$) within $0.055$ PPL on \texttt{s0,s1}. On hardware, $G = 1$ (scalar GDN) runs at the same single-kernel VMEM speed as $G = 4$ ($2.62$ vs.\ $2.85\text{ ms}$ at $8\text{K}$ and $14.16$ vs.\ $14.08\text{ ms}$ at $D = 1792, 32\text{K}$ in Table~\ref{tab:app_isoshape}) because $1$ or $4$ gate vectors both fit in VMEM vector registers without $[B, H_k, T/C, C, C, d_k]$ HBM materialization, whereas per-channel KDA ($G = d_k$) incurs a $1.62\times$--$2.45\times$ XLA slowdown (and $1.04\times$--$1.12\times$ in Pallas chunk kernels; Table~\ref{tab:app_isoshape}). On synthetic zero-shot length extrapolation without clamping, $G = 1$ ($b = 256$, $\tau_{\max} \le 56\text{K}$) reaches higher $32\text{K}$ $4/8$-Needle across $n = 6$ seeds ($70.3\% / 59.6\%$ vs.\ $46.3\% / 35.4\%$, $p = 0.017 / 0.048$, because unclamped $G = 4$ develops a slow $>100\text{K}$ tile on seeds \texttt{s2, s4, s5} while reaching $55.9\%$--$66.0\%$ / $45.7\%$--$61.7\%$ on \texttt{s0, s1, s3}) and $3$--$4$ points higher unclamped $64\text{K}$ QA ($68.2\% / 71.0\%$ vs.\ $64.0\% / 68.1\%$, $p = 0.20 / 0.22$; matched on \texttt{s0,s1} at $67.5\% / 71.2\%$), while finer groups ($G \ge 16$, $b \le 16$, $\tau_{\max} = 131\text{K}$) drop sharply to $19.0\%$--$27.1\%$ $8$-Needle and $50.8\%$--$61.9\%$ QA. Complementing the $n = 6$ MoE runs without routing noise, our $n = 6$ paired-seed dense LM control ($274.7\text{M}$ params, $104.9\text{M}$ FineWeb-Edu tokens, identical fused \texttt{SLA-R4} kernel, $\mu = 0$; Appendix~\ref{app:r87_controls}, Table~\ref{tab:app_g1_g4_mom}) separates $G = 4$ from $G = 1$ on $6/6$ seeds across every metric ($-1.15 \pm 0.49$ Val PPL, $t = -5.78$; $-5.28 \pm 2.61$ OOD PPL, $t = -4.96$; $-1.75 \pm 0.61$ $16\text{K}$ tail PPL, $t = -7.01$), captures $78.5\%$ of the $G = 1 \to G = 256$ in-domain gain, and beats unclamped $G = 256$ at $16\text{K}$ ($46.08$ vs.\ $46.47$ full, $46.32$ vs.\ $47.24$ tail). On a synthetic $2$-layer multi-timescale associative recall diagnostic ($n = 5$ seeds), per-channel $G = 256$ solves the task reliably ($84.1\%$ joint) while $G \in \{1, 2, 4\}$ remain bimodal or underfit, so we treat $G \in \{1, 4\}$ as a zero-overhead architectural knob ($G = 4$ for language-modeling perplexity, $G = 1$ for uniform unclamped synthetic passkey retrieval).

\begin{table}[ht]
\caption{Unconfounded $5$-point $[4, 256, 512]$ decay-granularity sweep on the identical \texttt{chunk\_kda} \texttt{Exact Inv.} FP32 kernel ($\mu = 0$, $12.5\text{B}$ MoE, \texttt{7L:1A}, $15.10\text{B}$ tok), reporting matched \texttt{s0,s1} $11$-set PPL (valid across all $5$ granularities) alongside multi-seed means ($n = 6$ paired seeds \texttt{s0..s5} for $G \in \{1, 4\}$; $n = 3$ \texttt{s0..s2} for $G \in \{16, 64\}$), paired $\Delta\text{PPL}$ vs.\ $G = 1$, $8\text{K} \to 32\text{K}$ Tail PPL, $32\text{K}$ $4/8$-N, \texttt{lm-eval} \texttt{norm9}, and $N = 512$ ($1\text{ SE} = 2.0$ pts) $64\text{K}$ \texttt{RULER} QA.}
\label{tab:app_gcurve}
\centering
\footnotesize
\renewcommand{\arraystretch}{1.08}
\setlength{\tabcolsep}{2.8pt}
\resizebox{\textwidth}{!}{%
\begin{tabular}{@{}lcccccccc@{}}
\toprule
\begin{tabular}[b]{@{}l@{}}$[4, 256, 512]$ \texttt{Exact Inv.}\\FP32 Arm ($\mu = 0$)\end{tabular} & $G$ ($b$) & \begin{tabular}[b]{@{}c@{}}Within-Tile\\Std\end{tabular} & \begin{tabular}[b]{@{}c@{}}\texttt{s0,s1} Mean PPL\\($n$-Seed Mean)$^\ast$\end{tabular} & \begin{tabular}[b]{@{}c@{}}Paired $\Delta\text{PPL}$\\vs.\ $G=1$\end{tabular} & \begin{tabular}[b]{@{}c@{}}$8\text{K} \to 32\text{K}$\\Tail\end{tabular} & \begin{tabular}[b]{@{}c@{}}$32\text{K}$\\$4$-N / $8$-N\end{tabular} & \texttt{norm9} & \begin{tabular}[b]{@{}c@{}}$N=512$ $64\text{K}$\\\texttt{SQuAD / Hotpot}\end{tabular} \\
\midrule
\textbf{Full-Tile Scalar Decay ($n = 6$)} & \textbf{1} ($256$) & \textbf{0.00} & $17.946$ ($17.822 \pm 0.606$) & $0.000$ (ref) & $11.64 \to \mathbf{11.34}$ & $\mathbf{70.3\% / 59.6\%}$ & $48.08\%$ & $\mathbf{68.2 \pm 2.5\% / 71.0 \pm 1.3\%}$ \\
\textbf{Systolic Sub-Tile Default ($n = 6$)} & \textbf{4} ($64$) & \textbf{0.00} & $17.749$ ($\mathbf{17.590 \pm 0.391}$) & $\mathbf{-0.232 \pm 0.235}$ ($6/6$, $t{=}-2.42$) & $\mathbf{11.57} \to 11.43$ & $46.3\% / 35.4\%^\ast$ & $\mathbf{48.56\%}$ & $64.0 \pm 4.6\% / 68.1 \pm 4.3\%^\ast$ \\
Intermediate Sub-Tile ($n = 3$) & 16 ($16$) & 4.34 & $17.873$ ($17.626$) & $-0.064$ ($-0.14, -0.00, -0.05$) & $11.75 \to 12.12 \pm 0.72$ & $41.4\% / 27.1\%$ & $48.35\%$ & $60.3 \pm 4.2\% / 61.9 \pm 5.4\%$ \\
Fine Sub-Tile Decay ($n = 3$) & 64 ($4$) & 5.05 & $17.748$ ($17.523$) & $-0.167$ ($-0.36, -0.04, -0.11$) & $11.53 \to 12.94 \pm 0.23$ & $27.7\% / 19.0\%$ & $48.61\%$ & $53.5 \pm 2.2\% / 50.8 \pm 1.8\%$ \\
Per-Channel Wide-Key ($n = 2$ PPL, $3$ OOD) & 256 ($1$) & 4.78 & $\mathbf{17.694}$ (\texttt{s0,s1}) & $\mathbf{-0.251}$ ($-0.36, -0.14$) & $11.53 \to 12.86 \pm 1.13$ & $41.0\% / 25.1\%$ & $48.54\%$ & $56.8\% / 58.9\%^\dagger$ \\
\bottomrule
\end{tabular}%
}
\par\vspace{2pt}\raggedright{\scriptsize $^\ast$Per-seed $11$-set PPL across \texttt{s0..s5} ($n = 6$, deterministic \texttt{OLMO\_DATA\_STEP\_OFFSET} stream slices $0\text{K}..500\text{K}$): $G=1$ ($18.210, 17.681, 17.178, 18.845, 17.445, 17.572$; \texttt{s0..s2} mean $17.690$), $G=4$ ($17.873, 17.625, 17.117, 18.189, 17.372, 17.363$; \texttt{s0..s2} mean $17.538$; paired $\Delta\text{PPL} = -0.34, -0.06, -0.06, -0.66, -0.07, -0.21$, two-sided $p = 0.060$, sign test $p = 0.031$). For $G \in \{16, 64, 256\}$ on \texttt{s0..s2}: $G=16$ ($18.067, 17.679, 17.132$), $G=64$ ($17.851, 17.644, 17.073$), $G=256$ ($17.850, 17.538$, \texttt{s2} $11$-set not run). On \texttt{s0, s1, s3}, unclamped $G=4$ reaches $55.9\%$--$66.0\%$ / $45.7\%$--$61.7\%$ $32\text{K}$ $4/8$-N and $66.2\%$--$69.0\%$ / $71.1\%$--$72.9\%$ $64\text{K}$ QA, whereas slow tiles ($\tau_{\max} > 100\text{K}$) on \texttt{s2, s4, s5} pull unclamped $32\text{K}$ $4/8$-N to $25.0\%$--$36.3\%$ / $15.2\%$--$19.5\%$ (full per-seed table in Appendix Table~\ref{tab:app_moe_eqwall_detail}c). $^\dagger$For $G=256$, $N=128$ $64\text{K}$ QA is $57.3\% / 59.1\%$.}
\end{table}

\paragraph{FP8 state.}
SLA states quantize well: zero-retrain row-scaled E4M3 state quantization changes $8\text{K}/32\text{K}$ loss by less than $10^{-4}$ nats (Section~\ref{app:pretraining}), and FP8 output QAT with \texttt{Exact Inv.} (\texttt{A6}) is within $0.09$ perplexity of BF16 on \texttt{s0} ($18.006$ vs.\ $17.913$).

\paragraph{Optional low-precision key whitening (\texttt{Polar-2+}).}
In hybrids with sparser softmax refresh such as \texttt{15L:1A}, middle linear layers hold $10\times$--$35\times$ more active key-value associations and develop poorly conditioned chunk key Gram matrices. Appendix~\ref{sec:whitening} introduces \texttt{SPD-SLA-W}, an optional extension that preconditions chunk keys with \texttt{Polar-2+}, a five-matmul BF16-safe inverse-square-root operator ($0.90\text{ ms}$ F+B on TPU v7x, zero NaNs across $524\text{K}$ tokens; Table~\ref{tab:whitening_kernels}). Because unwhitened SLA already matches or slightly improves on \texttt{SPD-SLA-W} in general language modeling by $60.4\text{B}$ tokens at lower step time, SLA remains unwhitened by default and reserves \texttt{SPD-SLA-W} for correlated key-overwrite workloads ($60.3\text{ dB}$ BF16 recall SNR; Appendix~\ref{app:gates}) and aggressive \texttt{15L:1A} hybrids (Table~\ref{tab:app_4rung}).

\paragraph{Shared-$\beta_t$ isolation, open-source solver audits, and exact vs.\ approximate inversion.}
Four targeted TPU v7x control studies (Appendix~\ref{app:v2_solver_audit}, Tables~\ref{tab:app_tokamax_audit}--\ref{tab:app_pretrain_solvers}) verify our solver attribution and design choices. First, a direct Tokamax input-precision audit confirms that BF16 QKV inputs activate Tokamax's MXU Neumann--Horner path, running $1.68\times$--$2.25\times$ faster in core forward ($2.41\times$--$2.58\times$ core F+B; $1.63\times$--$1.71\times$ full-layer F+B) than FP32 sequential substitution while matching outputs within $4.14\times 10^{-3}$--$4.19\times 10^{-3}$ relative $L_2$ error; Table~\ref{tab:v7x_kernels} already supplied BF16 QKV inputs to Tokamax, so its ratios measure SLA against Tokamax's fast MXU path. Second, against FP64 on real Qwen3.5-$0.8\text{B}/2\text{B}$ chunks and high-coherence $G = 4$ SLA chunks ($\beta_t \in (0, 2)$), FP32 block-doubling holds $1.4\times 10^{-7}$--$3.0\times 10^{-7}$ worst-case relative error from $C = 32$ to $C = 256$, whereas BF16 blocked Neumann--Horner reaches $10.6$--$12.7$ relative error at $C = 128$ ($16.7\%$--$17.4\%$ of Qwen3.5 chunks above $0.1$) and unblocked Neumann products diverge on $21.5\%$--$23.3\%$ of Qwen3.5 chunks and $100\%$ of high-coherence $C \ge 128$ chunks (Table~\ref{tab:app_solver_numerics}). Third, holding the layer shape fixed at $H_k = 2, d_k = 256, H_v = 8, G = 4, C = 128$ (\texttt{Exact Inv.}), sharing $\beta_t$ across the four packed value heads ($2$ solves vs.\ $8$ solves per chunk) accelerates core forward by $2.45\times$--$3.28\times$, core F+B by $2.25\times$--$3.00\times$, and full-layer F+B by $1.87\times$--$2.48\times$ across $D \in \{1792, 2048\}$ and $T \in \{8\text{K}, 32\text{K}\}$ (with $0.0$ output difference under equal gates), and speeds up $223\text{M}$-parameter $3$-seed pre-training by $1.68\times$--$1.69\times$ ($72.9$ vs.\ $122.4$--$123.4\text{ ms/step}$; validation perplexity $74.24 \pm 3.56$ vs.\ $72.66 \pm 1.45$ and $72.89 \pm 1.89$, beating the per-head control on $2/3$ seeds; Table~\ref{tab:app_shared_beta}). Fourth, once values are packed, exact inversion adds only $4.0\%$--$7.7\%$ to full-LM step time, gives the lowest mean perplexity and tightest seed spread at $204\text{M}$ ($69.94 \pm 1.15$, consistent with the $+0.22$ PPL cost of $1$-step Neumann at $12.5\text{B}$), and compiles cleanly at $C = 256$ once the lowering avoids the Mosaic \texttt{kVectorPack} assertion (Table~\ref{tab:app_c256_check}). For zero-shot inference on exact-trained Qwen3.5 weights, banded exact $16 \times 16$ solvers match exact $N = 1{,}425$ MMLU within $0.00$--$0.42$ points (KL $0.0029$--$0.0034$), whereas matmul-free $1$-step operators require healing and still lose $4.6$--$16.8$ MMLU points, keeping exact inversion as SLA's default (Appendix~\ref{app:v2_solver_audit}).

\section{Related Work}
\label{sec:related}

\paragraph{Decay Granularity in Linear Recurrences.}
Linear attentions replace the growing softmax KV cache with a fixed-size state updated once per token~\cite{Katharopoulos2020,Choromanski2021Performer}. Using the decay-group view of Section~\ref{subsec:gdn_prelim}, most follow-up models differ in how finely their forget gate is resolved across key channels: RetNet~\cite{Sun2023RetNet}, Mamba-2~\cite{Dao2024Mamba2}, and GDN~\cite{Yang2025GDN} use one decay rate per head, whereas GLA~\cite{Yang2024GLA}, RWKV-6/7~\cite{Peng2024RWKV6,Peng2025RWKV7}, KDA~\cite{Moonshot2025KimiLinear}, and GDN-2~\cite{NVIDIA2026GDN2} give every key channel its own rate. Mamba~\cite{Gu2023Mamba}, HGRN2~\cite{Qin2024HGRN2}, xLSTM~\cite{Beck2024xLSTM}, and Mamba-3~\cite{Gu2026Mamba3} explore related state and gating structures. SLA takes a middle point with four decay rates per $256$-channel head. This is coarse enough that the decay factors out of $64$-channel MXU matmuls. In our sweeps it lowers perplexity relative to one rate per head across all six paired $12.5\text{B}$ MoE and dense-LM seeds while trailing on unclamped synthetic $32\text{K}$ needle retrieval ($35.4\%$ vs.\ $59.6\%$), so one rate per head remains a strong zero-overhead alternative (Section~\ref{subsec:dolma_15b}).

\paragraph{Delta Rules, Higher-Order Memories, and Momentum.}
DeltaNet~\cite{Schlag2021DeltaNet,Yang2024DeltaNet}, test-time-trained memories~\cite{Sun2024TTT,Behrouz2025Titans,Behrouz2025Atlas}, and the MIRAS online-optimization framework~\cite{Behrouz2026MIRAS} view the state update as online regression with retention regularization, which improves associative recall over decay-only models~\cite{Arora2023Zoology,Jelassi2024RepeatAfterMe}. Later work adds multiple Householder reflections~\cite{Siems2025DeltaProduct}, Householder-diagonalized decay in HDLA~\cite{Zhang2026HDLA}, negative eigenvalues~\cite{Grazzi2025NegEig}, decoupled per-channel erase and write gates~\cite{NVIDIA2026GDN2}, second-order preconditioning~\cite{VonOswald2025MesaNet,Tumma2026PDN}, log-linear state hierarchies~\cite{Guo2025LogLinear}, and heavy-ball momentum~\cite{Polyak1964,Nguyen2022MomentumTransformer,Behrouz2025Titans}. Concurrent work, Muir~\cite{Gangavarapu2026Muir}, introduces sensitivity-allocated low-rank erase and write gates (reinvesting saved parameters in wider FFNs), a matched-filter readout gate that re-weights slow-decaying large-norm state rows, and a numerically stable large-chunk parallel update tailored to wide matrix units. Muir reports $1.8\%$--$2.0\%$ lower perplexity than GDN-2 across $166\text{M}$--$1.34\text{B}$ unpacked-head models and streaming extrapolation from $4\text{K}$ training up to $10\text{M}$ tokens; we do not run a head-to-head comparison against Muir in this work, and its gate allocation and readout selectivity are complementary to SLA's $4{:}1$ head sharing, tile-grouped decay, and single-kernel VMEM fusion across $6.5\text{B}$--$63.1\text{B}$ MoE models. More broadly, these algorithmic extensions enrich what the state update can express, whereas SLA keeps GDN's single-reflection update and reshapes it around systolic hardware. The two directions are complementary: GDN-2 reaches lower per-token perplexity than SLA in our $15.1\text{B}$-token MoE runs, and combining GDN-2's decoupled gates with SLA's systolic mapping inside one VMEM-fused kernel (\texttt{SLA-R4 + GDN-2}) beats open-source Tokamax KDA at equal chip-hours across $3/3$ paired seeds at both $12.5\text{B}$ and $63.1\text{B}$ scales (Table~\ref{tab:moe_eqwall_tokamax}).

\paragraph{Sharing Key Heads and Write Gates Across Value Heads.}
GQA~\cite{Ainslie2023GQA} shares key and value heads across query heads in softmax attention. For linear attention, Kimi Linear~\cite{Moonshot2025KimiLinear}, Qwen3-Next~\cite{Qwen2025Qwen3Next}, and \texttt{fla}~\cite{FLA2024Repo} share query-key heads across value heads at $d_k = 128$, while Gated Attention~\cite{Qwen2025GatedAttn} establishes sigmoid output gating; however, these layers keep independent decay and write gates $(g, \beta_t)$ per value head and still execute $H_v$ chunk solves per chunk. SLA widens keys to $d_k = 256$ and shares both key heads and $\beta_t$ across $R$ packed value heads (keeping per-value-head normalization), filling whole MXU tiles while cutting chunk solves by $R\times$.

\paragraph{Hardware-Aware Kernels and MXU Chunk Solvers.}
FlashAttention~\cite{Dao2022FlashAttention,Dao2023FlashAttention2,Dao2026FA4} re-architects softmax attention around the GPU memory hierarchy. For linear recurrences, \texttt{fla}~\cite{FLA2024Repo} provides chunkwise WY kernels~\cite{Bischof1987WY}, TFLA~\cite{Beck2025TFLA} introduces a second level of parallelism inside each chunk so linear RNN and xLSTM kernels scale to large chunk sizes, and FlexLinearAttention~\cite{Duanmu2026FlexLA} compiles a unified three-phase chunk abstraction on GPU tensor cores, while on TPUs and NPUs FLAT~\cite{Mishra2026FLAT} fuses Mamba-2 and GLA in Pallas, Sobczyk et al.~\cite{Sobczyk2026InvariantLinearAttn} analyze matmul-rich direct and iterative triangular inversion for delta-rule layers on NPUs, and open-source MaxText and Tokamax~\cite{MaxText2025Repo,Tokamax2025Repo} implement TPU GDN and KDA kernels that already move individual chunk solves onto the MXU (BF16 $8$-token Neumann plus Horner doubling with FP32 accumulation in Tokamax, validated at $C = 64$ for per-channel KDA alongside an FP32 sequential fallback and an inference op fusing 1D causal convolution with GDN/KDA state updates, and XLA block-doubling in MaxText's GDN path alongside our block-doubling Pallas KDA baseline). These systems optimize kernels for a fixed recurrence (with Tokamax's convolution fusion targeting single-step inference); SLA also co-designs the training recurrence itself ($R\times$ fewer solves via shared $\beta_t$, $G = 4$ tile-grouped decay, and full chunkwise WY forward-backward VMEM fusion with custom-VJP inverse reuse; distinct from works that abbreviate sparse-linear or softmax-linear hybrids as SLA).

\paragraph{Context Parallelism for Linear Recurrences.}
Head-sharded Ulysses~\cite{Jacobs2023Ulysses} requires at least as many key heads as devices, ring-based methods~\cite{Liu2023RingAttn,Sun2024LASP,Qin2024Lightning} pass the state through $\mathcal{O}(P)$ sequential hops, and LASP-2~\cite{Sun2025LASP2} gathers chunk states with one AllGather. Because every gated delta chunk transition with scalar $\beta_t$ is affine, chunks can also be combined with a parallel prefix scan~\cite{Blelloch1990Prefix,Martin2018ParallelLinearRNN}; we use this standard construction for both SLA and KDA (Section~\ref{subsec:sfa_cp}).

\section{Conclusion}
\label{sec:conclusion}

\paragraph{Summary.}
Recent gated-delta models were shaped around GPUs, where unpacked heads, per-channel gate rescaling, and per-value-head chunk solves (including sequential in-block fallbacks) are cheap relative to matmuls. On systolic accelerators, where matrix units outpace vector lanes by two orders of magnitude, kernel tuning or MXU chunk solves alone cannot hide an unpacked, vector-bound recurrence. Systolic Linear Attention (SLA) resolves this mismatch by co-designing the recurrence around the systolic array: sharing $256$-wide keys and write gates $\beta_t$ across value heads fills full tiles and cuts chunk solves by $R\times$, grouping forget gates into four tiles factors decay out of matmuls, and single-kernel VMEM fusion builds on open-source MXU block-doubling solvers~\cite{MaxText2025Repo,Tokamax2025Repo} to keep exact chunk inversion and its backward pass on-chip. Because value packing already makes exact inversion cheap ($4\%$--$8\%$ of step time), exact FP32 block-doubling stays optimal for pre-training, whereas zero-shot serving on exact-trained weights should use exact or banded-exact inversion rather than unhealed $1$-step approximations (Appendix~\ref{app:v2_solver_audit}). On TPU v7x, a $2$-stage \texttt{SLA-R4} layer with exact inversion runs $2.1\times$--$2.6\times$ faster than open-source MaxText KDA and $1.4\times$--$2.4\times$ faster than open-source Tokamax KDA ($1$-kernel fused \texttt{SLA-R4}: $2.5\times$--$2.8\times$ vs.\ MaxText and $1.5\times$--$2.4\times$ vs.\ Tokamax under single-chip protocols), with $49\%$--$61\%$ less temporary HBM; building a fused per-channel $G = d_k$ KDA kernel confirms that $4{:}1$ head packing ($1.28\times$--$1.41\times$), VMEM fusion ($1.10\times$--$1.12\times$, which actually regresses to $0.44\times$--$0.69\times$ without packing), and $G = 4$ tile grouping ($1.07\times$--$1.20\times$) each contribute multiplicatively (Appendix~\ref{app:r87_controls}). Across $6.5\text{B}$--$63.1\text{B}$ MoE models on up to $512$ chips per run, SLA trains $11\%$--$24\%$ faster than MaxText KDA ($1.08\times$ on the short $5\text{B}$-token $31.5\text{B}$ \texttt{SLA-R2} run; $1.14\times$--$1.24\times$ for \texttt{SLA-R4}) and $4\%$--$21\%$ faster end-to-end than open-source Tokamax Mosaic KDA under the unified $4$-scale protocol (Appendix Table~\ref{tab:app_single_protocol}b) at comparable per-token perplexity. Under controlled multi-seed equal-chip-hour pre-training on Dolma~3.5 across $64$-chip TPU v7x slices ($n = 3$ paired seeds per arm; Table~\ref{tab:moe_eqwall_tokamax}), $1$-kernel fused iso-parameter \texttt{SLA-R4} ($\mu = 0$) and fused \texttt{SLA-R4 + GDN-2} beat safe-clipped open-source Tokamax Mosaic KDA on all $3/3$ seeds at both $12.5\text{B}$ ($-0.37 \pm 0.11$ PPL, $t = -6.02, p = 0.027$, and $+0.92 \pm 0.31$ \texttt{lm-eval} points, $p = 0.036$, for \texttt{SLA-R4}; $-0.41 \pm 0.35$ PPL for \texttt{SLA-R4 + GDN-2}) and $63.1\text{B}$ ($-1.48 \pm 0.21$ PPL, $t = -12.08, p = 0.0068$, and $-1.41 \pm 0.16$ PPL, $t = -15.01, p = 0.0044$). Extended TPU v7x control studies (Appendix~\ref{app:r87_controls}) validate our analytical roofline out of sample ($18$ held-out shapes, zero-parameter Spearman $\rho = 0.893$, $13.9\%$ packed-regime MAPE, $+3.5\%$ error on held-out $(R = 8, d_k = 512, C = 256, d_v = 256)$), confirm that public \texttt{Kimi-Linear-48B-A3B} checkpoints exhibit the same slow-channel decay tail ($>1\text{M}$-token channels) and improve under zero-retrain clamping ($-0.44$ $16\text{K}$ tail PPL), evaluate $G = 4$ vs.\ $G = 1$ across $6/6$ paired seeds on both dense LMs ($t = -5.78$ Val / $t = -7.01$ $16\text{K}$ tail) and $12.5\text{B}$ MoE ($-0.23 \pm 0.24$ PPL, $6/6$ seed wins, sign-test $p = 0.031$), and establish $\mu = 0$ as SLA's recommended default with an optional Schur-stable guard ($5 \to 0$ unstable grid points).

\paragraph{Limitations and Code Availability.}
SLA targets accelerators with wide systolic tiles and a large matrix-to-vector throughput gap; on GPUs, smaller Tensor Core tiles and faster vector lanes already saturate narrow states. Several empirical boundaries warrant transparency: (1) our analytical roofline achieves $13.9\%$ calibrated MAPE inside the packed regime ($R \ge 4, d_k \ge 256$), whereas unpacked $R = 1$ shapes run $42\%$--$53\%$ slower than the single-$\eta$ packed extrapolation due to unmodeled VPU and scheduling overhead; (2) at fixed token count on $12.5\text{B}$ MoE, BF16 routing variance (${\approx}0.18$ PPL) yields per-token parity rather than a large per-token gap (though $G = 4$ beats $G = 1$ on $6/6$ paired $12.5\text{B}$ seeds by $-0.23 \pm 0.24$ PPL while $G = 1$ leads synthetic $32\text{K}$ passkey retrieval by $+24.3$ points, and on synthetic $2$-layer multi-timescale associative recall $G = 256$ outperforms $G \in \{1, 2, 4\}$; Tables~\ref{tab:app_gcurve} and~\ref{tab:app_g1_g4_mom}), and the $120.8\text{B}$-token $512$-chip $63.1\text{B}$ horizon runs remain single-seed; (3) zero-retraining gate clamping closes much of KDA's $8\text{K} \to 64\text{K}$ gap (and with \texttt{SSMax}, clamped $63.1\text{B}$ KDA beats \texttt{SLA-R4} on $64\text{K}$ QA while trailing on $32\text{K}$ $8$-Needle); (4) in our multi-seed equal-chip-hour comparisons against open-source Tokamax Mosaic KDA (Tables~\ref{tab:moe_eqwall_tokamax}, \ref{tab:app_eqwall_tokamax}, and~\ref{tab:app_moe_eqwall_detail}), Tokamax \texttt{0.0.14} required clipping $g \ge -3.5$ ($f \ge 0.030$) to prevent float32 \texttt{exp2} backward overflow, which bounds ultra-fast forgetting on the baseline, and two $12.5\text{B}$ \texttt{SLA-R4 + GDN-2} seeds plus late-stage $63.1\text{B}$ $64\text{K}$ RULER jobs exited after saving all in-loop validation checkpoints; and (5) chunk-boundary momentum introduces chunk-size dependence, doubles context-parallel transition size to $512 \times 512$, requires the $64\text{ MiB}$ scoped-VMEM cap at $D = 1792$, and showed no isolated perplexity gain over $\mu = 0$ on $3$ paired dense-LM seeds, so $\mu = 0$ is our recommended default while historical $12.5\text{B}/63.1\text{B}$ MoE runs used $\mu = 0.25$. Pallas kernels, verification tests, and evaluation scripts will be released publicly upon publication to enable full reproduction on TPU v7x.

\paragraph{Future Work.}
Extending our multi-seed $12.5\text{B}$ and $63.1\text{B}$ equal-chip-hour \texttt{SLA-R4} and \texttt{SLA-R4 + GDN-2} results to full $100\text{B}+$-token horizons and integrating concurrent sensitivity-allocated gating designs such as Muir~\cite{Gangavarapu2026Muir} onto SLA's systolic head-packing and tile-grouped mapping remain direct next steps. Scaling multi-seed long-context evaluations to larger $64\text{K}$--$128\text{K}$ benchmarks and testing exact block-doubling at $C = 256$ in full MoE pre-training runs also remain open directions.

\clearpage
\bibliographystyle{abbrvnat}
\bibliography{references}

\newpage
\appendix
\section{Conditioning Recurrent State in Low Precision: Oblique Whitening and \texttt{Polar-2+}}
\label{sec:whitening}
\label{subsec:capacity_law}
\label{subsec:polar2}

\begin{figure}[ht]
\centering
\includegraphics[width=0.92\textwidth]{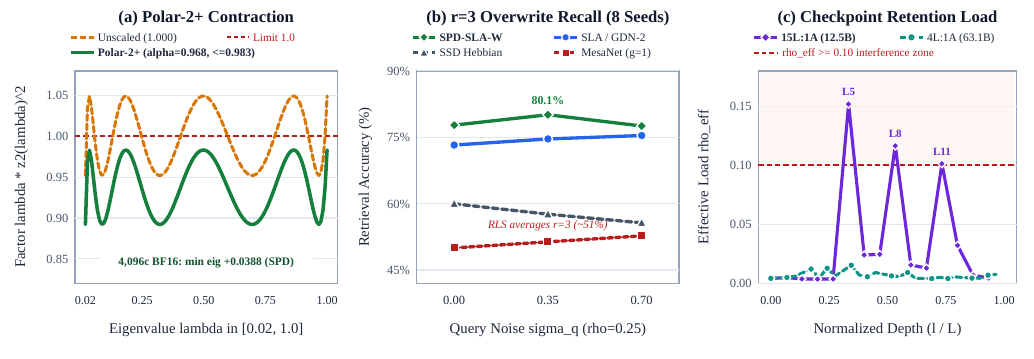}
\caption{\textbf{Polar-2+ minimax contraction, overwrite recall, and checkpoint retention audit.} \textbf{(a)} Scalar inverse-square-root contraction $\lambda z_2(\lambda)^2$ on normalized eigenvalue $\lambda \in [0.02, 1.0]$ (Proposition~\ref{prop:polar2_spd}): unscaled polynomial factor $\alpha = 1.000$ hits $1.0492 > 1$, whereas \texttt{Polar-2+} (safety factor $\alpha_{\text{safe}} = 0.968$) bounds the peak to $0.9831 < 1$, keeping minimum eigenvalue $\lambda_{\min}(\Sigma_c) \ge 0.0388$ of key preconditioner $\Sigma_c$ across $4{,}096$ BF16 chunks ($524\text{K}$ tok). \textbf{(b)} $8$-seed Hadamard overwrite accuracy (overwrite multiplicity $r = 3$) at memory load $\rho_{\text{eff}} = N / d_k = 0.25$ ($N = 16$ stored keys, $d_k = 64$). \textbf{(c)} Trained $63.1\text{B}$ (\texttt{4L:1A}) and $12.5\text{B}$ (\texttt{15L:1A}) checkpoint load $\rho_{\text{eff}}$.}
\label{fig:whitening_gram}
\end{figure}

This appendix details \texttt{SPD-SLA-W}, an optional key-whitening extension for aggressive \texttt{15L:1A} hybrids summarized at the end of Section~\ref{subsec:dolma_15b} (not used in the core SLA benchmarks of Section~\ref{sec:eval}). Reducing softmax frequency from \texttt{7L:1A} ($12.5\%$) to \texttt{15L:1A} ($6.25\%$) cuts $131\text{K}$ cache+state HBM to $16.47\text{ GiB}$ (Table~\ref{tab:cp_serving}), which forces middle linear layers (\texttt{L5, L8, L11}) to hold $10\times$--$35\times$ more active key-value associations $N_{\text{eff}}$: the memory load $\rho_{\text{eff}} = N_{\text{eff}} / d_k$ rises to $[0.101, 0.152]$ versus $\le 0.015$ at \texttt{4L:1A}, while chunk key Gram matrices become poorly conditioned ($\kappa_2(I_C + K_c K_c^\top) \in [20.6, 62.9]$; Figure~\ref{fig:whitening_gram}c). Whereas GPU second-order rules rely on $30$-step Conjugate Gradient (CG)~\cite{VonOswald2025MesaNet} or axis-aligned diagonal scaling~\cite{Tumma2026PDN}, a $256 \times 256$ MXU executes five $[128, 128]$ matmuls across $64$ chunks in $0.62\text{ ms}$ forward ($0.90\text{ ms}$ F+B). Below we analyze when key whitening improves BF16 recall and introduce \textbf{Polar-2+}, a five-matmul BF16-safe inverse-square-root primitive.

\begin{proposition}[Oblique Whitening vs.\ Euclidean \& RLS]
\label{prop:whitening}
Let $K \in \R^{N \times d_k}$ stack $N \le d_k$ unit keys $k_1, \dots, k_N \in \R^{d_k}$ with mutual coherence $\nu_{\text{coh}} = \max_{i \neq j} |\langle k_i, k_j\rangle|$ and Gram matrix $\mathcal{G} = K K^\top$ satisfying $\kappa_2(\mathcal{G}) > 1$ (proofs in Section~\ref{app:whitening_proofs}).
\begin{enumerate}[leftmargin=1.6em,itemsep=2pt]
\item \priortag{Prior Delta \cite{Yang2024DeltaNet,Tumma2026PDN}} For keys with coherence $\nu_{\text{coh}} > 0$, single-pass Euclidean delta lets each later write perturb the readout of earlier keys in proportion to their key overlap, and diagonal column-energy preconditioning $\diag(p(K))^{-1/2}k_t$ \cite{Tumma2026PDN} (where $p(K) \in \R^{d_k}$ has $m$-th entry $p_m(K) = \sum_{i=1}^N K_{i,m}^2$) fails under orthogonal key rotation $K \mapsto K Q$ ($Q \in \mathrm{O}(d_k)$).
\item \priortag{Prior RLS \cite{VonOswald2025MesaNet}} MesaNet (with recursive least-squares key covariance $H_t = \gamma H_{t-1} + k_t k_t^\top$ and scalar forget factor $\gamma \in (0, 1]$) averages $r \ge 2$ overwrites $v^{(1)}, \dots, v^{(r)} \in \R^{d_v}$ of the same key as $\frac{1}{r}\sum_{m=1}^r v^{(m)}$ at $\gamma = 1$, and suppresses stale values at $\gamma < 1$ only by decaying write-once keys stored $t - j$ steps earlier (at step $j < t$) at rate $\gamma^{t - j}$.
\item \textbf{[Ours]} Oblique whitened keys $\tilde{K}_c \in \R^{C \times d_k}$ and causal oblique keys $\tilde{k}_t \in \R^{d_k}$ (Eq.~\eqref{eq:oblique_key}) inside Eq.~\eqref{eq:sfa_intra} decorrelate intra-chunk writes while erasing selectively along $\tilde{k}_t$.
\end{enumerate}
\end{proposition}

\paragraph{Non-circular normalization and \texttt{Polar-2+} formulation.}
Matrix-sign polynomials (Polar Express \cite{Amsel2025PolarExpress}, CANS \cite{Grishina2025CANS}) collapse singular values to $1$, while coupled \texttt{NS-5} \cite{Higham2008MatrixFunctions,Anil2020Shampoo,Gupta2018Shampoo,Liu2025Muon} uses $10$ $[C, C]$ matmuls \priortag{Prior Polynomials}. Here $K_c \in \R^{C \times d_k}$ stacks the chunk's keys, $\Sigma_{c-1} \succ 0$ is the $d_k \times d_k$ key preconditioner carried across chunks (Eq.~\eqref{eq:woodbury_decay}; initialized at $\Sigma_0 = I_{d_k}$ and relaxed toward $I_{d_k}$ at rate $\rho_p \in [0, 1]$), $\eta_c \in (0, 1]$ is the whitening step size, and $\varepsilon_\beta > 0$ is the erasure safety margin (Section~\ref{subsec:recurrence}). To avoid circular regularization, we form the raw chunk Gram matrix $\mathcal{G}_0 \in \R^{C \times C}$, its Frobenius normalization $\bar{\mathcal{G}}_0$, spectral upper bound $\hat{u}_0$, and adaptive ridge $\varepsilon_{\text{reg}}$:
\[
\mathcal{G}_0 = I_C + \eta_c K_c \Sigma_{c-1} K_c^\top,
\qquad
\bar{\mathcal{G}}_0 = \frac{\mathcal{G}_0}{\norm{\mathcal{G}_0}_F},
\qquad
\hat{u}_0 = \max\!\big(\norm{\mathcal{G}_0}_F \norm{\bar{\mathcal{G}}_0^2}_F^{1/2},\, 1\big) \ge \lambda_{\max}(\mathcal{G}_0),
\]
and
\[
\varepsilon_{\text{reg}} = \max\!\left(0.02,\; \frac{0.0205}{0.9795}\hat{u}_0 - 1\right).
\]
Setting regularized Gram $\mathcal{G}_{\text{reg}} = \mathcal{G}_0 + \varepsilon_{\text{reg}} I_C$ and scale $u_0 = \hat{u}_0 + \varepsilon_{\text{reg}}$ maps the eigenvalues of $T_0 = \mathcal{G}_{\text{reg}} / u_0$ into $[0.02, 1.0]$ while reusing $\bar{\mathcal{G}}_0^2$ ($T_0^2$ adds zero matmuls; $5$ matmuls total), with step-size clamp
\[
\beta_t \le \frac{2 - \varepsilon_\beta}{\max(k_t^\top \tilde{k}_t,\, \varepsilon_\beta)}.
\]
Letting $K_{<t}$ stack the chunk keys strictly preceding token $t$, $\mathcal{M}_\ell \in \R^{C \times C}$ denote the degree-$2$ polynomial factors for stages $\ell \in \{0, 1\}$ with $T_1 = \mathcal{M}_0 T_0 \mathcal{M}_0$ and minimax coefficients $(a_0, b_0, e_0) = (4.92864, -11.29513, 7.70099)$ and $(a_1, b_1, e_1) = (2.00410, -1.38700, 0.38894)$, and $\mathcal{W}_{\text{isq}} \in \R^{C \times C}$ denote the safety-scaled ($\alpha_{\text{safe}} = 0.968$) inverse-square-root factor, \textbf{Polar-2+} computes:
\contribbox{Low-Precision Conditioning: Oblique Whitening \& BF16-Safe \texttt{Polar-2+}}{%
\begin{align}
\tilde{K}_c &= \sqrt{\eta_c}\, \mathcal{W}_{\text{isq}} K_c \Sigma_{c-1},
\qquad
\tilde{k}_t \propto \big(I_{d_k} - \eta_c \Sigma_{c-1} K_{<t}^\top K_{<t}\big) \Sigma_{c-1} k_t, \label{eq:oblique_key}\\
\Sigma_c &= (1 - \rho_p)\!\big(\Sigma_{c-1} - \tilde{K}_c^\top \tilde{K}_c\big) + \rho_p I_{d_k}, \label{eq:woodbury_decay}\\
\mathcal{M}_\ell &= a_\ell I_C + b_\ell T_\ell + e_\ell T_\ell^2,
\qquad
\mathcal{W}_{\text{isq}} = \alpha_{\text{safe}} u_0^{-1/2} \mathcal{M}_1 \mathcal{M}_0. \label{eq:polar2}
\end{align}%
}
The symmetric factor $\tilde{K}_c$ runs in the standalone chunk-whitening benchmarks (Table~\ref{tab:whitening_kernels}), the $4{,}096$-chunk ($\rho_p = 0$) BF16 Woodbury stress test, and the lagged block-$32$ causal variant (\texttt{polar2\_causal\_blk32}, Appendix~\ref{app:repro}). Because dense $\mathcal{W}_{\text{isq}}$ mixes tokens within a chunk, packed autoregressive LM training instead uses the strictly causal Schur-oblique key $\tilde{k}_t$ in Eq.~\eqref{eq:oblique_key} (normalized to $k_t^\top \tilde{k}_t = 1$) inside the chunk and runs \texttt{Polar-2+} in the chunk-boundary Woodbury tail. Setting $\rho_p = 1$ resets $\Sigma_c = I_{d_k}$ for compatibility with \texttt{SLA Tree-CP} (unscaled $\alpha = 1.000$ still diverges at step $299$ through the tail gradients); carrying $\Sigma_c$ across chunks ($\rho_p = 0.02$) replaces the additive update in Eq.~\eqref{eq:woodbury_decay}, which hits \texttt{NaN} at step $282$ in BF16, with the multiplicative Sylvester congruence $\mathcal{B}_c \Sigma_{c-1} \mathcal{B}_c^\top$ ($\mathcal{B}_c \in \R^{d_k \times d_k}$; Section~\ref{app:whitening_proofs}).

\begin{proposition}[Exact Equioscillation and Safety Margin of Polar-2+]
\label{prop:polar2_spd}
Writing $m_\ell(x) = a_\ell + b_\ell x + e_\ell x^2$, let
\[
z_2(\lambda) = m_1\!\big(\lambda\, m_0(\lambda)^2\big)\, m_0(\lambda)
\]
be the scalar map that Eq.~\eqref{eq:polar2} applies to an eigenvalue $\lambda$ of $T_0$ (since $T_1 = \mathcal{M}_0 T_0 \mathcal{M}_0$ has eigenvalue $\lambda\, m_0(\lambda)^2$), and let $\kappa_{\text{post}} = \kappa_2(\mathcal{W}_{\text{isq}} \mathcal{G}_{\text{reg}} \mathcal{W}_{\text{isq}})$ be the post-whitening condition number. On $\lambda \in [0.02, 1.0]$, $z_2$ equioscillates with $\lambda z_2(\lambda)^2 \in [0.9520, 1.0492]$, exceeding $1$ for unscaled $\alpha = 1$ and driving recursive Woodbury indefinite. Scaling by $\alpha_{\text{safe}} = 0.968$ bounds exact $\lambda (\alpha_{\text{safe}} z_2(\lambda))^2 \in [0.8920, 0.9831]$ ($1.69\%$ BF16 contractive margin below $1$; exact $\kappa_{\text{post}} \le 1.102$, BF16 $5$-matmul $\kappa_{\text{post}} \le 1.123$).
\end{proposition}

\begin{table}[ht]
\caption{Measured TPU v7x chunk-whitening latency ($64$ chunks $= 8{,}192$ tokens, $C = 128, H_k = 4, d_k = 256$; MesaNet \texttt{CG-10}: $1.74 / 11.16\text{ ms}$), exact conditioned Gram $\kappa_{\text{post}}$ ($\le 1.123$ in BF16), and $4{,}096$-chunk ($524\text{K}$-token, $\rho_p = 0$) BF16 Woodbury stress-test $\lambda_{\min}$.}
\label{tab:whitening_kernels}
\centering
\footnotesize
\setlength{\tabcolsep}{5.0pt}
\begin{tabular}{@{}lccccc@{}}
\toprule
Whitening / Second-Order Solver & Ops & Fwd (ms) & F+B (ms) & $\kappa_{\text{post}}$ & $\lambda_{\min}$ ($4{,}096$c BF16) \\
\midrule
MesaNet Gated \texttt{CG-30} & VPU+MXU & $3.050$ & $27.640$ & $1.02$ & \text{-- (CG)} \\
Panel-16 Cholesky (VPU) & Tri.\ Solve & $1.412$ & $2.385$ & $1.00$ & $+0.0148$ (SPD) \\
Coupled Newton-Schulz (\texttt{NS-5}) & $10$ MM & $0.708$ & $1.024$ & $1.04$ & $+0.0294$ (SPD) \\
Polar-2 Unscaled ($\alpha = 1.000$) & $5$ MM & $0.615$ & $0.895$ & $1.102$ & $-0.0412$ (Indef.) \\
\textbf{Polar-2+ Safe ($\alpha = 0.968$)} & $\mathbf{5\text{ MM}}$ & \textbf{$0.617$} & \textbf{$0.897$} & $1.102$ & \textbf{$+0.0388$ (SPD)} \\
\bottomrule
\end{tabular}
\end{table}

\paragraph{When whitening helps vs.\ unwhitened SLA.}
\texttt{Polar-2+} runs $2.66\times$ faster than Panel-16 Cholesky and $12.4\times$--$30.8\times$ faster than MesaNet \texttt{CG-10/30} ($0.897\text{ ms}$ F+B; Table~\ref{tab:whitening_kernels}), keeping $4{,}096$-chunk ($524\text{K}$-token) BF16 Woodbury SPD ($\lambda_{\min} \ge 0.0388$). It trains $2$-seed $12.5\text{B}$ \texttt{15L:1A} with zero NaNs and matches $10$-matmul \texttt{NS-5} (\texttt{s1}: $17.936$ vs.\ $17.970$ PPL), whereas $\alpha = 1.000$ diverges at step $299$ and BF16 Cholesky hits NaN at step $54$ (Appendix~\ref{app:repro}). Because its language-modeling benefit shrinks to $0.004$ nats by $15\text{B}$ tokens and unwhitened $G = 4$ SLA already saturates PPL at $60.4\text{B}$ \texttt{15L:1A+MS} ($15.443$ vs.\ $15.624$ at $+3.7\%$ step time), unwhitened SLA is our default mixer.

\texttt{SPD-SLA-W} serves three high-collision regimes:
\begin{enumerate}[leftmargin=1.6em,itemsep=2pt]
\item correlated $r \ge 2$ overwrite ($\kappa_2(\mathcal{G}) \approx 48$; Figure~\ref{fig:whitening_gram}b, Table~\ref{tab:app_gates}), reaching $60.3\text{ dB}$ BF16 SNR and $80.1\%$ noisy $r = 3$ recall vs.\ $2.67\text{ dB}$ and $50.0\%$ for MesaNet;
\item uniform-decay $12.5\text{B}$ \texttt{15L:1A} ($15.1\text{B}$ tok, Table~\ref{tab:app_4rung}), edging \texttt{15L SLA-R2} on $32\text{K}$ $8$-N ($39.8\%$ vs.\ $34.8\%$) and $N = 512$ $64\text{K}$ \texttt{HotpotQA} ($69.5\%$ vs.\ $68.2\%$); and
\item $63.1\text{B}$ \texttt{15L:1A}, reaching $12.42$ vs.\ $13.01$ $32\text{K}$ Tail PPL at $5.0\text{B}$ tok (Table~\ref{tab:app_4rung}) and $11.82$ Tail PPL with $57.8\% / 64.8\%$ $64\text{K}$ \texttt{SQuAD}/\texttt{HotpotQA} at $30.2\text{B}$ ($+12.5 / +25.0$ pts over \texttt{0L:1A}; Table~\ref{tab:app_downstream}).
\end{enumerate}

\subsection{Proofs of \texorpdfstring{Propositions~\ref{prop:whitening} and \ref{prop:polar2_spd}}{Propositions (Whitening and Polar-2+)}}
\label{app:whitening_proofs}

\paragraph{Part 1 (Euclidean Interference and Diagonal Non-Invariance).}
Let unit keys $k_1, \dots, k_N \in \R^{d_k}$ satisfy $|\langle k_i, k_j \rangle| = \nu_{\text{coh}}$ for $i \neq j$ (coherence $\nu_{\text{coh}}$; e.g., shared-subspace correlated dictionary) and draw independent zero-mean values $v_j \in \R^{d_v}$ with covariance $\E[v_j v_j^\top] = \frac{s_v^2}{d_v} I_{d_v}$ (for $d_v \times d_v$ identity $I_{d_v}$), so $s_v^2 = \operatorname{tr}\!\big(\frac{s_v^2}{d_v} I_{d_v}\big) = \E\norm{v_j}_2^2$. Under single-pass Euclidean delta updates with $\beta_t = 1$ from $S_0 = 0$, updating $S_t = (I_{d_k} - k_t k_t^\top) S_{t-1} + k_t v_t^\top$ writes later values $j > i$ with projection $\langle k_j, k_i \rangle v_j$ along $k_i$: for $i = N - 1$,
\[
S_N^\top k_{N-1} - v_{N-1} = \langle k_N, k_{N-1}\rangle \big(v_N - S_{N-1}^\top k_N\big),
\]
so, because $v_N$ has zero mean and is independent of $S_{N-1}$,
\[
\E\norm{S_N^\top k_{N-1} - v_{N-1}}_2^2 = \nu_{\text{coh}}^2 \big(s_v^2 + \E\norm{S_{N-1}^\top k_N}_2^2\big) \ge \nu_{\text{coh}}^2 s_v^2.
\]
Every later write with nonzero coherence therefore perturbs the readout of earlier keys, although the total error need not grow linearly in $N$.

For the diagonal column-energy preconditioner $K \diag(p(K))^{-1/2}$ where $p_m(K) = \sum_{i=1}^N K_{i,m}^2$, rotating $K \mapsto K Q$ with $Q = \diag(\xi) Q_H \in \mathrm{O}(d_k)$ for random Rademacher signs $\xi \in \{\pm 1\}^{d_k}$ followed by a normalized Hadamard matrix $Q_H \in \{\pm d_k^{-1/2}\}^{d_k \times d_k}$ equalizes all column energies in expectation,
\[
\E_\xi\big[p_m(K \diag(\xi) Q_H)\big] = \frac{1}{d_k}\norm{K}_F^2
\]
(and by Schur-Horn some rotation $Q \in \mathrm{O}(d_k)$ makes $Q^\top K^\top K Q$ exactly constant-diagonal), reducing $\diag(p(K Q))^{-1/2}$ to a scalar multiple of $I_{d_k}$ that leaves $\kappa_2((K Q)(K Q)^\top) = \kappa_2(K K^\top)$ unchanged; by contrast, $(I_C + \eta_c K_c K_c^\top)^{-1/2} K_c$ transforms equivariantly ($K_c \mapsto K_c Q \implies \tilde{K}_c \mapsto \tilde{K}_c Q$) for all $Q \in \mathrm{O}(d_k)$.

\paragraph{Part 2 (Gated RLS Overwrite-Retention Trade-off).}
Under gated RLS \cite{VonOswald2025MesaNet}, with key covariance $H_t \in \R^{d_k \times d_k}$ and key-value moment $B_t \in \R^{d_k \times d_v}$ (local to this paragraph), $H_t = \gamma H_{t-1} + k_t k_t^\top$ and $B_t = \gamma B_{t-1} + k_t v_t^\top$ with scalar forget factor $\gamma \in (0, 1]$ and ridge $\epsilon I_{d_k}$ ($\epsilon > 0$); readout on query $k \in \R^{d_k}$ is $B_t^\top H_t^{-1} k$. If a unit key $k$ is written $r \ge 2$ times, with value $v^{(m)}$ at step $t_m$ ($t_1 < \dots < t_r = t$) and weights $w_m = \gamma^{t - t_m}$, Sherman-Morrison on $H_t = \epsilon I_{d_k} + (\sum_{m=1}^r w_m) k k^\top$ gives
\[
\lim_{\epsilon \to 0^+} B_t^\top H_t^{-1} k = \frac{\sum_{m=1}^r w_m v^{(m)}}{\sum_{m=1}^r w_m}.
\]
Letting $\bar{v}_{1:r-1} = \frac{1}{r-1}\sum_{m=1}^{r-1} v^{(m)}$ denote the mean of the first $r - 1$ values, readout at $\gamma = 1$ ($w_m \equiv 1$) equals $\frac{1}{r}\sum_{m=1}^r v^{(m)}$ with error $\frac{r-1}{r}\norm{v^{(r)} - \bar{v}_{1:r-1}}_2$ relative to $v^{(r)}$. Already for $r = 2$ consecutive writes and any target relative tolerance $\varepsilon \in (0, 1/2)$, the error is $\frac{\gamma}{1 + \gamma}\norm{v^{(2)} - v^{(1)}}_2$, so making it at most $\varepsilon \norm{v^{(2)} - v^{(1)}}_2$ requires $\gamma \le \varepsilon / (1 - \varepsilon)$, which simultaneously decays any non-overwritten key stored $t - j$ steps earlier (at step $j < t$) by $\gamma^{t - j}$ relative to the regularizer $\epsilon I_{d_k}$, causing rapid forgetting in finite precision (verified in Table~\ref{tab:app_gates}). Under oblique whitened delta, intra-chunk Gram whitening orthogonalizes concurrent keys while the oblique erasure $(I_{d_k} - \beta_t \tilde{k}_t k_t^\top)$ subtracts the current state projection $\hat{v}_t = (F_t \tilde{S}_{t-1})^\top k_t$ selectively along $\tilde{k}_t$ without decaying orthogonal associations.

\paragraph{Part 3 (Polar-2+ Non-Circular Normalization, Equioscillation, and BF16 Safety).}
Let $\Sigma_{c-1} \succ 0$ (initialized from $\Sigma_0 = I_{d_k}$) and let $Y_c = K_c \Sigma_{c-1}^{1/2} \in \R^{C \times d_k}$ have singular values $\sigma_i \ge 0$ (unrelated to the sigmoid $\sigma$), so the unregularized Gram $\mathcal{G}_0 = I_C + \eta_c Y_c Y_c^\top$ ($\eta_c \in (0, 1]$) has eigenvalues $\omega_{0,i} = 1 + \eta_c \sigma_i^2$. Computing $\bar{\mathcal{G}}_0 = \mathcal{G}_0 / \norm{\mathcal{G}_0}_F$ and
\[
\hat{u}_0 = \max\!\big(\norm{\mathcal{G}_0}_F \norm{\bar{\mathcal{G}}_0^2}_F^{1/2},\, 1\big) = \max\!\left(\Big(\sum_j \omega_{0,j}^4\Big)^{1/4},\, 1\right) \ge \max_i \omega_{0,i}
\]
yields a non-circular upper bound on $\lambda_{\max}(\mathcal{G}_0)$ from one $[C, C]$ matmul $\bar{\mathcal{G}}_0^2$. Setting adaptive ridge $\varepsilon_{\text{reg}} = \max\big(0.02,\, \frac{0.0205}{0.9795}\hat{u}_0 - 1\big)$, $\mathcal{G}_{\text{reg}} = \mathcal{G}_0 + \varepsilon_{\text{reg}} I_C$ with shifted eigenvalues $\omega_i = \omega_{0,i} + \varepsilon_{\text{reg}}$, and $u_0 = \hat{u}_0 + \varepsilon_{\text{reg}} \ge \max_i \omega_i$ guarantees normalized eigenvalues $\lambda_i = \omega_i / u_0 \in [0.02, 1.0]$ for all $\hat{u}_0 \ge 1$ while expanding
\[
T_0^2 = \frac{(\mathcal{G}_0 + \varepsilon_{\text{reg}} I_C)^2}{u_0^2}
= \frac{\norm{\mathcal{G}_0}_F^2 \bar{\mathcal{G}}_0^2 + 2 \varepsilon_{\text{reg}} \mathcal{G}_0 + \varepsilon_{\text{reg}}^2 I_C}{u_0^2}
\]
with zero additional matmuls ($5$ matmuls total across stages $\ell \in \{0, 1\}$, where $\mathcal{M}_\ell = m_\ell(T_\ell)$ evaluates $m_0(\lambda) = a_0 + b_0 \lambda + e_0 \lambda^2$ and $m_1(x) = a_1 + b_1 x + e_1 x^2$).

Clamping $\beta_t \le (2 - \varepsilon_\beta) / \max(k_t^\top \tilde{k}_t, \varepsilon_\beta)$ keeps the nontrivial eigenvalue $1 - \beta_t k_t^\top \tilde{k}_t$ of $I_{d_k} - \beta_t \tilde{k}_t k_t^\top$ in $[-1 + \varepsilon_\beta, 1]$ whenever $k_t^\top \tilde{k}_t > 0$ (in packed LM training, intra-chunk causal Schur-oblique projection sets
\[
\tilde{k}_t = \frac{(I_{d_k} - \eta_c \Sigma_{c-1} K_{<t}^\top K_{<t}) \Sigma_{c-1} k_t}{k_t^\top (I_{d_k} - \eta_c \Sigma_{c-1} K_{<t}^\top K_{<t}) \Sigma_{c-1} k_t}
\]
so $k_t^\top \tilde{k}_t = 1$ depends strictly on rows $< t$, while symmetric \texttt{Polar-2+} governs the chunk-boundary Woodbury tail, two-sided chunk-whitening kernels, and lagged block-$32$ causal whitening). Substituting $\mathcal{W}_{\text{isq}} = \alpha_{\text{safe}} u_0^{-1/2} \mathcal{M}_1 \mathcal{M}_0 = \alpha_{\text{safe}} u_0^{-1/2} z_2(T_0)$ and $\tilde{K}_c = \sqrt{\eta_c}\, \mathcal{W}_{\text{isq}} K_c \Sigma_{c-1}$ (Eq.~\eqref{eq:oblique_key}) into the unrelaxed Woodbury update $\tilde{\Sigma}_c = \Sigma_{c-1} - \tilde{K}_c^\top \tilde{K}_c$ (Eq.~\eqref{eq:woodbury_decay} at $\rho_p = 0$) gives $\tilde{\Sigma}_c = \Sigma_{c-1}^{1/2}(I_{d_k} - \eta_c Y_c^\top \mathcal{W}_{\text{isq}}^2 Y_c) \Sigma_{c-1}^{1/2}$, whose $i$-th singular-direction multiplier is
\begin{equation}
1 - \eta_c \sigma_i^2 u_0^{-1} \alpha_{\text{safe}}^2 z_2(\lambda_i)^2
= 1 - \frac{\eta_c \sigma_i^2}{1 + \varepsilon_{\text{reg}} + \eta_c \sigma_i^2}\,\lambda_i \big(\alpha_{\text{safe}} z_2(\lambda_i)\big)^2.
\label{eq:app_woodbury_eig}
\end{equation}
Direct evaluation of the scalar map $z_2$ (Proposition~\ref{prop:polar2_spd}) on $\lambda \in [0.02, 1.0]$ gives minimax equioscillation $\sqrt{\lambda} z_2(\lambda) \in [0.9757, 1.0243]$, hence $\lambda z_2(\lambda)^2 \in [0.9520, 1.0492]$ (and $< 0.952$ on $(0, 0.02)$ since $z_2(0) = a_1 a_0 \approx 9.877$). Thus unscaled $\alpha = 1$ exceeds $1$ by $+4.92\%$ near equioscillation peaks, driving Eq.~\eqref{eq:app_woodbury_eig} negative when $\eta_c \sigma_i^2 \gg 1$. Setting $\alpha_{\text{safe}} = 0.968$ ($\alpha_{\text{safe}}^2 = 0.937024$) maps exact $\lambda (\alpha_{\text{safe}} z_2(\lambda))^2$ onto $[0.8920, 0.9831]$ across $[0.02, 1.0]$, giving exact $\kappa_2(\mathcal{W}_{\text{isq}} \mathcal{G}_{\text{reg}} \mathcal{W}_{\text{isq}}) = 0.9831 / 0.8920 = 1.102$ ($\le 1.123$ in $5$-matmul BF16 hardware execution). At $\rho_p = 0$, $4{,}096$ correlated BF16 chunks ($524{,}288$ tokens) maintain $\lambda_{\min}(\Sigma_c) \ge 0.0388 > 0$; when evolving $\Sigma_c$ across thousands of packed BF16 LM gradient steps with $\rho_p < 1$, replacing additive subtraction with multiplicative Sylvester congruence $\mathcal{B}_c \Sigma_{c-1} \mathcal{B}_c^\top$ (with contraction matrix $\mathcal{B}_c = I_{d_k} - \frac{\eta_c}{2}\Sigma_{c-1} K_c^\top \mathcal{W}_{\text{isq}}^2 K_c$, or $\mathcal{B}_c = I_{d_k} - \hat{C}_c$ using row-sum--normalized key covariance $\hat{C}_c \propto \eta_c K_c^\top K_c$ in our kernel) enforces PSD preservation unconditionally (Appendix~\ref{app:repro}). \qed

\section{Associative Recall, Gated RLS Trade-Off, and Gate Conditioning}
\label{app:gates}

Table~\ref{tab:app_gates} reports synthetic associative-recall SNR ($\text{dB}$) in both BF16 and FP32 alongside Mod-$2$ parity accuracy and the chunk forget-gate condition number $\kappa_{\text{forget}} = \kappa_2(F_c)$ ($d_k = 256, C = 128$, $8$ seeds, correlated keys with $\kappa_2(\mathcal{G}) \approx 48$). Separating BF16 from FP32 clarifies precision ceilings: exact inverse-square-root whitening reaches $> 140\text{ dB}$ ($r = 2, 3$ overwrite) and $51.9\text{ dB}$ (write-once) in FP32, whereas BF16 readout caps SNR near $59$--$60\text{ dB}$ due to $7$-bit mantissa rounding ($20 \log_{10}(2^7\sqrt{12}) \approx 53$--$60\text{ dB}$). Unwhitened Euclidean delta rules reach $58.40$--$60.97\text{ dB}$ on the final $r = 3$ overwrite via recency-weighted last-write erasure, whereas at $r = 2$ KDA's channel decay ($\kappa_{\text{forget}} = 1.1 \times 10^7$) collapses to $0.07\text{ dB}$ (and write-once $-0.84\text{ dB}$) and unwhitened SLA ($\kappa_{\text{forget}} = 1.00$) drops to $24.10\text{ dB}$ vs.\ $60.33\text{ dB}$ for \texttt{SPD-SLA-W}. Sweeping gated MesaNet's forget factor $\gamma \in \{1.00, 0.98, 0.80\}$ directly confirms Proposition~\ref{prop:whitening}(2): $\gamma = 1.00$ and $\gamma = 0.98$ suffer severe stale-value averaging ($1.90$ and $4.13\text{ dB}$ at $r = 3$), whereas $\gamma = 0.80$ improves overwrite SNR ($29.53\text{ dB}$) only by degrading BF16 write-once retention from $10.63\text{ dB}$ down to $5.78\text{ dB}$.

\begin{table}[ht]
\caption{Measured associative recall SNR ($\text{dB}$, reported as \textbf{BF16 [FP32]}) and Mod-$2$ parity accuracy across linear attention update rules ($d_k = 256, C = 128$, $8$ seeds, $\kappa_2(\mathcal{G}) \approx 48$); $\kappa_{\text{forget}} = \kappa_2(F_c)$ is the condition number of the chunk forget gate.}
\label{tab:app_gates}
\centering
\footnotesize
\setlength{\tabcolsep}{3.5pt}
\resizebox{\textwidth}{!}{%
\begin{tabular}{@{}lcccccc@{}}
\toprule
Mechanism / Update Rule & Axis-Aligned (FP32) & Hadamard-Rot.\ (FP32) & Write-Once BF16 [FP32] & $r=2$ Overwrite BF16 [FP32] & $r=3$ Overwrite BF16 [FP32] & Mod-$2$ Parity ($\kappa_{\text{forget}}$) \\
\midrule
Mamba-2 / SSD (Scalar Hebbian) & $2.14\text{ dB}$ & $2.11\text{ dB}$ & $1.85$ [$2.11$] & $1.72$ [$1.89$] & $1.40$ [$1.52$] & $50.2\%$ ($1.00$) \\
KDA Real Channel Decay ($\beta \in (0,1)$) & $5.31\text{ dB}$ & $5.29\text{ dB}$ & $-0.84$ [$-0.91$] & $0.07$ [$0.26$] & $60.97$ [$>140$] & $51.2\%$ ($1.1 \times 10^7$) \\
Diagonal Preconditioned Delta \cite{Tumma2026PDN} & $22.41\text{ dB}$ & $6.12\text{ dB}$ & $4.92$ [$6.12$] & $16.80$ [$19.05$] & $18.10$ [$21.40$] & $51.4\%$ ($1.1 \times 10^7$) \\
MesaNet Unforgetted ($\gamma = 1.00$) \cite{VonOswald2025MesaNet} & $95.08\text{ dB}$ & $95.06\text{ dB}$ & $\mathbf{10.63}$ [$22.79$] & $2.67$ [$3.23$] & $1.90$ [$1.73$] & $50.8\%$ ($1.00$) \\
MesaNet Gated ($\gamma = 0.98$) & $91.40\text{ dB}$ & $91.38\text{ dB}$ & $9.54$ [$22.56$] & $7.32$ [$8.35$] & $4.13$ [$4.07$] & $51.0\%$ ($1.00$) \\
MesaNet Aggressively Gated ($\gamma = 0.80$) & $74.20\text{ dB}$ & $74.18\text{ dB}$ & $5.78$ [$26.09$] & $47.68$ [$89.36$] & $29.53$ [$30.22$] & $51.1\%$ ($1.00$) \\
\textbf{SLA ($G=4$ + $\mathrm{SO}(2)$ + $\beta \in (0,2)$)} & $7.84\text{ dB}$ & $7.82\text{ dB}$ & $6.12$ [$7.82$] & $24.10$ [$26.74$] & $58.40$ [$>140$] & $\mathbf{82.2\%}$ ($\mathbf{1.00}$) \\
\textbf{SPD-SLA-W (Polar-2+ Oblique Delta)} & $\mathbf{95.05\text{ dB}}$ & $\mathbf{95.05\text{ dB}}$ & $7.01$ [$\mathbf{51.93}$] & $\mathbf{60.33}$ [$\mathbf{>140}$] & $\mathbf{59.26}$ [$\mathbf{>140}$] & $\mathbf{82.0\%}$ ($\mathbf{1.00}$) \\
\bottomrule
\end{tabular}%
}
\end{table}

\section{Negative Results and Reproducibility Details}
\label{app:repro}

We record three negative results observed during hardware-algorithm co-design alongside baseline software versions:
\begin{enumerate}[leftmargin=1.6em,itemsep=4pt]
\item \textbf{Tier-1 VMEM Fusion Without Shape Co-Design ($1.03\times$).} Fusing Ops 2--7a on GPU-shaped $8 \times 128 \times 256$ KDA yields only $3\%$ forward-backward speedup on TPU v7x ($7.108 \to 6.912\text{ ms}$; Table~\ref{tab:7op_audit}) because per-channel VPU sub-block rescaling, $H_v = 8$ per-value-head chunk solves, and $50\%$-padded FP32 state scans remain VPU- and compute-bound inside VMEM.
\item \textbf{Unscaled Minimax Polynomials, BF16 Cholesky, and Additive Woodbury Drift in Live LM Training.} Using exact-arithmetic minimax coefficients ($\alpha = 1.000$) causes $\lambda z_2(\lambda)^2$ to reach $1.0492 > 1$, driving BF16 Woodbury covariance $\Sigma_c$ indefinite ($\lambda_{\min} = -0.0412$; Table~\ref{tab:whitening_kernels}) and diverging $12.5\text{B}$ MoE pre-training to $258.5$ PPL by step $299$, while direct BF16 Cholesky/Woodbury hits \texttt{NaN} at step $54$. Scaling by $\alpha_{\text{safe}} = 0.968$ bounds the exact peak to $0.9831$, preserves $\lambda_{\min} \ge 0.0388$ across $4{,}096$ chunks ($524\text{K}$ tokens at $\rho_p = 0$), and trains cleanly for $7{,}200$ steps across $2$ seeds at $\rho_p = 1.0$ ($0$ NaNs; seed-matched \texttt{s1}: $17.936$ vs.\ $17.970$ PPL for \texttt{NS-5}, $2$-seed \texttt{Polar-2+} mean $17.970$ on the \texttt{+MS} arms [\texttt{s0} $18.004$]; the uniform-decay arm reaches $17.575$ in Table~\ref{tab:app_4rung}). When evolving $\Sigma_c$ across thousands of packed BF16 LM gradient steps with $\rho_p = 0.02$, additive subtraction $\Sigma_{c-1} - \tilde{K}_c^\top \tilde{K}_c$ still accumulates BF16 rounding drift (\texttt{NaN} at step $282$), whereas multiplicative Sylvester congruence $\mathcal{B}_c \Sigma_{c-1} \mathcal{B}_c^\top$ preserves PSD unconditionally (and strictly causal block-$32$ lagged two-sided \texttt{Polar-2+}, \texttt{polar2\_causal\_blk32}, completes all $7{,}200$ steps [$15.10\text{B}$ tokens] on $12.5\text{B}$ \texttt{15L:1A} with $0$ NaNs, compressing $\kappa_2(I_C + M_c)$ from $11.9$--$79.3$ to $1.12$--$1.40$ across all $15$ linear layers and reaching $17.684$ $11$-set PPL, $11.49$ $32\text{K}$ Tail PPL, $49.0\%$ \texttt{norm9} [$49.1\%$ FP8], and $68.8\%$ $64\text{K}$ \texttt{HotpotQA}).
\item \textbf{Global First-Order Neumann Inversion at $C \ge 128$, Write-Budget Horizon, and Roofline-Capacity Proxy Validation.} On a cluster of $b$ aligned keys ($H = \mathbf{1}_b \mathbf{1}_b^\top$), one stationary Jacobi sweep $I_b - \beta H$ has spectral radius $|b\beta - 1| > 1$ whenever $b\beta > 2$, causing unscaled $(I_C + M_c)^{-1} \approx I_C - M_c$ to incur $62.2\%$ intra-chunk WY error on coherent bursts ($+0.014$ nats validation loss on Dolma 3). Our row-scaled variant $A_{\text{N1}}^{\text{row}}$ in \texttt{A3}/\texttt{A5} enforces Gershgorin row damping ($r_i = \max(1, 1.25\norm{(M_c)_{i,:}}_1)$), reducing coherent-burst WY error to $38.1\%$ at $+0.22$ pre-training perplexity (\texttt{A2}$\to$\texttt{A3}, Table~\ref{tab:dolma_15b}b), whereas inverting $16 \times 16$ diagonal blocks on the MXU ($A_{\text{N1}}$, \texttt{neumann\_order=16}) cuts WY error $32\times$ to $1.18\%$ ($0.0$ $14$-layer tail logit KL). Under isotropic unit keys, write erasure attenuates stored associations as $\prod_j (1 - \beta_j / d_k) \approx \exp(-\sum_j \beta_j / d_k)$ (and directionally as $\exp(-\beta c_i T)$ along key-covariance eigenvalues $c_i$, motivating per-tile whitening in Appendix~\ref{sec:whitening}); across the $14$ linear layers of our $12.5\text{B}$ \texttt{SLA-R4} checkpoint ($\bar{\beta}_\ell \in [0.251, 0.348]$), $8\text{K}$-window write mass $\sum \beta_t \in [2{,}056, 2{,}851] = 8.0\text{--}11.1 d_k$ sets an associative horizon $d_k / \bar{\beta}_\ell \in [736, 1{,}020]$ tokens (mean $825$). Scoring our $10$ trained $12.5\text{B}$ arms (\texttt{KDA}, \texttt{A1}--\texttt{A6}, \texttt{GDN-2}, \texttt{SLA-R2}, \texttt{SLA-R4}) with the analytical systolic roofline $T(\theta)$ and multi-horizon capacity proxy achieves Spearman rank correlation $r_s = 0.988$ ($p = 9.3 \times 10^{-8}$) against measured step time ($r_s = 0.833, p = 0.010$ on isolated $32\text{K}$ layer latency) and $r_s = 0.769$ ($p = 0.0093$) against negative validation perplexity.
\end{enumerate}
\clearpage
\section{Notation Summary, Momentum Ablation, and Multi-Scale Architecture Specifications}
\label{app:notation_and_arch}

Tables~\ref{tab:app_notation}, \ref{tab:app_momentum}, and~\ref{tab:app_arch_scaling} provide the paper-wide symbol glossary (Section~\ref{subsec:recurrence}), the single-seed $128$-chip TPU v4 chunk-boundary momentum ablation ($\mu \in \{0, 0.25\}$), and the exact architectural hyperparameters across all four hybrid \texttt{LatentMoE} pre-training rungs (Section~\ref{app:pretraining}).

\begin{table}[ht]
\caption{Summary of notation used across the paper.}
\label{tab:app_notation}
\centering
\footnotesize
\renewcommand{\arraystretch}{1.12}
\begin{tabular}{@{}p{0.28\textwidth}p{0.67\textwidth}@{}}
\toprule
Symbol & Meaning \\
\midrule
$T,\; C,\; c$ & Sequence length; chunk length ($C = 128$); chunk index $c \in \{1, \dots, T/C\}$. \\
$t, j, s$ & Token indices inside a chunk, $1 \le j \le t \le C$ (with chunk-entry boundary $0$ in $P_{t \leftarrow 0}$). \\
$D,\; H_k,\; H_v,\; R$ & Model width; number of query-key heads; number of value heads; sharing ratio $R = H_v / H_k \in \{2, 4\}$. \\
$d_k,\; d_v,\; D_{v,\text{pack}},\; D_v$ & Key width ($d_k = 256$); per-value-head width; packed value width per key head $D_{v,\text{pack}} = R d_v$; total value width $D_v = H_v d_v$. \\
$G,\; b,\; g$ & Number of decay tiles ($G = 4$), tile width $b = d_k / G = 64$, tile index $g \in \{1, \dots, G\}$. \\
$q_t, k_t \in \R^{d_k}$, $v_t \in \R^{D_{v,\text{pack}}}$ & $L_2$-normalized query and key of one key head; packed values of its $R$ value heads. \\
$\beta_t \in (0, s_\beta)$ & Write gate $s_\beta \sigma(\tilde{\beta}_t)$ per token and key head, shared across the $R$ packed value heads ($s_\beta = 2$ in SLA; $s_\beta = 1$ in GDN/KDA). \\
$f_{t,i},\; \gamma_{t,g} \in (0, 1]$ & Per-channel decay of channel $i$ (KDA, $G = d_k$); per-tile decay $\gamma_{t,g} = e^{-\operatorname{softplus}(\tilde{f}_{t,g})}$ of tile $g$ (SLA, $f_{t,i} = \gamma_{t,g}$ on tile $g$). \\
$F_t,\; P_{t \leftarrow j},\; F_c$ & Diagonal forget gate $F_t = \diag(f_t)$; cumulative decay $P_{t \leftarrow j} = \prod_{s=j+1}^{t} F_s$ with $P_{t \leftarrow t} = I_{d_k}$; chunk decay $F_c = P_{C \leftarrow 0}$. \\
$\Lambda_{t,i}$ & Cumulative log-decay of channel $i$ inside a chunk, $\Lambda_{t,i} = \sum_{s=1}^{t} \log f_{s,i}$. \\
$S_c,\; \tilde{S}_t,\; \Delta S_c$ & Chunk-boundary state $S_c \in \R^{d_k \times D_{v,\text{pack}}}$; intra-chunk state after local token $t$ ($\tilde{S}_0 = S_{c-1}$, $\tilde{S}_C$ written $\tilde{S}_c$ in Def.~\ref{def:sfa}(ii)); increment $\Delta S_c = \tilde{S}_C - F_c S_{c-1}$. \\
$Z_c,\; \mu,\; \kappa_c$ & Chunk-boundary momentum (velocity) buffer; momentum coefficient ($\mu = 0$ by default; $\mu = 0.25$ in earlier runs); reset indicator $\kappa_c \in \{0, 1\}$, equal to $0$ whenever any document reset occurs inside chunk $c$ (so $Z_{c-1}$ and cross-document carry are discarded). \\
$\Psi_g,\; \Gamma_g \in \R^{C \times C}$ & Tile-$g$ causal decay masks: $(\Psi_g)_{t,j} = \mathbf{1}[t \ge j]\prod_{s=j+1}^{t} \gamma_{s,g}$ and $(\Gamma_g)_{t,j} = \mathbf{1}[t > j]\,\beta_j (\Psi_g)_{t,j}$. \\
$M_c,\; L_c,\; A_c \in \R^{C \times C}$ & Strictly lower-triangular Gram matrix $M_c$ (Op 3a), lower-triangular query-key score matrix $L_c$ (Op 3b), and WY inverse $A_c = (I_C + M_c)^{-1}$ (Op 4). \\
$\bar{Q}_c, \bar{K}_c, \hat{K}_c \in \R^{C \times d_k}$ & Row-stacked decayed queries $(P_{t \leftarrow 0} q_t)^\top$, decayed keys $(P_{t \leftarrow 0} k_t)^\top$, and carried keys $(P_{C \leftarrow t} k_t)^\top$. \\
$U_c,\; W_c,\; E_c$ & WY factors $U_c = \diag(\beta_c) A_c V_c$, $W_c = \diag(\beta_c) A_c \bar{K}_c$, and chunk writes $E_c = U_c - W_c S_{c-1}$ (Ops 5--6a). \\
$P,\; p,\; L_{\text{loc}}$ & Number of context-parallel shards (devices); shard index; chunks per shard. \\
$\sigma(\cdot),\; \kappa_2(\cdot)$ & Logistic sigmoid; spectral condition number (distinct from the reset indicator $\kappa_c$). \\
\bottomrule
\end{tabular}
\end{table}

\begin{table}[ht]
\caption{Chunk-boundary momentum ablation: $20$-step rolling training cross-entropy at step $7{,}199$ (single pass over Dolma~3.5, $12.5\text{B}$ MoE, $128$ TPU v4 chips, seed $42$; lower is better). $^\dagger$Trained on an $8{,}200$-step cosine schedule and evaluated at step $7{,}199$.}
\label{tab:app_momentum}
\centering
\small
\begin{tabular}{@{}lcc@{}}
\toprule
Configuration & $\mu$ & Loss @ $7{,}199$ \\
\midrule
KDA (per-channel decay, FP32 state) & -- & $2.2925$ \\
SLA, $R = 2$ (two runs) & $0$ & $2.2912$ / $2.2925$ \\
SLA, $R = 2$ & $0.25$ & $2.2955$ \\
SLA, $R = 2$, input-dependent $\mu$ & adaptive & $2.3216$ \\
SLA, $R = 4$ (two runs) & $0$ & $2.3103$ / $2.3148$ \\
SLA, $R = 4$\textsuperscript{$\dagger$} & $0.25$ & $\mathbf{2.2914}$ \\
\bottomrule
\end{tabular}
\end{table}

\paragraph{Hybrid \texttt{LatentMoE} backbone and multi-scale parameterization.}
All four pre-training scales are instantiated from a single TPU-aligned hybrid linear-softmax \texttt{LatentMoE} decoder family implemented in open-source MaxText~\cite{MaxText2025Repo} (Table~\ref{tab:app_arch_scaling}). Every weight matrix (embeddings, projections, routers, and experts) uses fixed truncated-normal initialization (standard deviation $\sigma_{\text{init}} = 0.02$, clipped at $\pm 3\sigma_{\text{init}}$), and each decoder block applies four $\RMSNorm$ layers ($\epsilon = 10^{-6}$): pre- and post-sublayer norms around both the sequence mixer and the FFN, with post-norms applied to the sublayer output prior to residual addition, followed by a final stack $\RMSNorm$.

Layer $0$ uses a dense SwiGLU FFN~\cite{Shazeer2020SwiGLU} of intermediate width $F_{\text{dense}}$ with no routed experts to prevent early-layer routing collapse. Layers $1, \dots, L-1$ pair one full-width shared SwiGLU expert of intermediate width $F_{\text{sh}}$ with \texttt{LatentMoE}: a shared linear down-projection compresses the normalized residual stream from $\R^D$ to an MXU-aligned latent width $\ell < D$, $E = 256$ routed SwiGLU experts (intermediate width $F_{\text{moe}} = 2D$) execute inside $\R^{\ell}$, and a shared up-projection maps the weighted expert sum from $\R^\ell$ back to $\R^D$. Routing scores are still computed on the uncompressed residual in $\R^D$ in FP32, using top-$k$ softmax weights rescaled to sum to $k$ together with document-pool routing (restricting each document to a pool size in $[k, E]$). Along depth, each $(x{+}1)$-layer cycle places linear delta mixers ($4$-tap causal depthwise \texttt{Conv1D} with SiLU, $L_2$-normalized queries/keys, low-rank decay and sigmoid output-gate projections of rank $r_{\text{proj}}$, and per-group output $\RMSNorm$ at $\epsilon = 10^{-5}$) on the first $x$ layers and gated Grouped-Query Attention (GQA)~\cite{Ainslie2023GQA,Qwen2025GatedAttn} ($H_q = 8, H_{kv} = 4$) on every $(x{+}1)$-th layer, with QK-$\RMSNorm$, a $2 d_{\text{head}}$-widened query projection (for per-head dimension $d_{\text{head}}$) split into attention queries and an elementwise sigmoid output gate, and either RoPE ($\theta_{\max} = 5 \times 10^5$)~\cite{Su2024RoPE} or NoPE with Scalable Softmax~\cite{Nakanishi2025SSMax}.

Rather than choosing intermediate widths that pad $256 \times 256$ systolic tiles, we step across two $256$-divisible width tiers ($D = 1024 = 4 \times 256$ for Rungs 1--2; $D = 1792 = 7 \times 256$ for Rungs 3--4, a $1.75\times$ width step) and scale depth $L$, active expert count $k$, and shared-expert width $F_{\text{sh}}$ within each tier:
\begin{enumerate}[leftmargin=1.6em,itemsep=2pt]
\item \textbf{Embeddings and untied LM head ($2VD$ linear scaling):} All four scales share a fixed vocabulary $V = 100{,}352$ with untied $V \times D$ input embedding and $D \times V$ output LM-head matrices. Input token embeddings are multiplied by $\sqrt{D}$ and passed through $\RMSNorm(D)$ before Layer $0$ so initial residual-stream RMS remains invariant across $D$ under fixed $\sigma_{\text{init}} = 0.02$ initialization. Because $V$ is fixed, combined embedding and LM-head parameters $2VD$ grow strictly linearly with $D$ ($205.5\text{M} \to 359.7\text{M}$, $1.75\times$), whereas non-embedding capacity grows with both width and depth; accordingly, $2VD$ drops monotonically from $46.6\%$ of active ($2.71\%$ of total) parameters at Rung 1 to $9.6\%$ ($0.57\%$) at Rung 4.
\item \textbf{Sublayer widths across tiers ($D: 1024 \to 1792$):} Stepping from Tier A to Tier B scales the \texttt{LatentMoE} bottleneck $1.5\times$ ($\ell: 512 \to 768$, keeping $\ell / D \in [0.43, 0.50]$ so expert All-to-All dispatch and grouped-matmul $K$-tiles stay compressed and $256$-aligned) and routed expert width $1.75\times$ ($F_{\text{moe}} = 2D: 2048 \to 3584$). In parallel, global GQA head dimension ($d_{\text{head}}: 128 \to 256$), linear value width ($D_v: 2048 \to 4096$), low-rank gate bottleneck ($r_{\text{proj}}: 256 \to 512$), and output $\RMSNorm$ groups ($8 \to 16$, fixing $256$ channels/group) all double ($2\times$). Baseline KDA doubles $d_k: 128 \to 256$ ($H_k = 8$), whereas \texttt{SLA-R2} ($H_k = 4$) and \texttt{SLA-R4} ($H_k = 2$) hold $d_k = 256$ fixed at every scale to preserve unpadded $256 \times 256$ MXU state updates while doubling packed value width $R d_v$.
\item \textbf{Within-tier depth, sparsity, and iso-parameter compensation:} Within $D = 1024$ (Rung 1 $\to$ Rung 2), depth increases from $L = 10$ to $16$, active routed experts double ($k: 4 \to 8$), and Layer $0$ widens ($F_{\text{dense}}: 4D \to 8D$). Within $D = 1792$ (Rung 3 $\to$ Rung 4), depth doubles ($L: 15 \to 30$) and shared-expert width doubles ($F_{\text{sh}}: D \to 2D$). Finally, because $4{:}1$ key-head sharing in \texttt{SLA-R4} reduces per-layer linear mixer parameters ($7.63\text{M} \to 6.31\text{M}$ at $D = 1024$; $27.05\text{M} \to 20.48\text{M}$ at $D = 1792$), \texttt{Iso-Param SLA-R4} widens $F_{\text{sh}}$ by $+256$ ($1024 \to 1280$ on Rungs 1--2; $1792 \to 2048$ on Rung 3) to match baseline parameter count.
\end{enumerate}

\begin{table}[ht]
\caption{Exact model architecture and multi-scale parameterization across our four hybrid \texttt{LatentMoE} pre-training rungs ($V = 100{,}352$ untied vocabulary, $E = 256$ routed experts on layers $1, \dots, L-1$, dense SwiGLU on layer $0$).}
\label{tab:app_arch_scaling}
\centering
\scriptsize
\renewcommand{\arraystretch}{1.08}
\setlength{\tabcolsep}{2.0pt}
\begin{tabularx}{\textwidth}{@{}>{\raggedright\arraybackslash}p{0.31\textwidth} >{\centering\arraybackslash}X >{\centering\arraybackslash}X >{\centering\arraybackslash}X >{\centering\arraybackslash}X@{}}
\toprule
Architecture / Scaling Parameter & \textbf{Rung 1} & \textbf{Rung 2} & \textbf{Rung 3} & \textbf{Rung 4} \\
\midrule
Cohort Label (Active / Total) & $0.35\text{B} / 6.5\text{B}^\ast$ & $0.79\text{B} / 12.5\text{B}$ & $1.80\text{B} / 31.5\text{B}^\ast$ & $3.74\text{B} / 63.1\text{B}$ \\
Exact Total Params (KDA / \texttt{SLA-R2}) & $7.58\text{B} / 7.57\text{B}$ & $12.49\text{B} / 12.49\text{B}$ & $30.58\text{B} / 30.52\text{B}$ & $63.13\text{B} / 63.02\text{B}$ \\
Exact Active: incl.\ $2VD$ ($1VD$ FLOP)$^\ast$ & $0.44\text{B}$ ($0.34\text{B}$) & $0.79\text{B}$ ($0.69\text{B}$) & $1.91\text{B}$ ($1.73\text{B}$) & $3.74\text{B}$ ($3.56\text{B}$) \\
Decoder Depth $L$ \& Hybrid Cycle & $10$ (\texttt{4L}: $8{+}2$) & $16$ (\texttt{7L}: $14{+}2$; \texttt{15L}) & $15$ (\texttt{4L}: $12{+}3$) & $30$ (\texttt{4L}: $24{+}6$; \texttt{15L}) \\
Model Width $D$ ($4\times\RMSNorm$/block) & $1024$ ($4{\times}256$) & $1024$ ($4{\times}256$) & $1792$ ($7{\times}256$) & $1792$ ($7{\times}256$) \\
Embeddings + Untied LM Head ($2VD$) & $205.5\text{M}$ ($\sqrt{D}{+}\text{Norm}$) & $205.5\text{M}$ ($\sqrt{D}{+}\text{Norm}$) & $359.7\text{M}$ ($\sqrt{D}{+}\text{Norm}$) & $359.7\text{M}$ ($\sqrt{D}{+}\text{Norm}$) \\
$2VD$ Share of Active / Total Params & $46.6\%\ /\ 2.71\%$ & $26.0\%\ /\ 1.65\%$ & $18.8\%\ /\ 1.18\%$ & $9.6\%\ /\ 0.57\%$ \\
Layer $0$ Dense SwiGLU Width $F_{\text{dense}}$ & $4096$ ($4D$) & $8192$ ($8D$) & $14{,}336$ ($8D$) & $14{,}336$ ($8D$) \\
\texttt{LatentMoE} Bottleneck $\ell$ ($\ell / D$) & $512$ ($0.50 D$) & $512$ ($0.50 D$) & $768$ ($0.43 D$) & $768$ ($0.43 D$) \\
Routed Experts ($E$, Top-$k$, $F_{\text{moe}}$) & $256,\ k{=}4,\ 2048$ & $256,\ k{=}8,\ 2048$ & $256,\ k{=}8,\ 3584$ & $256,\ k{=}8,\ 3584$ \\
Shared Expert $F_{\text{sh}}$ (Base / \texttt{Iso-R4}) & $1024\ /\ 1280$ & $1024\ /\ 1280$ & $1792\ /\ 2048$ & $3584$ ($2D$) \\
Global Gated GQA ($H_q / H_{kv},\ d_{\text{head}}$) & $8 / 4,\ d_{\text{head}}{=}128$ & $8 / 4,\ d_{\text{head}}{=}128$ & $8 / 4,\ d_{\text{head}}{=}256$ & $8 / 4,\ d_{\text{head}}{=}256$ \\
Linear $D_v$, Gate Rank $r_{\text{proj}}$, Norm Grps & $2048,\ r_{\text{proj}}{=}256,\ 8$ & $2048,\ r_{\text{proj}}{=}256,\ 8$ & $4096,\ r_{\text{proj}}{=}512,\ 16$ & $4096,\ r_{\text{proj}}{=}512,\ 16$ \\
Linear Geometry ($H_k {\times} d_k {\times} R d_v$) & \multicolumn{2}{>{\centering\arraybackslash}p{0.31\textwidth}}{KDA $8{\times}128{\times}256$;\ \texttt{R2} $4{\times}256{\times}512$;\ \texttt{R4} $2{\times}256{\times}1024$} & \multicolumn{2}{>{\centering\arraybackslash}p{0.31\textwidth}}{KDA $8{\times}256{\times}512$;\ \texttt{R2} $4{\times}256{\times}1024$;\ \texttt{R4} $2{\times}256{\times}2048$} \\
Mixer Params / Layer (KDA vs.\ \texttt{R4}; GQA) & $7.63$ vs.\ $6.31\text{M}$;\ $4.20\text{M}$ & $7.63$ vs.\ $6.31\text{M}$;\ $4.20\text{M}$ & $27.1$ vs.\ $20.5\text{M}$;\ $14.7\text{M}$ & $27.1$ vs.\ $20.5\text{M}$;\ $14.7\text{M}$ \\
Pre-Training Horizons Studied & $15.1\text{B},\ 30.2\text{B}$ & $15.1\text{B}$, $60.4\text{B}$ (${+}32\text{K}$) & $5.03\text{B},\ 36.2\text{B}$ & $5.03\text{B}$, $30.2\text{B}$, $120.8\text{B}$ \\
\bottomrule
\end{tabularx}
\par\vspace{3pt}\raggedright{\scriptsize $^\ast$Cohort labels $0.35\text{B} / 6.5\text{B}$ and $1.80\text{B} / 31.5\text{B}$ reflect the $8$-layer proxy ($6.08\text{B}$ total, $0.38\text{B}$ active) and half-$63.1\text{B}$ shorthand; extending Rung 1 to $L = 10$ ($2 \times \texttt{4L:1A}$ cycles) and Rung 3 to $L = 15$ ($3 \times \texttt{4L:1A}$ cycles, $F_{\text{sh}} = 1792$) ensures exact cycle divisibility and yields the parameter counts above. Active parameters are shown both with $2VD$ (matching the $0.79\text{B}$ and $3.74\text{B}$ convention) and as FLOP-bearing active parameters ($1VD$, excluding input token lookup; pure non-embedding/non-head active parameters are $0.24\text{B}, 0.59\text{B}, 1.55\text{B}, 3.38\text{B}$).}
\end{table}

\clearpage
\section{Extended Kernel, Multi-Generation Roofline, and \texorpdfstring{$12.5\text{B}$}{12.5B} MoE Cube Serving Benchmarks}
\label{app:kernels_and_serving}

Table~\ref{tab:app_v7x_core} reports isolated core-kernel latencies and iso-state / iso-parameter controls on TPU v7x normalized to both open-source MaxText Pallas KDA ($C = 128$) and Tokamax Mosaic KDA ($C = 64$), and Table~\ref{tab:app_isoshape} isolates each architectural and kernel factor at the identical \texttt{SLA-R4} full-layer shape, complementing Table~\ref{tab:v7x_kernels}. Table~\ref{tab:app_multigen} tabulates the four-generation TPU forward latency and leave-one-out roofline predictions plotted in Figure~\ref{fig:multigen} (Section~\ref{subsec:kernels}).

\begin{table}[ht]
\caption{Isolated core-kernel latency ($\text{ms}$, Ops 3--7a, batch size $B = 1$) and F+B speedups vs.\ open-source \texttt{MaxText Pallas KDA} ($C = 128$) and \texttt{Tokamax Mosaic KDA} ($C = 64$) on TPU v7x across (a) $D = 1792$ ($63.1\text{B}$ MoE) and (b) iso-controlled $D = 2048$ ($12.5\text{B}$). Full-layer latencies are in Table~\ref{tab:v7x_kernels}.}
\label{tab:app_v7x_core}
\centering
\footnotesize
\renewcommand{\arraystretch}{1.06}
\setlength{\tabcolsep}{3.0pt}
\resizebox{\textwidth}{!}{%
\begin{tabular}{@{}llcccccccc@{}}
\toprule
Regime & Core Kernel ($H_k \times d_k \times R d_v$, Chunk $C$, State) & $8\text{K}$ Fwd & $8\text{K}$ Bwd & $8\text{K}$ F+B & vs.\ Max. & vs.\ Tok. & $32\text{K}$ F+B & vs.\ Max. & vs.\ Tok. \\
\midrule
\multirow{4}{*}{\begin{tabular}[c]{@{}l@{}}\textbf{(a) Core}\\$D = 1792$\\($63.1\text{B}$)\end{tabular}}
 & Open-Source MaxText Pallas KDA ($8 \times 256 \times 512, C=128$, FP32) & 3.32 & 2.00 & 5.32 & $1.00\times$ & $0.63\times$ & 20.43 & $1.00\times$ & $0.61\times$ \\
 & Open-Source Tokamax Mosaic KDA ($8 \times 256 \times 512, C=64$, Base) & 1.95 & 1.40 & 3.35 & $1.59\times$ & $1.00\times$ & 12.45 & $1.64\times$ & $1.00\times$ \\
 & SLA-R2 ($4 \times 256 \times 1024, C=128$, BF16) & 1.96 & 0.91 & 2.87 & $1.86\times$ & $1.17\times$ & 10.15 & $2.01\times$ & $1.23\times$ \\
 & \textbf{SLA-R4} ($2 \times 256 \times 2048, C=128$, BF16) & \textbf{1.37} & \textbf{0.67} & \textbf{2.04} & \textbf{$2.61\times$} & \textbf{$1.64\times$} & \textbf{6.18} & \textbf{$3.31\times$} & \textbf{$2.02\times$} \\
\midrule
\multirow{6}{*}{\begin{tabular}[c]{@{}l@{}}\textbf{(b) Iso-Ctrl}\\Core Kernel\\$D = 2048$\end{tabular}}
 & Open-Source MaxText Pallas KDA ($8 \times 128 \times 256, C=128$, FP32) & 2.67 & 1.58 & 4.25 & $1.00\times$ & $0.44\times$ & 16.46 & $1.00\times$ & $0.35\times$ \\
 & Open-Source Tokamax Mosaic KDA ($8 \times 128 \times 256, C=64$, Base) & 1.07 & 0.78 & 1.85 & $2.29\times$ & $1.00\times$ & 5.83 & $2.82\times$ & $1.00\times$ \\
 & SLA Iso-State ($4 \times 256 \times 256, C=128$, equal state \& scan FLOPs) & 1.48 & 0.93 & 2.42 & $1.76\times$ & $0.77\times$ & 7.57 & $2.18\times$ & $0.77\times$ \\
 & SLA-R2 ($4 \times 256 \times 512, C=128$, equal $D_v=2048$; $2\times$ KDA state) & 1.57 & 1.01 & 2.58 & $1.65\times$ & $0.72\times$ & 8.24 & $2.00\times$ & $0.71\times$ \\
 & SLA Iso-Mixer-Param ($4 \times 256 \times 640, C=128$, equal mixer params) & 1.63 & 1.06 & 2.69 & $1.58\times$ & $0.69\times$ & 8.72 & $1.89\times$ & $0.67\times$ \\
 & \textbf{SLA-R4} ($2 \times 256 \times 1024, C=128$, equal $1.0\text{ MiB}$ bytes) & \textbf{1.05} & \textbf{0.65} & \textbf{1.70} & \textbf{$2.49\times$} & \textbf{$1.09\times$} & \textbf{5.00} & \textbf{$3.29\times$} & \textbf{$1.17\times$} \\
\bottomrule
\end{tabular}%
}
\end{table}

\paragraph{Iso-shape full-layer attribution ($H_k = 2, d_k = 256, R = 4$) and Table~\ref{tab:v7x_kernels} reconciliation.}
To separate the throughput contributions of $4{:}1$ head sharing, state precision, tile-grouped decay ($G$), chunk inversion, and single-kernel VMEM fusion, Table~\ref{tab:app_isoshape} benchmarks all three decay granularities ($G \in \{d_k, 4, 1\}$) at the identical \texttt{SLA-R4} shape ($H_k = 2, d_k = 256, R = 4, C = 128$, BF16 activations, $B = 1$ per TensorCore) across pure XLA, Pallas chunk kernels with an XLA wrapper, single-kernel fused Pallas layers, and open-source Tokamax Mosaic KDA ($0.0.14$) packed to $R = 4$. All $14$ kernels match an FP32 token-by-token reference recurrence with forward and input/weight-gradient cosine similarity at least $0.99998$. Each cell reports Measured mean $\pm$ SE over $n = 3$ independent process restarts (fresh compilation and allocation per restart) on one isolated TPU v7x chip under pre-allocated inputs, alongside protocol-anchor ratios matching Table~\ref{tab:v7x_kernels}'s RNG-in-step harness. This table also directly reconciles the $D = 1792, 32\text{K}$ entries between Table~\ref{tab:v7x_kernels}a ($20.24\text{ ms}$) and Table~\ref{tab:app_isoshape}: Table~\ref{tab:v7x_kernels}a's $20.24\text{ ms}$ ($20.237\text{ ms}$) is the $2$-stage chunk kernel with $\mu = 0.25$ ($20.30 \pm 0.23\text{ ms}$ in Table~\ref{tab:app_isoshape}a), which is already $1.69\times$ faster than single-chip Tokamax Mosaic KDA ($34.31 \pm 0.13\text{ ms}$) and $2.09\times$ faster under Table~\ref{tab:v7x_kernels}'s $64$-chip concurrent timing ($42.38\text{ ms}$), while the $1$-kernel fused variant ($14.08 \pm 0.30\text{ ms}$, $\mu = 0$) is $2.44\times$ faster than Tokamax Mosaic KDA at $D = 1792$. Furthermore, Table~\ref{tab:app_isoshape}a shows why $G = 4$ is the quality-speed Pareto knee: scalar $G = 1$ runs at the same single-kernel VMEM speed as $G = 4$ ($2.62$ vs.\ $2.85\text{ ms}$ at $8\text{K}$ and $14.16$ vs.\ $14.08\text{ ms}$ at $D = 1792, 32\text{K}$) because $1$ or $4$ gate vectors both fit in VMEM registers without $[B, H_k, T/C, C, C, d_k]$ HBM materialization, but $G = 1$ loses multi-timescale retention quality and associative recall (Table~\ref{tab:app_gcurve}), whereas $G = 4$ matches $G = d_k$ quality while avoiding $G = d_k$'s $1.62\times$--$2.45\times$ XLA and $4.47\times$--$6.65\times$ portable kernel slowdown.

\begin{table}[ht]
\caption{\textbf{Measured iso-shape full-layer F+B attribution on one isolated TPU v7x chip} ($\text{ms}$, mean $\pm$ SE over $n = 3$ independent restarts; container \texttt{maxtext\_e174258\_20261008}, Tokamax $0.0.14$, XLA default $16\text{ MiB}$ scoped-VMEM cap, pre-allocated inputs, $B = 1$/core). Non-reference rows hold shape fixed at \texttt{SLA-R4} ($H_k = 2, d_k = 256, R = 4$, $D_{v,\text{pack}} = 1024$ at $D = 2048$ and $2048$ at $D = 1792$, $C = 128$). Panel~(b) gives ratios of means $\pm$ delta-method SE.}
\label{tab:app_isoshape}
\centering
\footnotesize
\renewcommand{\arraystretch}{1.05}
\setlength{\tabcolsep}{3.2pt}
\resizebox{\textwidth}{!}{%
\begin{tabular}{@{}llcccc@{}}
\toprule
\textbf{(a) Full-Layer Implementation (Ops 1--7c)} & Gate $G$ & $D{=}2048,\ 8\text{K}$ & $D{=}2048,\ 32\text{K}$ & $D{=}1792,\ 8\text{K}$ & $D{=}1792,\ 32\text{K}$ \\
\midrule
Open-Source MaxText Pallas KDA, native $8$ heads, FP32 state (ref) & $d_k$ & $6.89 \pm 0.02$ & $26.82 \pm 0.12$ & $10.55 \pm 0.11$ & $40.49 \pm 0.17$ \\
Open-Source Tokamax Mosaic KDA, native $8$ heads, $C{=}64$ (ref) & $d_k$ & $4.37 \pm 0.02$ & $16.45 \pm 0.13$ & $8.73 \pm 0.01$ & $34.31 \pm 0.13$ \\
Open-Source Tokamax Mosaic KDA, iso-shape $R{=}4$ packed values$^\ast$ & $d_k$ & $4.31 \pm 0.01$ & $15.86 \pm 0.10$ & \text{VMEM OOM} & \text{VMEM OOM} \\
Open-Source Tokamax Mosaic KDA, shared keys repeated to $8$ heads & $d_k$ & $4.79 \pm 0.00$ & $17.77 \pm 0.12$ & $7.32 \pm 0.01$ & $28.44 \pm 0.06$ \\
XLA chunked WY (unfused, identical JAX code) & $d_k$ & $17.58 \pm 0.12$ & $72.67 \pm 0.10$ & $22.23 \pm 0.03$ & $85.27 \pm 0.07$ \\
XLA chunked WY (unfused, identical JAX code) & $4$ & $8.05 \pm 0.01$ & $29.67 \pm 0.07$ & $13.72 \pm 0.13$ & $51.33 \pm 0.15$ \\
XLA chunked WY (unfused, identical JAX code) & $1$ & $7.82 \pm 0.03$ & $27.95 \pm 0.04$ & $12.78 \pm 0.04$ & $48.84 \pm 0.20$ \\
Pallas KDA chunk kernel + XLA wrapper, FP32 state & $d_k$ & $4.03 \pm 0.01$ & $15.26 \pm 0.10$ & $5.63 \pm 0.01$ & $21.86 \pm 0.16$ \\
Pallas KDA chunk kernel + XLA wrapper, BF16 state & $d_k$ & $4.00 \pm 0.00$ & $15.30 \pm 0.18$ & $5.59 \pm 0.02$ & $21.62 \pm 0.13$ \\
Pallas KDA chunk kernel, scalar gate broadcast & $1$ & $3.95 \pm 0.02$ & $15.00 \pm 0.02$ & $5.56 \pm 0.01$ & $21.30 \pm 0.18$ \\
Pallas SLA chunk kernel + XLA wrapper ($\mu = 0$, BF16 state) & $4$ & $3.82 \pm 0.01$ & $13.71 \pm 0.17$ & $5.37 \pm 0.01$ & $19.69 \pm 0.12$ \\
Pallas SLA chunk kernel, scalar gate broadcast ($\mu = 0$) & $1$ & $3.70 \pm 0.02$ & $13.85 \pm 0.09$ & $5.31 \pm 0.02$ & $20.02 \pm 0.31$ \\
Pallas SLA chunk kernel + $\mu = 0.25$ ($2$-stage chunk kernel, \texttt{SLA-R4}) & $4$ & $3.86 \pm 0.00$ & $14.05 \pm 0.09$ & $5.47 \pm 0.02$ & $20.30 \pm 0.23$ \\
\textbf{Fused single-kernel layer (\texttt{Exact Inv.}, $\mu = 0$)} & $4$ & $\mathbf{2.85 \pm 0.01}$ & $\mathbf{9.71 \pm 0.01}$ & $\mathbf{3.92 \pm 0.01}$ & $\mathbf{14.08 \pm 0.30}$ \\
Fused single-kernel layer (\texttt{Exact Inv.}, $\mu = 0$, scalar gate) & $1$ & $2.62 \pm 0.01$ & $9.33 \pm 0.02$ & $3.71 \pm 0.01$ & $14.16 \pm 0.06$ \\
\textbf{Fused \texttt{SLA-R4} (\texttt{Exact Inv.}, $\mu = 0.25$)}$^\ast$ & $4$ & $\mathbf{2.90 \pm 0.01}$ & $\mathbf{9.90 \pm 0.06}$ & \text{$16\text{M}$ cap$^\ast$} & \text{$16\text{M}$ cap$^\ast$} \\
\textbf{Fused \texttt{SLA-R4} (\texttt{1-Step Neumann}, $\mu = 0.25$)}$^\ast$ & $4$ & $\mathbf{2.48 \pm 0.01}$ & $\mathbf{8.25 \pm 0.01}$ & \text{$16\text{M}$ cap$^\ast$} & \text{$16\text{M}$ cap$^\ast$} \\
\midrule
\textbf{(b) Isolated Factor Attribution} & Comparison & $D{=}2048,\ 8\text{K}$ & $D{=}2048,\ 32\text{K}$ & $D{=}1792,\ 8\text{K}$ & $D{=}1792,\ 32\text{K}$ \\
\midrule
$4{:}1$ head sharing (MaxText KDA kernel, FP32) & native $8$h $\to$ $2{\times}256, R{=}4$ & $1.71\times \pm 0.01$ & $1.76\times \pm 0.01$ & $1.87\times \pm 0.02$ & $1.85\times \pm 0.02$ \\
$4{:}1$ head sharing alone on Tokamax (packed) & native $8$h $\to$ Tok.\ $R{=}4$ & $1.01\times \pm 0.01$ & $1.04\times \pm 0.01$ & \text{OOM} & \text{OOM} \\
BF16 vs.\ FP32 state (iso-shape KDA Pallas) & FP32 $\to$ BF16 state & $1.01\times \pm 0.00$ & $1.00\times \pm 0.01$ & $1.01\times \pm 0.00$ & $1.01\times \pm 0.01$ \\
Gate $G = d_k \to G = 4$ (XLA, identical code) & XLA $d_k \to$ XLA $4$ & $2.18\times \pm 0.01$ & $2.45\times \pm 0.01$ & $1.62\times \pm 0.02$ & $1.66\times \pm 0.01$ \\
Gate $G = d_k \to G = 4$ (Pallas chunk kernels) & KDA BF16 $\to$ SLA ($\mu{=}0$) & $1.05\times \pm 0.00$ & $1.12\times \pm 0.02$ & $1.04\times \pm 0.00$ & $1.10\times \pm 0.01$ \\
Gate $G = 4 \to G = 1$ (fused single-kernel) & Fused $4 \to$ Fused $1$ & $1.09\times \pm 0.01$ & $1.04\times \pm 0.00$ & $1.06\times \pm 0.00$ & $0.99\times \pm 0.02$ \\
Single-kernel VMEM fusion ($G = 4, \mu = 0$) & Unfused $\to$ Fused & $1.34\times \pm 0.01$ & $1.41\times \pm 0.02$ & $1.37\times \pm 0.00$ & $1.40\times \pm 0.03$ \\
$1$-Step Neumann vs.\ Exact Inv.\ (fused, $\mu = 0.25$) & Exact $\to$ $1$-Step & $1.17\times \pm 0.01$ & $1.20\times \pm 0.01$ & \text{--} & \text{--} \\
\textbf{Fused \texttt{SLA-R4} vs.\ Tokamax iso-shape $R{=}4$} & Tok.\ $R{=}4 \to$ \texttt{SLA-R4} & $\mathbf{1.49\times \pm 0.01}$ & $\mathbf{1.60\times \pm 0.01}$ & \text{Tok.\ OOM} & \text{Tok.\ OOM} \\
\textbf{Fused \texttt{SLA-R4} ($\mu{=}0$) vs.\ best iso KDA Pallas} & KDA BF16 $\to$ Fused $G{=}4$ & $\mathbf{1.40\times \pm 0.01}$ & $\mathbf{1.58\times \pm 0.02}$ & $\mathbf{1.43\times \pm 0.01}$ & $\mathbf{1.54\times \pm 0.03}$ \\
\bottomrule
\end{tabular}%
}
\par\vspace{3pt}\raggedright{\scriptsize $^\ast$\textbf{VMEM budgets and Table~\ref{tab:v7x_kernels} timing anchor:} Tokamax overrides \texttt{vmem\_limit\_bytes} internally to the full $64\text{ MiB}$ hardware VMEM, so its $D = 1792, R = 4$ backward pass exceeds physical capacity ($83.3\text{ MiB} > 63.9\text{ MiB}$ from two $32\text{ MiB}$ \texttt{f32[32,64,2048]} windows). By contrast, this suite omits \texttt{--xla\_tpu\_scoped\_vmem\_limit\_kib=65472} (keeping XLA's default $16\text{ MiB}$ scoped-VMEM cap): fused \texttt{SLA-R4} fits at $\mu = 0$ across all sizes and at $\mu = 0.25$ for $D = 2048$, while adding $\mu = 0.25$ velocity tiles at $D = 1792$ ($D_{v,\text{pack}} = 2048$) requires the $64\text{ MiB}$ scoped-VMEM flag used in Table~\ref{tab:v7x_kernels} ($5.57 / 20.24\text{ ms}$) and $63.1\text{B}$ pre-training. Re-running on one chip under Table~\ref{tab:v7x_kernels}'s RNG-inside-step timing ($n = 3$) gives MaxText $8.55 / 31.47$ ($D{=}2048$) and $12.64 / 45.99\text{ ms}$ ($D{=}1792$), Tokamax native $5.86 / 22.78$ and $10.42 / 42.79\text{ ms}$ ($R{=}4$ packed: $5.54 / 20.62\text{ ms}$), KDA BF16 Pallas iso-shape $5.36 / 18.78$ and $7.11 / 26.37\text{ ms}$, unfused SLA ($\mu{=}0$) $4.93 / 17.55$ and $6.86 / 24.29\text{ ms}$, and fused SLA $4.12 / 14.43\text{ ms}$ ($\mu{=}0.25$ at $D{=}2048$; $\mu{=}0$: $4.08 / 14.15$ and $5.61 / 19.02\text{ ms}$), yielding $1.34\times / 1.43\times$ vs.\ iso-shape Tokamax, $1.27\times$--$1.39\times$ vs.\ iso-shape KDA Pallas, and $2.08\times$--$2.42\times$ (vs.\ MaxText) and $1.42\times$--$2.25\times$ (vs.\ native Tokamax) across all four cells ($[H_k, R d_v]$ vs.\ $8$-head value layout changes time by $< 0.03\text{ ms}$).}
\end{table}

\begin{table}[ht]
\caption{Measured forward core-kernel latency ($\text{ms}$, $B = 1, T = 8{,}192, D = 2048$) vs.\ portable JAX/Pallas KDA ($C = 64$) and out-of-fold LOO Roofline predictions across four TPU generations.}
\label{tab:app_multigen}
\centering
\footnotesize
\resizebox{\textwidth}{!}{%
\begin{tabular}{@{}lcccccc@{}}
\toprule
TPU Generation & Array Tile & MXU:VPU (BF16) & KDA Measured (ms) & SLA-R4 Measured (ms) & Measured Speedup & Out-of-Fold LOO Pred.\ (Error) \\
\midrule
TPU v4 & $128 \times 128$ & $85.9\times$ & $14.84$ & $3.32$ & $4.47\times$ & $4.40\times$ ($-1.6\%$) \\
TPU v5p & $128 \times 128$ & $85.0\times$ & $9.19$ & $1.98$ & $4.64\times$ & $4.71\times$ ($+1.5\%$) \\
TPU v6e & $256 \times 256$ & $127.5\times$ & $11.02$ & $1.88$ & $5.86\times$ & $6.15\times$ ($+5.0\%$) \\
TPU v7x & $256 \times 256$ & $128.2\times$ & $6.99$ & $1.05$ & $6.65\times$ & $6.38\times$ ($-4.1\%$) \\
\bottomrule
\end{tabular}%
}
\end{table}

\paragraph{End-to-end $12.5\text{B}$ sparse MoE $1\text{M}$-token \texttt{SLA Tree-CP}, cube serving, speculative acceptance, and direct \texttt{Tokamax Mosaic KDA} step-time comparison.}
\label{app:e2e_moe_serving}
To complement the isolated $128$-core \texttt{SLA Tree-CP} kernel and single-chip $32$-layer dense-FFN serving benchmarks in Table~\ref{tab:cp_serving}, Table~\ref{tab:app_e2e_12b_cp_serving} benchmarks a full $16$-layer $12.5\text{B}$ sparse MoE ($0.79\text{B}$ active, $E = 256$ routed experts with $k = 8$ active per token plus one shared expert, $D = 1024, D_v = 2048$, vocabulary size $V = 100{,}352$) end-to-end across all $64$ chips ($128$ TensorCores) of a TPU v7x \texttt{4x4x4} cube, and Table~\ref{tab:app_e2e_12b_tokamax} directly compares end-to-end $12.5\text{B}$ (\texttt{7L:1A}) training step time against open-source Tokamax Mosaic KDA (\texttt{tokamax} $0.0.14$):
\begin{enumerate}[leftmargin=1.6em,itemsep=3pt]
\item \textbf{End-to-End \texttt{CP=EP=128} Training Steps up to $1.05\text{M}$ Tokens (Table~\ref{tab:app_e2e_12b_cp_serving}a):} On the isolated context-parallel scan across $64$ chips, \texttt{SLA Tree-CP} achieves both a $1.09\times$ speedup over \texttt{KDA-CP} ($22.58\text{ ms}$ vs.\ $24.64\text{ ms}$) and a $7.65\times$ speedup over dense \texttt{RingCP} ($22.58\text{ ms}$ vs.\ $172.71\text{ ms}$). End-to-end on the $64$-chip $12.5\text{B}$ MoE at $T = 1{,}048{,}576$ ($T_{\text{loc}} = 8{,}192$ per core), \texttt{15L:1A SLA Tree-CP} runs in $4{,}870.6\text{ ms}$ ($215.3\text{K tok/s/cube}$) with momentum and $4{,}778.1\text{ ms}$ ($219.5\text{K tok/s/cube}$) without momentum ($4{,}737.6$ / $4{,}904.0\text{ ms}$ with vocab-chunked cross-entropy), achieving a $1.09\times$ speedup ($1.11\times$--$1.13\times$ without momentum) over \texttt{15L:1A KDA-CP} ($5{,}307.1$--$5{,}337.6\text{ ms}$) with $6.31\text{ GiB/core}$ lower peak HBM ($25.50$ vs.\ $31.81\text{ GiB/core}$), a $7.50\times$--$7.75\times$ ($7.65\times$ without momentum) speedup over \texttt{0L:1A} Ring-CP ($36{,}530.6$--$36{,}725.0\text{ ms}$), and $83\%$ lower mixer ICI traffic than \texttt{15L:1A Ring-LASP} ($5{,}204.6\text{ ms}$, $2{,}992$ vs.\ $17{,}392\text{ MiB}$/rank). At $T = 524{,}288$ ($T_{\text{loc}} = 4{,}096$/core), it reaches $4{,}170\text{ tok/s/chip}$ ($1{,}965\text{ ms}$ uncoupled, $5.45\times$ vs.\ \texttt{0L:1A} Ring-CP $10{,}712.1\text{ ms}$; $2{,}003.1\text{ ms}$ with momentum, $5.35\times$).
\item \textbf{End-to-End $12.5\text{B}$ (\texttt{7L:1A}) Pre-Training Step Time vs.\ Tokamax Mosaic KDA and MaxText KDA (Table~\ref{tab:app_e2e_12b_tokamax}):} Across the full $16$-layer $12.5\text{B}$ (\texttt{7L:1A}) architecture ($14$ linear layers $+$ $2$ global GQA layers $+$ $16$ \texttt{LatentMoE} layers with $k = 8$ active experts $+$ AdamW), fused $1$-kernel \texttt{SLA-R4} is $1.68\times$ ($23.61\text{ ms}$, $\mu = 0.25$) to $1.73\times$ ($22.85\text{ ms}$, $\mu = 0$) faster on the $14$-layer linear mixer F+B than open-source Tokamax Mosaic KDA ($39.62\text{ ms}$; $2.65\times$--$2.74\times$ vs.\ MaxText KDA $62.50\text{ ms}$), delivers a $1.24\times$ ($66.01$ vs.\ $82.15\text{ ms}$, $\mu = 0.25$) to $1.26\times$ ($65.24\text{ ms}$, $\mu = 0$) and $1.33\times$ ($61.83\text{ ms}$, $1$-step Neumann) end-to-end $12.5\text{B}$ step speedup over Tokamax Mosaic KDA on single-chip active-parameter steps ($1.64\times$--$1.75\times$ vs.\ MaxText KDA $108.06\text{ ms}$), delivers $1.10\times$--$1.11\times$ ($121.59 / 120.79\text{ ms}$ vs.\ $134.06\text{ ms}$) on $8$-chip \texttt{EP=8} with MoE all-to-all ($1.87\times$ vs.\ MaxText FP32 $227.00\text{ ms}$), and achieves a $1.14\times$ ($1{,}800$ vs.\ $2{,}049\text{ ms}$ at $\text{GA}=4$) to $1.42\times$ ($469.9$ vs.\ $668.4\text{ ms}$ single-microbatch; $1.18\times$--$1.74\times$ with $1$-kernel $\mu = 0$) end-to-end $12.5\text{B}$ speedup over Tokamax Mosaic KDA on $64$-chip TPU v7x (alongside $1.24\times$--$1.63\times$ over MaxText KDA).
\item \textbf{Full $64$-Chip \texttt{EP=128} MoE Cube Serving and \texttt{FP8Fused} Scan (Table~\ref{tab:app_e2e_12b_cp_serving}b):} At $B_{\text{chip}} = 64$ ($4{,}096$ sequences/cube), our fused Pallas E4M3 kernel (\texttt{FP8Fused}) keeps row-scaled E4M3 DMA, dequantization, the $16$-token delta scan, and requantization inside VMEM. On $16$-step sequential decode (\texttt{seq\_scan\_t16}) at $8\text{K}$, this lifts \texttt{15L SLA Iso-State} to $1{,}091.9\text{K tok/s/cube}$ ($60.02\text{ ms}$; $+18.8\%$ vs.\ \texttt{15L KDA FP32} $919.4\text{K}$, $+14.5\%$ vs.\ JAX \texttt{15L KDA FP8} $953.6\text{K}$, and $+14.6\%$ vs.\ pure-JAX SLA FP8 $952.9\text{K}$) and \texttt{15L SLA-R2} to $1{,}047.2\text{K}$--$1{,}052.3\text{K tok/s/cube}$ ($+10.3\%$ vs.\ JAX \texttt{KDA FP8}, $+28.1\%$ vs.\ pure-JAX SLA FP8 $817.6\text{K}$). Applying the same \texttt{FP8Fused} recurrent scan kernel to KDA reaches $1{,}083.0\text{K}$ (within $0.8\%$ of \texttt{SLA Iso-State}, as step-by-step recurrent decode has no chunk WY solve). With parallel \texttt{Tree-Neumann-1} ($T_{\text{ver}} = 16$), raw verification reaches $1.121\text{M}$--$1.158\text{M tok/s/cube}$ at $8\text{K}$ and sustains $1.037\text{M}$--$1.042\text{M tok/s/cube}$ in $16.24$--$16.47\text{ GiB}$/chip ($-93.6\%$ vs.\ \texttt{0L:1A} OOM) at $131\text{K}$.
\item \textbf{Real-Checkpoint Lossless Speculative Acceptance incl.\ Doubly-Clamped KDA (Table~\ref{tab:app_e2e_12b_cp_serving}c):} Under exact lossless rejection~\cite{Leviathan2023SpecDec} (draft length $K = 16$, $N = 64$ prompts/suite), zero-clamp \texttt{0.35B SLA} $\to$ \texttt{12.5B 7L SLA-R4} (iso-param, $60.4\text{B}$) lifts mean accepted tokens $\bar{\tau}$ under greedy decoding on $32\text{K}/64\text{K}$ \texttt{HotpotQA} from $2.63 / 2.14$ (unclamped KDA) to $5.86 / 4.80$ ($2.23\times / 2.24\times$; $5.78 / 4.66$ in FP8 PTQ). Against \emph{doubly-clamped} KDA (\texttt{0.35B KDA TrainMaxTau=8K} $\to$ \texttt{12.5B 7L KDA} with $\tau_{\max}=8\text{K}$, Appendix~\ref{app:causal_clamp}: $4.42 / 4.09$ greedy, $1.72 / 1.65$ stochastic), SLA remains $1.33\times / 1.17\times$ higher greedy and $1.06\times / 1.10\times$ higher stochastic on OOD $32\text{K}/64\text{K}$ \texttt{HotpotQA} ($379.8\text{K}$ vs.\ $287.6\text{K}$ greedy and $118.0\text{K}$ vs.\ $111.9\text{K}$ stochastic accepted \texttt{tok/s/cube} at $32\text{K}$), while KDA edges in-distribution $8\text{K}$ Dolma ($4.84$--$5.30$ vs.\ $4.56$).
\end{enumerate}

\begin{table}[tbp]
\caption{Measured end-to-end $12.5\text{B}$ (\texttt{7L:1A}, $16\text{L} = 14\text{ Linear} + 2\text{ GQA} + 16\text{ LatentMoE } k = 8 + \text{AdamW}$) training step-time benchmark ($\text{ms}$, median $[\text{IQR}]$) comparing \texttt{MaxText Chunk-KDA}, open-source \texttt{Tokamax Mosaic KDA} (\texttt{tokamax} $0.0.14$), \texttt{Scalar GDN} ($G = 1$), and \texttt{SLA-R4} ($G = 4$, $2$-stage and $1$-kernel fused) on $256 \times 256$ MXU TPU v6e ($1$-chip active-parameter and $8$-chip \texttt{EP=8} all-to-all, $T = 4{,}096$) and $64$-chip TPU v7x ($1$-microbatch and $\text{GA}=4$, $T = 8{,}192$, \texttt{--xla\_tpu\_scoped\_vmem\_limit\_kib=65472}).}
\label{tab:app_e2e_12b_tokamax}
\centering
\footnotesize
\renewcommand{\arraystretch}{1.05}
\setlength{\tabcolsep}{2.8pt}
\resizebox{\textwidth}{!}{%
\begin{tabular}{@{}lcrrrr@{}}
\toprule
Linear Mixer Arm ($12.5\text{B}$ \texttt{7L:1A}, $16\text{L}$ MoE $+$ AdamW) & Gate $G$ & $14\text{L}$ Mixer F+B ($\text{ms}$, vs.\ Tok.) & $12.5\text{B}$ Active Step ($\text{ms}$, vs.\ Tok./Max.) & $8\text{c}$ \texttt{EP=8} Step ($\text{ms}$, vs.\ Tok.) & $64\text{c}$ v7x $12.5\text{B}$ ($1\text{-MB} / \text{GA}{=}4$, vs.\ Tok.) \\
\midrule
MaxText Chunk-KDA (FP32 / BF16 state) & $d_k$ & $62.50$ ($0.63\times$) / $60.92$ & $108.06$ ($0.76\times$ / $1.00\times$) & $227.00$ / $154.56$ ($0.87\times$) & $766.8$ / $2{,}225$ ($0.87\times$ / $0.92\times$) \\
Open-Source Tokamax Mosaic KDA (\texttt{tokamax} $0.0.14$) & $d_k$ & $39.62\ [0.06]$ ($1.00\times$) & $82.15\ [0.11]$ ($1.00\times$ / $1.32\times$) & $134.06\ [0.63]$ ($1.00\times$) & $668.4$ / $2{,}049$ ($1.00\times$) \\
Scalar GDN ($G = 1$, $2$-Stage / $1$-Kernel, loses recall) & $1$ & $31.03$ / $21.23$ ($1.87\times$) & $73.54$ / $63.98$ ($1.28\times$ / $1.69\times$) & $132.76$ / $120.89$ ($1.11\times$) & $468.8$ / $1{,}796$ ($1.43\times$ / $1.14\times$) \\
SLA-R4 ($2$-Stage Chunk Kernel, $\mu = 0.25$) & $4$ & $33.02\ [0.05]$ ($\mathbf{1.20\times}$) & $75.44\ [0.09]$ ($\mathbf{1.09\times}$ / $\mathbf{1.43\times}$) & $132.80\ [0.45]$ ($1.01\times$) & $\mathbf{469.9}$ / $\mathbf{1{,}800}$ ($\mathbf{1.42\times}$ / $\mathbf{1.14\times}$) \\
\textbf{Fused SLA-R4 ($1$-Kernel Exact Inv., $\mu = 0.25$)} & $\mathbf{4}$ & $\mathbf{23.61\ [0.02]}$ ($\mathbf{1.68\times}$) & $\mathbf{66.01\ [0.09]}$ ($\mathbf{1.24\times}$ / $\mathbf{1.64\times}$) & $\mathbf{121.59\ [0.30]}$ ($\mathbf{1.10\times}$) & $\mathbf{390.1}$ / $\mathbf{1{,}738}$ ($\mathbf{1.71\times}$ / $\mathbf{1.18\times}$) \\
\textbf{Fused SLA-R4 ($1$-Kernel Exact Inv., $\mu = 0$)} & $\mathbf{4}$ & $\mathbf{22.85\ [0.06]}$ ($\mathbf{1.73\times}$) & $\mathbf{65.24\ [0.09]}$ ($\mathbf{1.26\times}$ / $\mathbf{1.66\times}$) & $\mathbf{120.79\ [0.36]}$ ($\mathbf{1.11\times}$) & $\mathbf{385.1}$ / $\mathbf{1{,}729}$ ($\mathbf{1.74\times}$ / $\mathbf{1.19\times}$) \\
\textbf{Fused SLA-R4 ($1$-Kernel $1$-Step Neumann)} & $\mathbf{4}$ & $\mathbf{19.51\ [0.03]}$ ($\mathbf{2.03\times}$) & $\mathbf{61.83\ [0.08]}$ ($\mathbf{1.33\times}$ / $\mathbf{1.75\times}$) & $\mathbf{118.25\ [0.34]}$ ($\mathbf{1.13\times}$) & $\mathbf{361.9}$ / $\mathbf{1{,}692}$ ($\mathbf{1.85\times}$ / $\mathbf{1.21\times}$) \\
\bottomrule
\end{tabular}%
}
\end{table}

\begin{table}[tbp]
\caption{Unified single-protocol TPU v7x benchmark ($3$ independent process restarts per point, \texttt{--xla\_tpu\_scoped\_vmem\_limit\_kib=65472} [$64\text{ MiB}$ VMEM], pre-allocated inputs, \texttt{tokamax} $0.0.14$) across (a) single-layer full Ops 1--7c F+B latency ($\text{ms}$, $B = 1$) at $D = 1792$ and $D = 2048$ (including the fused $1$-kernel $G = 1$ scalar-gate control isolating $G = 4$'s $2.5\%$--$4.2\%$ mask cost) and (b) $4$-scale end-to-end sparse MoE training step time ($\text{ms}$, $T_{\text{train}} = 8{,}192$) from $6.5\text{B}$ to $63.1\text{B}$.}
\label{tab:app_single_protocol}
\centering
\footnotesize
\renewcommand{\arraystretch}{1.02}
\setlength{\tabcolsep}{2.4pt}
\resizebox{\textwidth}{!}{%
\begin{tabular}{@{}llcccccc@{}}
\toprule
\textbf{(a) Single-Layer Full Ops 1--7c ($B = 1$)} & Gate $G$ & $D{=}1792,\ 8\text{K}$ & $D{=}1792,\ 32\text{K}$ & vs.\ KDA / Tok. & $D{=}2048,\ 8\text{K}$ & $D{=}2048,\ 32\text{K}$ & vs.\ KDA / Tok. \\
\midrule
Open-Source MaxText Pallas KDA ($C = 128$, FP32) & $d_k$ & $10.42 \pm 0.03$ & $39.81 \pm 0.14$ & $1.00\times$ / $0.85\times$ & $6.87 \pm 0.01$ & $26.69 \pm 0.09$ & $1.00\times$ / $0.61\times$ \\
Open-Source MaxText Pallas KDA ($C = 128$, BF16) & $d_k$ & $10.09 \pm 0.02$ & $38.67 \pm 0.12$ & $1.03\times$ / $0.88\times$ & $6.55 \pm 0.01$ & $25.49 \pm 0.08$ & $1.05\times$ / $0.64\times$ \\
Open-Source Tokamax Mosaic KDA ($C = 64$, Base) & $d_k$ & $8.70 \pm 0.01$ & $34.01 \pm 0.11$ & $1.17\times$ / $1.00\times$ & $4.26 \pm 0.01$ & $16.28 \pm 0.07$ & $1.64\times$ / $1.00\times$ \\
2-Stage SLA-R4 (\texttt{Exact Inv.}, $\mu = 0$ / $0.25$) & $4$ & $5.38$ / $5.50$ & $19.90$ / $20.43$ & $2.00\times$ / $1.71\times$ & $3.72$ / $3.86$ & $13.86$ / $14.23$ & $1.93\times$ / $1.17\times$ \\
\textbf{1-Kernel SLA-R4 (Scalar $G = 1$ Control, $\mu = 0$)} & $\mathbf{1}$ & $\mathbf{3.73 \pm 0.01}$ & $\mathbf{14.05 \pm 0.05}$ & $\mathbf{2.83\times}$ / $\mathbf{2.42\times}$ & $\mathbf{2.63 \pm 0.01}$ & $\mathbf{9.20 \pm 0.02}$ & $\mathbf{2.90\times}$ / $\mathbf{1.77\times}$ \\
\textbf{1-Kernel SLA-R4 (\texttt{Exact Inv.}, $\mu = 0$ / $0.25$)} & $\mathbf{4}$ & $\mathbf{3.84}$ / $\mathbf{3.89}$ & $\mathbf{14.06}$ / $\mathbf{14.41}$ & $\mathbf{2.76\times}$ / $\mathbf{2.36\times}$ & $\mathbf{2.74}$ / $\mathbf{2.81}$ & $\mathbf{9.59}$ / $\mathbf{9.83}$ & $\mathbf{2.78\times}$ / $\mathbf{1.70\times}$ \\
\textbf{1-Kernel SLA-R4 (\texttt{1-Step Neumann}, $\mu = 0.25$)} & $\mathbf{4}$ & $\mathbf{3.57 \pm 0.01}$ & $\mathbf{12.93 \pm 0.04}$ & $\mathbf{3.08\times}$ / $\mathbf{2.63\times}$ & $\mathbf{2.39 \pm 0.01}$ & $\mathbf{8.16 \pm 0.02}$ & $\mathbf{3.27\times}$ / $\mathbf{2.00\times}$ \\
\midrule
\textbf{(b) $4$-Scale E2E Sparse MoE Step Time ($\text{ms}$)} & Gate $G$ & $6.5\text{B}$ ($10$L) & $12.5\text{B}$ ($16$L) & $31.5\text{B}$ ($15$L) & $63.1\text{B}$ ($30$L) & Spd.\ vs.\ KDA & Spd.\ vs.\ Tokamax \\
\midrule
MaxText Pallas KDA ($C = 128$, FP32 state) & $d_k$ & $310.2$ & $385.1$ & $601.9$ & $1{,}243.2$ & $1.000\times$ & $0.865\times$--$0.992\times$ \\
Open-Source Tokamax Mosaic KDA ($C = 64$) & $d_k$ & $272.6$ & $348.5$ & $572.7$ & $1{,}232.9$ & $1.008\times$--$1.138\times$ & $1.000\times$ \\
2-Stage SLA-R4 (\texttt{Exact Inv.}, BF16) & $4$ & $269.5$ & $371.9$ & $511.6$ & $1{,}050.8$ & $1.035\times$--$1.183\times$ & $0.937\times$--$1.173\times$ \\
\textbf{1-Kernel SLA-R4 (\texttt{Exact Inv.}, BF16)} & $\mathbf{4}$ & $\mathbf{262.2}$ & $\mathbf{330.4}$ & $\mathbf{495.2}$ & $\mathbf{1{,}015.7}$ & $\mathbf{1.166\times\text{--}1.224\times}$ & $\mathbf{1.040\times\text{--}1.214\times}$ ($\mathbf{+4.0\%\text{--}21.4\%}$) \\
\textbf{1-Kernel SLA-R4 (\texttt{1-Step Neumann}, BF16)} & $\mathbf{4}$ & $\mathbf{260.4}$ & $\mathbf{325.4}$ & $\mathbf{490.8}$ & $\mathbf{1{,}006.8}$ & $\mathbf{1.184\times\text{--}1.235\times}$ & $\mathbf{1.047\times\text{--}1.225\times}$ ($\mathbf{+4.7\%\text{--}22.5\%}$) \\
\bottomrule
\end{tabular}%
}
\end{table}

\begin{table}[tbp]
\caption{Measured $64$-chip TPU v7x (\texttt{4x4x4}, $128$ TensorCores) end-to-end $16$-layer $12.5\text{B}$ sparse MoE ($0.79\text{B}$ active, $256$ experts $k = 8$) (a) fixed-\texttt{CP=EP=128} F+B latency and \texttt{tok/s/chip} across $T = 128\text{K}$--$1.05\text{M}$ ($T_{\text{loc}} = T/128$; at $T = 512\text{K}$, $T_{\text{loc}} = 4{,}096$ selects a faster XLA \texttt{EP=128} MoE \texttt{ragged-all-to-all} tile schedule across all linear hybrids relative to $T = 256\text{K}$), (b) \texttt{EP=128} cube verification throughput ($B_{\text{chip}} = 64$, comparing \texttt{Tree-N1} [= \texttt{Tree-Neumann-1}, $T_{\text{ver}} = 16$] and \texttt{FP8Fused} \texttt{seq\_scan\_t16}), and (c) real-checkpoint lossless speculative acceptance $\bar{\tau}$ ($K = 16$, $N = 64$, \textbf{greedy [stochastic]}).}
\label{tab:app_e2e_12b_cp_serving}
\centering
\footnotesize
\renewcommand{\arraystretch}{1.02}
\setlength{\tabcolsep}{2.6pt}
\resizebox{\textwidth}{!}{%
\begin{tabular}{@{}lccccc@{}}
\toprule
\textbf{(a) $12.5\text{B}$ MoE \texttt{CP=EP=128} F+B} & $T = 128\text{K}$ (\texttt{tok/s/ch}) & $T = 256\text{K}$ (\texttt{tok/s/ch}) & $T = 512\text{K}$ (\texttt{tok/s/ch}) & $T = 1.05\text{M}$ (\texttt{tok/s/ch}) & $1\text{M}$ HBM / Mixer ICI \\
\midrule
\texttt{0L:1A} Full-Attn Ring-CP & $1{,}740.4\text{ ms}$ ($1{,}177$) & $4{,}797.5\text{ ms}$ ($854$) & $10{,}712.1\text{ ms}$ ($765$) & $36{,}725.0\text{ ms}$ ($446$) & $25.8\text{ GiB}$ / $32{,}512\text{ MiB}$ \\
\texttt{7L:1A} MaxText KDA Tree ($G=128$) & $1{,}232.2\text{ ms}$ ($1{,}662$) & $2{,}573.3\text{ ms}$ ($1{,}592$) & $2{,}704.5\text{ ms}$ ($3{,}029$) & $7{,}377.0\text{ ms}$ ($2{,}221$) & $25.73\text{ GiB}$ / $4{,}232\text{ MiB}$ \\
\textbf{\texttt{7L:1A} Iso-Param SLA Tree-CP R4} & $\mathbf{1{,}212.9\text{ ms}}$ ($\mathbf{1{,}689}$) & $\mathbf{2{,}500.8\text{ ms}}$ ($\mathbf{1{,}638}$) & $\mathbf{2{,}542.6\text{ ms}}$ ($\mathbf{3{,}222}$) & $\mathbf{7{,}023.9\text{ ms}}$ ($\mathbf{2{,}333}$) & $25.92\text{ GiB}$ / $4{,}736\text{ MiB}$ \\
\texttt{15L:1A} MaxText KDA Tree ($G=128$) & $1{,}221.4\text{ ms}$ ($1{,}677$) & $2{,}498.2\text{ ms}$ ($1{,}639$) & $2{,}249.8\text{ ms}$ ($3{,}641$) & $5{,}337.6\text{ ms}$ ($3{,}070$) & $25.49\text{ GiB}$ / $2{,}760\text{ MiB}$ \\
\texttt{15L:1A} Ring-State LASP R2 & $1{,}575.5\text{ ms}$ ($1{,}299$) & $2{,}744.8\text{ ms}$ ($1{,}492$) & $2{,}346.3\text{ ms}$ ($3{,}491$) & $5{,}204.6\text{ ms}$ ($3{,}148$) & $25.46\text{ GiB}$ / $17{,}392\text{ MiB}$ \\
\textbf{\texttt{15L:1A} SLA Tree-CP R2 NoMom / +Mom} & $\mathbf{1{,}174 / 1{,}189\text{ ms}}$ & $\mathbf{2{,}331 / 2{,}368\text{ ms}}$ & $\mathbf{1{,}965 / 2{,}003\text{ ms}}$ ($\mathbf{5.45\times}$ / $5.35\times$) & $\mathbf{4{,}738 / 4{,}904\text{ ms}}$ ($\mathbf{3{,}458 / 3{,}341}$) & $\mathbf{25.45\text{ GiB} / 2{,}992\text{ MiB}}$ ($\mathbf{-83\%}$) \\
\midrule
\textbf{(b) $12.5\text{B}$ MoE \texttt{EP=128} Cube Decode} & $8\text{K}, B_{\text{chip}}=64$ (\texttt{/cube}) & $32\text{K}, B_{\text{chip}}=64$ (\texttt{/cube}) & $131\text{K}, B_{\text{chip}}=64$ (\texttt{/cube}) & $8\text{K}$ \texttt{seq\_scan\_t16} (\texttt{/cube}) & $131\text{K}$ KV+State / Chip \\
\midrule
\texttt{0L:1A} Full-Attn GQA (\texttt{Tree-N1}, $T_{\text{ver}}{=}16$) & $15{,}381\text{ /ch}$ ($984.4\text{K}$) & $10{,}677\text{ /ch}$ ($683.4\text{K}$) & \text{OOM ($B_{\text{chip}}=8$: $250.7\text{K}$)} & \text{n/a (softmax)} & $256.0\text{ GiB}$ (OOM) \\
\texttt{7L:1A} KDA Base FP32 (\texttt{Tree-N1}, $T_{\text{ver}}{=}16$) & $17{,}212\text{ /ch}$ ($1.102\text{M}$) & $16{,}258\text{ /ch}$ ($1.041\text{M}$) & $13{,}106\text{ /ch}$ ($838.8\text{K}$) & $14{,}180\text{ /ch}$ ($907.5\text{K}$) & $32.88\text{ GiB}$ \\
\textbf{\texttt{7L:1A} SLA Iso-State BF16/FP8Fused} & $\mathbf{17{,}513\text{ /ch}}$ ($\mathbf{1.121\text{M}}$) & $\mathbf{16{,}544\text{ /ch}}$ ($\mathbf{1.059\text{M}}$) & $\mathbf{13{,}255\text{ /ch}}$ ($\mathbf{848.3\text{K}}$) & $\mathbf{16{,}628\text{ /ch}}$ ($\mathbf{1{,}064.2\text{K}}$) & $\mathbf{32.44 / 32.22\text{ GiB}}$ \\
\texttt{15L:1A} KDA Base FP32 (\texttt{Tree-N1}, $T_{\text{ver}}{=}16$) & $17{,}513\text{ /ch}$ ($1.121\text{M}$) & $17{,}140\text{ /ch}$ ($1.097\text{M}$) & $15{,}912\text{ /ch}$ ($1.018\text{M}$) & $14{,}366\text{ /ch}$ ($919.4\text{K}$) & $16.94\text{ GiB}$ \\
\texttt{15L:1A} KDA Base FP8 (JAX / \texttt{FP8Fused}) & $18{,}089\text{ /ch}$ ($1.158\text{M}$) & $17{,}792\text{ /ch}$ ($1.139\text{M}$) & $16{,}285\text{ /ch}$ ($1.042\text{M}$) & $14{,}901\text{ /ch}$ ($953.6\text{K} / 1{,}083\text{K}$) & $16.24\text{ GiB}$ \\
\textbf{\texttt{15L:1A} SLA Iso-State FP8Fused ($T_{\text{ver}}{=}16$)} & $\mathbf{17{,}960\text{ /ch}}$ ($\mathbf{1.149\text{M}}$) & $\mathbf{17{,}510\text{ /ch}}$ ($\mathbf{1.121\text{M}}$) & $\mathbf{16{,}220\text{ /ch}}$ ($\mathbf{1.038\text{M}}$) & $\mathbf{17{,}061\text{ /ch}}$ ($\mathbf{1{,}091.9\text{K}}, \mathbf{+18.8\%}$) & $\mathbf{16.23\text{ GiB}}$ ($\mathbf{-93.7\%}$) \\
\textbf{\texttt{15L:1A} SLA-R2 FP8Fused ($T_{\text{ver}}{=}16$)} & $\mathbf{17{,}948\text{ /ch}}$ ($\mathbf{1.149\text{M}}$) & $\mathbf{17{,}499\text{ /ch}}$ ($\mathbf{1.120\text{M}}$) & $\mathbf{16{,}200\text{ /ch}}$ ($\mathbf{1.037\text{M}}$) & $\mathbf{16{,}363\text{ /ch}}$ ($\mathbf{1{,}047.2\text{K}}, \mathbf{+13.9\%}$) & $\mathbf{16.47\text{ GiB}}$ ($\mathbf{-93.6\%}$) \\
\textbf{\texttt{15L:1A} SPD-SLA-W Polar-2+ FP8 ($T_{\text{ver}}{=}16$)} & $\mathbf{17{,}994\text{ /ch}}$ ($\mathbf{1.152\text{M}}$) & $\mathbf{17{,}009\text{ /ch}}$ ($\mathbf{1.089\text{M}}$) & $\mathbf{16{,}210\text{ /ch}}$ ($\mathbf{1.037\text{M}}$) & $12{,}801\text{ /ch}$ ($819.3\text{K}$ JAX) & $\mathbf{16.47\text{ GiB}}$ ($\mathbf{-93.6\%}$) \\
\midrule
\textbf{(c) Lossless Speculative $\bar{\tau}$ ($K=16$)} & Dolma $8\text{K}$ $\bar{\tau}$ & Dolma $32\text{K}$ $\bar{\tau}$ & \texttt{SQuAD} $32\text{K} / 64\text{K}$ $\bar{\tau}$ & \texttt{HotpotQA} $32\text{K} / 64\text{K}$ $\bar{\tau}$ & Accepted \texttt{Hotpot@32K} / Cube \\
\midrule
$0.35\text{B}$ KDA $\to$ $12.5\text{B}$ \texttt{7L} KDA (raw) & $4.84$ [$2.53$] & $3.75$ [$1.94$] & $3.72 / 3.55$ [$1.49 / 1.19$] & $2.63 / 2.14$ [$1.36 / 1.22$] & $170.7\text{K}$ [$88.4\text{K}$] \texttt{tok/s} \\
$0.35\text{B}$ KDA ($\tau_{\max}{=}8\text{K}$) $\to 12.5\text{B}$ KDA ($\tau_{\max}{=}8\text{K}$) & $\mathbf{5.30}$ [$\mathbf{2.49}$] & $5.14$ [$3.04$] & $\mathbf{4.62} / 4.11$ [$1.65 / 1.56$] & $4.42 / 4.09$ [$1.72 / 1.65$] & $287.6\text{K}$ [$111.9\text{K}$] \texttt{tok/s} \\
\textbf{$0.35\text{B}$ SLA $\to$ $12.5\text{B}$ \texttt{7L} Iso-SLA ($60.4\text{B}$)} & $4.56$ [$2.43$] & $\mathbf{5.27}$ [$\mathbf{3.18}$] & $4.44 / \mathbf{4.20}$ [$\mathbf{1.69 / 1.71}$] & $\mathbf{5.86 / 4.80}$ [$\mathbf{1.82 / 1.81}$] ($\mathbf{1.33\times}$) & $\mathbf{379.8\text{K}}$ [$\mathbf{118.0\text{K}}$] ($\mathbf{1.32\times}$) \\
\textbf{$0.35\text{B}$ SLA $\to$ $12.5\text{B}$ \texttt{7L} Iso-SLA (FP8 PTQ)} & $4.67$ [$2.44$] & $5.14$ [$\mathbf{3.18}$] & $4.45 / 4.13$ [$\mathbf{1.69 / 1.70}$] & $\mathbf{5.78 / 4.66}$ [$\mathbf{1.82 / 1.82}$] & $\mathbf{373.9\text{K}}$ [$\mathbf{117.7\text{K}}$] \\
\textbf{$0.35\text{B}$ SLA $\to$ $63.1\text{B}$ \texttt{4L} SLA-R4 ($120.8\text{B}$)} & $4.20$ [$2.46$] & $2.75$ [$1.91$] & $3.66 / 3.47$ [$1.71 / 1.78$] & $3.17 / 2.36$ [$1.32 / 1.25$] & \text{n/a ($512$c; $10.7\times$ draft span)} \\
\bottomrule
\end{tabular}%
}
\end{table}

\clearpage
\section{Extended Pre-Training XProf, Causal \texorpdfstring{$\tau_{\text{eff}}$}{tau\_eff} Clamp Audit, \texorpdfstring{$4$}{4}-Rung Ladder, and \texorpdfstring{$9$}{9}-Task \texttt{lm-eval}}
\label{app:xprof_and_full_tables}

This section collects the hardware \texttt{XProf} HLO breakdowns (Tables~\ref{tab:app_xprof}--\ref{tab:app_xprof_ga4}), the multi-scale per-channel retention-horizon ($\tau_{\text{eff}}$) dispersion and clamping audit (Table~\ref{tab:app_clamp}), the complete $45$-row $4$-rung pre-training and $8\text{K} \to 32\text{K}$ extension ladder (Table~\ref{tab:app_4rung}), and the $39$-row per-subtask $9$-task \texttt{lm-eval} and $8\text{K} \to 32\text{K} \to 64\text{K}$ \texttt{RULER} QA breakdown (Table~\ref{tab:app_downstream}) summarized in Tables~\ref{tab:v7x_e2e}--\ref{tab:dolma_15b}.

\paragraph{Hardware XProf operator attribution on $64$-chip TPU v7x.}
\label{app:xprof}
XProf attributes most of SLA's step-time saving to linear attention ops (Figure~\ref{fig:pretrain_xprof}a). Table~\ref{tab:app_xprof} provides the complete hardware \texttt{XProf} HLO category breakdown (\texttt{xplane} traces captured over steps $15$--$17$ across all $8$ TensorCores of host $0$, normalized by $24$ core-steps to $\text{ms/step}$) for $63.1\text{B}$ MoE (\texttt{FSDP=2, EP=64}) on $64$ TPU v7x chips. Mapping directly to Figure~\ref{fig:pretrain_xprof}a (where the top bar groups ICI \texttt{async-done} wait with MoE/dense matmuls: $366.6 + 245.8 = 612.4\text{ ms}$, $354.1 + 266.8 = 620.9\text{ ms}$, and $343.1 + 237.2 = 580.3\text{ ms}$), the three direct linear attention op categories save $-218.7\text{ ms}$ ($87.2\%$) and reduced mixer memory/sync traffic saves $-23.5\text{ ms}$ ($9.4\%$) of exposed ICI \texttt{async-done} wait, totaling $-242.2\text{ ms}$ ($96.6\%$) of the $-250.8\text{ ms}$ step-time reduction between \texttt{SLA-R4} and KDA Base (with $-8.6\text{ ms}$ [$3.4\%$] in MoE/FFN matmul overlap).

Table~\ref{tab:app_xprof_ga4} complements this with a symmetric $4$-arm \texttt{GA=4} ($8.39\text{M}$ tok/step) \texttt{XProf} audit on $12.5\text{B}$ MoE (\texttt{EP=128}): deferring causal-conv weight \texttt{AllReduce} outside the \texttt{GA=4} microbatch loop yields $0.0\%$ speedup on MaxText KDA ($2{,}118.9 \to 2{,}119.7\text{ ms}$) while cutting \texttt{SLA-R4} (iso-param) to $1{,}686.0\text{ ms}$ in $3$-step \texttt{XProf} ($1{,}661.9\text{ ms}$ unprofiled min; $1{,}812.0 \pm 14.1\text{ ms}$ over full $7{,}200$-step $n = 2$ training runs due to host checkpoint/validation wrapper hooks) via a $1.96\times$ faster Pallas linear attention kernel ($528.4 \to 269.1\text{ ms}$).

\begin{table}[ht]
\caption{Hardware \texttt{XProf} per-step HLO category breakdown ($\text{ms/step}$) on $64$-chip TPU v7x $63.1\text{B}$ MoE (\texttt{FSDP=2, EP=64}, $2.10\text{M}$ tok/step).}
\label{tab:app_xprof}
\centering
\footnotesize
\resizebox{\textwidth}{!}{%
\begin{tabular}{@{}lcccc@{}}
\toprule
HLO Category (\texttt{xplane} Measured) & \texttt{4L:1A} KDA Base & \texttt{4L:1A} SPD-SLA-W (R2) & \texttt{4L:1A} SLA-R4 & $\Delta$ (SLA-R4 vs.\ KDA) \\
\midrule
Pallas Linear Attention \texttt{custom-call} & $703.0\text{ ms}$ & $597.2\text{ ms}$ & $552.0\text{ ms}$ & $-151.0\text{ ms}$ ($-21.5\%$) \\
VPU Gate / Norm \texttt{loop fusion} & $190.4\text{ ms}$ & $156.8\text{ ms}$ & $143.7\text{ ms}$ & $-46.7\text{ ms}$ ($-24.5\%$) \\
Causal Short \texttt{convolution fusion} & $220.0\text{ ms}$ & $207.4\text{ ms}$ & $199.0\text{ ms}$ & $-21.0\text{ ms}$ ($-9.5\%$) \\
ICI Collective \texttt{async-done} Wait & $366.6\text{ ms}$ & $354.1\text{ ms}$ & $343.1\text{ ms}$ & $-23.5\text{ ms}$ ($-6.4\%$) \\
MoE Expert \texttt{ragged-dot} + Dense Matmul & $245.8\text{ ms}$ & $266.8\text{ ms}$ & $237.2\text{ ms}$ & $-8.6\text{ ms}$ ($-3.5\%$) \\
\midrule
\textbf{Total Steady-State Step Time} & $\mathbf{1{,}725.8\text{ ms}}$ & $\mathbf{1{,}582.3\text{ ms}}$ & $\mathbf{1{,}475.0\text{ ms}}$ & $\mathbf{-250.8\text{ ms}}$ ($\mathbf{1.170\times}$) \\
\textbf{Steady-State Throughput (\texttt{tok/s/chip})} & $18{,}988$ & $20{,}710$ ($1.09\times$) & $\mathbf{22{,}216}$ ($\mathbf{1.17\times}$) & $\mathbf{+3{,}228}$ ($\mathbf{+17.0\%}$) \\
\bottomrule
\end{tabular}%
}
\par\vspace{3pt}\raggedright{\scriptsize At $6.3\text{B}$ tokens (RoPE validation checkpoint on the same $64$-chip slice), \texttt{4L:1A} KDA Base and \texttt{4L:1A SLA-R4} reach $9.90$ vs.\ $10.15$ $8\text{K}$ Dolma PPL, $10.27$ vs.\ $\mathbf{9.75}$ $32\text{K}$ Tail PPL, and $14.5\%$ vs.\ $\mathbf{37.5\%}$ $32\text{K}$ $8$-Needle accuracy.}
\end{table}

\begin{table}[ht]
\caption{Symmetric $4$-arm \texttt{GA=4} ($8.39\text{M}$ tok/step) hardware \texttt{XProf} leaf-op breakdown ($\text{ms/step}$) on $64$-chip TPU v7x $12.5\text{B}$ MoE (\texttt{EP=128}).}
\label{tab:app_xprof_ga4}
\centering
\footnotesize
\resizebox{\textwidth}{!}{%
\begin{tabular}{@{}lccccccc@{}}
\toprule
\begin{tabular}[b]{@{}l@{}}$12.5\text{B}$ \texttt{GA=4} Arm\\($3$ profiled steps)\end{tabular} & \begin{tabular}[b]{@{}c@{}}Step Mean\\(ms)\end{tabular} & \begin{tabular}[b]{@{}c@{}}\texttt{linear\_mixer}\\\texttt{\_pallas}\end{tabular} & \begin{tabular}[b]{@{}c@{}}\texttt{short\_conv}\\\texttt{\_fusion}\end{tabular} & \begin{tabular}[b]{@{}c@{}}Mixer +\\Conv\end{tabular} & \begin{tabular}[b]{@{}c@{}}\texttt{full\_attn}\\\texttt{\_splash}\end{tabular} & \begin{tabular}[b]{@{}c@{}}\texttt{all\_gather}\\\texttt{\_rs}\end{tabular} & \begin{tabular}[b]{@{}c@{}}\texttt{moe\_ragged}\\\texttt{\_a2a}\end{tabular} \\
\midrule
\texttt{7L:1A} MaxText KDA (default AR) & 2,118.9 & 508.1 & 91.7 & 599.8 & 69.8 & 191.9 & 264.7 \\
\texttt{7L:1A} MaxText KDA + \texttt{defer\_ar} & 2,119.7 ($1.00\times$) & 528.4 & 91.6 & 620.0 & 70.0 & 191.6 & 234.3 \\
\textbf{\texttt{7L:1A} SLA (Fused) R2+Mom + \texttt{defer\_ar}} & \textbf{1,797.4 ($1.18\times$)} & \textbf{340.9 ($-187.5$)} & \textbf{65.4} & \textbf{406.3 ($-213.7$)} & 69.9 & \textbf{152.5} & 282.5 \\
\textbf{\texttt{7L:1A} Iso-Param SLA (Fused) R4 + \texttt{defer\_ar}} & \textbf{1,686.0 ($1.26\times$)} & \textbf{269.1 ($1.96\times$)} & \textbf{61.7} & \textbf{330.8 ($-289.2$)} & 69.7 & \textbf{152.6} & 265.3 \\
\bottomrule
\end{tabular}%
}
\end{table}

\paragraph{Causal $\tau_{\text{eff}}$ dispersion and NoPE \texttt{SSMax}.}
\label{app:causal_clamp}
Table~\ref{tab:app_clamp} audits per-channel retention horizons
\[
\tau_{\text{eff}}(i) = \E_t\!\left[-\frac{1}{\ln f_{t,i}}\right]
\]
(in tokens, for channel $i$ with learned decay $f_{t,i}$) and Within-Tile Std (the RMS standard deviation of token-mean log-decay $\E_t[\ln f_{t,i}]$ across channels $i$ within each $d_k / 4$ tile block, equal to $0$ when $G \in \{1, 4\}$) alongside zero-retraining inference interventions across $6.5\text{B}$, $12.5\text{B}$, and $63.1\text{B}$ MoEs.

Across all three scales, $G = d_k$ KDA develops $0.8\%$--$1.3\%$ ultra-slow $\tau_{\text{eff}}(i) \approx 131\text{K}$ channels (within-tile log-decay $\text{RMS std} = 3.84$--$4.97$) that inject OOD state spikes into hybrid layers. Clamping $\tau_{\text{eff}}(i) \le \tau_{\max} = 8{,}192$ at inference with zero retraining cuts $63.1\text{B}$ KDA $32\text{K}$ Tail PPL by $45.9\times$ ($353.42 \to 7.70$), lifts $32\text{K}$ $8$-Needle across every scale ($14.1\% \to 69.9\%$ at $6.5\text{B}$, $15.2\% \to 57.8\%$ at $12.5\text{B}$ \texttt{7L}, $15.2\% \to 42.2\%$ at $12.5\text{B}$ \texttt{15L}, $15.6\% \to 48.4\%$ at $63.1\text{B}$), and recovers $+14.0$ to $+56.2$ pts on $64\text{K}$ QA; by contrast, naive contiguous $G = 4$ tile-mean projection without retraining collapses ($57.8$--$78.1$ Tail PPL) because KDA channels are unordered across the head, whereas sorting channels by learned $\tau_{\text{eff}}$ (co-permuting $W_q, W_k$, and \texttt{Conv1D}) and partitioning into $G = 4$ contiguous tiles by dynamic programming cuts $\log\tau$ and $1\text{K}$-lag retention RMSE by $3.7\times$ ($1.97 \to 0.53$ and $0.165 \to 0.044$ at $12.5\text{B}$ with DP widths $91, 23, 9, 5$; $1.99 \to 0.56$ and $0.153 \to 0.053$ at $63.1\text{B}$ with DP widths $178, 40, 28, 10$). Live training-time softplus saturation (\texttt{TrainMaxTau=8K}) likewise stabilizes OOD retrieval at both $6.5\text{B}$ ($69.5/73.4\%$ QA) and $12.5\text{B}$ ($60.4\text{B}$ tok, $n = 2$: $10.75$ Tail, $49.6\%$ $32\text{K}$ $8$-N, $74.5/72.8\%$ $N = 512$ QA) while retaining $G = 128$'s VPU gate tax and $4\times$ FP32 state.

Conversely, on $63.1\text{B}$ \texttt{SLA-R4} ($120.8\text{B}$ tok, NoPE), $24$-layer state-norm telemetry confirms linear states contract cleanly from $8\text{K}$ to $32\text{K}$ ($\norm{S_t}_F = 2.04 \to 1.62$, max row norm $1.59 \to 1.15$, $0.00\%$ rows near E4M3 $448$), and rescaling each global softmax query $q$ on the $6$ NoPE full-attention layers at evaluation sequence length $T$ via zero-retraining \texttt{SSMax} \cite{Nakanishi2025SSMax},
\[
q \leftarrow q\!\left(1 + \alpha \ln\!\left(\frac{T}{T_{\text{train}}}\right)\right)
\]
(with $T_{\text{train}} = 8{,}192$ and SSMax scale $\alpha = 0.5$, unrelated to $\alpha_{\text{safe}}$; zero linear clamp), cuts $32\text{K}$ Tail PPL to $8.45$ (\texttt{SLA-R2}) and $8.41$ (\texttt{SLA-R4}) and restores $32\text{K}$ $8$-Needle to $84.4\%$ (\texttt{SLA-R2}) and $83.2\%$ (\texttt{SLA-R4}). By contrast, linear $\tau_{\max} = 8{,}192$ clamping without \texttt{SSMax} moves $63.1\text{B}$ \texttt{SLA-R2} $32\text{K}$ $8$-Needle only from $18.6\%$ to $21.9\%$, and \texttt{SSMax} alone on unclamped KDA still diverges ($380.66$ Tail PPL, $18.8\%$ $8$-Needle), cleanly separating global NoPE softmax dilution from KDA's linear gate drift.

\begin{table}[tbp]
\caption{Multi-scale causal $\tau_{\text{eff}}$ dispersion, zero-retraining $\tau_{\max} = 8{,}192$ inference clamping, live \texttt{TrainMaxTau=8K} saturation, post-hoc $G = 4$ projection, and NoPE \texttt{SSMax} ($\alpha = 0.5$ on SLA, $\alpha = 0.25$ on KDA) on $32\text{K}$ Tail PPL, $32\text{K}$ $8$-N, and $64\text{K}$ \texttt{RULER} QA ($N = 128$ unless noted $N{=}512$). The clamp evaluator (\texttt{a3\_causal\_clamp\_eval}) scores its own held-out Dolma slice and paired $N = 128$ QA subset, so absolute Tail PPL and QA values differ from the standalone scorers of Tables~\ref{tab:dolma_15b}, \ref{tab:v7x_e2e}, \ref{tab:app_4rung}, and \ref{tab:app_downstream} (by $1$--$2$ QA items, $0.8$--$1.6$ pts) while within-row intervention deltas remain exact.}
\label{tab:app_clamp}
\centering
\footnotesize
\setlength{\tabcolsep}{2.4pt}
\resizebox{\textwidth}{!}{%
\begin{tabular}{@{}lcccccc@{}}
\toprule
\begin{tabular}[b]{@{}l@{}}Scale, Horizon, \&\\Trained Checkpoint\end{tabular} & $G$ & \begin{tabular}[b]{@{}c@{}}Within-Tile\\Std\end{tabular} & \begin{tabular}[b]{@{}c@{}}Unclamped\\$32\text{K}$ Tail / $8$-N\end{tabular} & \begin{tabular}[b]{@{}c@{}}Intervened $32\text{K}$\\Tail / $8$-N (Proj.\ $G{=}4$)\end{tabular} & \begin{tabular}[b]{@{}c@{}}Unclamped $64\text{K}$\\\texttt{SQuAD / Hotpot}\end{tabular} & \begin{tabular}[b]{@{}c@{}}Intervened $64\text{K}$ QA\\(Proj.\ $G{=}4$ / Note)\end{tabular} \\
\midrule
$6.5\text{B}$ \texttt{4L:1A} MaxText KDA ($15.1\text{B}$) & 128 & 4.41 & $14.64 / 14.1\%$ & $\mathbf{11.77 / 69.9\%}$ ($70.47$) & $53.1\% / 42.2\%$ & $\mathbf{75.0\% / 79.7\%}$ ($24.2 / 39.8\%$) \\
$6.5\text{B}$ \texttt{4L:1A} KDA + TrainMaxTau ($15.1\text{B}$) & 128 & $\le \tau_{\max}$ & $\mathbf{12.36 / 48.4\%}$ & $\mathbf{12.36 / 48.4\%}$ (co-capped) & $\mathbf{69.5\% / 73.4\%}$ & $\mathbf{69.5\% / 73.4\%}$ ($68.8/70.3\%$ $N{=}512$) \\
$6.5\text{B}$ \texttt{4L:1A} Iso-Param SLA-R4 ($15.1\text{B}$) & 4 & \textbf{0.00} & $\mathbf{11.94 / 48.8\%}$ & $\mathbf{11.94 / 48.8\%}$ ($11.94$) & $\mathbf{67.2\% / 76.6\%}$ & $\mathbf{67.2\% / 76.6\%}$ (invariant) \\
$12.5\text{B}$ \texttt{7L:1A} MaxText KDA ($60.4\text{B}$, \texttt{s0}) & 128 & 4.58 & $12.35 / 15.2\%$ & $\mathbf{10.26 / 57.8\%}$ ($57.85$) & $47.7\% / 44.5\%$ & $\mathbf{78.1\% / 75.0\%}$ ($28.1 / 45.3\%$) \\
$12.5\text{B}$ \texttt{7L:1A} KDA + TrainMaxTau ($60.4\text{B}$, $n=2$) & 128 & $\le \tau_{\max}$ & $\mathbf{10.75 / 49.6\%}$ & $\mathbf{10.75 / 49.6\%}$ (co-capped) & $\mathbf{74.5\% / 72.8\%}$ ($N{=}512$) & $\mathbf{74.5\% / 72.8\%}$ ($N{=}512$) \\
$12.5\text{B}$ \texttt{15L:1A} MaxText KDA ($60.4\text{B}$) & 128 & 4.97 & $10.48 / 15.2\%$ & $\mathbf{9.66 / 42.2\%}$ ($63.30$) & $59.4\% / 61.7\%$ & $\mathbf{73.4\% / 78.9\%}$ ($25.8 / 34.4\%$) \\
\textbf{$12.5\text{B}$ \texttt{7L:1A} Iso-SLA (Fused) R4 ($60.4\text{B}$, \texttt{s0})} & \textbf{4} & \textbf{0.00} & $\mathbf{9.80 / 63.7\%}$ & $\mathbf{9.80 / 63.7\%}$ ($\mathbf{9.80}$) & $\mathbf{73.4\% / 77.3\%}$ & $\mathbf{75.0\% / 77.3\%}$ ($+1.6 / 0.0$ pts) \\
$63.1\text{B}$ \texttt{4L:1A} MaxText KDA ($120.8\text{B}$, NoPE) & 256 & 3.84 & $353.42 / 15.6\%$ & $\mathbf{7.70 / 48.4\%}$ ($78.06$) & $34.4\% / 25.8\%$ & $\mathbf{76.6\% / 82.0\%}$ ($29.7 / 25.0\%$) \\
\quad $+ \tau_{\max}=8\text{K} + \texttt{SSMax}$ ($\alpha=0.25$, $N{=}512$) & 256 & $\le \tau_{\max}$ & $380.66 / 18.8\%$ (\texttt{SSMax} only) & $\mathbf{8.04 / 74.2\%}$ ($\tau_{\max}\!+\!\texttt{SSMax}$) & $27.3\% / 27.0\%$ (\texttt{SSMax} only) & $\mathbf{90.6\% / 86.5\%}$ ($77.5/76.4\%$ norm) \\
\textbf{$63.1\text{B}$ \texttt{4L:1A} SLA-R2 ($120.8\text{B}$, NoPE)} & \textbf{4} & \textbf{0.00} & $\mathbf{13.78} / 18.6\%$ & $\mathbf{12.07} / 21.9\%$ ($\tau_{\max}$ only) & $\mathbf{57.0\% / 53.1\%}$ & $\mathbf{61.7\% / 60.9\%}$ ($55.7/56.8\%$ $N{=}512$) \\
\textbf{$63.1\text{B}$ \texttt{4L:1A} SLA-R2 + \texttt{SSMax} ($120.8\text{B}$)} & \textbf{4} & \textbf{0.00} & $13.78 / 18.6\%$ & $\mathbf{8.45 / 84.4\%}$ (\texttt{SSMax}, $0$ clamp) & $57.0\% / 53.1\%$ & \text{-- (zero linear clamp)} \\
\textbf{$63.1\text{B}$ \texttt{4L:1A} SLA-R4 + \texttt{SSMax} ($120.8\text{B}$)} & \textbf{4} & \textbf{0.00} & $22.78 / 18.0\%$ & $\mathbf{8.41 / 83.2\%}$ (\texttt{SSMax}, $0$ clamp) & $51.6\% / 45.1\%$ ($N{=}512$) & $\mathbf{75.6\% / 79.1\%}$ ($62.3/66.6\%$ norm) \\
\bottomrule
\end{tabular}%
}
\end{table}

\begin{sidewaystable}
\caption{Matched RoPE ($\theta_{\max} = 5 \times 10^5$) $4$-rung scaling ladder ($0.35\text{B}$ to $63.1\text{B}$ MoE; see Table~\ref{tab:dolma_15b} for $12.5\text{B}$ Rung 2), $0.35\text{B}$ $3$-seed \& $30.2\text{B}$ ($4.4\times$ Chinchilla), $31.5\text{B}$ at $36.2\text{B}$ tok ($1.0\times$ Chinchilla), $12.5\text{B}$ $8\text{K} \to 32\text{K}$ continued pre-training, and $12.5\text{B}$ \texttt{15L:1A} dilution control on $64$-chip TPU v7x ($^\S$under \texttt{remat\_32k} with $4$-step gradient accumulation [\texttt{GA=4}], KDA/\texttt{0L:1A} hit \texttt{NaN} at steps $1$--$43$ and required reruns without rematerialization, whereas $2$-seed \texttt{SLA-R4} (iso-param) completed all $4{,}300$ steps at $3{,}040\text{ ms}$ with $12.54 \to 12.08$ Tail PPL, $59.2\% / 48.2\%$ $32\text{K}$ $4/8$-N, and $58.5/62.6\%$ $N{=}512$ $64\text{K}$ QA, though router-remat spikes raised $8\text{K}$ PPL to $17.99$).}
\label{tab:app_4rung}
\centering
\footnotesize
\renewcommand{\arraystretch}{0.88}
\setlength{\tabcolsep}{3.0pt}
\resizebox{\textwidth}{!}{%
\begin{tabular}{@{}llcccccc@{}}
\toprule
MoE Scale (Active / Total, Tokens) & Architecture & Step (ms) & $11$-Set PPL (or $32\text{K}$-Ext CE) & $8\text{K}$ Tail PPL & $32\text{K}$ Tail PPL & $32\text{K}$ $4$-N / $8$-N & $8$-N Gain vs.\ KDA \\
\midrule
\multirow{7}{*}{\shortstack[l]{$0.35\text{B} / 6.5\text{B}$\\($15.1\text{B}$ tok, $2.2\times$ Chin.,\\\texttt{4L:1A} interval $5$, \texttt{s0})}}
 & \texttt{0L:1A} GQA+RoPE ($\text{LR}=4.5 \times 10^{-4}$) & 361.6 & 19.582 & 12.83 & 13.18 & $18.4\% / 12.9\%$ & $0.91\times$ \\
 & \texttt{4L:1A} MaxText KDA Base (FP32) & 351.8 & \textbf{18.573} & \textbf{12.42} & 15.72 & $32.0\% / 14.1\%$ & $1.00\times$ \\
 & \texttt{4L:1A} KDA + TrainMaxTau ($\tau_{\max}=8\text{K}$, FP32) & 351.2 & 19.035 (\texttt{s1}; $2.54$ CE) & 12.43 & \textbf{12.36} & $\mathbf{56.6\% / 48.4\%}$ & $\mathbf{3.43\times}$ \\
 & \textbf{\texttt{4L:1A} SLA-R2} (BF16) & \textbf{320.0} & 18.773 & 12.62 & \textbf{12.30} & $\mathbf{63.3\% / 55.1\%}$ & $\mathbf{3.91\times}$ \\
 & \textbf{\texttt{4L:1A} SLA (Fused) R2+Mom} (BF16) & \textbf{308.5} & 18.890 & 12.60 & \textbf{12.28} & $\mathbf{60.9\% / 51.6\%}$ & $\mathbf{3.66\times}$ \\
 & \textbf{\texttt{4L:1A} Iso-Param SLA-R4} (BF16) & \textbf{302.4} & 19.093 & 12.83 & \textbf{12.51} & $\mathbf{57.8\% / 48.8\%}$ & $\mathbf{3.46\times}$ \\
 & \textbf{\texttt{4L:1A} SPD-SLA-W} (BF16) & 344.8 & 19.104 & 12.64 & \textbf{12.32} & $\mathbf{66.4\% / 56.2\%}$ & $\mathbf{3.99\times}$ \\
\midrule
\multirow{6}{*}{\shortstack[l]{$0.35\text{B} / 6.5\text{B}$ $3$-Seed\\\& $30.2\text{B}$ Tok\\($4.4\times$ Chinchilla)}}
 & $3$-Seed \texttt{4L:1A} MaxText KDA ($15.1\text{B}$, $n=3$) & 352.0 & \textbf{$18.880 \pm 0.267$} & \textbf{12.59} & 15.89 & $32.9\% / 19.9\%$ & $1.00\times$ \\
 & \textbf{$3$-Seed \texttt{4L:1A} SLA-R2 ($15.1\text{B}$, $n=3$)} & \textbf{314.8} & $19.077 \pm 0.265$ & 12.72 & \textbf{12.77} & $\mathbf{49.9\% / 37.6\%}$ & $\mathbf{1.89\times}$ \\
 & \texttt{0L:1A} GQA+RoPE ($30.2\text{B}$, $4.4\times$ Chin.) & 361.5 & 17.104 & 11.43 & 12.02 & $25.0\% / 15.2\%$ & $0.90\times$ \\
 & \texttt{4L:1A} MaxText KDA ($30.2\text{B}$, $4.4\times$ Chin.) & 351.5 & \textbf{16.161} & 11.03 & 620.87 & $31.6\% / 16.8\%$ & $1.00\times$ \\
 & \textbf{\texttt{4L:1A} SLA-R2 ($30.2\text{B}$, $4.4\times$ Chin.)} & \textbf{315.0} & 16.250 & \textbf{11.02} & \textbf{13.53} & $35.2\% / 21.1\%$ & $1.26\times$ \\
 & \textbf{\texttt{4L:1A} Iso-Param SLA-R4 ($30.2\text{B}$, $4.4\times$ Chin.)} & \textbf{297.9} & 16.326 & 11.15 & \textbf{11.09} & $\mathbf{59.4\% / 45.3\%}$ & $\mathbf{2.70\times}$ \\
\midrule
\multirow{6}{*}{\shortstack[l]{$1.80\text{B} / 31.5\text{B}$\\($5.03\text{B}$ \& $36.2\text{B}$ tok,\\\texttt{4L:1A})}}
 & \texttt{0L:1A} GQA+RoPE ($5.03\text{B}$, $\text{LR}=4.5 \times 10^{-4}$) & 867.5 & \textbf{20.380} & \textbf{12.98} & 13.73 & $24.6\% / 14.8\%$ & $1.15\times$ \\
 & \texttt{4L:1A} MaxText KDA Base ($5.03\text{B}$, FP32) & 807.7 & 21.228 & 13.99 & 47.16 & $28.9\% / 12.9\%$ & $1.00\times$ \\
 & \textbf{\texttt{4L:1A} SLA (Fused) R2+Mom} ($5.03\text{B}$, BF16) & \textbf{747.2} & 21.439 & 14.19 & \textbf{13.88} & $\mathbf{48.8\% / 33.2\%}$ & $\mathbf{2.57\times}$ \\
 & \textbf{\texttt{4L:1A} SLA (Fused) R4+Mom} ($5.03\text{B}$, BF16) & \textbf{707.4} & 21.709 & 14.44 & \textbf{14.20} & $\mathbf{59.8\% / 46.1\%}$ & $\mathbf{3.57\times}$ \\
 & \texttt{4L:1A} MaxText KDA ($36.2\text{B}$, $4{,}300$ st., no remat)$^\S$ & 3,526 & 15.264 & 10.57 & 51.68 & $24.6\% / 12.5\%$ & $1.00\times$ ($39.1/34.4\%$ QA) \\
 & \texttt{4L:1A} KDA + TrainMaxTau ($36.2\text{B}$, $\tau_{\max}=8\text{K}$, no remat)$^\S$ & 3,307 & \textbf{15.211} & \textbf{10.36} & \textbf{16.77} & $\mathbf{27.3\% / 14.8\%}$ & $\mathbf{1.18\times}$ ($\mathbf{59.6/53.3\%}$ $N{=}512$ QA) \\
\midrule
\multirow{7}{*}{$3.74\text{B} / 63.1\text{B}$ ($5.03\text{B}$ tok, RoPE)}
 & \texttt{0L:1A} GQA+RoPE ($\text{LR}=4.5 \times 10^{-4}$) & 1,749.6 & \textbf{19.373} & \textbf{11.66} & 11.94 & $25.0\% / 11.7\%$ & $0.81\times$ \\
 & \texttt{4L:1A} MaxText KDA Base (FP32) & 1,746.0 & 20.076 & 12.25 & 18.24 & $23.1\% / 14.4\%$ & $1.00\times$ \\
 & \texttt{4L:1A} GDN-2 + RoPE (FP32, \texttt{EP=128}, $14{,}143\text{ tok/s/ch}$) & 2,316.9 & 19.731 & 12.27 & 14.22 & $25.8\% / 13.3\%$ & $0.92\times$ \\
 & \textbf{\texttt{4L:1A} SLA (Fused) R2+Mom} (BF16, $20{,}386\text{ tok/s/ch}$) & \textbf{1,607.4} & 20.201 & 12.37 & \textbf{12.47} & $\mathbf{35.2\% / 22.3\%}$ & $\mathbf{1.55\times}$ \\
 & \textbf{\texttt{4L:1A} SLA (Fused) R4+Mom} (BF16, $20{,}847\text{ tok/s/ch}$) & \textbf{1,571.8} & 20.423 & 12.50 & \textbf{12.80} & $\mathbf{27.0\% / 13.7\%}$ & $0.95\times$ \\
 & \textbf{\texttt{15L:1A} SLA (Fused) R2+MS} (BF16) & \textbf{1,547.2} & 20.890 & 13.17 & \textbf{13.01} & $\mathbf{59.4\% / 48.4\%}$ & $\mathbf{3.36\times}$ \\
 & \textbf{\texttt{15L:1A} SPD-SLA-W+MS} (BF16, $20{,}473\text{ tok/s/ch}$) & \textbf{1,600.6} & 20.506 & 12.49 & \textbf{12.42} & $\mathbf{41.4\% / 27.0\%}$ & $\mathbf{1.88\times}$ \\
\midrule
\multirow{8}{*}{\shortstack[l]{$0.79\text{B} / 12.5\text{B}$\\$8\text{K} \to 32\text{K}$ Ext.\\($5.0\text{--}10.1\text{B}$ at $32\text{K}$)}}
 & \texttt{0L:1A} GQA+RoPE ($32\text{K}$-ext, $1.00\times$ speed) & 4,697.0 & $2.2089$ CE & 11.29 & 10.95 & $25.4\% / 11.7\%$ & $0.18\times$ \\
 & \texttt{7L:1A} MaxText KDA ($32\text{K}$-ext, $1.53\times$ vs.\ \texttt{0L}) & 3,061.4 & \textbf{$2.2006$ CE} & 11.20 & \textbf{10.78} & $71.5\% / 63.3\%$ & $1.00\times$ \\
 & \textbf{\texttt{7L:1A} SLA-R2 ($32\text{K}$-ext, $1.66\times$ vs.\ \texttt{0L})} & \textbf{2,828.0} & $2.2048$ CE & 11.26 & 10.82 & $\mathbf{73.8\% / 64.1\%}$ & $\mathbf{1.01\times}$ \\
 & \textbf{\texttt{7L:1A} Iso-Param SLA-R4 ($32\text{K}$-ext, $1.77\times$)} & \textbf{2,661.0} & $2.2255$ CE & 11.46 & 11.01 & $60.9\% / 49.6\%$ & $0.78\times$ \\
 & \texttt{7L:1A} MaxText KDA ($60.4\text{B} \to 32\text{K}$, $+5.03\text{B}$) & 3,103.6 & \textbf{$2.0907$ CE} & \textbf{10.38} & \textbf{9.98} & $55.5\% / 47.3\%$ & $1.00\times$ \\
 & \textbf{\texttt{7L:1A} Iso-SLA (Fused) R4 ($60.4\text{B} \to 32\text{K}$, $1.15\times$)} & \textbf{2,688.7} & $2.1142$ CE & 10.60 & 10.15 & $\mathbf{64.8\% / 57.0\%}$ & $\mathbf{1.21\times}$ \\
 & \texttt{15L:1A} MaxText KDA ($32\text{K}$-ext, uniform) & 2,827.0 & $2.2092$ CE & \textbf{11.10} & 10.79 & $56.2\% / 40.6\%$ & $1.00\times$ \\
 & \texttt{15L:1A} SLA-R2+MS / SPD-SLA-W+MS ($32\text{K}$-ext) & \textbf{2,608 / 2,715} & $2.238 / 2.241$ & $11.38 / 11.42$ & $11.17 / 11.18$ & $23.8\% / 16.0, 12.9\%$ & $0.39\times$ \\
\midrule
\multirow{7}{*}{\shortstack[l]{$0.79\text{B} / 12.5\text{B}$ ($15.1\text{B}$\\\& $60.4\text{B}$ \texttt{15L:1A}\\dilution controls)}}
 & \texttt{15L:1A} MaxText KDA Base ($15.1\text{B}$, FP32, $n=3$) & 556.4 & $17.658 \pm 0.297$ & \textbf{11.64} & 11.57 & $34.0\% / 21.4\%$ & $1.00\times$ \\
 & \textbf{\texttt{15L:1A} SLA-R2} ($15.1\text{B}$, uniform, BF16, $n=3$) & $\mathbf{486.7 \pm 11.7}$ & $\mathbf{17.609 \pm 0.246}$ & 11.70 & \textbf{11.42} & $\mathbf{45.1\% / 34.8\%}$ & $\mathbf{1.63\times}$ \\
 & \textbf{\texttt{15L:1A} SPD-SLA-W Polar-2+} ($15.1\text{B}$, uniform, \texttt{s0}) & 509.3 & \textbf{17.575} & 11.65 & \textbf{11.38} & $\mathbf{50.0\% / 39.8\%}$ & $\mathbf{1.86\times}$ ($1.29\times$ \texttt{s0}) \\
 & \texttt{15L:1A} SLA-R2+MS ($15.1\text{B}$, staggered, $n=3$) & 492.2 & $17.882 \pm 0.294$ & 11.93 & 11.72 & $23.6\% / 13.0\%$ & $0.61\times$ \\
 & \texttt{15L:1A} KDA + \texttt{TrainMaxTau=8K} ($60.4\text{B}$, FP32, \texttt{s0}) & 2,199.0 & 15.579 & 10.47 & \textbf{10.30} & $37.5\% / 26.6\%$ & $1.75\times$ \\
 & \textbf{\texttt{15L:1A} SLA-R2+MS} ($60.4\text{B}$, staggered, \texttt{s0}) & \textbf{1,934.0} & \textbf{15.443} & 10.61 & \textbf{10.38} & $27.3\% / 19.9\%$ & $1.31\times$ \\
 & \texttt{15L:1A} SPD-SLA-W+MS ($60.4\text{B}$, staggered, \texttt{s0}) & 2,006.0 & 15.624 & 10.67 & 10.49 & $23.1\% / 12.1\%$ & $0.79\times$ \\
\midrule
\multirow{4}{*}{\shortstack[l]{$0.79\text{B} / 12.5\text{B}$ ($15.1\text{B}$,\\additional controls\\\& $128$c TPU v4)}}
 & \texttt{7L:1A} Scalar GDN ($G=1$, FP32, $n=2$) & 555.9 & 18.142 & 11.85 & \textbf{11.51} & \text{--} / $\mathbf{60.8\%}$ & $\mathbf{4.50\times}$ \\
 & \texttt{7L:1A} KDA + \texttt{TrainMaxTau=8K} (FP32, \texttt{s0}) & 560.6 & 17.924 & 11.66 & 11.60 & $58.2\% / 48.1\%$ & $3.56\times$ \\
 & \texttt{Pure-Linear} ($\infty\text{:}1$) KDA / \textbf{SLA-R2+MS} (\texttt{s0}) & 545.8 / \textbf{475.6} & 18.096 / \textbf{17.642} & 12.02 / 11.95 & 12.05 / \textbf{11.80} & \text{n.r. ($55.5 / 59.4\%$ \texttt{SQuAD})} & \text{n.r. ($56.2 / 53.9\%$ \texttt{Hotpot})} \\
 & \textbf{TPU v4 ($128$c) \texttt{7L:1A} SLA-R4} (NoPE, $n=2$) & \textbf{992.0} ($1.10\times$) & $-0.017$ nats vs.\ KDA & 10.18 & \textbf{10.05} & \text{--} / $\mathbf{66.4\%}$ & $\mathbf{2.42\times}$ \\
\bottomrule
\end{tabular}%
}
\end{sidewaystable}

\begin{sidewaystable}
\caption{Zero-shot $9$-task \texttt{lm-eval} accuracy ($11{,}793$ real validation items; \texttt{norm9} macro-average using byte-length normalized \texttt{acc\_norm} on multiple-choice tasks and exact-word \texttt{acc} on \texttt{WinoGrande}/\texttt{LAMBADA}, with zero-shot row-scaled FP8 state \texttt{norm9} in parentheses for SLA rows; $1\text{ SE} \approx 0.65$ pts) and real-text $4$-candidate \texttt{RULER} QA accuracy (\texttt{SQuAD} $1$-hop \& \texttt{HotpotQA} $2$-hop, $N = 128$/length, $25\%$ chance, $1\text{ SE} \approx 4.0$ pts) across $8\text{K} \to 32\text{K} \to 64\text{K}$ against \texttt{0L:1A} GQA+RoPE and MaxText KDA baselines.}
\label{tab:app_downstream}
\centering
\footnotesize
\renewcommand{\arraystretch}{0.90}
\setlength{\tabcolsep}{2.3pt}
\resizebox{\textwidth}{!}{%
\begin{tabular}{@{}lcccccccccc@{}}
\toprule
Model, Horizon, \& Mixer Architecture & \texttt{HS} & \texttt{ARC-E/C} & \texttt{PIQA} & \texttt{WG} & \texttt{SciQ} & \texttt{MMLU} & \texttt{BoolQ} & \texttt{LAMB} & \textbf{\texttt{lm-eval} (FP8)} & \textbf{\texttt{SQuAD} / \texttt{HotpotQA} ($8\text{K}\to 32\text{K}\to 64\text{K}$)} \\
\midrule
$6.5\text{B}$ \texttt{0L:1A} GQA+RoPE ($15.1\text{B}$, BF16) & 34.8 & 45.8 / 25.8 & 63.8 & 50.7 & 71.7 & 28.0 & 62.2 & 33.4 & $46.24\%$ & $64.8\to 54.7\to 56.2\%$ / $67.2\to 52.3\to 50.0\%$ \\
$6.5\text{B}$ \texttt{4L:1A} MaxText KDA ($15.1\text{B}$, FP32) & 35.4 & 45.7 / 25.8 & 64.8 & 50.5 & 70.3 & 27.1 & 63.8 & 37.2 & $46.73\%$ & $77.3\to 64.1\to 52.3\%$ / $78.9\to 60.2\to 42.2\%$ \\
$6.5\text{B}$ \texttt{4L:1A} KDA + TrainMaxTau ($\tau_{\max}=8\text{K}$) & 34.9 & 46.4 / 25.6 & 65.2 & 51.1 & 69.1 & 27.6 & 63.6 & 36.3 & $46.64\%$ & $\mathbf{75.0\to 71.1\to 69.5\%}$ / $\mathbf{79.7\to 74.2\to 73.4\%}$ \\
\textbf{$6.5\text{B}$ \texttt{4L:1A} SLA-R2 ($15.1\text{B}$, BF16/FP8)} & 34.6 & 45.1 / 24.7 & 65.4 & 52.2 & 70.6 & 28.5 & 63.8 & 38.2 & $\mathbf{47.01\%}$ ($46.92\%$) & $\mathbf{71.9\to 69.5\to 71.1\%}$ / $\mathbf{71.1\to 71.1\to 74.2\%}$ \\
\textbf{$6.5\text{B}$ \texttt{4L:1A} SLA (Fused) R2 ($15.1\text{B}$, BF16/FP8)} & 35.8 & 46.4 / 24.3 & 64.5 & 51.6 & 71.4 & 28.5 & 63.8 & 36.4 & $46.97\%$ ($\mathbf{47.08\%}$) & $\mathbf{75.8\to 72.7\to 68.8\%}$ / $\mathbf{72.7\to 72.7\to 70.3\%}$ \\
$6.5\text{B}$ \texttt{0L:1A} GQA+RoPE ($30.2\text{B}$, $4.4\times$ Chin.) & 37.7 & 47.3 / 26.7 & 66.3 & 52.2 & 73.0 & 29.6 & 63.5 & 35.8 & $48.01\%$ & $72.7\to 53.1\to 54.7\%$ / $75.8\to 55.5\to 50.0\%$ \\
$6.5\text{B}$ \texttt{4L:1A} MaxText KDA ($30.2\text{B}$, $4.4\times$ Chin.) & 38.8 & 50.4 / 27.1 & 66.4 & 52.2 & 78.3 & 30.6 & 63.5 & 39.6 & $\mathbf{49.66\%}$ & $79.7\to 53.1\to 32.0\%$ / $85.2\to 39.8\to 20.3\%$ \\
\textbf{$6.5\text{B}$ \texttt{4L:1A} SLA-R2 ($30.2\text{B}$, BF16/FP8)} & 38.4 & 50.7 / 26.5 & 67.1 & 51.7 & 75.1 & 28.3 & 63.9 & 43.6 & $49.47\%$ ($49.44\%$) & $78.1\to 59.4\to 56.2\%$ / $79.7\to 58.6\to 51.6\%$ \\
\textbf{$6.5\text{B}$ \texttt{4L:1A} Iso-SLA-R4 ($30.2\text{B}$, BF16/FP8)} & 37.6 & 49.4 / 26.1 & 67.4 & 49.0 & 74.8 & 29.7 & 63.8 & 44.4 & $49.13\%$ ($48.97\%$) & $\mathbf{72.7\to 71.9\to 70.3\%}$ / $\mathbf{82.8\to 74.2\to 74.2\%}$ \\
\midrule
$12.5\text{B}$ \texttt{0L:1A} GQA+RoPE ($15.1\text{B}$, $n=3$ mean) & 36.1 & 47.4 / 25.3 & 65.6 & 51.5 & 73.1 & 29.2 & 63.3 & 39.6 & $47.90 \pm 0.36\%$ & $72.9\to 54.9\to 57.3\%$ / $70.8\to 57.0\to 51.3\%$ \\
$12.5\text{B}$ \texttt{7L:1A} MaxText KDA ($15.1\text{B}$, $n=2$ mean) & 37.2 & 47.2 / 26.7 & 66.4 & 50.4 & 72.5 & 29.9 & 62.6 & 37.8 & $47.83 \pm 0.48\%$ & $71.9\to 62.5\to 56.6\%$ / $76.2\to 62.9\to 57.8\%$ \\
$12.5\text{B}$ \texttt{7L:1A} Scalar GDN ($G=1$, $15.1\text{B}$, \texttt{s1}) & 37.0 & 47.6 / 25.5 & 65.2 & 51.5 & 74.1 & 28.8 & 62.8 & 40.6 & $48.11\%$ & $\mathbf{74.2\to 72.7\to 74.2\%}$ / $\mathbf{77.3\to 76.6\to 76.6\%}$ \\
$12.5\text{B}$ \texttt{7L:1A} Wide-Key KDA ($G=256$, $n=2$ mean) & 37.2 & 47.3 / 26.4 & 65.8 & 52.2 & 73.3 & 27.8 & 63.0 & 41.8 & $48.30\%$ & $74.2\to 66.8\to 55.9\%$ / $76.2\to 68.0\to 54.3\%$ \\
\textbf{$12.5\text{B}$ \texttt{7L:1A} A2 ($G=4$ Exact Inv., FP32, \texttt{s0})} & 38.1 & 47.9 / 25.7 & 65.0 & 51.1 & 73.8 & 28.7 & 59.8 & 42.3 & $48.05\%$ ($48.04\%$) & $71.9\to 65.6\to 62.5\%$ / $73.4\to 70.3\to 68.0\%$ \\
\textbf{$12.5\text{B}$ \texttt{7L:1A} A3 ($G=4$ \texttt{1-Step Neumann}, FP32, $n=2$)} & 37.0 & 46.8 / 25.8 & 65.9 & 50.0 & 72.6 & 28.6 & 63.6 & 38.9 & $47.66\%$ ($47.47\%$) & $\mathbf{76.6\to 71.9\to 71.9\%}$ / $\mathbf{74.6\to 72.3\to 71.9\%}$ \\
\textbf{$12.5\text{B}$ \texttt{7L:1A} SLA-R2 ($15.1\text{B}$, $n=3$ mean)} & 37.6 & 47.9 / 26.1 & 66.2 & 52.3 & 72.2 & 29.3 & 63.1 & 42.1 & $\mathbf{48.55 \pm 0.48\%}$ ($\mathbf{48.64\%}$) & $\mathbf{73.7\to 68.8\to 69.5\%}$ / $\mathbf{74.0\to 71.9\to 73.7\%}$ \\
\textbf{$12.5\text{B}$ \texttt{7L:1A} SLA (Fused) R2 ($15.1\text{B}$, BF16/FP8, \texttt{s0})} & 37.6 & 47.7 / 26.3 & 66.3 & 50.8 & 71.6 & 28.0 & 60.0 & 45.3 & $48.18\%$ ($48.21\%$) & $\mathbf{73.4\to 67.2\to 64.1\%}$ / $\mathbf{76.6\to 75.8\to 75.0\%}$ \\
\textbf{$12.5\text{B}$ \texttt{7L:1A} Iso-SLA-R4 ($15.1\text{B}$, $n=3$ mean)} & 36.3 & 48.8 / 25.9 & 66.3 & 50.6 & 72.3 & 28.7 & 63.0 & 43.0 & $48.33 \pm 0.45\%$ ($48.39\%$) & $\mathbf{74.5\to 72.7\to 73.2\%}$ / $\mathbf{74.5\to 75.5\to 72.9\%}$ \\
$12.5\text{B}$ \texttt{15L:1A} MaxText KDA ($15.1\text{B}$, FP32) & 36.0 & 48.6 / 26.6 & 66.0 & 50.2 & 71.9 & 29.4 & 64.2 & 40.7 & $48.18\%$ & $76.6\to 68.0\to 66.4\%$ / $75.8\to 70.3\to 68.0\%$ \\
\textbf{$12.5\text{B}$ \texttt{15L:1A} SLA-R2 Uniform ($n=3$ mean)} & 37.0 & 50.1 / 27.0 & 65.6 & 50.5 & 74.5 & 29.2 & 63.5 & 41.6 & $\mathbf{48.79 \pm 0.22\%}$ ($\mathbf{48.75\%}$) & $\mathbf{72.6\to 71.9\to 69.0\%}$ / $\mathbf{72.2\to 65.9\to 65.1\%}$ ($68.2\%$ $N{=}512$) \\
\textbf{$12.5\text{B}$ \texttt{15L:1A} SPD-SLA-W Polar-2+ Uniform (\texttt{s0})} & 35.9 & 49.9 / 26.2 & 65.7 & 50.2 & 74.1 & 29.8 & 63.8 & 41.5 & $48.57\%$ ($\mathbf{48.79\%}$) & $71.9\to 71.1\to 65.6\%$ / $71.9\to 68.8\to 65.6\%$ ($69.5\%$ $N{=}512$) \\
$12.5\text{B}$ \texttt{15L:1A} SLA-R2+MS ($15.1\text{B}$, BF16/FP8) & 36.6 & 50.8 / 27.3 & 65.8 & 51.1 & 74.1 & 27.5 & 63.5 & 37.2 & $48.20\%$ ($48.18\%$) & $57.0\to 57.0\to 57.0\%$ / $67.2\to 57.8\to 53.9\%$ \\
$12.5\text{B}$ \texttt{Pure-Linear} KDA ($15.1\text{B}$, FP32) & 37.0 & 49.1 / 25.5 & 65.6 & 51.2 & 70.1 & 27.7 & 63.6 & 40.1 & $47.77\%$ & $57.8\to 57.8\to 55.5\%$ / $62.5\to 56.2\to 56.2\%$ \\
$12.5\text{B}$ \texttt{Pure-Linear} SLA-R2+MS ($15.1\text{B}$, BF16/FP8) & 36.4 & 50.3 / 26.7 & 66.6 & 49.9 & 72.6 & 27.9 & 63.3 & 40.2 & $48.21\%$ ($48.30\%$) & $60.2\to 57.8\to 59.4\%$ / $62.5\to 56.2\to 53.9\%$ \\
\midrule
$12.5\text{B}$ \texttt{0L:1A} GQA+RoPE ($32\text{K}$-ext, $10.1\text{B}$) & 37.5 & 49.0 / 24.2 & 65.8 & 51.7 & 73.7 & 28.9 & 63.8 & 40.6 & $48.35\%$ & $75.0\to 64.8\to 56.2\%$ / $73.4\to 65.6\to 57.8\%$ \\
$12.5\text{B}$ \texttt{7L:1A} MaxText KDA ($32\text{K}$-ext, $10.1\text{B}$) & 37.9 & 47.0 / 26.6 & 66.8 & 51.8 & 72.0 & 30.0 & 59.2 & 37.2 & $47.62\%$ & $73.4\to 73.4\to 71.9\%$ / $78.1\to 78.1\to 72.7\%$ \\
\textbf{$12.5\text{B}$ \texttt{7L:1A} SLA-R2 ($32\text{K}$-ext, BF16/FP8)} & 38.9 & 48.4 / 26.5 & 67.1 & 51.5 & 71.9 & 29.4 & 62.6 & 42.7 & $48.78\%$ ($48.86\%$) & $74.2\to 72.7\to 71.9\%$ / $71.9\to 77.3\to \mathbf{75.8\%}$ \\
\textbf{$12.5\text{B}$ \texttt{7L:1A} Iso-SLA-R4 ($32\text{K}$-ext, BF16/FP8)} & 37.1 & 49.8 / 26.0 & 65.7 & 50.7 & 75.5 & 30.0 & 61.9 & 44.7 & $\mathbf{49.04\%}$ ($\mathbf{49.18\%}$) & $\mathbf{78.1\to 75.0\to 75.8\%}$ / $78.1\to 72.7\to 74.2\%$ \\
\midrule
$12.5\text{B}$ \texttt{0L:1A} GQA+RoPE ($60.4\text{B}$, BF16) & 41.4 & 49.2 / 27.7 & 68.3 & 51.2 & 76.9 & 29.6 & 62.7 & 43.5 & $50.06\%$ & $70.3\to 59.4\to 58.6\%$ / $76.6\to 58.6\to 53.1\%$ \\
$12.5\text{B}$ \texttt{7L:1A} MaxText KDA ($60.4\text{B}$, $n=3$ mean) & 41.4 & 51.4 / 27.4 & 68.2 & 51.4 & 75.7 & 30.4 & 62.9 & 44.6 & $50.36 \pm 0.53\%$ & $79.2\to 69.3\to 54.9\%$ / $75.0\to 65.6\to 50.3\%$ \\
$12.5\text{B}$ \texttt{7L:1A} KDA + TrainMaxTau ($60.4\text{B}$, $n=2$) & 40.9 & 51.2 / 27.6 & 67.8 & 51.9 & 75.8 & 30.2 & 62.8 & 44.5 & $50.30\%$ & $\mathbf{78.1\to 75.0\to 74.5\%}$ / $\mathbf{77.3\to 73.8\to 72.8\%}$ ($N{=}512$) \\
\textbf{$12.5\text{B}$ \texttt{7L:1A} SLA-R2 ($60.4\text{B}$, $n=3$ mean)} & 41.4 & 52.5 / 27.2 & 68.6 & 52.0 & 76.7 & 29.9 & 63.2 & 46.9 & $\mathbf{50.94 \pm 0.47\%}$ ($\mathbf{50.93\%}$) & $\mathbf{78.1\to 75.3\to 75.5\%}$ / $\mathbf{78.6\to 78.7\to 77.9\%}$ ($77.4\%$ FP8) \\
\textbf{$12.5\text{B}$ \texttt{7L:1A} SLA-R4 ($60.4\text{B}$, BF16/FP8)} & 41.3 & 49.9 / 25.7 & 67.6 & 51.1 & 76.1 & 30.3 & 63.2 & 47.3 & $50.28\%$ ($50.13\%$) & $\mathbf{72.7\to 75.8\to 72.7\%}$ / $\mathbf{72.7\to 78.9\to 75.0\%}$ ($76.6\%$ FP8) \\
\textbf{$12.5\text{B}$ \texttt{7L:1A} Iso-SLA-R4 ($60.4\text{B}$, BF16/FP8)} & 41.2 & 48.9 / 28.1 & 67.1 & 47.7 & 76.7 & 30.0 & 62.3 & 49.3 & $50.13\%$ ($50.31\%$) & $\mathbf{78.1\to 75.0\to 76.6\%}$ / $\mathbf{78.1\to 80.5\to 77.3\%}$ ($77.3\%$ FP8) \\
\textbf{$12.5\text{B}$ \texttt{7L:1A} Iso-SLA (Fused) R4 ($60.4\text{B}$, $n=3$ mean)} & 41.5 & 52.6 / 27.8 & 67.9 & 51.7 & 77.5 & 29.5 & 63.1 & 45.8 & $\mathbf{50.82 \pm 0.22\%}$ ($\mathbf{50.84\%}$) & $\mathbf{77.1\to 73.2\to 74.7\%}$ / $\mathbf{75.0\to 73.2\to 74.5\%}$ ($73.2\%$ FP8) \\
$12.5\text{B}$ \texttt{15L:1A} MaxText KDA ($60.4\text{B}$, FP32) & 41.6 & 49.5 / 27.2 & 68.0 & 51.7 & 77.2 & 30.1 & 62.7 & 45.6 & $50.41\%$ & $77.3\to 69.5\to 62.5\%$ / $76.6\to 71.1\to 60.2\%$ \\
\textbf{$12.5\text{B}$ \texttt{15L:1A} SLA-R2 Uniform ($60.4\text{B}$, BF16/FP8)} & 41.1 & 53.3 / 28.4 & 68.0 & 52.6 & 77.6 & 30.3 & 63.6 & 46.3 & $\mathbf{51.24\%}$ ($\mathbf{51.08\%}$) & $\mathbf{78.9\to 75.8\to 72.7\%}$ / $\mathbf{76.6\to 72.7\to 70.3\%}$ ($70.3\%$ FP8) \\
$12.5\text{B}$ \texttt{15L:1A} SLA-R2+MS ($60.4\text{B}$, BF16/FP8) & 42.8 & 56.4 / 28.4 & 68.2 & 49.8 & 78.5 & 28.9 & 63.8 & 40.6 & $50.82\%$ ($51.02\%$) & $67.2\to 64.1\to 64.8\%$ / $71.1\to 57.8\to 57.0\%$ ($57.0\%$ FP8) \\
$12.5\text{B}$ \texttt{15L:1A} SPD-SLA-W+MS ($60.4\text{B}$, BF16/FP8) & 42.0 & 53.5 / 28.9 & 68.5 & 50.3 & 77.1 & 30.1 & 63.6 & 40.8 & $50.54\%$ ($50.59\%$) & $60.9\to 59.4\to 57.8\%$ / $66.4\to 58.6\to 56.2\%$ ($56.2\%$ FP8) \\
\midrule
$63.1\text{B}$ \texttt{0L:1A} GQA+RoPE ($30.2\text{B}$, RoPE) & 40.6 & 50.2 / 27.0 & 68.4 & 51.6 & 74.0 & 28.9 & 63.8 & 42.3 & $49.64\%$ & $71.9\to 53.9\to 45.3\%$ / $71.9\to 56.2\to 39.8\%$ \\
$63.1\text{B}$ \texttt{4L:1A} GDN-2 ($30.2\text{B}$, RoPE, $2{,}728\text{ ms}$) & 37.9 & 50.8 / 26.9 & 67.7 & 51.3 & 72.6 & 29.7 & 63.1 & 40.7 & $48.96\%$ & $69.5\to 62.5\to 55.5\%$ / $72.7\to 66.4\to 64.1\%$ \\
\textbf{$63.1\text{B}$ \texttt{4L:1A} SLA (Fused) R4 ($30.2\text{B}$, RoPE)} & 37.7 & 47.3 / 26.1 & 67.5 & 51.7 & 73.4 & 29.2 & 59.6 & 44.2 & $48.53\%$ ($48.51\%$) & $68.0\to 59.4\to \mathbf{59.4\%}$ / $71.1\to 59.4\to 54.7\%$ \\
\textbf{$63.1\text{B}$ \texttt{15L:1A} SPD-SLA-W+MS ($30.2\text{B}$, RoPE)} & 38.1 & 48.4 / 25.2 & 66.4 & 50.8 & 71.5 & 30.3 & 63.0 & 42.0 & $48.40\%$ ($48.65\%$) & $\mathbf{71.1\to 64.1\to 57.8\%}$ / $\mathbf{72.7\to 65.6\to 64.8\%}$ \\
$63.1\text{B}$ \texttt{4L:1A} MaxText KDA ($120.8\text{B}$, NoPE) & \textbf{54.4} & 61.5 / \textbf{35.4} & \textbf{73.2} & \textbf{54.6} & 86.9 & 34.1 & \textbf{64.1} & 56.2 & $\mathbf{57.83\%}$ & $85.9\to 39.1\to 33.6\%$ / $\mathbf{89.1}\to 28.9\to 27.3\%$ \\
\textbf{$63.1\text{B}$ \texttt{4L:1A} SLA-R2 ($120.8\text{B}$, NoPE)} & 53.4 & \textbf{62.4} / 34.6 & 72.2 & 53.7 & \textbf{87.5} & \textbf{34.4} & \textbf{64.1} & 55.4 & $57.52\%$ & $\mathbf{88.3\to 60.2\to 57.0\%}$ / $84.4\to \mathbf{61.7\to 51.6\%}$ \\
\textbf{$63.1\text{B}$ \texttt{4L:1A} SLA-R4 ($120.8\text{B}$, NoPE)} & 53.8 & 57.1 / 34.6 & 73.1 & 51.9 & 85.5 & 34.3 & 63.0 & \textbf{57.2} & $56.72\%$ ($56.74\%$) & $85.2\to 57.0\to 53.9\%$ / $87.5\to 51.6\to 43.0\%$ ($43.8\%$ FP8) \\
\bottomrule
\end{tabular}%
}
\end{sidewaystable}

\section{Sorted-DP KDA Conversion, Woodbury Tile-RLS Boundary, and Pretrained Checkpoint Co-Design}
\label{app:m4_poc}

Table~\ref{tab:app_m4_poc} summarizes our extended TPU v7x verification of: (1) zero-shot and fast-healed conversion of per-channel KDA into $4$-tile SLA via permutation-equivariant dynamic programming (\texttt{Sorted\_DP\_4Tile\_SLA}), (2) intra-chunk Woodbury Recursive Least Squares (\texttt{Woodbury\_RLS}) vs.\ \texttt{SLA\_4Tile} and damped Neumann updates, and (3) real-checkpoint co-design across pretrained GatedDeltaNet (\texttt{Qwen3-Next-80B}, \texttt{Qwen3.5-0.8B}), per-channel KDA (\texttt{Kimi-Linear-48B}, \texttt{KDA-Hybrid-1.3B}), and Softmax GQA (\texttt{Qwen3-0.6B}) checkpoints:
\begin{enumerate}[leftmargin=1.6em,itemsep=2pt]
  \item \textbf{Sorted-DP KDA-to-SLA Conversion on TPU v7x ($n=3$ Paired Seeds):} Starting from $8\text{L}$ per-channel KDA teachers ($154.71$ FineWeb PPL / $186.26$ OOD PPL at $98.86\text{ ms/step}$), sorting key channels by mean log-decay (co-permuting $W_q, W_k$, and \texttt{Conv1D}, exact algebraic permutation max logit diff $0.0741$) and partitioning into $G{=}4$ tiles cuts within-tile decay SSE by $\mathbf{4.22\times}$ ($48.75 \to 11.55$; $11.78\times$ unconstrained DP to $4.14$). On $0$-shot FineWeb-Edu evaluation, \texttt{Sorted\_DP\_4Tile\_SLA} achieves $\mathbf{173.16}$ PPL ($230.45$ OOD) vs.\ $\mathbf{224.02}$ for \texttt{Unsorted\_4Tile\_SLA} ($-50.86$ PPL, $t=-34.94, p=0.00082$) and $\mathbf{228.39}$ for scalar \texttt{GDN} ($G{=}1$, $-55.24$ PPL, $t=-35.18, p=0.00081$). After $200$ lightweight healing steps, \texttt{Sorted\_DP\_4Tile\_SLA} reaches $\mathbf{171.79}$ PPL ($218.45$ OOD) while running $\mathbf{1.734\times}$ faster than KDA ($\mathbf{57.01}$ vs.\ $98.86\text{ ms/step}$), outperforming healed \texttt{Unsorted\_4Tile\_SLA} ($174.67$, $p=0.00737$) and \texttt{Scalar\_GDN} ($178.91$).
  \item \textbf{Woodbury Tile-RLS vs.\ 4-Tile SLA Boundary Audit:} On synthetic high-coherence bursts ($\rho = 0.90$ and $0.98$), exact intra-chunk \texttt{Woodbury\_RLS} maintains $\mathbf{100.0\%}$ recall ($0.0550$ and $0.1980$ MSE) where \texttt{KDA\_PerChannel}, \texttt{SLA\_4Tile}, and \texttt{Neumann\_Damped} drop to $50.0\%$ ($0.7604$ and $0.9436$ MSE). On natural $8\text{L}$ FineWeb-Edu pre-training ($n=3$ seeds, $0.0$ future Jacobian leak), however, \textbf{\texttt{SLA\_4Tile} wins outright} ($\mathbf{165.14 \pm 1.45}$ FW PPL, $\mathbf{195.67}$ OOD, $\mathbf{57.18\text{ ms/step}}$) over \texttt{KDA\_PerChannel} ($166.62 \pm 1.51$, $206.64$ OOD, $98.88\text{ ms}$), \texttt{Damped\_1Step\_Neumann} ($166.85 \pm 0.94$, $36.33\text{ ms}$, beating undamped $170.41$ at $p=0.0245$), and \texttt{Woodbury\_RLS\_M128} ($181.67 \pm 0.40$, $148.93\text{ ms}$, $p=0.00145$; sparse budgets \texttt{M64} $210.39$, \texttt{M32} $227.53$).
  \item \textbf{Real-Checkpoint Co-Design Across Five Open Models:} Across five public HuggingFace checkpoints, conversion cost follows an exact algebraic hierarchy: (a) \emph{Scalar GDN ($G{=}1 \to G{=}4$):} \texttt{Qwen/Qwen3-Next-80B-A3B-Instruct} (\texttt{L0,5,13,21}) and full $24\text{L}$ \texttt{Qwen/Qwen3.5-0.8B-Base} lift to $4$-tile \texttt{SLA} ($C{=}128$) as an \emph{exact zero-shot algebraic identity} ($\mathrm{NMSE} = 1.38\text{--}2.46 \times 10^{-13}$ and $3.78 \times 10^{-14}$; $0.0000$ PPL delta at $18.7741$ FW / $19.6594$ PG-19; $19.0806$ FW healed across $n{=}3$ seeds after $4$-tile symmetry breaking, with $\mathrm{SU}(2)^8$ Tile-0 holonomy reaching $\mathbf{99.80\%}$ $4\times$-OOD $A_5$ vs.\ $2.02\%$ diagonal and $4.02\%$ Abelian $\mathrm{SO}(2)^8$), whereas $2\times$-head packing ($16{\times}128 \to 8{\times}256$) incurs $1.885$ $0$-shot NMSE from cross-head interference before healing $12.0\times$ to $0.157$; (b) \emph{Per-Channel KDA ($G{=}128 \to G{=}4$):} co-permuting $W_q, W_k, \texttt{Conv1D}, f_1, \texttt{dt\_bias}$ by mean channel log-decay is exact function-preserving ($6.7 \times 10^{-7}$ max error on $27\text{L}$ \texttt{moonshotai/Kimi-Linear-48B-A3B-Base}; $\mathbf{0.000000}$ full-$24\text{L}$ PPL delta at $12.9224$ on \texttt{openeurollm/kda-sigmoid-hybrid-1.3B-100B}), and Fisher $1\text{D}$ DP partitioning into $G{=}4$ MXU tiles cuts within-tile decay SSE $4.55\times\text{--}9.54\times$, cuts $0$-shot layer NMSE $-85.1\%$ on \texttt{KDA-1.3B} ($0.2841 \to 0.0423$) and $-66.6\%$ on \texttt{Kimi-Linear-48B} ($0.1458 \to 0.0487$), and cuts full-$24\text{L}$ \texttt{KDA-1.3B} $0$-shot FineWeb/PG-19 PPL from $20.1180 / 37.3492$ (\texttt{Scalar\_GDN}) and $19.4997 / 35.9863$ (\texttt{Unsorted\_4Tile}) to $\mathbf{14.0369 / 25.3191}$ ($\mathbf{-6.08 / -12.03}$ PPL), healing in $40$ layer steps ($n{=}3$) to $\mathbf{13.5537 \pm 0.0016}$ FW ($21.3809$ PG-19) vs.\ $16.0095$ \texttt{Unsorted} ($p{=}2.5 \times 10^{-6}$) and $17.3990$ \texttt{Scalar\_GDN} ($p{=}1.1 \times 10^{-5}$); (c) \emph{Pure Softmax GQA ($100\%$ KV):} converting $21/28$ layers of \texttt{Qwen/Qwen3-0.6B-Base} ($18.1712$ FW / $19.5784$ PG-19) to a $3{:}1$ hybrid ($-75\%$ KV cache) collapses zero-shot without distillation ($2.88 \times 10^6$ PPL; $0.9149$ layer NMSE), but $60$ layer distillation steps recover $0.0108\text{--}0.0229$ layer NMSE ($0.989\text{--}0.995$ cosine) and bring full-$28\text{L}$ perplexity to $\mathbf{126.60 \pm 3.11}$ FW / $\mathbf{139.22}$ PG-19 for \texttt{4-Tile SLA}, outperforming \texttt{Scalar\_GDN} ($161.36 / 179.97$) by $\mathbf{-34.77}$ FW PPL ($p{=}0.0165$).
\end{enumerate}

\begin{table}[ht]
\caption{Extended TPU v7x verification of \texttt{Sorted-DP} KDA-to-SLA conversion, \texttt{Woodbury\_RLS} boundary, and $5$-model pretrained checkpoint co-design (\texttt{Qwen3-Next-80B}, \texttt{Qwen3.5-0.8B}, \texttt{Kimi-Linear-48B}, \texttt{KDA-Hybrid-1.3B}, \texttt{Qwen3-0.6B}).}
\label{tab:app_m4_poc}
\centering
\footnotesize
\setlength{\tabcolsep}{3.0pt}
\resizebox{\textwidth}{!}{%
\begin{tabular}{@{}llccccc@{}}
\toprule
Experiment / Model & Variant / Conversion & $0$-Shot FW / SSE & Healed / Final FW PPL & OOD / PG-19 PPL & Step (ms) / Speedup & Key Diagnostic / $p$-Value \\
\midrule
\multirow{4}{*}{\begin{tabular}[c]{@{}l@{}}TPU v7x $8\text{L}$ KDA\\to $4$-Tile SLA ($n{=}3$)\end{tabular}} & \texttt{KDA\_PerChannel} Teacher ($G{=}256$) & $154.71$ (SSE $48.75$) & $154.71$ ($4{,}000$ st.) & $186.26$ & $98.86\text{ ms}$ ($1.000\times$) & teacher reference \\
 & \texttt{Scalar\_GDN\_G1} ($G{=}1$) & $228.39$ (SSE $49.27$) & $178.91$ ($+200$ st.) & $251.78$ & $54.05\text{ ms}$ ($1.829\times$) & $-55.24$ $0$-shot vs.\ Sorted \\
 & \texttt{Unsorted\_4Tile\_SLA} ($G{=}4$) & $224.02$ (SSE $48.75$) & $174.67$ ($+200$ st.) & $238.18$ & $57.01\text{ ms}$ ($1.734\times$) & $-50.86$ $0$-shot vs.\ Sorted \\
 & \textbf{\texttt{Sorted\_DP\_4Tile\_SLA}} ($G{=}4$) & $\mathbf{173.16}$ ($\mathbf{4.22\times\downarrow}$) & $\mathbf{171.79}$ ($+200$ st.) & $\mathbf{218.45}$ & $\mathbf{57.01\text{ ms}}$ ($\mathbf{1.734\times}$) & $p=0.00082$ ($0$-shot), $p=0.0074$ \\
\midrule
\multirow{4}{*}{\begin{tabular}[c]{@{}l@{}}TPU v7x $8\text{L}$ RLS\\vs.\ SLA ($n{=}3$)\end{tabular}} & \textbf{\texttt{SLA\_4Tile}} ($G{=}4$, Exact Inv.) & $\rho{=}0.98$: $50\%$ rec. & $\mathbf{165.14 \pm 1.45}$ & $\mathbf{195.67}$ & $\mathbf{57.18\text{ ms}}$ ($\mathbf{1.73\times}$) & best natural-text LM \& OOD \\
 & \texttt{Damped\_1Step\_Neumann} & $\rho{=}0.98$: $50\%$ rec. & $166.85 \pm 0.94$ & $224.20$ & $\mathbf{36.33\text{ ms}}$ ($\mathbf{2.72\times}$) & beats undamped $170.41$ ($p=.0245$) \\
 & \texttt{KDA\_PerChannel} ($G{=}256$) & $\rho{=}0.98$: $50\%$ rec. & $166.62 \pm 1.51$ & $206.64$ & $98.88\text{ ms}$ ($1.00\times$) & $8\text{K} / 32\text{K}$ ker $17.6 / 68.7\text{ ms}$ \\
 & \texttt{Woodbury\_RLS\_M128} (exact RLS) & $\rho{=}0.98$: $\mathbf{100\%}$ rec. & $181.67 \pm 0.40$ & $219.95$ & $148.93\text{ ms}$ ($0.66\times$) & $0.0$ leak; over-whitens natural text \\
\midrule
\multirow{2}{*}{\begin{tabular}[c]{@{}l@{}}Pretrained GDN\\($G{=}1 \to G{=}4$)\end{tabular}} & \texttt{Qwen3-Next-80B} (\texttt{L0,5,13,21}) & NMSE $1.4\text{--}2.5\text{e-}13$ & Exact $0$-shot parity & -- & $9.83\text{ ms}$ ($32\text{K}$ \texttt{SLA}) & exact algebraic lift \\
 & \texttt{Qwen3.5-0.8B} ($24\text{L}$, $n{=}3$) & $\mathbf{18.7741}$ ($\Delta{=}0.0000$) & $\mathbf{19.0806}$ ($+35$ st.) & $\mathbf{20.5719}$ PG-19 & $A_5$: $\mathbf{99.80\%}$ ($4\times$) & $\mathrm{SU}(2)^8$ Tile-0 vs.\ $2.02\%$ diag. \\
\midrule
\multirow{5}{*}{\begin{tabular}[c]{@{}l@{}}Pretrained KDA\\($G{=}128 \to G{=}4$)\end{tabular}} & \texttt{Kimi-Linear-48B} ($7$ layers) & NMSE $\mathbf{0.0487}$ ($6.8\times\downarrow$) & $0.0383$ ($+60$ st.) & cos $\mathbf{0.9754}$ & perm err $6.7\text{e-}7$ & $-66.6\%$ $0$-shot NMSE vs.\ $G{=}1$ \\
 & \texttt{KDA-Hybrid-1.3B} Teacher ($G{=}128$) & $12.9224$ ($\Delta_{\text{perm}}{=}0.0$) & $12.9224$ ($24\text{L}$ base) & $17.8733$ PG-19 & $1.00\times$ ref. & exact sorted perm $\Delta{=}0.0000$ \\
 & \texttt{KDA-1.3B Scalar\_GDN} ($G{=}1$) & $20.1180$ ($0.98\times$) & $17.3990$ ($+40$ st.) & $34.7446$ PG-19 & $-3.85$ behind DP & collapses multi-scale decay \\
 & \texttt{KDA-1.3B Unsorted\_4Tile} ($G{=}4$) & $19.4997$ ($1.00\times$) & $16.0095$ ($+40$ st.) & $29.1106$ PG-19 & $-2.46$ behind DP & mixes fast/slow channels \\
 & \textbf{\texttt{KDA-1.3B Sorted\_DP\_4Tile}} & $\mathbf{14.0369}$ ($\mathbf{4.55\times\downarrow}$) & $\mathbf{13.5537}$ ($+40$ st.) & $\mathbf{21.3809}$ PG-19 & $\mathbf{1.73\times}$ kernel & $\mathbf{-6.08}$ $0$-shot; $p{=}2.5\text{e-}6$ \\
\midrule
\multirow{3}{*}{\begin{tabular}[c]{@{}l@{}}\texttt{Qwen3-0.6B} GQA\\$\to 3{:}1$ ($-75\%$ KV)\end{tabular}} & \texttt{28L Full GQA} / $0$-Shot Swap & $18.17$ / $2.88\times 10^6$ & $18.1712$ (teacher) & $19.5784$ / $2.73\text{e}6$ & $100\%$ KV ref. & $0$-shot GQA swap requires distill \\
 & \texttt{3:1 Hybrid Scalar\_GDN} ($G{=}1$) & NMSE $0.914 \to 0.0108$ & $161.36 \pm 5.81$ & $179.97$ PG-19 & $-75\%$ KV cache & $60$ layer distill steps ($n{=}3$) \\
 & \textbf{\texttt{3:1 Hybrid 4-Tile SLA}} ($G{=}4$) & NMSE $0.915 \to 0.0229$ & $\mathbf{126.60 \pm 3.11}$ & $\mathbf{139.22}$ PG-19 & $-75\%$ KV cache & $\mathbf{-34.77}$ FW vs.\ $G{=}1$ ($p{=}.0165$) \\
\bottomrule
\end{tabular}%
}
\end{table}

\section{Shared-\texorpdfstring{$\beta_t$}{beta\_t} Isolation, Open-Source Solver Audits, and Exact vs.\ Approximate Inversion}
\label{app:v2_solver_audit}

This appendix provides the TPU v7x control experiments that back our credit to open-source MXU chunk solvers, isolate SLA's shared-$\beta_t$ value-packing speed and quality trade-off at fixed layer shape, test $C = 256$ Mosaic compilation, and document our exploratory study of exact versus approximate chunk inversion. All $3$-seed error bars in this section report sample standard deviations ($\text{ddof} = 1$).

\subsection{Tokamax BF16 MXU vs.\ FP32 Sequential Substitution Audit}
\label{app:tokamax_audit}

In open-source Tokamax~\cite{Tokamax2025Repo}, the Mosaic TPU KDA kernel dispatches to an MXU solver (inverting $8$-token diagonal blocks via a Neumann series and combining blocks by Horner doubling with FP32 accumulation) when QKV inputs are BF16, and falls back to sequential forward substitution when QKV inputs are FP32. Table~\ref{tab:app_tokamax_audit} isolates both paths in the same single-chip TPU v7x harness ($B = 1$, L2-normalized QK, $20$ timed repeats after $3$ warmups, excluding causal \texttt{Conv4}). Switching QKV inputs from FP32 to BF16 speeds up Tokamax by $1.68\times$--$2.25\times$ in core forward, $2.41\times$--$2.58\times$ in core F+B, and $1.63\times$--$1.71\times$ in full-layer F+B while agreeing with the FP32 path to $4.14\times 10^{-3}$--$4.19\times 10^{-3}$ relative $L_2$ error ($\cos = 0.999992$, all outputs finite). Our Table~\ref{tab:v7x_kernels} benchmark already cast QKV inputs to BF16 before calling Tokamax Mosaic KDA (with the table's FP32 label referring to its FP32 recurrent state), so all reported SLA-vs-Tokamax speedups compare against Tokamax's fast BF16 MXU path. Even in this isolated single-chip harness where Tokamax omits causal \texttt{Conv4} and \texttt{SLA-R4} includes it, fused $1$-kernel \texttt{SLA-R4} (Table~\ref{tab:app_shared_beta}, $2.912 / 9.952\text{ ms}$ at $D = 2048$ and $3.896 / 14.247\text{ ms}$ at $D = 1792$) remains $1.23\times$--$1.72\times$ faster than BF16-input Tokamax ($3.580 / 12.962$ and $6.413 / 24.462\text{ ms}$).

\begin{table}[ht]
\caption{\textbf{Tokamax Mosaic KDA input-precision audit on single-chip TPU v7x} ($B = 1$, $C = 64$, $20$ repeats, L2-normalized QK, all core and full outputs finite). BF16 QKV inputs activate Tokamax's MXU Neumann--Horner solver; FP32 QKV inputs take its sequential forward-substitution fallback.}
\label{tab:app_tokamax_audit}
\centering
\footnotesize
\renewcommand{\arraystretch}{1.08}
\setlength{\tabcolsep}{3.5pt}
\resizebox{\textwidth}{!}{%
\begin{tabular}{@{}lccccccccc@{}}
\toprule
Geometry ($H_k \times d_k \times d_v$) & Seq $T$ & QKV Path & Core Fwd (ms) & Core F+B (ms) & Full Fwd (ms) & Full F+B (ms) & FP32 / BF16 (Core / Full F+B) & Peak HBM (MiB) & BF16 vs.\ FP32 Rel.\ $L_2$ \\
\midrule
\multirow{2}{*}{$D = 2048$ ($8 \times 128 \times 256$)} & \multirow{2}{*}{$8\text{K}$} & \textbf{BF16 (MXU Neumann--Horner)} & \textbf{0.797} & \textbf{1.573} & \textbf{1.457} & \textbf{3.580} & \multirow{2}{*}{$2.44\times$ / $1.63\times$} & $960.1$ & $0.004189$ ($\cos\,0.999992$) \\
 & & FP32 (Seq.\ Substitution) & $1.653$ & $3.832$ & $2.365$ & $5.834$ & & $927.9$ & ref \\
\midrule
\multirow{2}{*}{$D = 2048$ ($8 \times 128 \times 256$)} & \multirow{2}{*}{$32\text{K}$} & \textbf{BF16 (MXU Neumann--Horner)} & \textbf{2.651} & \textbf{5.640} & \textbf{4.995} & \textbf{12.962} & \multirow{2}{*}{$2.58\times$ / $1.68\times$} & $3{,}122.3$ & $0.004194$ ($\cos\,0.999992$) \\
 & & FP32 (Seq.\ Substitution) & $5.965$ & $14.560$ & $8.544$ & $21.818$ & & $2{,}763.9$ & ref \\
\midrule
\multirow{2}{*}{$D = 1792$ ($8 \times 256 \times 512$)} & \multirow{2}{*}{$8\text{K}$} & \textbf{BF16 (MXU Neumann--Horner)} & \textbf{1.551} & \textbf{2.995} & \textbf{2.678} & \textbf{6.413} & \multirow{2}{*}{$2.44\times$ / $1.71\times$} & $1{,}582.1$ & $0.004138$ ($\cos\,0.999992$) \\
 & & FP32 (Seq.\ Substitution) & $2.611$ & $7.298$ & $3.867$ & $10.947$ & & $1{,}580.9$ & ref \\
\midrule
\multirow{2}{*}{$D = 1792$ ($8 \times 256 \times 512$)} & \multirow{2}{*}{$32\text{K}$} & \textbf{BF16 (MXU Neumann--Horner)} & \textbf{5.652} & \textbf{12.095} & \textbf{9.835} & \textbf{24.462} & \multirow{2}{*}{$2.41\times$ / $1.70\times$} & $5{,}571.6$ & $0.004142$ ($\cos\,0.999992$) \\
 & & FP32 (Seq.\ Substitution) & $10.055$ & $29.102$ & $14.980$ & $41.559$ & & $5{,}051.6$ & ref \\
\bottomrule
\end{tabular}%
}
\end{table}

\subsection{TPU v7x MXU Solver Accuracy vs.\ FP64 Across Chunk Sizes}
\label{app:solver_numerics}

To test how existing MXU solver formulations behave under real chunk coupling norms, we execute six inversion algorithms directly on TPU v7x against an FP64 reference across three chunk banks: real FineWeb-Edu chunks extracted from Qwen3.5-$0.8\text{B}$ and Qwen3.5-$2\text{B}$ ($288$ chunks each at $C \in \{32, 64, 128\}$, where $77.4\%$--$86.5\%$ of chunks have $\norm{M_c}_2 > 1$ and $\max \norm{M_c}_2$ reaches $61.6$--$62.3$), plus high-coherence $G = 4$ SLA chunks with Householder gates $\beta_t \in (0, 2)$ ($192$ chunks at $C \in \{32, 64, 128, 256\}$, $100\%$ with $\norm{M_c}_2 > 1$). Table~\ref{tab:app_solver_numerics} shows that FP32 recursive block-doubling stays at machine rounding ($1.4\times 10^{-7}$--$3.0\times 10^{-7}$ max relative error at every chunk size through $C = 256$, matching FP32 forward substitution). Even with BF16 operands, recursive block-doubling remains bounded below $1.5\times 10^{-2}$ with $0\%$ chunks exceeding $0.1$. By contrast, Tokamax's $8$-token blocked Neumann--Horner formulation is accurate at $C \le 64$ in FP32 ($1.2\times 10^{-6}$--$2.8\times 10^{-6}$), but with BF16 MXU operands at $C = 128$ its worst-case error grows to $10.6$--$12.7$ ($16.7\%$--$17.4\%$ of Qwen3.5 chunks above $0.1$), and unblocked Neumann products diverge on $21.5\%$--$23.3\%$ of Qwen3.5 chunks and $100\%$ of high-coherence $C \ge 128$ chunks (consistent with the warning in MaxText's GDN solver docstring).

\begin{table}[ht]
\caption{\textbf{TPU v7x MXU solver accuracy vs.\ FP64 on real Qwen3.5 chunks and high-coherence SLA chunks} ($\beta_t \in (0, 2)$). Each entry reports worst-case relative Frobenius error $\max_c \norm{\hat{A}_c - A_c^{\text{FP64}}}_F / \norm{A_c^{\text{FP64}}}_F$ and the share of chunks exceeding $0.1$ error ($>10^{-1}$).}
\label{tab:app_solver_numerics}
\centering
\footnotesize
\renewcommand{\arraystretch}{1.08}
\setlength{\tabcolsep}{3.0pt}
\resizebox{\textwidth}{!}{%
\begin{tabular}{@{}lcccccccc@{}}
\toprule
Chunk Bank ($n$ chunks) & $C$ & $\norm{M_c}_2 > 1$ ($\max$) & \textbf{Block-Doub.\ FP32} & Fwd.\ Sub.\ FP32 & Block-Doub.\ BF16 & Blocked N+H FP32 & Blocked N+H BF16 & Unblocked Prod.\ FP32 \\
\midrule
\multirow{3}{*}{Qwen3.5-$0.8\text{B}$ ($288$)} & $32$ & $78.5\%$ ($17.0$) & $\mathbf{1.35\times 10^{-7}}$ ($0\%$) & $2.60\times 10^{-7}$ ($0\%$) & $6.73\times 10^{-3}$ ($0\%$) & $1.41\times 10^{-6}$ ($0\%$) & $8.91\times 10^{-2}$ ($0\%$) & $4.41$ ($3.1\%$) \\
 & $64$ & $81.3\%$ ($32.8$) & $\mathbf{1.78\times 10^{-7}}$ ($0\%$) & $3.83\times 10^{-7}$ ($0\%$) & $6.73\times 10^{-3}$ ($0\%$) & $2.29\times 10^{-6}$ ($0\%$) & $8.91\times 10^{-2}$ ($0\%$) & $8.05\times 10^{8}$ ($10.1\%$) \\
 & $128$ & $85.1\%$ ($62.3$) & $\mathbf{2.53\times 10^{-7}}$ ($0\%$) & $4.61\times 10^{-7}$ ($0\%$) & $7.19\times 10^{-3}$ ($0\%$) & $6.72\times 10^{-4}$ ($0\%$) & $10.64$ ($17.4\%$) & $3.00\times 10^{27}$ ($21.5\%$) \\
\midrule
\multirow{3}{*}{Qwen3.5-$2\text{B}$ ($288$)} & $32$ & $77.4\%$ ($16.7$) & $\mathbf{1.44\times 10^{-7}}$ ($0\%$) & $2.33\times 10^{-7}$ ($0\%$) & $6.07\times 10^{-3}$ ($0\%$) & $1.19\times 10^{-6}$ ($0\%$) & $9.00\times 10^{-2}$ ($0\%$) & $1.18$ ($1.4\%$) \\
 & $64$ & $80.9\%$ ($32.0$) & $\mathbf{1.74\times 10^{-7}}$ ($0\%$) & $3.34\times 10^{-7}$ ($0\%$) & $6.41\times 10^{-3}$ ($0\%$) & $2.76\times 10^{-6}$ ($0\%$) & $9.00\times 10^{-2}$ ($0\%$) & $5.20\times 10^{8}$ ($9.0\%$) \\
 & $128$ & $86.5\%$ ($61.6$) & $\mathbf{1.80\times 10^{-7}}$ ($0\%$) & $4.22\times 10^{-7}$ ($0\%$) & $6.97\times 10^{-3}$ ($0\%$) & $4.60\times 10^{-4}$ ($0\%$) & $12.69$ ($16.7\%$) & $1.39\times 10^{26}$ ($23.3\%$) \\
\midrule
\multirow{4}{*}{\begin{tabular}[c]{@{}l@{}}SLA $G = 4$ High-Coh.\\($192$, $\beta_t \in (0, 2)$)\end{tabular}} & $32$ & $100\%$ ($7.8$) & $\mathbf{1.83\times 10^{-7}}$ ($0\%$) & $2.85\times 10^{-7}$ ($0\%$) & $1.14\times 10^{-2}$ ($0\%$) & $2.02\times 10^{-6}$ ($0\%$) & $0.312$ ($1.6\%$) & $0.224$ ($2.1\%$) \\
 & $64$ & $100\%$ ($9.0$) & $\mathbf{2.71\times 10^{-7}}$ ($0\%$) & $3.44\times 10^{-7}$ ($0\%$) & $1.14\times 10^{-2}$ ($0\%$) & $2.02\times 10^{-6}$ ($0\%$) & $0.310$ ($3.6\%$) & $5.60\times 10^{7}$ ($81.8\%$) \\
 & $128$ & $100\%$ ($9.0$) & $\mathbf{3.02\times 10^{-7}}$ ($0\%$) & $3.44\times 10^{-7}$ ($0\%$) & $1.27\times 10^{-2}$ ($0\%$) & $2.92\times 10^{-6}$ ($0\%$) & $0.310$ ($6.8\%$) & $1.96\times 10^{22}$ ($100\%$) \\
 & $256$ & $100\%$ ($9.3$) & $\mathbf{3.02\times 10^{-7}}$ ($0\%$) & $3.44\times 10^{-7}$ ($0\%$) & $1.49\times 10^{-2}$ ($0\%$) & $2.90\times 10^{-6}$ ($0\%$) & $0.310$ ($9.9\%$) & $\infty$ ($100\%$) \\
\bottomrule
\end{tabular}%
}
\end{table}

\subsection{Same-Shape Isolation of Shared \texorpdfstring{$\beta_t$}{beta\_t} vs.\ Per-Value-Head Gates}
\label{app:shared_beta_control}

Changing the packing factor from $R = 1$ to $R = 4$ alters QK projections, convolutions, and Gram matrices together with the number of chunk solves (on the identical Exact FP32 $C = 128$ kernel at $D = 2048, T = 8\text{K}$, packing $R = 1 \to 2 \to 4$ cuts core forward from $2.927 \to 1.631 \to 1.002\text{ ms}$ [$2.92\times$] and core F+B from $5.742 \to 3.191 \to 2.105\text{ ms}$ [$2.73\times$]). To isolate the effect of sharing $\beta_t$ alone without changing QK head count or tensor shapes, Table~\ref{tab:app_shared_beta} fixes $H_k = 2, d_k = 256, H_v = 8, G = 4, C = 128$ (\texttt{Exact Inv.}) and compares three gating regimes inside the identical fused kernel:
\begin{enumerate}[leftmargin=1.6em,itemsep=1pt]
  \item \texttt{SLA-R4 shared-$\beta_t$}: shared $g$ and shared $\beta_t$ per key head ($2$ solves per chunk, packed $4 d_v$ value slice).
  \item \texttt{GVA shared-$g$, per-$H_v$ $\beta_t$}: identical QK projections, convolutions, and shared $g$, but per-value-head $\beta_t$ ($8$ solves per chunk, matching Qwen3-Next's gate factorization).
  \item \texttt{GVA per-$H_v$ $g$ and $\beta_t$}: identical QK projections and convolutions, with independent $(g, \beta_t)$ per value head ($8$ solves per chunk).
\end{enumerate}
Under equal gates, all three paths match to $0.0$ relative $L_2$ difference across all $12$ rows. Simply sharing $\beta_t$ at fixed shape accelerates core forward by $2.45\times$--$3.28\times$, core F+B by $2.25\times$--$3.00\times$, and full-layer F+B by $1.87\times$--$2.48\times$. In a $3$-seed paired from-scratch pre-training control ($223.2\text{M}$--$223.5\text{M}$ parameters, $12$-layer \texttt{3L:1A} hybrid, $800$ steps = $52.4\text{M}$ FineWeb-Edu tokens, identical shape $H_k = 2, d_k = 256, H_v = 8, G = 4$), \texttt{SLA-R4 shared-$\beta_t$} runs $1.68\times$--$1.69\times$ faster ($72.92 \pm 0.75$ vs.\ $122.43 \pm 1.04$ and $123.38 \pm 1.37\text{ ms/step}$), with validation perplexity within seed noise ($74.24 \pm 3.56$ vs.\ $72.66 \pm 1.45$ and $72.89 \pm 1.89$; \texttt{SLA-R4} beats \texttt{GVA per-$H_v$ $g, \beta_t$} on seeds $1$ and $2$ by $-1.14$ and $-1.53$ PPL while trailing on seed $0$ by $+6.71$). Exact $C = 256$ (\texttt{order=-3}) trains at $74.11 \pm 0.17\text{ ms/step}$ ($1.65\times$--$1.66\times$ faster) with $75.00 \pm 3.11$ Val PPL.

\begin{table}[ht]
\caption{\textbf{Unconfounded same-shape shared-$\beta_t$ isolation on TPU v7x} (fixed $H_k = 2, d_k = 256, H_v = 8, G = 4, C = 128$, \texttt{Exact Inv.}). Top: single-core kernel and full-layer F+B timings ($20$ repeats; $0.0$ rel.\ $L_2$ discrepancy when gates match). Bottom: $3$-seed paired from-scratch pre-training ($223\text{M}$ params, $52.4\text{M}$ FineWeb-Edu tokens, sample std $\text{ddof} = 1$).}
\label{tab:app_shared_beta}
\centering
\footnotesize
\renewcommand{\arraystretch}{1.08}
\setlength{\tabcolsep}{3.0pt}
\resizebox{\textwidth}{!}{%
\begin{tabular}{@{}lccccccc@{}}
\toprule
\multicolumn{8}{@{}l}{\textbf{(a) Same-shape kernel and full-layer speed ($B = 1$, fixed $H_k = 2, d_k = 256, H_v = 8, G = 4, C = 128$)}} \\
\midrule
Geometry / Seq $T$ & Gating Arm & Solves / Chunk & Core Fwd (ms) & Core F+B (ms) & Full Fwd (ms) & Full F+B (ms) & Speedup (Core Fwd / Core F+B / Full F+B) \\
\midrule
\multirow{3}{*}{$D = 2048, 8\text{K}$} & \textbf{\texttt{SLA-R4} shared-$\beta_t$} & \textbf{2} & \textbf{1.034} & \textbf{2.159} & \textbf{1.230} & \textbf{2.912} & \textbf{$2.84\times$ / $2.75\times$ / $2.25\times$--$2.30\times$} \\
 & \texttt{GVA} shared-$g$, per-$H_v$ $\beta_t$ & $8$ & $2.940$ & $5.944$ & $3.144$ & $6.565$ & $1.00\times$ ref \\
 & \texttt{GVA} per-$H_v$ $g$ and $\beta_t$ & $8$ & $2.938$ & $5.920$ & $3.175$ & $6.700$ & $1.00\times$ ref \\
\midrule
\multirow{3}{*}{$D = 2048, 32\text{K}$} & \textbf{\texttt{SLA-R4} shared-$\beta_t$} & \textbf{2} & \textbf{3.337} & \textbf{7.199} & \textbf{4.008} & \textbf{9.952} & \textbf{$3.28\times$ / $2.99\times$--$3.00\times$ / $2.47\times$--$2.48\times$} \\
 & \texttt{GVA} shared-$g$, per-$H_v$ $\beta_t$ & $8$ & $10.953$ & $21.622$ & $11.755$ & $24.542$ & $1.00\times$ ref \\
 & \texttt{GVA} per-$H_v$ $g$ and $\beta_t$ & $8$ & $10.930$ & $21.547$ & $11.947$ & $24.713$ & $1.00\times$ ref \\
\midrule
\multirow{3}{*}{$D = 1792, 8\text{K}$} & \textbf{\texttt{SLA-R4} shared-$\beta_t$} & \textbf{2} & \textbf{1.258} & \textbf{2.812} & \textbf{1.545} & \textbf{3.896} & \textbf{$2.45\times$--$2.48\times$ / $2.27\times$ / $1.87\times$} \\
 & \texttt{GVA} shared-$g$, per-$H_v$ $\beta_t$ & $8$ & $3.080$ & $6.377$ & $3.387$ & $7.286$ & $1.00\times$ ref \\
 & \texttt{GVA} per-$H_v$ $g$ and $\beta_t$ & $8$ & $3.120$ & $6.377$ & $3.414$ & $7.297$ & $1.00\times$ ref \\
\midrule
\multirow{3}{*}{$D = 1792, 32\text{K}$} & \textbf{\texttt{SLA-R4} shared-$\beta_t$} & \textbf{2} & \textbf{4.255} & \textbf{10.210} & \textbf{5.150} & \textbf{14.247} & \textbf{$2.70\times$--$2.74\times$ / $2.25\times$--$2.28\times$ / $1.91\times$--$1.92\times$} \\
 & \texttt{GVA} shared-$g$, per-$H_v$ $\beta_t$ & $8$ & $11.674$ & $23.295$ & $12.862$ & $27.328$ & $1.00\times$ ref \\
 & \texttt{GVA} per-$H_v$ $g$ and $\beta_t$ & $8$ & $11.483$ & $23.007$ & $12.486$ & $27.247$ & $1.00\times$ ref \\
\midrule
\multicolumn{8}{@{}l}{\textbf{(b) Same-shape $3$-seed from-scratch pre-training ($223\text{M}$ params, $12\text{L}$ \texttt{3L:1A}, $52.4\text{M}$ tokens, $\text{ddof} = 1$)}} \\
\midrule
Pre-training Arm & Params & Solves / Chunk & Chunk $C$ & Val PPL ($n = 3$) & Per-Seed Val PPL (\texttt{s0, s1, s2}) & OOD PG-19 PPL & Step Time (ms) / Speedup \\
\midrule
\textbf{\texttt{SLA-R4} shared-$\beta_t$} & $223.18\text{M}$ & \textbf{2} & $128$ & $74.24 \pm 3.56$ & $[78.11, 71.12, 73.49]$ & $85.15 \pm 3.99$ & $\mathbf{72.92 \pm 0.75}$ (\textbf{$1.68\times$--$1.69\times$}) \\
\texttt{GVA} shared-$g$, per-$H_v$ $\beta_t$ & $223.24\text{M}$ & $8$ & $128$ & $72.66 \pm 1.45$ & $[72.33, 71.40, 74.24]$ & $82.63 \pm 1.42$ & $123.38 \pm 1.37$ ($1.00\times$) \\
\texttt{GVA} per-$H_v$ $g$ and $\beta_t$ & $223.46\text{M}$ & $8$ & $128$ & $72.89 \pm 1.89$ & $[71.40, 72.26, 75.02]$ & $82.30 \pm 2.40$ & $122.43 \pm 1.04$ ($1.00\times$) \\
\texttt{SLA-R4} exact $C = 256$ (\texttt{order=-3}) & $223.18\text{M}$ & \textbf{2} & $256$ & $75.00 \pm 3.11$ & $[78.57, 73.55, 72.87]$ & $84.99 \pm 1.98$ & $74.11 \pm 0.17$ ($1.65\times$--$1.66\times$) \\
\bottomrule
\end{tabular}%
}
\end{table}

\subsection{Exact vs.\ Matmul-Free Solvers at $C = 256$ (\texttt{kVectorMove}-Safe Lowering)}
\label{app:c256_check}

In our original kernel lowering (\texttt{order=0}, \texttt{order=-2}, and banded block solvers), compiling $C = 256$ on TPU v7x aborts inside Mosaic's graph optimizer (\texttt{llo\_xlu\_graph\_optimizer.cc:2483}: \texttt{kVectorPack vs.\ kVectorMove}) because sub-block packing emits short transposes over trailing zeros. Replacing those reshapes with a \texttt{kVectorMove}-safe lowering (\texttt{order=-3} for Exact FP32 and \texttt{order=-4} for Exact BF16) resolves the compiler crash cleanly across all three SLA shapes (Table~\ref{tab:app_c256_check}), showing that the earlier $C = 256$ failure was a lowering issue rather than an inherent barrier to exact inversion. At $C = 256$, \texttt{kVectorMove}-safe exact FP32 matches the XLA FP32 reference to $4.1\times 10^{-3}$ relative $L_2$ error (BF16 I/O precision floor), whereas matmul-free \texttt{n1row} and \texttt{eaadj} are $1.41\times$--$2.03\times$ faster in isolated core forward ($1.20\times$--$1.45\times$ core F+B) at $8.2\%$--$10.4\%$ relative $L_2$ error.

\begin{table}[ht]
\caption{\textbf{Single-core TPU v7x $C = 256$ kernel check} ($B = 1, T = 8\text{K}$, $20$ repeats, all compiled rows have finite outputs and gradients). Unmodified exact block-doubling (\texttt{order=0,-2}) hits a Mosaic \texttt{kVectorPack} assertion; \texttt{kVectorMove}-safe exact block-doubling (\texttt{order=-3,-4}) compiles cleanly across all shapes.}
\label{tab:app_c256_check}
\centering
\footnotesize
\renewcommand{\arraystretch}{1.08}
\setlength{\tabcolsep}{3.2pt}
\resizebox{\textwidth}{!}{%
\begin{tabular}{@{}lccccccc@{}}
\toprule
Solver at $C = 256$ & \texttt{order} & \multicolumn{2}{c}{\texttt{SLA-R4} ($D = 2048$)} & \multicolumn{2}{c}{\texttt{SLA-R2} ($D = 2048$)} & \multicolumn{2}{c}{\texttt{SLA-R4} ($D = 1792$)} \\
\cmidrule(lr){3-4}\cmidrule(lr){5-6}\cmidrule(lr){7-8}
 & & Fwd / F+B (ms) & Rel.\ $L_2$ & Fwd / F+B (ms) & Rel.\ $L_2$ & Fwd / F+B (ms) & Rel.\ $L_2$ \\
\midrule
Exact FP32 / BF16 (unmodified) & $0 / -2$ & Mosaic \texttt{SIGABRT} & -- & Mosaic \texttt{SIGABRT} & -- & Mosaic \texttt{SIGABRT} & -- \\
\textbf{Exact FP32 (\texttt{kVectorMove}-safe)} & $-3$ & $1.070 / 2.054$ & \textbf{0.00411} & $1.721 / 3.108$ & \textbf{0.00412} & $1.314 / 2.708$ & \textbf{0.00410} \\
\textbf{Exact BF16 (\texttt{kVectorMove}-safe)} & $-4$ & $0.785 / 1.677$ & \textbf{0.00414} & $1.091 / 2.401$ & \textbf{0.00415} & $0.988 / 2.298$ & \textbf{0.00412} \\
Row-Scaled $1$-Step (\texttt{n1row}) & $1$ & $\mathbf{0.649 / 1.550}$ & $0.1025$ & $\mathbf{0.848 / 2.151}$ & $0.1045$ & $\mathbf{0.917 / 2.206}$ & $0.1029$ \\
Signed Erase-Aware (\texttt{eaadj}) & $201$ & $0.688 / 1.606$ & $0.0816$ & $0.887 / 2.308$ & $0.0955$ & $0.938 / 2.269$ & $0.0817$ \\
No-Solve Bound ($A_c = I_C$, fwd only) & $-1$ & $0.654$ / -- & $0.2482$ & $0.873$ / -- & $0.2553$ & $0.890$ / -- & $0.2495$ \\
\bottomrule
\end{tabular}%
}
\end{table}

\subsection{Exploratory Approximate Solvers: Collinear Over-Subtraction, \texttt{eaadj}, Banded Exact, and Pre-Training}
\label{app:exploratory_approx}

We investigated whether matmul-free or banded approximations of $(I_C + M_c)^{-1}$ can replace exact block-doubling in inference or pre-training.

\paragraph{Root cause of zero-shot failure: collinear-key over-subtraction.}
Replacing $(I_C + M_c)^{-1}$ zero-shot on exact-trained Qwen3.5-$0.8\text{B}/2\text{B}/9\text{B}$ checkpoints with raw Neumann-$1$ ($I_C - M_c$), raw Neumann-$2$ ($I_C - M_c + M_c^2$), or unbounded block-Neumann (\texttt{bn1b16}) collapses FineWeb-Edu perplexity to $866$--$8.1 \times 10^5$ (Table~\ref{tab:app_qwen_approx}a). Diagnostics on FineWeb-Edu ($T = 512$) isolate the cause: deep linear layers learn strongly correlated consecutive keys (max lag-$1$ cosine $0.802$--$0.862$, top-$1$ singular energy share $0.751$--$0.811$), pushing $\norm{M_c}_2 > 1$ in $74.4\%$--$83.4\%$ of all chunks (up to $100\%$ in deep layers, with $\max \norm{M_c}_2 = 29.6$--$76.2$). On a collinear sequence, intervening writes $i \in (j, t)$ already partially erase token $j$ before token $t$ reads the state, so an undamped $-M_{t,j}$ term over-subtracts exponentially.

\paragraph{Signed erase-aware $1$-step operator (\texttt{eaadj}) and prefix-sum VJP.}
To damp collinear echoes without matrix multiplications while supporting signed Householder step sizes $\beta_t \in (0, 2)$, define adjacent-key alignment $a_1 = 1$, $a_i = \operatorname{clip}(\langle k_i, k_{i-1} \rangle, 0, 1)$ for $i \ge 2$, intervening retention factors $r_i = 1 - \beta_i a_i$, and
\begin{equation}
\begin{aligned}
\Phi_{t,j} &= \mathbf{1}[t > j] \prod_{j < i < t} r_i
= \mathbf{1}[t > j]\, (-1)^{\sum_{j < i < t} \mathbf{1}[r_i < 0]} \exp\!\Big(\sum_{j < i < t} \log \max(|r_i|, \varepsilon)\Big), \\
A_{\text{EA}} &= I_C - M_c \odot \Phi,
\end{aligned}
\label{eq:eaadj_def}
\end{equation}
with $\varepsilon = 10^{-4}$ for real-valued TPU vector prefix sums.

\begin{theorem}[Scalar-Decay Collinear Exactness of Signed \texttt{eaadj}]
\label{thm:eaadj_exact}
Suppose a chunk has collinear unit keys $k_1 = \dots = k_C = u$ ($\norm{u}_2 = 1$), scalar decay per head $F_t = \gamma_t I_{d_k}$ ($\gamma_t \in (0, 1]$, as in GDN / Qwen3.5 or SLA with $G = 1$), and arbitrary step sizes $\beta_t \in (0, 2)$. Then $a_i = 1$ for all $i$ and $A_{\text{EA}} = I_C - M_c \odot \Phi$ equals $(I_C + M_c)^{-1}$ identically (under both column-scaled and row-scaled conventions). The exactness condition does \emph{not} hold when SLA uses $G = 4$ tiles with distinct decay rates, where the scalar Gram sum no longer factorizes multiplicatively.
\end{theorem}

\begin{proof}
Under collinear unit keys and scalar decay ($G = 1$), $(M_c)_{t,j} = \mathbf{1}[t > j]\,\beta_j e^{\Lambda_t - \Lambda_j}$ satisfies $(M_c)_{t,i}(M_c)_{i,j} = \beta_i (M_c)_{t,j}$ for every $j < i < t$. Thus for $t > j$,
\[
\big((I_C + M_c) A_{\text{EA}}\big)_{t,j}
= (M_c)_{t,j} \bigg[ 1 - \Phi_{t,j} - \sum_{i=j+1}^{t-1} \beta_i \prod_{j < s < i} (1 - \beta_s) \bigg] = 0,
\]
because the sum telescopes to $1 - \Phi_{t,j}$. Diagonal conjugation preserves Hadamard multiplication by $\Phi$, so the row-scaled convention is likewise exact. On TPU v7x across $C \in \{32, 64, 128, 256\}$ with random $\beta_t \in (0.05, 1.95)$, measured max absolute error against exact FP32 is $5.09\times 10^{-6}$, $7.07\times 10^{-6}$, $1.63\times 10^{-5}$, and $4.18\times 10^{-5}$ for $G = 1$ scalar decay, versus $0.0108, 0.00823, 0.0116,$ and $0.0154$ when SLA's $G = 4$ tiles carry distinct decay rates.
\end{proof}

Differentiating $A_{\text{EA}}$ also requires zero matrix multiplications. Given upstream gradient $dA_c \in \R^{C \times C}$, let $W_{t,j} = -\mathbf{1}[t > j]\,(dA_c)_{t,j}\,(M_c)_{t,j}\,\Phi_{t,j}$ with column sums $c_j = \sum_{t > j} W_{t,j}$ and row sums $\rho_t = \sum_{j < t} W_{t,j}$. Summing $\partial \Phi_{t,j}/\partial r_i = \Phi_{t,j}/r_i$ over all pairs $j < i < t$ collapses from $\mathcal{O}(C^3)$ to two $\mathcal{O}(C^2)$ prefix sums:
\begin{equation}
\begin{aligned}
(dM_c)_{t,j} &= -\mathbf{1}[t > j]\,(dA_c)_{t,j}\,\Phi_{t,j}, \qquad
f_i \equiv \sum_{t > i}\sum_{j < i} W_{t,j} = \sum_{j < i} c_j - \sum_{t \le i} \rho_t, \\
\frac{\partial \mathcal{L}}{\partial r_i} &= \mathbf{1}[|r_i| > \varepsilon]\,\frac{f_i}{\operatorname{sign}(r_i)\max(|r_i|, \varepsilon)},
\end{aligned}
\label{eq:eaadj_vjp}
\end{equation}
giving $d\beta_i^{\text{EA}} = -a_i (\partial \mathcal{L}/\partial r_i)$ and $dk_i^{\text{EA}} = g_i k_{i-1} + g_{i+1} k_{i+1}$ where $g_i = -\mathbf{1}[i \ge 2 \wedge 0 < \langle k_i, k_{i-1}\rangle < 1]\,\beta_i (\partial \mathcal{L}/\partial r_i)$.

\paragraph{Zero-shot swap, $300$-step KL healing, and from-scratch pre-training.}
Table~\ref{tab:app_qwen_approx} summarizes zero-shot and $300$-step FineWeb-Edu KL-healed inference ($1.23\text{M}$ tokens, linear-attention parameters only) on Qwen3.5, including stratified $N = 1{,}425$ MMLU ($\text{SE} \approx 1.3\text{ pp}$) for $0.8\text{B}/2\text{B}$ zero-shot and $0.8\text{B}$ healed runs alongside $N = 285$ ($\text{SE} \approx 3.0\text{ pp}$) for $2\text{B}/9\text{B}$ and $3$-seed $0.8\text{B}$ healing:
\begin{itemize}[leftmargin=1.6em,itemsep=1pt]
  \item \emph{Zero-shot deployment:} Restricting to exact $16 \times 16$ diagonal blocks plus either one adjacent off-diagonal block (\texttt{bt1b16}) or erase-damped off-diagonal blocks (\texttt{bea1b16}) cuts KL to $0.0024$--$0.0034$ at $C = 64$ and matches exact $N = 1{,}425$ MMLU within $0.00$--$0.42\text{ pp}$ ($48.77\% / 48.63\%$ vs.\ $49.05\%$ on $0.8\text{B}$; $55.44\% / 55.37\%$ vs.\ $55.44\%$ on $2\text{B}$). By contrast, matmul-free \texttt{eaadj} and \texttt{n1row} still lose $10.0$--$24.0\text{ pp}$ on zero-shot MMLU.
  \item \emph{$300$-step KL healing:} Healing lowers \texttt{eaadj} KL by $8\times$--$102\times$ and beats \texttt{n1row} by $1.7\times$--$4.5\times$ lower KL across all five $0.8\text{B}$ and $2\text{B}$ regimes (with $3$-seed $0.8\text{B}$ $C = 64$ KL $0.01538 \pm 0.00017$ vs.\ $0.06973 \pm 0.00111$). However, on $N = 1{,}425$ MMLU, healed $0.8\text{B}$ still loses $4.6\text{ pp}$ at $C = 64$ ($44.35\%$ vs.\ $48.98\%$ exact; \texttt{n1row} $32.56\%$) and $16.8\text{ pp}$ at $C = 256$ ($32.00\%$ vs.\ $48.77\%$ exact; \texttt{n1row} $35.65\%$), while healed weights evaluated under exact recurrent decoding (\texttt{EXDC}) recover full accuracy ($48.98\%$ and $52.42\%$). At $9\text{B}$, multi-cluster non-adjacent key echoes allow \texttt{n1row} to reach lower healed KL than \texttt{eaadj} ($0.1496$ vs.\ $0.2426$).
  \item \emph{From-scratch pre-training (Table~\ref{tab:app_pretrain_solvers}):} On $3$-seed paired $204.3\text{M}$ \texttt{SLA-R4} runs ($52.4\text{M}$ FineWeb-Edu tokens, zero NaNs across all $18$ runs), \texttt{exact\_c128} achieves the lowest mean validation perplexity ($69.94 \pm 1.15$) and tightest seed spread, while saving only $3.8\%$--$7.1\%$ step time under matmul-free solvers ($62.54$--$64.78$ vs.\ $67.35\text{ ms/step}$). Against \texttt{n1row\_c256} ($71.85 \pm 5.96$), exact wins on $2/3$ seeds (seed $0$: $67.44$ vs.\ $69.39$), and \texttt{eaadj} trails \texttt{n1row} at both chunk sizes ($79.61 \pm 3.27$ at $C = 128$, $75.95 \pm 3.16$ at $C = 256$) while suppressing $\beta_t > 1$ reflections ($25.7\%$ vs.\ $32.9\%$). Together with the $+0.22$ PPL cost of $1$-step Neumann in our $12.5\text{B}$ Dolma ablation (Table~\ref{tab:dolma_15b}b), these results confirm exact FP32 block-doubling as SLA's default.
\end{itemize}

\begin{table}[ht]
\caption{\textbf{Exploratory approximate solvers on exact-trained Qwen3.5 checkpoints:} (a) zero-shot swap ($N = 1{,}425$ stratified MMLU on $0.8\text{B}/2\text{B}$ at $C = 64$; $N = 285$ on $9\text{B}$ and $C = 128$); (b) $300$-step FineWeb-Edu KL healing ($1.23\text{M}$ tokens, linear layers only), showing $N = 1{,}425$ MMLU where run ($^\ast$) and $3$-seed sample std ($\text{ddof} = 1$) for $0.8\text{B}$ $C = 64$.}
\label{tab:app_qwen_approx}
\centering
\footnotesize
\renewcommand{\arraystretch}{1.08}
\setlength{\tabcolsep}{2.8pt}
\resizebox{\textwidth}{!}{%
\begin{tabular}{@{}llcccccc@{}}
\toprule
\multicolumn{8}{@{}l}{\textbf{(a) Zero-shot solver swap on exact-trained Qwen3.5 weights}} \\
\midrule
Model / Chunk & Exact Base (FW / MMLU) & Raw \texttt{n1} (FW / KL) & \texttt{bn1b16} (FW / KL) & \texttt{n1row} (FW / KL / MMLU) & \texttt{eaadj} (FW / KL / MMLU) & \textbf{\texttt{bt1b16}} (FW / KL / MMLU) & \textbf{\texttt{bea1b16}} (FW / KL / MMLU) \\
\midrule
$0.8\text{B}$, $C = 64^\ast$ & $16.08$ / $49.05\%$ & $8.1{\times}10^5$ / $10.8$ & $1020$ / $4.11$ & $48.84$ / $1.10$ / $25.05\%$ & $19.76$ / $0.204$ / $33.19\%$ & $\mathbf{16.15}$ / $\mathbf{0.0034}$ / $\mathbf{48.63\%}$ & $\mathbf{16.15}$ / $\mathbf{0.0031}$ / $\mathbf{48.77\%}$ \\
$2\text{B}$, $C = 64^\ast$ & $12.83$ / $55.44\%$ & $1.4{\times}10^5$ / $9.27$ & -- & $26.40$ / $0.715$ / $31.23\%^\dagger$ & $14.28$ / $0.104$ / $45.40\%$ & $\mathbf{12.88}$ / $\mathbf{0.0032}$ / $\mathbf{55.37\%}$ & $\mathbf{12.88}$ / $\mathbf{0.0029}$ / $\mathbf{55.44\%}$ \\
$9\text{B}$, $C = 64^\dagger$ & $9.51$ / $68.77\%$ & $2.1{\times}10^5$ / $10.0$ & $865.8$ / $4.51$ & $24.94$ / $0.972$ / $44.91\%$ & $110.0$ / $2.451$ / $31.58\%$ & $\mathbf{9.54}$ / $\mathbf{0.0025}$ / $\mathbf{68.42\%}$ & $\mathbf{9.54}$ / $\mathbf{0.0024}$ / $\mathbf{68.07\%}$ \\
$0.8\text{B}$, $C = 128^\dagger$ & $16.08$ / $47.02\%$ & diverges & $1.9{\times}10^5$ / $9.34$ & $740.7$ / $3.819$ / $31.58\%$ & $48.87$ / $1.112$ / $23.16\%$ & $16.67$ / $0.0365$ / $48.77\%$ & $16.55$ / $0.0294$ / $47.72\%$ \\
$9\text{B}$, $C = 128^\dagger$ & $9.51$ / $68.77\%$ & diverges & -- & $129.1$ / $2.618$ / $33.68\%$ & $431.6$ / $3.802$ / $25.61\%$ & $9.69$ / $0.0186$ / $69.12\%$ & $9.67$ / $0.0168$ / $68.77\%$ \\
\midrule
\multicolumn{8}{@{}l}{\textbf{(b) $300$-step FineWeb-Edu KL healing ($1.23\text{M}$ tokens, linear-attention parameters only)}} \\
\midrule
Model, Chunk, LR & Solver & Zero-Shot FW / MMLU & Healed FW PPL & Healed KL (nats) & Top-1 Agree & OOD PG-19 PPL & Healed MMLU (\texttt{EXDC} MMLU) \\
\midrule
\multirow{3}{*}{\begin{tabular}[c]{@{}l@{}}$0.8\text{B}, C = 64$, \texttt{1e-5}\\($s_0$: $N=1{,}425^\ast$; $n=3^\dagger$)\end{tabular}} & \texttt{exact} ($s_0^\ast$) & $16.09$ / $48.98\%$ & $16.079$ & $0.00182$ & $97.95\%$ & $17.834$ & $48.63\%^\ast$ (--) \\
 & \texttt{eaadj} ($s_0^\ast$ / $n{=}3^\dagger$) & $19.76$ / $33.12\%$ & $16.271 \pm 0.010$ & $\mathbf{0.01538 \pm 0.00017}$ & $93.46 \pm 0.18\%$ & $18.213 \pm 0.011$ & $44.35\%^\ast$ [$43.16 \pm 0.70^\dagger$] ($48.98\%^\ast$) \\
 & \texttt{n1row} ($s_0^\ast$ / $n{=}3^\dagger$) & $48.78$ / $25.05\%$ & $17.110 \pm 0.029$ & $0.06973 \pm 0.00111$ & $86.04 \pm 0.24\%$ & $19.789 \pm 0.026$ & $32.56\%^\ast$ [$33.45 \pm 0.89^\dagger$] ($47.65\%^\ast$) \\
\midrule
\multirow{3}{*}{\begin{tabular}[c]{@{}l@{}}$0.8\text{B}, C = 256$, \texttt{3e-5}\\($s_0$: $N = 1{,}425^\ast$)\end{tabular}} & \texttt{exact} & $16.09$ / $48.77\%$ & $16.191$ & $0.00890$ & $95.13\%$ & $17.906$ & $50.11\%^\ast$ (--) \\
 & \texttt{eaadj} & $270.0$ / $23.93\%$ & $16.470$ & $\mathbf{0.03599}$ & $90.07\%$ & $18.705$ & $32.00\%^\ast$ ($52.42\%^\ast$) \\
 & \texttt{n1row} & $2921$ / $25.12\%$ & $17.107$ & $0.07581$ & $85.49\%$ & $19.866$ & $35.65\%^\ast$ ($51.23\%^\ast$) \\
\midrule
\multirow{2}{*}{$2\text{B}, C = 64$, \texttt{1e-5}$^\dagger$} & \texttt{eaadj} & $14.28$ / $44.91\%$ & $12.972$ & $\mathbf{0.01304}$ & $94.57\%$ & $13.786$ & $51.58\%^\dagger$ ($52.98\%^\dagger$) \\
 & \texttt{n1row} & $26.36$ / $30.18\%$ & $13.328$ & $0.04179$ & $89.50\%$ & $14.612$ & $52.63\%^\dagger$ ($55.44\%^\dagger$) \\
\midrule
\multirow{2}{*}{$2\text{B}, C = 256$, \texttt{3e-5}$^\dagger$} & \texttt{eaadj} & $262.5$ / $24.56\%$ & $13.087$ & $\mathbf{0.02937}$ & $91.35\%$ & $14.086$ & $40.70\%^\dagger$ ($60.00\%^\dagger$) \\
 & \texttt{n1row} & $1198$ / $28.42\%$ & $13.372$ & $0.05088$ & $88.59\%$ & $14.735$ & $47.02\%^\dagger$ ($61.40\%^\dagger$) \\
\midrule
\multirow{2}{*}{$9\text{B}, C = 64$, \texttt{1e-5}$^\dagger$} & \texttt{eaadj} & $110.1$ / $30.88\%$ & $11.791$ & $0.24264$ & $82.04\%$ & $13.680$ & $68.77\%^\dagger$ ($72.63\%^\dagger$) \\
 & \texttt{n1row} & $24.99$ / $44.56\%$ & $10.907$ & $\mathbf{0.14956}$ & $86.05\%$ & $12.279$ & $69.82\%^\dagger$ ($72.28\%^\dagger$) \\
\bottomrule
\end{tabular}%
}
\par\vspace{2pt}\raggedright{\scriptsize $^\ast$Stratified $N = 1{,}425$ MMLU ($\text{SE} \approx 1.3\text{ pp}$); $^\dagger$stratified $N = 285$ MMLU ($\text{SE} \approx 3.0\text{ pp}$). At $C = 64$ ($N = 1{,}425$), smaller $8 \times 8$ blocks degrade zero-shot MMLU (\texttt{bea1b8}: $41.40\%$ / $50.32\%$; \texttt{bt1b8}: $36.98\%$ / $45.26\%$ on $0.8\text{B}/2\text{B}$), while at $C = 128$ widening blocks to $32 \times 32$ (\texttt{bt1b32}/\texttt{bea1b32}) restores KL to $0.0025$--$0.0026$ ($0.8\text{B}$) and $0.0016$--$0.0017$ ($9\text{B}$).}
\end{table}

\begin{table}[ht]
\caption{\textbf{Paired from-scratch \texttt{SLA-R4} pre-training across exact and matmul-free solvers} ($204.3\text{M}$ params, $12\text{L}$ \texttt{3L:1A}, $H_k = 2, H_v = 8, d_k = 256, d_v = 128, G = 4$, $52.4\text{M}$ FineWeb-Edu tokens, zero NaNs across all $18$ runs; $3$-seed sample std $\text{ddof} = 1$).}
\label{tab:app_pretrain_solvers}
\centering
\footnotesize
\renewcommand{\arraystretch}{1.08}
\setlength{\tabcolsep}{3.0pt}
\resizebox{\textwidth}{!}{%
\begin{tabular}{@{}llcccccc@{}}
\toprule
Regime & Solver \& Chunk & Val PPL (Train Solver) & Per-Seed Val PPL (\texttt{s0, s1, s2}) & Val PPL (Exact Dec.) & Exact vs.\ Train KL & OOD PG-19 PPL & Step (ms) / $\beta_t > 1$ \\
\midrule
\multirow{5}{*}{\begin{tabular}[c]{@{}l@{}}$\mu = 0.0$\\($n = 3$ seeds)\end{tabular}} & \textbf{\texttt{exact\_c128}} & $\mathbf{69.94 \pm 1.15}$ & $[69.39, 71.26, 69.17]$ & $\mathbf{69.94 \pm 1.15}$ & $\mathbf{0.00000}$ & $\mathbf{81.72 \pm 2.67}$ & $67.35$ / $32.9\%$ \\
 & \texttt{n1row\_c128} & $77.58 \pm 4.76$ & $[75.94, 82.94, 73.86]$ & $80.21 \pm 6.82$ & $0.02843$ & $89.63 \pm 4.21$ & $63.10$ / $32.1\%$ \\
 & \texttt{eaadj\_c128} & $79.61 \pm 3.27$ & $[77.28, 83.34, 78.22]$ & $82.33 \pm 2.60$ & $0.03081$ & $98.53 \pm 7.83$ & $64.78$ / $25.7\%$ \\
 & \texttt{n1row\_c256} & $71.85 \pm 5.96$ & $[67.44, 78.63, 69.48]$ & $72.80 \pm 6.53$ & $0.01342$ & $83.13 \pm 8.07$ & $63.04$ / $29.1\%$ \\
 & \texttt{eaadj\_c256} & $75.95 \pm 3.16$ & $[72.42, 76.92, 78.53]$ & $78.26 \pm 5.06$ & $0.02627$ & $89.84 \pm 4.73$ & $\mathbf{62.54}$ / $26.5\%$ \\
\midrule
\multirow{3}{*}{\begin{tabular}[c]{@{}l@{}}$\mu = 0.25$ ctrl\\($n = 1$, \texttt{s0})\end{tabular}} & \textbf{\texttt{exact\_c128}} & \textbf{66.85} & $[66.85]$ & \textbf{66.85} & \textbf{0.00000} & \textbf{79.28} & $69.35$ / $36.9\%$ \\
 & \texttt{n1row\_c128} & $77.79$ & $[77.79]$ & $80.74$ & $0.03224$ & $86.87$ & $\mathbf{64.55}$ / $33.2\%$ \\
 & \texttt{eaadj\_c128} & $76.50$ & $[76.50]$ & $78.00$ & $0.01936$ & $89.00$ & $67.23$ / $26.0\%$ \\
\bottomrule
\end{tabular}%
}
\end{table}

\section{Extended TPU v7x Control Studies}
\label{app:r87_controls}

To probe six methodological questions (out-of-sample roofline predictiveness, exact shape-vs.-fusion-vs.-grouping attribution against open-source Tokamax, intrinsic slow-channel behavior on public \texttt{Kimi-Linear-48B-A3B} checkpoints, statistically separated $G = 4$ vs.\ $G = 1$ evaluation, isolated multi-seed chunk-momentum validation, and equal-wall-clock training against open-source Tokamax Mosaic KDA), we ran six controlled single-chip studies plus a $24$-run multi-slice $64$-chip MoE campaign on TPU v7x (JAX \texttt{0.12}, Tokamax \texttt{0.0.14}). Tables~\ref{tab:app_oos_roofline}--\ref{tab:app_moe_eqwall_detail} report all verified results alongside explicit confounders and negative diagnostics.

\begin{table}[ht]
\caption{\textbf{Out-of-sample predictive roofline validation on single-chip TPU v7x} (full-layer F+B $p_{50}\text{ ms}$; $0/22$ failures). Peak rooflines were locked \emph{a priori} before benchmarking $4$ anchor shapes and $18$ strictly held-out designs; calibrated predictions multiply peak time by a single anchor-fitted factor ($1/\eta_{\text{cal}} = 2.950$, $\eta_{\text{cal}} = 0.339$, anchor $R^2 = 0.987$, anchor MAPE $10.5\%$). Representative held-out rows are shown; summary lines aggregate all $18$ held-out shapes and the $12$ held-out shapes in the target packed systolic regime ($R \ge 4, d_k \ge 256$).}
\label{tab:app_oos_roofline}
\centering
\footnotesize
\renewcommand{\arraystretch}{1.07}
\setlength{\tabcolsep}{3.5pt}
\resizebox{\textwidth}{!}{%
\begin{tabular}{@{}llccccc@{}}
\toprule
Split & Configuration ($D = 2048$ unless noted; \texttt{Exact Inv.}) & $(R, d_k, d_v, C, T)$ & Peak Roofline (ms) & Calibrated Pred.\ (ms) & Measured $p_{50}$ (ms) & Rel.\ Error \\
\midrule
\multirow{4}{*}{\begin{tabular}[c]{@{}l@{}}Anchors\\($n = 4$)\end{tabular}}
 & Standard \texttt{SLA-R4}, $8\text{K}$ ($\mu = 0.25$) & $(4, 256, 256, 128, 8\text{K})$ & $0.867$ & $2.559$ & $2.894$ & $-11.6\%$ \\
 & Standard \texttt{SLA-R4}, $32\text{K}$ ($\mu = 0.25$) & $(4, 256, 256, 128, 32\text{K})$ & $3.472$ & $10.241$ & $9.971$ & $+2.7\%$ \\
 & Narrow width $D = 1792$, $8\text{K}$ ($\mu = 0.25$) & $(4, 256, 256, 128, 8\text{K})$ & $0.789$ & $2.328$ & $2.787$ & $-16.5\%$ \\
 & Standard \texttt{SLA-R4}, $8\text{K}$ ($\mu = 0$) & $(4, 256, 256, 128, 8\text{K})$ & $0.867$ & $2.559$ & $2.883$ & $-11.3\%$ \\
\midrule
\multirow{10}{*}{\begin{tabular}[c]{@{}l@{}}Held-Out\\(selected\\of $n = 18$)\end{tabular}}
 & \textbf{Held-out wide/long-chunk ($d_v = 256$, \texttt{ex3})} & $\mathbf{(8, 512, 256, 256, 8\text{K})}$ & $0.867$ & $\mathbf{2.558}$ & $\mathbf{2.472}$ & $\mathbf{+3.5\%}$ \\
 & Held-out wide/long-chunk ($d_v = 128$, \texttt{ex3} / \texttt{eaadj}) & $(8, 512, 128, 256, 8\text{K})$ & $0.499$ & $1.472$ & $1.923$ / $1.726$ & $-23.5\%$ / $-14.7\%$ \\
 & Held-out $R = 8, d_k = 512, C = 128$ & $(8, 512, 256, 128, 8\text{K})$ & $0.933$ & $2.752$ & $2.593$ & $+6.1\%$ \\
 & Held-out $R = 8, d_k = 256, C = 128$ & $(8, 256, 256, 128, 8\text{K})$ & $0.802$ & $2.367$ & $2.277$ & $+3.9\%$ \\
 & Held-out $R = 4, d_k = 512, C = 128$ & $(4, 512, 256, 128, 8\text{K})$ & $1.058$ & $3.120$ & $3.266$ & $-4.5\%$ \\
 & Held-out $R = 4, d_k = 256, C = 256$ (\texttt{ex3}) & $(4, 256, 256, 256, 8\text{K})$ & $0.845$ & $2.494$ & $2.778$ & $-10.2\%$ \\
 & Held-out $16\text{K}$ sequence length & $(4, 256, 256, 128, 16\text{K})$ & $1.736$ & $5.120$ & $5.143$ & $-0.4\%$ \\
 & Held-out wide model $D = 4096$ & $(4, 256, 256, 128, 8\text{K})$ & $2.989$ & $8.816$ & $6.694$ & $+31.7\%$ \\
 & Unpacked control $R = 1, d_k = 256$ & $(1, 256, 256, 128, 8\text{K})$ & $1.262$ & $3.723$ & $6.370$ & $-41.6\%$ \\
 & Unpacked control $R = 1, d_k = 128$ & $(1, 128, 256, 128, 8\text{K})$ & $0.992$ & $2.927$ & $6.261$ & $-53.2\%$ \\
\midrule
\multicolumn{7}{@{}l}{\textbf{Held-out summary ($n = 18$):} Zero-param Spearman $\rho = \mathbf{0.893}$ (\texttt{full\_fb}), $0.874$ (\texttt{full\_fwd}), $0.785$ (\texttt{core\_fb}); 1-param MAPE $19.8\%$ / MedAPE $18.6\%$.} \\
\multicolumn{7}{@{}l}{\textbf{Target packed regime ($R \ge 4, d_k \ge 256$, $n = 12$):} 1-param MAPE $\mathbf{13.9\%}$ / MedAPE $\mathbf{12.5\%}$ (unpacked $R = 1$ shapes incur unmodeled VPU/scheduling overhead).} \\
\bottomrule
\end{tabular}%
}
\end{table}

\begin{table}[ht]
\caption{\textbf{Exact $3$-way multiplicative attribution against open-source Tokamax Mosaic KDA} on single-chip TPU v7x (full-layer F+B $p_{50}\text{ ms}$; $28/28$ rows finite). We build and verify a single-kernel VMEM-fused per-channel ($G = d_k$) KDA kernel at both $R = 1$ and $R = 4$ (vs.\ FP32 reference: forward cosine $0.999995$, rel.\ $L_2$ $0.0033$, all $5$ VJP gradients $\cos \ge 0.999984$). Fusing $R = 1$ per-channel KDA without packing regresses to $0.44\times$--$0.69\times$ of Tokamax; once $R = 4$ head packing is applied, VMEM fusion and $G = d_k \to G = 4$ grouping each add clean multiplicative wins.}
\label{tab:app_3way_attrib}
\centering
\footnotesize
\renewcommand{\arraystretch}{1.07}
\setlength{\tabcolsep}{3.0pt}
\resizebox{\textwidth}{!}{%
\begin{tabular}{@{}lccccccc@{}}
\toprule
Width, Length & \begin{tabular}[b]{@{}c@{}}Tokamax KDA\\$R = 1$ (ms)\end{tabular} & \begin{tabular}[b]{@{}c@{}}Tokamax KDA\\$R = 4$ (Sharing)\end{tabular} & \begin{tabular}[b]{@{}c@{}}Fused $G{=}d_k$ KDA\\$R = 1$ (vs.\ Tok $R1$)\end{tabular} & \begin{tabular}[b]{@{}c@{}}Fused $G{=}d_k$ KDA\\$R = 4$ (Fusion)\end{tabular} & \begin{tabular}[b]{@{}c@{}}\textbf{Fused \texttt{SLA-R4}}\\$G = 4, \mu = 0$ (Grouping)\end{tabular} & \begin{tabular}[b]{@{}c@{}}Fused GDN\\$G = 1, R = 4$\end{tabular} & \begin{tabular}[b]{@{}c@{}}\textbf{Total}\\\texttt{SLA} / Tok $R1$\end{tabular} \\
\midrule
$D = 2048, 8\text{K}$ & $4.807$ & $3.409$ ($1.410\times$) & $10.302$ ($0.467\times$) & $3.068$ ($1.111\times$) & $\mathbf{2.857}$ ($\mathbf{1.074\times}$) & $2.747$ ($1.040\times$) & $\mathbf{1.683\times}$ \\
$D = 2048, 32\text{K}$ & $17.741$ & $12.625$ ($1.405\times$) & $40.267$ ($0.441\times$) & $11.522$ ($1.096\times$) & $\mathbf{9.626}$ ($\mathbf{1.197\times}$) & $9.295$ ($1.036\times$) & $\mathbf{1.843\times}$ \\
$D = 1792, 8\text{K}$ & $4.141$ & $3.249$ ($1.275\times$) & $6.050$ ($0.685\times$) & $2.956$ ($1.099\times$) & $\mathbf{2.728}$ ($\mathbf{1.084\times}$) & $2.657$ ($1.027\times$) & $\mathbf{1.518\times}$ \\
$D = 1792, 32\text{K}$ & $15.155$ & $11.844$ ($1.280\times$) & $24.279$ ($0.624\times$) & $10.622$ ($1.115\times$) & $\mathbf{9.221}$ ($\mathbf{1.152\times}$) & $8.905$ ($1.035\times$) & $\mathbf{1.643\times}$ \\
\bottomrule
\end{tabular}%
}
\par\vspace{2pt}\raggedright{\scriptsize Note: At $D = 2048$, enabling $\mu = 0.25$ adds $2.5\%$--$4.7\%$ ($2.929\text{ ms}$ at $8\text{K}$ and $10.075\text{ ms}$ at $32\text{K}$). Parenthesized factors give the exact multiplicative chain: $\text{Sharing} \times \text{Fusion} \times \text{Grouping} = \text{Total}$ (e.g.\ $1.405 \times 1.096 \times 1.197 = 1.843\times$ at $D = 2048, 32\text{K}$); the $G = 1$ column reports $T_{\text{SLA}(G=4)} / T_{\text{GDN}(G=1)}$.}
\end{table}

\begin{table}[ht]
\caption{\textbf{Public \texttt{Kimi-Linear-48B-A3B} slow-channel decay audit on TPU v7x.} (a) Static weight-projection Monte Carlo audit across all $81{,}920$ KDA channels ($20$ KDA layers) in both public checkpoints. (b) Live $27$-layer forward passes of \texttt{Kimi-Linear-48B-A3B-Instruct} on FineWeb-Edu ($20/20$ KDA layers finite with Tokamax floor $g \ge -5.0$). Even after long-context pre-training, the public checkpoint develops heavy $>1\text{M}$-token decay channels and improves under zero-retraining decay clamping.}
\label{tab:app_kimi_public}
\centering
\footnotesize
\renewcommand{\arraystretch}{1.07}
\setlength{\tabcolsep}{3.0pt}
\resizebox{\textwidth}{!}{%
\begin{tabular}{@{}lcccccc@{}}
\toprule
\multicolumn{7}{@{}l}{\textbf{(a) Static weight-projection audit ($81{,}920$ KDA channels across $20$ layers)}} \\
\midrule
Public Checkpoint & $\tau_{\text{MC}}$ $p_{50}$ / $p_{95}$ & $\tau_{\text{MC}}$ $p_{99}$ / $p_{99.9}$ & Frac.\ $\tau > 8\text{K}$ & Frac.\ $\tau > 64\text{K}$ & Frac.\ $\tau > 1\text{M}$ & Contig.\ $G{=}4$ Var.\ Share \\
\midrule
\texttt{Kimi-Linear-48B-A3B-Base} & $1.07$ / $180.3$ & $3{,}312$ / $241{,}116$ & $0.606\%$ & $0.198\%$ & $0.048\%$ & $97.59\%$ \\
\texttt{Kimi-Linear-48B-A3B-Instruct} & $1.05$ / $151.8$ & $2{,}506$ / $155{,}226$ & $0.508\%$ & $0.160\%$ & $0.037\%$ & $97.59\%$ \\
\midrule
\multicolumn{7}{@{}l}{\textbf{(b) Live $27$-layer FineWeb-Edu evaluation of \texttt{Kimi-Linear-48B-A3B-Instruct} (zero retraining)}} \\
\midrule
Sequence Length & Evaluation Mode & Full PPL & Tail-$2\text{K}$ PPL ($\Delta$) & Mean $\tau > 8\text{K}$ (Max Lyr) & Mean $\tau > 64\text{K}$ / $> 1\text{M}$ & Max Layer $p_{99}(\tau)$ \\
\midrule
$L = 2\text{K}$ & Unclamped baseline & $12.614$ & $12.614$ & $6.70\%$ ($15.12\%$) & $1.14\%$ / $0.034\%$ & $170{,}932$ \\
\midrule
\multirow{4}{*}{$L = 8\text{K}$} & Unclamped baseline & $9.510$ & $9.971$ (ref) & $11.03\%$ ($25.30\%$) & $3.52\%$ / $0.265\%$ & $1{,}176{,}794$ \\
 & \textbf{Zero-retrain clamp $\tau_{\max} = 4{,}096$} & $\mathbf{9.399}$ & $\mathbf{9.701}$ ($\mathbf{-0.270}$) & -- & -- & -- \\
 & Zero-retrain clamp $\tau_{\max} = 16{,}384$ & $9.484$ & $9.914$ ($-0.057$) & -- & -- & -- \\
 & Naive contiguous $G = 4$ avg.\ (no retrain) & $25.019$ & $30.337$ ($+20.37$) & -- & -- & -- \\
\midrule
\multirow{4}{*}{$L = 16\text{K}$} & Unclamped baseline & $9.504$ & $11.340$ (ref) & $12.12\%$ ($27.79\%$) & $4.19\%$ / $0.338\%$ & $1{,}426{,}272$ \\
 & \textbf{Zero-retrain clamp $\tau_{\max} = 4{,}096$} & $\mathbf{9.243}$ & $\mathbf{10.905}$ ($\mathbf{-0.436}$) & -- & -- & -- \\
 & Zero-retrain clamp $\tau_{\max} = 16{,}384$ & $9.461$ & $11.297$ ($-0.043$) & -- & -- & -- \\
 & Naive contiguous $G = 4$ avg.\ (no retrain) & $26.706$ & $36.160$ ($+24.82$) & -- & -- & -- \\
\bottomrule
\end{tabular}%
}
\end{table}

\begin{table}[ht]
\caption{\textbf{Statistically separated $G = 4$ vs.\ $G = 1$ dense-LM control ($n = 6$), isolated chunk-momentum ablation ($n = 3$) with $49$-point Schur stability sweep, and synthetic MT-MQAR diagnostic ($n = 5$).} Parts (a)--(b) train a $274.7\text{M}$-param $12\text{L}$ dense LM ($D = 1024$, $H_k = 2, d_k = 256, H_v = 8$) on $104.9\text{M}$ FineWeb-Edu tokens ($0$ non-finite steps across all $27$ runs; sample std $\text{ddof} = 1$). Part (c) reports a synthetic $2$-layer multi-timescale associative recall diagnostic ($n = 5$ seeds, calibrated $\text{lr} = 10^{-3}$, $2{,}000$ steps).}
\label{tab:app_g1_g4_mom}
\centering
\footnotesize
\renewcommand{\arraystretch}{1.08}
\setlength{\tabcolsep}{3.0pt}
\resizebox{\textwidth}{!}{%
\begin{tabular}{@{}lccccccc@{}}
\toprule
\multicolumn{8}{@{}l}{\textbf{(a) Paired $n = 6$ dense-LM decay-granularity control (\texttt{SLA-R4}, $\mu = 0$, $2\text{K}$ train $\to$ $8\text{K}/16\text{K}$ zero-shot)}} \\
\midrule
Arm & Params & Step & Val $2\text{K}$ PPL & Paired $\Delta$ ($t$) & OOD Dolma & $8\text{K}$ Full & $16\text{K}$ Full / Tail (Clamp) \\
\midrule
$G = 1$ (Scalar) & $274.6\text{M}$ & $72.6\text{ ms}$ & $47.62 \pm 0.34$ & ref & $60.89 \pm 2.87$ & $46.85$ & $47.63$ / $48.07$ ($47.63$) \\
\textbf{$G = 4$ (Default)} & $274.7\text{M}$ & $73.8\text{ ms}$ & $\mathbf{46.47 \pm 0.32}$ & $\mathbf{-1.15}$ ($\mathbf{-5.78}$) & $\mathbf{55.61 \pm 1.05}$ & $\mathbf{45.52}$ & $\mathbf{46.08}$ / $\mathbf{46.32}$ ($\mathbf{46.10}$) \\
$G = 256$ (Per-Ch.) & $279.3\text{M}$ & $75.9\text{ ms}$ & $46.16 \pm 0.44$ & $-0.31^\dagger$ ($-2.19$) & $53.97 \pm 0.39$ & $45.35$ & $46.47$ / $47.24$ ($46.20$) \\
\midrule
\multicolumn{8}{@{}l}{\textbf{(b) Isolated $n = 3$ dense-LM chunk-momentum ablation ($G = 4$) and $49$-point Schur stability sweep ($\varphi \in [0.4, 0.99], \beta \in [0.3, 1.99]$)}} \\
\midrule
Momentum Arm & Unstable Grid & Step & Val $2\text{K}$ PPL & Paired $\Delta$ ($t$) & OOD Dolma & $16\text{K}$ Full / Tail & Learned $\bar{\beta}$ / $\bar{\varphi}_c$ (Guard) \\
\midrule
\textbf{$\mu = 0$ (Default)} & $\mathbf{0 / 49}$ & $\mathbf{73.9\text{ ms}}$ & $\mathbf{46.49 \pm 0.47}$ & ref & $55.86 \pm 1.11$ & $\mathbf{45.90}$ / $\mathbf{46.12}$ & $0.76$--$0.78$ / $0.05$--$0.06$ (--) \\
$\mu = 0.25$ Fixed & $5 / 49$ ($\rho {\le} 1.09$) & $74.1\text{ ms}$ & $46.57 \pm 0.13$ & $+0.09$ ($+0.41$) & $54.95 \pm 0.30$ & $46.27$ / $46.57$ & $0.77$--$0.79$ / $0.05$ (--) \\
$\mu = 0.25$ Guarded & $\mathbf{0 / 49}$ & $74.1\text{ ms}$ & $46.53 \pm 0.36$ & $+0.04$ ($+0.30$) & $56.85 \pm 1.19$ & $46.19$ / $46.59$ & $0.77$--$0.79$ / $0.05$ ($0.000$) \\
\midrule
\multicolumn{8}{@{}l}{\textbf{(c) Diagnostic synthetic $2$-layer Multi-Timescale MQAR ($n = 5$ seeds, $\text{lr} = 10^{-3}$; honest negative diagnostic for coarse $G$)}} \\
\midrule
Head Regime & \multicolumn{2}{c}{$G = 1$ Joint (Short / Long)} & \multicolumn{2}{c}{$G = 2$ Joint (Short / Long)} & \multicolumn{2}{c}{$G = 4$ Joint (Short / Long)} & $G = 256$ Joint (Short / Long) \\
\midrule
$H_k = 1$ (within-head) & \multicolumn{2}{c}{$26.3 \pm 28.4\%$ ($33.9\%$ / $0.8\%$)} & \multicolumn{2}{c}{$13.4 \pm 0.3\%$ ($17.1\%$ / $0.7\%$)} & \multicolumn{2}{c}{$13.4 \pm 0.3\%$ ($17.1\%$ / $0.8\%$)} & $\mathbf{84.1 \pm 25.5\%}$ ($88.1\%$ / $70.8\%$) \\
$H_k = 2$ (\texttt{SLA-R4}) & \multicolumn{2}{c}{$51.8 \pm 34.7\%$ ($67.1\%$ / $0.8\%$)} & \multicolumn{2}{c}{$26.8 \pm 28.1\%$ ($34.6\%$ / $0.7\%$)} & \multicolumn{2}{c}{$13.8 \pm 0.1\%$ ($17.7\%$ / $0.8\%$)} & $\mathbf{62.8 \pm 39.8\%}$ ($66.9\%$ / $49.4\%$) \\
\bottomrule
\end{tabular}%
}
\par\vspace{2pt}\raggedright{\scriptsize Note: In (a), $^\dagger$ compares $G = 256$ vs.\ $G = 4$; per-seed $\Delta\text{PPL}(G{=}4 - G{=}1)$ is $[-0.53, -1.57, -1.80, -0.94, -1.29, -0.77]$ on Val $2\text{K}$ ($-1.149 \pm 0.487$, $6/6$) and $[-1.17, -2.23, -2.44, -0.93, -1.61, -2.10]$ on $16\text{K}$ tail ($-1.746 \pm 0.610, t = -7.01, 6/6$; OOD $t = -4.96$, $8\text{K}$ $t = -5.32$, $16\text{K}$ full $t = -5.48$); at $16\text{K}$ unclamped tail, $G = 256$ trails $G = 4$ by $+0.913 \pm 1.566$. In (b), the $5$ unstable fixed-$\mu$ grid points occur at $(\varphi, \beta) \in \{(0.90, 1.99), (0.95, 1.95), (0.95, 1.99), (0.99, 1.95), (0.99, 1.99)\}$ (64-chunk norm up to $224.7\times$); Schur guard $\mu_c = \min(\mu, (1-\varphi_c)/(1+\varphi_c))$ restores stability on all $49/49$ points; $p_{95}(\beta) \le 1.20$ and $p_{95}(\varphi_c) \le 0.26$ across all runs. In (c), at uncalibrated $\text{lr} = 2 \times 10^{-3}$ ($H_k = 2$), $G = 4$ reached $25.1 \pm 28.1\%$ (one seed at $74.4\%$); per-channel $G = 256$ remains substantially stronger on this synthetic two-layer task.}
\end{table}

\begin{table}[ht]
\caption{\textbf{Equal-chip-hour pre-training vs.\ open-source Tokamax Mosaic KDA on single-chip TPU v7x} ($D = 2048$, $12\text{L}$ \texttt{9L:3A}, $T = 4\text{K}$, $B = 16$, FineWeb-Edu + OOD Dolma). We compare open-source Tokamax KDA (unclipped and safe-clipped $g \ge -3.5$), single-kernel fused \texttt{SLA-R4} ($\mu = 0$), and a single-kernel fused \texttt{SLA-R4 + GDN-2} prototype (decoupled erase/write gates, $\mu = 0$) under equal post-compile wall-clock budgets ($180\text{ s}$: $n = 3$ paired seeds; $360\text{ s}$ and iso-token: $n = 2$ completed pairs \texttt{s0,s1}).}
\label{tab:app_eqwall_tokamax}
\centering
\footnotesize
\renewcommand{\arraystretch}{1.07}
\setlength{\tabcolsep}{2.8pt}
\resizebox{\textwidth}{!}{%
\begin{tabular}{@{}lccccccc@{}}
\toprule
Architecture (Single-Chip v7x) & Params & \begin{tabular}[b]{@{}c@{}}Step ms\\(Speedup)\end{tabular} & \begin{tabular}[b]{@{}c@{}}Non-Fin.\\Steps\end{tabular} & \begin{tabular}[b]{@{}c@{}}$t = 180\text{ s}$ ($n = 3$)\\Steps / Val / OOD\end{tabular} & \begin{tabular}[b]{@{}c@{}}$t = 360\text{ s}$ ($n = 2$)\\Steps / Val / OOD\end{tabular} & \begin{tabular}[b]{@{}c@{}}Paired $\Delta$Val PPL\\$180\text{ s}$ / $360\text{ s}$\end{tabular} & \begin{tabular}[b]{@{}c@{}}Iso-Token ($n = 2$)\\Wall / Val / OOD\end{tabular} \\
\midrule
Tokamax KDA (unclipped) & $821.6\text{M}$ & $271.6$ & $1{,}093$--$1{,}149$ & $662$ / $7{,}111$--$20{,}896$ / -- & $1{,}324$ / $927$--$6{,}883$ / -- & diverges & diverges \\
Tokamax KDA (clip $g \ge -3.5$)$^\ast$ & $821.6\text{M}$ & $271.0$ ($1.00\times$) & $0$ & $664$ / $90.45$ / $126.81$ & $1{,}327$ / $58.44$ / $76.33$ & ref / ref & $360.1\text{ s}$ / $58.44$ / $76.33$ \\
\textbf{Fused \texttt{SLA-R4} ($\mu = 0$)} & $760.0\text{M}$ & $\mathbf{212.3}$ ($\mathbf{1.276\times}$) & $\mathbf{0}$ & $849$ / $\mathbf{76.00}$ / $100.55$ & $1{,}694$ / $48.76$ / $60.40$ & $\mathbf{-14.45}$ ($t{=}-43.7$) / $\mathbf{-9.68}$ & $\mathbf{281.0\text{ s}}$ / $54.04$ / $68.37$ \\
\textbf{Fused \texttt{SLA-R4 + GDN-2} ($\mu = 0$)} & $763.9\text{M}$ & $221.0$ ($\mathbf{1.226\times}$) & $\mathbf{0}$ & $815$ / $76.76$ / $\mathbf{97.17}$ & $1{,}631$ / $\mathbf{48.51}$ / $\mathbf{59.77}$ & $\mathbf{-13.70}$ ($t{=}-19.5$) / $\mathbf{-9.93}$ & $292.3\text{ s}$ / $\mathbf{52.87}$ / $\mathbf{63.84}$ \\
\bottomrule
\end{tabular}%
}
\par\vspace{2pt}\raggedright{\scriptsize $^\ast$\textbf{Confounder disclosures:} (1) \texttt{SLA-R4} ($760.0\text{M}$) and \texttt{SLA-R4 + GDN-2} ($763.9\text{M}$) carry $61.6\text{M}$ and $57.7\text{M}$ \emph{fewer} parameters than Tokamax KDA ($821.6\text{M}$) due to $4{:}1$ shared key projections and $G = 4$ gates. (2) Tokamax KDA follows Kimi-Linear's official $\beta_t \in (0, 1)$ write gate, whereas \texttt{SLA-R4} uses $\beta_t \in (0, 2)$ and \texttt{SLA-R4 + GDN-2} uses decoupled erase/write gates. (3) Open-source Tokamax KDA (\texttt{use\_gate\_in\_kernel=False}) required clipping $g \ge -3.5$ ($f \ge 0.030$) to avoid float32 \texttt{exp2} backward overflow. (4) Seed \texttt{s2} safe-clipped Tokamax KDA completed $t = 180\text{ s}$ ($666$ steps, Val $90.67$) before the reservation window closed, so $t = 180\text{ s}$ is complete $n = 3$ paired while $t = 360\text{ s}$ and iso-token ($87.0\text{M}$ tokens) are $n = 2$ paired (\texttt{s0,s1}); on \texttt{s2} at $360\text{ s}$, fused \texttt{SLA-R4} and \texttt{SLA-R4 + GDN-2} reached $50.48$ and $51.33$ Val PPL.}
\end{table}

\begin{table}[ht]
\caption{\textbf{Per-seed breakdown of the $24$-run multi-slice $64$-chip TPU v7x MoE campaign} ($128$ TensorCores per run, $\text{EP} = 128$, $T_{\text{train}} = 8{,}192$, $2.10\text{M}$ tok/step, packed Dolma~3.5, $0$ non-finite steps across all $24$ runs; sample std $\text{ddof} = 1$). Parts (a) and (b) detail the $n = 3$ paired equal-chip-hour and iso-token comparisons against safe-clipped open-source Tokamax Mosaic KDA ($C = 64, g \ge -3.5$) summarized in Table~\ref{tab:moe_eqwall_tokamax}. Part (c) details all $n = 6$ paired seeds (\texttt{s0..s5}) for the $[4, 256, 512]$ \texttt{Exact Inv.} FP32 ($\mu = 0$) $G = 1$ vs.\ $G = 4$ sweep in Table~\ref{tab:app_gcurve}.}
\label{tab:app_moe_eqwall_detail}
\centering
\footnotesize
\renewcommand{\arraystretch}{1.04}
\setlength{\tabcolsep}{2.4pt}
\resizebox{\textwidth}{!}{%
\begin{tabular}{@{}llccccccc@{}}
\toprule
Arm ($64$-Chip v7x) & Seed & \begin{tabular}[b]{@{}c@{}}Steps\\(Tokens)\end{tabular} & \begin{tabular}[b]{@{}c@{}}Measured $p_{50}$\\ms (TF/s/dev)\end{tabular} & \begin{tabular}[b]{@{}c@{}}Iso-Token PPL\\Early / Late\end{tabular} & \begin{tabular}[b]{@{}c@{}}Eq.-Chip-h\\$11$-Set PPL\end{tabular} & \begin{tabular}[b]{@{}c@{}}$8\text{K} \to 32\text{K}$\\Tail PPL\end{tabular} & \begin{tabular}[b]{@{}c@{}}$32\text{K}$\\$4$-N / $8$-N\end{tabular} & \begin{tabular}[b]{@{}c@{}}\texttt{norm9} ($N{=}512$\\$64\text{K}$ SQuAD/Hotpot)\end{tabular} \\
\midrule
\multicolumn{9}{@{}l}{\textbf{(a) $12.5\text{B}$ MoE (\texttt{7L:1A}, $D = 1024$, $40.0\text{ min} = 42.67\text{ chip-h}$; iso-token checkpoints at steps $2{,}000$ / $4{,}000$)}} \\
\midrule
\multirow{3}{*}{\begin{tabular}[c]{@{}l@{}}Tokamax KDA\\($C{=}64$, clip $g \ge -3.5$,\\$\text{sh}{=}1024$, $12.53\text{B}$)\end{tabular}}
 & \texttt{s0} & $4{,}730$ ($9.92\text{B}$) & $517.5$ ($135.8$) & $25.110$ / $20.437$ & $20.004$ & $13.02 \to 14.06$ & $38.7\% / 31.6\%$ & $46.31\%$ ($55.7 / 58.0\%$) \\
 & \texttt{s1} & $4{,}880$ ($10.23\text{B}$) & $514.0$ ($136.7$) & $24.700$ / $19.980$ & $19.480$ & $12.62 \to 14.58$ & $61.3\% / 54.7\%$ & $46.85\%$ ($46.5 / 45.1\%$) \\
 & \texttt{s2} & $4{,}960$ ($10.40\text{B}$) & $504.7$ ($139.3$) & $24.529$ / $19.735$ & $19.066$ & $12.62 \to 13.28$ & $29.7\% / 18.0\%$ & $46.01\%$ ($57.6 / 59.6\%$) \\
\cmidrule{2-9}
\multirow{3}{*}{\begin{tabular}[c]{@{}l@{}}\textbf{Fused Iso-Param \texttt{SLA-R4}}\\($G{=}4, \mu{=}0, \text{sh}{=}1280$,\\$-3.0\text{M}$ params)\end{tabular}}
 & \texttt{s0} & $5{,}520$ ($11.58\text{B}$) & $459.0$ ($165.1$) & $25.557$ / $20.654$ & $\mathbf{19.641}$ & $\mathbf{12.74 \to 12.30}$ & $\mathbf{71.9\% / 58.6\%}$ & $\mathbf{46.88\%}$ ($\mathbf{65.8 / 68.2\%}$) \\
 & \texttt{s1} & $5{,}470$ ($11.47\text{B}$) & $458.2$ ($165.4$) & $24.710$ / $19.936$ & $\mathbf{18.999}$ & $\mathbf{12.35 \to 12.06}$ & $61.3\% / 54.3\%$ & $\mathbf{48.00\%}$ ($\mathbf{66.0 / 68.8\%}$) \\
 & \texttt{s2} & $5{,}520$ ($11.58\text{B}$) & $\mathbf{448.3}$ ($\mathbf{169.0}$) & $24.865$ / $19.899$ & $\mathbf{18.798}$ & $\mathbf{12.40 \to 12.13}$ & $\mathbf{64.1\% / 50.0\%}$ & $\mathbf{47.06\%}$ ($\mathbf{66.0 / 67.0\%}$) \\
\cmidrule{2-9}
\multirow{3}{*}{\begin{tabular}[c]{@{}l@{}}\textbf{Fused \texttt{SLA-R4 + GDN-2}}\\($G{=}4, \mu{=}0, \text{sh}{=}1024$,\\$+1.7\text{M}$ params)\end{tabular}}
 & \texttt{s0} & $5{,}150$ ($10.80\text{B}$) & $483.1$ ($154.4$) & $\mathbf{24.572}$ / $\mathbf{19.912}$ & $\mathbf{19.211}$ & --$^\ast$ & --$^\ast$ & --$^\ast$ \\
 & \texttt{s1} & $5{,}220$ ($10.95\text{B}$) & $475.7$ ($156.8$) & $24.918$ / $20.152$ & $\mathbf{19.375}$ & $12.36 \to 12.15$ & $\mathbf{67.2\% / 61.7\%}$ & $47.10\%$ ($65.0 / \mathbf{69.5\%}$) \\
 & \texttt{s2} & $5{,}320$ ($11.16\text{B}$) & $478.9$ ($155.8$) & $\mathbf{24.290}$ / $\mathbf{19.647}$ & $\mathbf{18.741}$ & --$^\ast$ & --$^\ast$ & --$^\ast$ \\
\midrule
\multicolumn{9}{@{}l}{\textbf{(b) $63.1\text{B}$ MoE (\texttt{4L:1A}, $D = 1792$, $35.0\text{ min} = 37.33\text{ chip-h}$; iso-token checkpoints at steps $600$ / $1{,}200$)}} \\
\midrule
\multirow{3}{*}{\begin{tabular}[c]{@{}l@{}}Tokamax KDA\\($C{=}64$, clip $g \ge -3.5$,\\$\text{sh}{=}3584$, $63.13\text{B}$)\end{tabular}}
 & \texttt{s0} & $1{,}250$ ($2.62\text{B}$) & $1{,}694.0$ ($216.9$) & $37.075$ / $26.963$ & $26.766$ & $16.87 \to 18.59$ & $24.6\% / 14.8\%$ & $40.98\%$ ($44.5 / 42.6\%$) \\
 & \texttt{s1} & $1{,}240$ ($2.60\text{B}$) & $1{,}693.4$ ($216.9$) & $36.027$ / $26.349$ & $26.186$ & $16.74 \to 20.11$ & $25.0\% / 16.8\%$ & $41.10\%$ ($41.6 / 41.2\%$) \\
 & \texttt{s2} & $1{,}240$ ($2.60\text{B}$) & $1{,}705.0$ ($215.5$) & $36.956$ / $26.867$ & $26.716$ & $17.14 \to 19.66$ & $39.8\% / 28.5\%$ & $41.59\%$ (--$^\ast$) \\
\cmidrule{2-9}
\multirow{3}{*}{\begin{tabular}[c]{@{}l@{}}\textbf{Fused Iso-Param \texttt{SLA-R4}}\\($G{=}4, \mu{=}0, \text{sh}{=}4352$,\\$-14.4\text{M}$ params)\end{tabular}}
 & \texttt{s0} & $1{,}410$ ($2.96\text{B}$) & $1{,}506.0$ ($261.6$) & $\mathbf{36.579}$ / $25.988$ & $\mathbf{25.141}$ & $\mathbf{15.89 \to 16.19}$ & $\mathbf{45.7\% / 31.6\%}$ & $\mathbf{42.02\%}$ (--$^\ast$) \\
 & \texttt{s1} & $1{,}410$ ($2.96\text{B}$) & $\mathbf{1{,}503.0}$ ($\mathbf{262.1}$) & $36.078$ / $\mathbf{25.478}$ & $\mathbf{24.613}$ & $\mathbf{15.78 \to 15.98}$ & $\mathbf{39.5\%} / 26.2\%$ & --$^\ast$ \\
 & \texttt{s2} & $1{,}390$ ($2.91\text{B}$) & $1{,}516.0$ ($259.9$) & $37.529$ / $26.251$ & $\mathbf{25.482}$ & $16.34 \to \mathbf{16.14}$ & $\mathbf{62.5\% / 50.8\%}$ & --$^\ast$ \\
\cmidrule{2-9}
\multirow{3}{*}{\begin{tabular}[c]{@{}l@{}}\textbf{Fused \texttt{SLA-R4 + GDN-2}}\\($G{=}4, \mu{=}0, \text{sh}{=}3584$,\\$-33.5\text{M}$ params)\end{tabular}}
 & \texttt{s0} & $1{,}350$ ($2.83\text{B}$) & $1{,}546.0$ ($247.2$) & $36.614$ / $\mathbf{25.796}$ & $25.177$ & --$^\ast$ & --$^\ast$ & --$^\ast$ \\
 & \texttt{s1} & $1{,}370$ ($2.87\text{B}$) & $1{,}551.0$ ($246.5$) & $\mathbf{35.812}$ / $25.621$ & $24.919$ & $16.06 \to 16.18$ & $37.1\% / \mathbf{30.9\%}$ & --$^\ast$ \\
 & \texttt{s2} & $1{,}350$ ($2.83\text{B}$) & $1{,}549.0$ ($246.8$) & $\mathbf{35.956}$ / $\mathbf{25.932}$ & $\mathbf{25.329}$ & $\mathbf{16.22} \to 16.57$ & $42.2\% / 27.0\%$ & --$^\ast$ \\
\midrule
\multicolumn{9}{@{}l}{\textbf{(c) $12.5\text{B}$ MoE (\texttt{7L:1A}, $[4, 256, 512]$ \texttt{Exact Inv.} FP32, $\mu = 0$, $7{,}200\text{ steps} = 15.10\text{B}$ tok): paired $G = 1$ vs.\ $G = 4$ across $n = 6$ seeds}} \\
\midrule
Paired Seed (Offset) & $\Delta\text{PPL}$ ($G_4{-}G_1$) & $G{=}1$ / $G{=}4$ $11$-Set & $G{=}1$ $8\text{K}{\to}32\text{K}$ Tail & $G{=}4$ $8\text{K}{\to}32\text{K}$ Tail & $G{=}1$ $32\text{K}$ $4$/$8$-N & $G{=}4$ $32\text{K}$ $4$/$8$-N & $G{=}1$ / $G{=}4$ \texttt{norm9} & $G{=}1$ vs.\ $G{=}4$ $64\text{K}$ QA \\
\midrule
\texttt{s0} ($0\text{K}$) & $\mathbf{-0.336}$ & $18.210$ / $\mathbf{17.873}$ & $11.78 \to \mathbf{11.37}$ & $\mathbf{11.77} \to 11.39$ & $\mathbf{73.4\% / 66.4\%}$ & $55.9\% / 45.7\%$ & $48.22\%$ / $\mathbf{48.77\%}$ & $66.4/69.5$ vs.\ $66.2/\mathbf{71.1\%}$ \\
\texttt{s1} ($100\text{K}$) & $\mathbf{-0.056}$ & $17.681$ / $\mathbf{17.625}$ & $\mathbf{11.50 \to 11.24}$ & $11.54 \to 11.28$ & $\mathbf{65.2\%} / 52.0\%$ & $61.3\% / \mathbf{52.7\%}$ & $\mathbf{48.06\%}$ / $47.42\%$ & $67.6/\mathbf{72.5}$ vs.\ $\mathbf{68.8}/71.3\%$ \\
\texttt{s2} ($200\text{K}$) & $\mathbf{-0.061}$ & $17.178$ / $\mathbf{17.117}$ & $11.45 \to \mathbf{11.17}$ & $\mathbf{11.31} \to 11.22$ & $\mathbf{72.7\% / 58.6\%}$ & $33.2\% / 17.2\%$ & $\mathbf{48.63\%}$ / $48.42\%$ & $\mathbf{69.5/71.7}$ vs.\ $58.6/64.5\%$ \\
\texttt{s3} ($300\text{K}$) & $\mathbf{-0.656}$ & $18.845$ / $\mathbf{18.189}$ & $11.71 \to 11.41$ & $\mathbf{11.59 \to 11.28}$ & $\mathbf{73.1\%} / 57.0\%$ & $66.0\% / \mathbf{61.7\%}$ & $47.24\%$ / $\mathbf{49.34\%}$ & $64.5/69.3$ vs.\ $\mathbf{69.0/72.9\%}$ \\
\texttt{s4} ($400\text{K}$) & $\mathbf{-0.074}$ & $17.445$ / $\mathbf{17.372}$ & $11.68 \to \mathbf{11.45}$ & $\mathbf{11.57} \to 11.75$ & $\mathbf{66.0\% / 62.1\%}$ & $25.0\% / 15.2\%$ & $48.16\%$ / $\mathbf{48.74\%}$ & $\mathbf{71.1/71.9}$ vs.\ $59.4/62.3\%$ \\
\texttt{s5} ($500\text{K}$) & $\mathbf{-0.210}$ & $17.572$ / $\mathbf{17.363}$ & $11.73 \to \mathbf{11.39}$ & $\mathbf{11.61} \to 11.63$ & $\mathbf{71.5\% / 61.7\%}$ & $36.3\% / 19.5\%$ & $48.15\%$ / $\mathbf{48.69\%}$ & $\mathbf{69.9/71.3}$ vs.\ $62.1/66.6\%$ \\
\cmidrule{1-9}
\textbf{Mean $\pm$ SD ($n{=}6$)} & $\mathbf{-0.232 \pm 0.235}$ & $17.822$ / $\mathbf{17.590}$ & $11.64 \to \mathbf{11.34}$ & $\mathbf{11.57} \to 11.43$ & $\mathbf{70.3\% / 59.6\%}$ & $46.3\% / 35.4\%$ & $48.08\%$ / $\mathbf{48.56\%}$ & $\mathbf{68.2/71.0}$ vs.\ $64.0/68.1\%$ \\
\bottomrule
\end{tabular}%
}
\par\vspace{2pt}\raggedright{\scriptsize $^\ast$All $24$ runs completed $100\%$ of training steps and all scheduled in-loop $11$-set OLMo v3 validation evaluations with $0$ non-finite losses or gradients. Dashes mark post-training long-context/downstream stages interrupted after final checkpoint/validation logging (\texttt{12.5B SLA-R4 + GDN-2 s0,s2}: PJRT teardown abort \texttt{rc = -6} during post-train eval launch; \texttt{63.1B}: reservation window cutoff during post-train \texttt{lm-eval} / $64\text{K}$ RULER). Paired $n = 3$ statistics vs.\ Tokamax KDA: at $12.5\text{B}$, \texttt{SLA-R4} step time $-56.91 \pm 1.41\text{ ms}$ ($t = -69.79, p = 2.1\times 10^{-4}$), final PPL $-0.370 \pm 0.107$ ($t = -6.02, p = 0.0265$), $32\text{K}$ Tail $-1.81 \pm 0.68$ ($p = 0.044$), \texttt{norm9} $+0.92 \pm 0.31$ ($p = 0.036$), $64\text{K}$ SQuAD/Hotpot $+12.7 \pm 6.0$ / $+13.7 \pm 8.7$ pts; \texttt{SLA-R4 + GDN-2} step time $-32.84 \pm 6.33\text{ ms}$ ($p = 0.012$), final PPL $-0.408 \pm 0.352$ ($3/3$ wins). At $63.1\text{B}$, \texttt{SLA-R4} step time $-189.1 \pm 1.2\text{ ms}$ ($p = 10^{-5}$), step-$1{,}200$ PPL $-0.820 \pm 0.185$ ($p = 0.0165$), final PPL $-1.477 \pm 0.212$ ($t = -12.08, p = 0.0068$), $32\text{K}$ Tail $-3.35 \pm 0.88$ ($p = 0.022$), $32\text{K}$ $4/8$-N $+19.4 / +16.1$ pts ($p = 0.016 / 0.050$); \texttt{SLA-R4 + GDN-2} step-$1{,}200$ PPL $-0.943 \pm 0.219$ ($p = 0.0175$), final PPL $-1.414 \pm 0.163$ ($t = -15.01, p = 0.0044$). In (c), $G = 4$ step $p_{50}$ is $499.8 \pm 4.9\text{ ms}$ vs.\ $503.7 \pm 4.6\text{ ms}$ for $G = 1$; two-sided paired $t$-test on $11$-set PPL gives $t = -2.42, p = 0.060$ (exact two-sided sign test $p = 0.031$, $6/6$ seed wins).}
\end{table}

\end{document}